\documentclass[10pt]{article} % For LaTeX2e
\usepackage[accepted]{tmlr}
\usepackage{amsmath,amsfonts,bm}

\def\eqref#1{equation~\ref{#1}}
\def\1{\bm{1}}

\DeclareMathAlphabet{\mathsfit}{\encodingdefault}{\sfdefault}{m}{sl}
\SetMathAlphabet{\mathsfit}{bold}{\encodingdefault}{\sfdefault}{bx}{n}

\usepackage{hyperref}
\usepackage{url}

\usepackage{graphicx}
\usepackage{amsmath, amssymb}  % For mathematical symbols and environments
\usepackage{amsthm}            % For theorem-like environments
\usepackage{booktabs}          % For improved table formatting
\usepackage{multirow}          % For tables with multi-row cells
\usepackage{makecell}          % For custom cell formatting in tables
\usepackage{hhline}            % For custom horizontal lines in tables
\usepackage{float}             % For improved figure placement
\usepackage{subfigure}         % For subfigures within a figure
\usepackage{graphbox}          % For alignment of graphics
\usepackage{wrapfig}           % For wrapping text around figures
\usepackage{epstopdf}          % For EPS to PDF conversion
\usepackage{algorithmic}
\usepackage{algorithm}
\usepackage{comment}
\usepackage{mathrsfs}

\newtheorem{prop}{Proposition}
\usepackage[capitalise]{cleveref}
\usepackage{wrapfig}

\title{A Distance-based Anomaly Detection Framework for Deep Reinforcement Learning}

\author{\name Hongming Zhang${^*}$ \email hongmin2@ualberta.ca \\
      \addr Department of Computing Science, University of Alberta\\
       Alberta Machine Intelligence Institute (Amii), University of Alberta %Canada
      \and
      \name Ke Sun${^*}$ \email ksun6@ualberta.ca \\
      \addr Department of Mathematical and Statistical Sciences, University of Alberta \\
      Alberta Machine Intelligence Institute (Amii), University of Alberta
      \and
      \name Bo Xu \email boxu@ia.ac.cn\\
      \addr Institute of Automation, Chinese Academy of Sciences
      \and 
      \name Linglong Kong \email lkong@ualberta.ca \\
      \addr Department of Mathematical and Statistical Sciences, University of Alberta \\
      Alberta Machine Intelligence Institute (Amii), University of Alberta
      \and
      \name Martin M{\"u}ller \email mmueller@ualberta.ca \\
      \addr Department of Computing Science, University of Alberta\\
      Alberta Machine Intelligence Institute (Amii), University of Alberta % Canada
      }

\newcommand{\algname}{{MDX}}

\def\month{10}  % Insert correct month for camera-ready version
\def\year{2024} % Insert correct year for camera-ready version
\def\openreview{\url{https://openreview.net/forum?id=TNKhDBV6PA}} % Insert correct link to OpenReview for camera-ready version

\newcommand{\hongming}[1]{{\color{black}#1}}
\newcommand{\ke}[1]{{\color{black}#1}}

\begin{document}

\maketitle
\def\thefootnote{*}\footnotetext{These authors contributed equally.}\def\thefootnote{\arabic{footnote}}

\begin{abstract}
In deep reinforcement learning (RL) systems, abnormal states pose significant risks by potentially triggering unpredictable behaviors and unsafe actions, thus impeding the deployment of RL systems in real-world scenarios. It is crucial for reliable decision-making systems to have the capability to cast an alert whenever they encounter unfamiliar observations that they are not equipped to handle. In this paper, we propose a novel Mahalanobis distance-based (MD) anomaly detection framework, called \textit{\algname}, for deep RL algorithms. \algname \ simultaneously addresses random, adversarial, and out-of-distribution (OOD) state outliers in both offline and online settings. It utilizes Mahalanobis distance within class-conditional distributions for each action and operates within a statistical hypothesis testing framework under the Gaussian assumption. We further extend it to robust and distribution-free versions by incorporating Robust MD and conformal inference techniques. Through extensive experiments on {\ke{classical control environments}}, Atari games, and autonomous driving scenarios, we demonstrate the effectiveness of our MD-based detection framework. \algname \ offers a simple, unified, and practical anomaly detection tool for enhancing the safety and reliability of RL systems in real-world applications.
\end{abstract}

\section{Introduction}

Deep reinforcement learning~(RL) algorithms vary considerably in their performance and are highly sensitive to a wide range of factors, including the environment, state observations, and hyper-parameters~\citep{jordan2020evaluating,patterson2023empirical}. 
% The lack of robustness of RL algorithms hinders their deployment in real-world scenarios, particularly in safety-critical applications, such as autonomous driving~\citep{kiran2021deep}. Recently, the reliability of RL algorithms has garnered substantial attention~\citep{chan2019measuring,gu2022review}, emphasizing the need for anomaly detection-based strategies to build trustworthy RL systems~\citep{haider2023out, danesh2021out, sedlmeier2019uncertainty}. 
\hongming{The lack of robustness in RL algorithms and raised safety concerns surrounding learned policies hinder their deployment in real-world scenarios, particularly in safety-critical applications such as autonomous driving~\citep{kiran2021deep,NEURIPS2022_8be9c134,hu2023potential}. Recently, the reliability of RL algorithms has garnered substantial attention~\citep{chan2019measuring,gu2022review}. Several studies have highlighted the importance of anomaly detection as a crucial component for enabling safe RL systems~\citep{garcia2015comprehensive,hendrycks2021unsolved,muller2022towards}, emphasizing the need for anomaly detection-based strategies to build trustworthy and safe RL systems~\citep{sedlmeier2019uncertainty,danesh2021out,haider2023out}.}

\noindent \textbf{Practical Scenarios.} Observed states often contain natural measurement errors~(random noises), adversarial perturbations, and out-of-distribution (OOD) observations. For instance, consider an autonomous vehicle with malfunctioning or unreliable sensors or cameras. Under such circumstances, the collected data, such as the vehicle's observed location, can be contaminated by random measurement errors. Furthermore, an autonomous car can encounter sensory inputs that have been adversarially manipulated regarding traffic signs. For example, a stop sign maliciously altered to be misclassified as a speed limit sign~\citep{chen2019shapeshifter}, increases the risk of traffic accidents. Regarding OOD samples, an RL policy trained to drive only on sunny days will struggle with observations from rainy days, which are beyond its trained experience. Such OOD observations can lead to safety violations, performance degradation, and potentially catastrophic failures. All these scenarios highlight the necessity of detecting inaccurate sensor signals from noisy state observations to ensure a vehicle's accurate and reliable operation. Beyond autonomous driving, anomaly detection is critical in many other applications involving sequential decision-making. In healthcare, the RL agent might adjust treatment recommendations if it detects a sudden change in the patient's health condition~\citep{hu2022doubly}. Similarly, detecting fraud and anomalous market states in financial systems is becoming increasingly instrumental in preventing substantial financial losses from market manipulation and fraudulent activities~\citep{hilal2022financial}.

\begin{figure}[b!]
	\centering
 \vskip -0.3in
 \subfigure[Unsafe behavior in autonomous driving under noisy sensor signals.]{\includegraphics[width=0.33\textwidth,trim=200 80 300 60,clip]{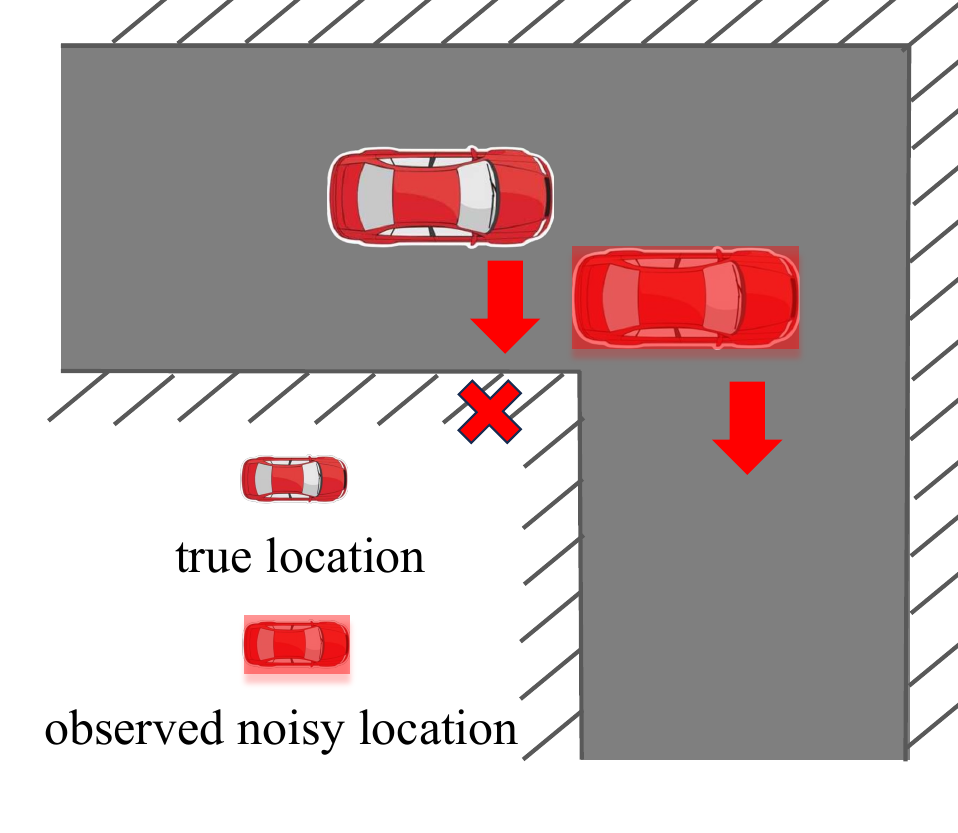} \label{toy example a}}
	\subfigure[Performance degradation when noises injected in policy deployment.]{\includegraphics[width=0.295\textwidth]{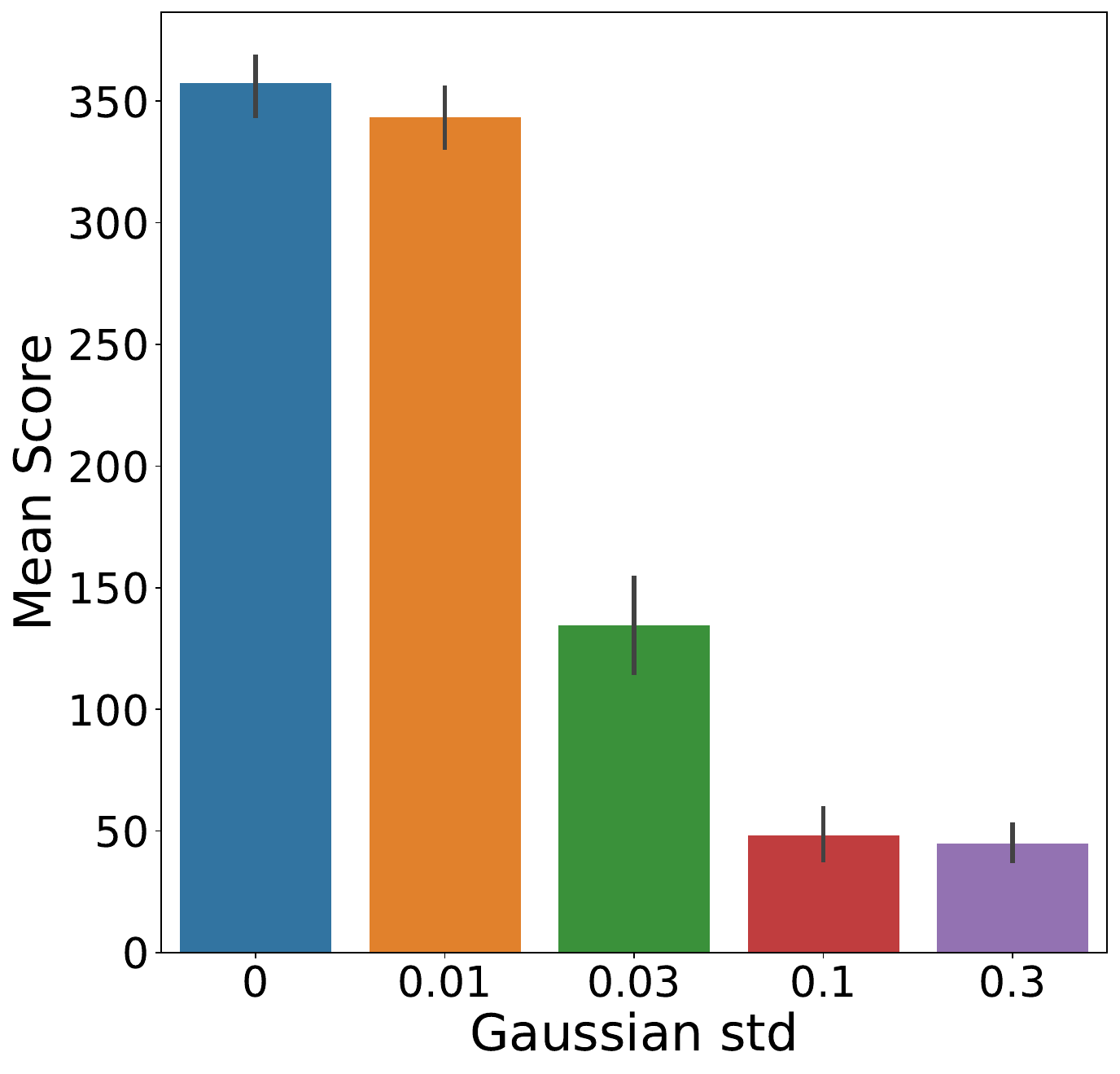} \label{toy example b}}
    \subfigure[Performance degradation when noises injected during policy learning.]{\includegraphics[width=0.305\textwidth]{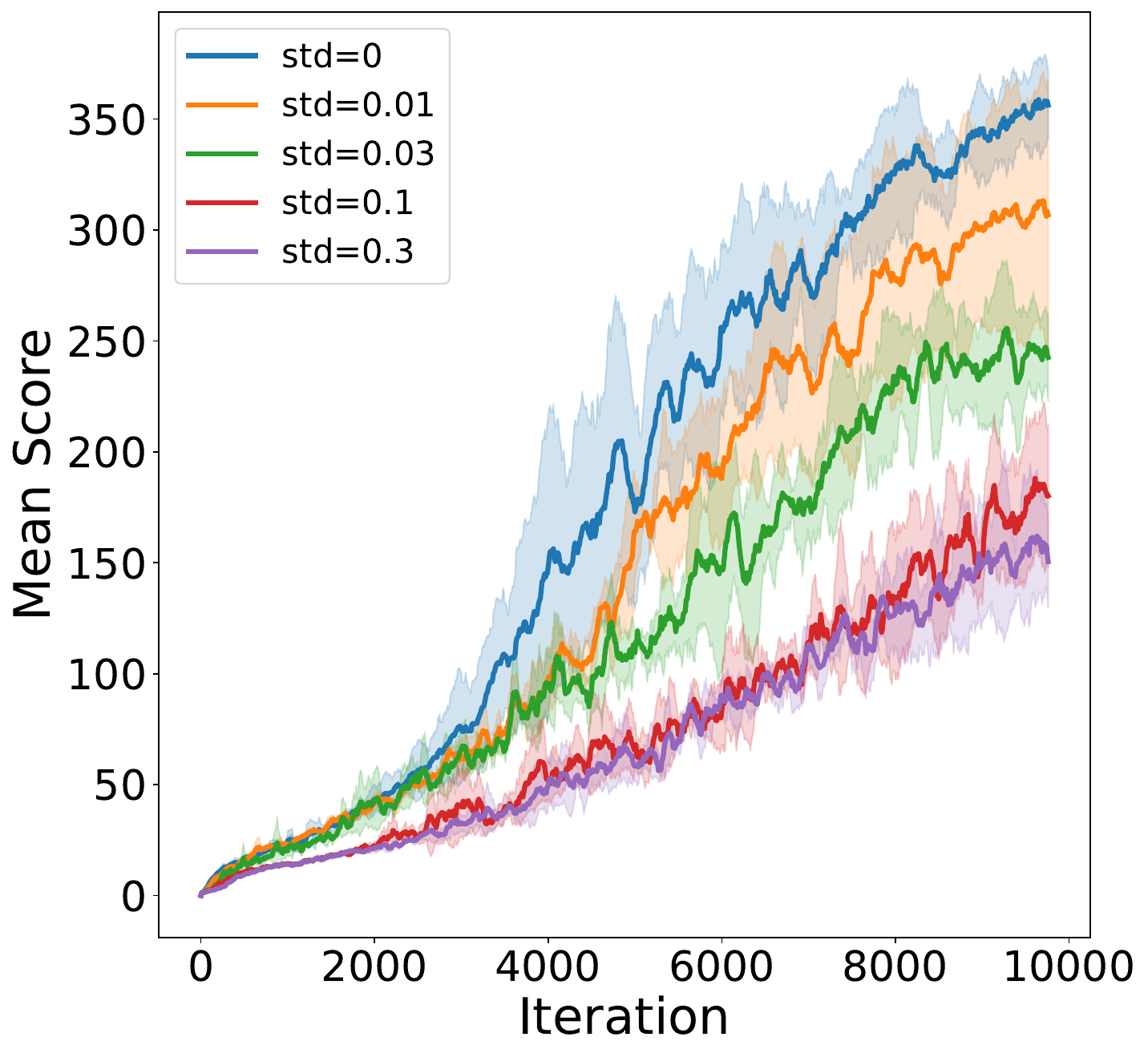} \label{toy example c}}
	\vskip -0.1in
	\caption{(a) An autonomous car navigates using location data observed from sensors such as GPS. Without an effective anomaly detection mechanism, inaccuracies or malfunctions in these sensors can cause the car to prematurely turn right, leading to a collision.
	% based on the observed noisy locations. This unsafe policy can potentially result in the vehicle crashing into the wall. 
	(b) and (c): 
	Performance degradation occurs when noisy states are observed in the Breakout environment. Gaussian noises with increasing standard deviations are injected into the state observations during policy deployment (b) and policy learning (c).
	% (b) The performance of the well-trained policy in a clean environment is evaluated in a noisy environment. In (c), we inject noises for states the agent observes during the learning process.
	}
	\label{toy example}
    \vskip -0.1in
\end{figure}

\noindent \textbf{Motivating Examples.} \cref{toy example a} illustrates a potential collision scenario where an autonomous car, relying on noisy location data
in the red region~(such as GPS coordinate errors), turns right prematurely, risking an accident. Without anomaly detection, 
% Without being equipped with an anomaly detection technique, 
the car reacts incorrectly due to the location error. \cref{toy example b} highlights how increasing measurement errors, represented by the standard deviation of Gaussian noises, dramatically degrade policy performance. For instance, autonomous cars with RL systems may take sub-optimal or unsafe actions when processing noisy sensory signals in deployment. In addition, incorporating excessive noise during online training~(\cref{toy example c}) can severely impair policy learning and diminish performance. These motivating examples underscore the importance of detecting different types of abnormal states for developing trustworthy RL systems in real-world applications.

Our research aims to provide a general framework for applying anomaly detection in deep RL problems, including problem formulation, detection algorithms, and evaluation scenarios. {\ke{This study contributes to anomaly detection, particularly within the context of safe RL, which falls under the broader research field of managing distribution shift in RL; see~\cref{sec:related_work} for detailed discussions.}} Specifically, we strive to develop an effective and unified anomaly detection framework for deep RL in \textit{both offline and online settings}. 

\begin{enumerate}
    \item \textbf{Offline Setting.} In this setting, a dataset is fixed without additional online data collection. Given a pre-trained policy, our objective is to utilize a fixed dataset to develop a distance-based anomaly detector tailored for a pre-trained policy. This detector aims to effectively identify whether a state is an outlier~\footnote{Compared with the classical tasks of policy evaluation and learning in offline RL, our offline setting also utilizes a fixed dataset but specifically focuses on developing detection methods given a fixed policy.}, {\ke{ensuring the reliable operation and stable performance of decision-making systems in deployment.}}
    % when the policy interacts with the (noisy) real world~
    \item \textbf{Online Setting.} In this setting, the RL agent interacts with a noisy environment and continuously updates its policy. Our goal is to develop a detection strategy that identifies state outliers, which are outside the RL system's training experience. Removing these outliers can prevent them from interfering with policy training, {\ke{leading to a robust learning process of RL systems.}}.
\end{enumerate}

Methodologically, we first design an RL outlier detection approach using Mahalanobis Distance~(MD)~\citep{de2000mahalanobis} within a statistical hypothesis test framework and extend it to a robust MD version~\citep{butler1993asymptotics}. 
These strategies are applied \textit{in a parametric manner} under the Gaussian assumption for state features in each class, which may not always be accurate in practice. To address this limitation, we introduce a \textit{non-parametric conformal version} of MD detection to relax the Gaussian assumption. We empirically investigate the effectiveness of these proposed detection approaches in both offline and online settings across a representative set of RL environments, including {\ke{classical control environments,}} Atari games, and autonomous driving. Our contributions can be summarized as follows:

\begin{itemize}
    \item Our primary technical contribution is the design of RL outlier detection strategies based on the concepts of Mahalanobis Distance~(MD), robust MD, and conformal inference. The anomaly detection strategies are specially developed for deep RL within a hypothesis test framework, accommodating both parametric~(Gaussian assumption) and non-parametric~(conformal calibration) approaches.
    \item  Secondly, in our online setting, our anomaly detection can be applied to a dynamic dataset, where the RL policy continually improves when interacting with the environment. This dynamic setting contrasts with the simpler anomaly detection in supervised learning with a static dataset. To address this challenge, we particularly develop \textit{moving window estimation} and \textit{double self-supervised detectors} for anomaly detection in the online RL setting.
    \item To our best knowledge, we are the first to conduct a comprehensive study on distance-based anomaly detection in deep RL, covering all typical types of outliers. Our anomaly detectors can simultaneously identify random, adversarial, and out-of-distribution state outliers.  We perform extensive experiments to verify the effectiveness of our proposed methods in both offline and online settings.
\end{itemize}

\section{Related Work}
\label{sec:related_work}

\noindent \textbf{Anomaly Detection in Reinforcement Learning.} Anomaly detection has yet to be extensively explored in RL. The connection between anomaly detection and RL was first established in~\citep{muller2022towards}; however, their work is mainly conceptual and does not propose practical detection algorithms. Change point detection has been investigated in the tabular setting of RL, particularly in environments described as doubly inhomogeneous under temporal non-stationarity and subject heterogeneity~\citep{hu2022doubly}. They focus on identifying ``best data chunks” within the environment that exhibit similar dynamics for policy learning, while our detection focuses on anomaly detection in \textit{deep} RL scenarios. Prior studies have also probed anomaly detection in specific RL contexts, such as the offline imitation learning with a transformer-based policy network~\citep{wang2024oil} and detecting adversarial attacks within cooperative multi-agent RL~\citep{kazari2023decentralized}. However, these studies are limited to specific scenarios that do not address general anomaly detection, even in single-agent RL. {\ke{\citet{sedlmeier2020policy} introduced a simple policy entropy based out-of-distribution detector in one-class classification problems.}}  \citet{haider2023out} proposed a model-based method using probabilistic dynamics models and bootstrapped ensembles, but this approach {\ke{highly relies on the capability of the learned environment model}} and is also computationally expensive. {\ke{Unlike the previously mentioned detection methods tailored for specific RL areas, our research aims to further enhance this field by developing a distance-based anomaly detection framework applicable to a broad range of deep RL scenarios.}}
% identification of abnormal state observations, particularly prevalent in safety-critical scenarios, such as autonomous driving. Our proposed anomaly detection framework, based on distance measurements, seeks to address these limitations in practical deep RL scenarios.

\noindent \textbf{Distance-based Anomaly Detection.} Recently, there has been a growth of interest in developing anomaly detection strategies in deep learning scenarios~\citep{elmrabit2020evaluation,pang2021deep}. In image classification, Mahalanobis distance~(MD) was effectively applied by~\citep{lee2018simple}, who constructed a Mahalanobis confidence score by training a logistic regression detector using validation samples. This score was evaluated in a supervised way, relying on \textit{a validation set}, and thus it is unsuitable for the RL setting. The ``tied'' covariance assumption used by~\citep{lee2018simple}, where class-conditional distributions of pre-trained features share the same covariance, was criticized as implausible by~\citep{kamoi2020mahalanobis} based on Gaussian discriminant analysis~\citep{klecka1980discriminant}. In contrast, our detection framework \algname \ avoids the unrealistic ``tied covariance'' assumption by estimating variance for each class using quadratic discriminant analysis. This approach extends linear boundaries to quadratic ones between classes, offering a more flexible and accurate detection~\citep{hastie2009elements}. {\ke{Additionally, we have developed a distribution-free detection strategy using conformal prediction, which eliminates the reliance on the Gaussian assumption and potentially extends applicability across a wider range of practices. }}

% cabana2021multivariate
\noindent \textbf{Robust Statistics for RL.} Deep RL algorithms inherently face challenges related to instability and divergence due to the use of function approximation, bootstrapping, and off-policy learning~\citep{sutton2018reinforcement}. Employing Mahalanobis distance~(MD) for anomaly detection can be particularly sensitive during unstable learning phases. The computation of MD is based on Maximum Likelihood Estimate~(MLE), which is susceptible to outliers or noisy data~\citep{rousseeuw1990unmasking}. Robust statistics~\citep{huber2004robust} have been developed to address these robustness problems, especially leveraging robust estimation techniques that are not unduly affected by outliers. For example, Robust MD is a robust version of MD that employs robust estimators, e.g., Minimum Covariance Determinant (MCD)~\citep{rousseeuw1984least, grubel1988minimal}, for location and covariance estimation~\citep{maronna2014robust}. {\ke{Our study enhances the understanding of robust statistical approaches' applicability across a variety of areas in RL, particularly in anomaly detection.}}

\noindent \textbf{Conformal Prediction and Conformal Anomaly Detection.} Conformal anomaly detection~\citep{laxhammar2011sequential,ishimtsev2017conformal} is based on the conformal prediction~\citep{angelopoulos2020uncertainty,teng2022predictive}, a popular, modern technique for providing valid prediction intervals for arbitrarily machine learning models. Conformal prediction has garnered increasing attention as it can provide a simple, distribution-free, and computationally effective way of tuning the distribution threshold. Its validity relies on the data exchangeability condition~\citep{shafer2008tutorial}, where different orderings of samples are equally likely, but recent studies have verified its applicability in scenarios involving distribution shift~\citep{tibshirani2019conformal,barber2023conformal} and off-policy evaluation~\citep{zhang2023conformal}. These examples justify the potential of using conformal inference to detect outliers in the context of RL.

{\ke{
\noindent \textbf{Distribution Shift in RL.} Developing reliable decision-making systems requires effectively addressing distribution shifts in the RL regime. Pertinent research areas include meta RL~\citep{nagabandi2018learning, xu2018meta,ajay2022distributionally}, transfer RL~\citep{taylor2009transfer,parisotto2015actor,zhu2023transfer,bai2024efficient}, continual RL~\citep{khetarpal2022towards, anand2024prediction, abel2024definition}, and robust generalization in RL~\citep{boyan1994generalization, pinto2017robust,zhang2020robust}. While anomaly detection and these subfields all need to handle distribution shifts to create trustworthy RL systems, our work specifically focuses on detecting outliers to ensure reliable decision-making within the capacity of the learned policy. This focus distinguishes our study from the other related subfields to tackle distribution shifts. For a deeper discussion on their differences, please refer to~\citep{muller2022towards}.

}}

\section{Background}\label{sec:pre}

\noindent \textbf{Markov Decision Process.} The interaction of an agent with its environment can be modeled as a Markov Decision Process~(MDP), a 5-tuple ($\mathcal{S}, \mathcal{A}, R, P, \gamma$). $\mathcal{S}$ and $\mathcal{A}$ are the state and action spaces, 
$P:\mathcal{S} \times \mathcal{A} \times \mathcal{S} \rightarrow [0, 1]$ is the environment transition dynamics, $R: \mathcal{S} \times \mathcal{A} \times \mathcal{S} \rightarrow \mathbb{R}$ is the reward function and $\gamma \in (0,1)$ is the discount factor. The policy $\pi$ is continually updated in this online interaction paradigm. 
Compared to the online setting, a recent popular paradigm for reinforcement learning is offline RL~\citep{levine2020offline}. In the offline setting, RL algorithms utilize previously collected data to extract policies without additional online data collection.
% In the offline setting, the parameterized policy $\pi_\theta$ remains fixed, whereas, in the online policy learning scenario, $\pi_\theta$ is continually updated.

\noindent \textbf{Proximal Policy Optimization~(PPO).} The policy gradient algorithm of Proximal Policy Optimization~(PPO)~\citep{schulman2017proximal} has achieved state-of-the-art or competitive performance on Atari games~\citep{bellemare13arcade} and MuJoCo robotic tasks~\citep{todorov2012mujoco}. Typical policy gradient algorithms optimize the expected reward function $\rho\left(\theta, s_{0}\right)=\mathbb{E}_{\pi_{\theta}}\left[\sum_{t=0}^{\infty} \gamma^{t} r\left(s_{t}\right) \mid s_{0}\right]$ by using the policy gradient theorem~\citep{sutton2018reinforcement}. Here $\pi_{\theta}$ is the $\theta$-parameterized policy function. Trust Region Policy Optimization~(TRPO)~\citep{schulman2015trust} and PPO~\citep{schulman2017proximal} utilize constraints and advantage estimation to perform the update by reformulating the original optimization problem with the surrogate loss $L(\theta)$ as:
% \begin{equation}
% 	\begin{aligned}
%  L(\theta)&=\mathbb{E}_{t}\left[\frac{\pi_{\theta}\left(s_{t}, a_{t}\right)}{\pi_{\theta_{\text {old }}}\left(s_{t}, a_{t}\right)} A_{\pi_{\theta_{\text {old }}}}\left(s_{t}, a_{t}\right)\right] =\mathbb{E}_{t}\left[r_t(\theta) A_{t}\left(s_{t}, a_{t}\right)\right],
% 	\end{aligned}
% \end{equation}
\begin{equation}
	\begin{aligned}
 L(\theta)&=\mathbb{E}_{t}\left[\frac{\pi_{\theta}\left(s_{t}, a_{t}\right)}{\pi_{\theta_{\text{old}}}\left(s_{t}, a_{t}\right)} A_{\pi_{\theta_{\text{old}}}}\left(s_{t}, a_{t}\right)\right],
	\end{aligned}
\end{equation}
where $A_{\pi_{\theta_{\text{old}}}}$ is the generalized advantage function (GAE)~\citep{schulman2015high}. PPO introduces clipping in the objective function in order to penalize changes to the policy that make $\pi_{\theta}$ vastly different from $\pi_{\theta_{\text{old}}}$:
% $r_t(\theta)$ vastly different from 1:
% \begin{equation}
% 	\begin{aligned}
% 		\mathbb{E}_{t}\left[\min \left(r_t(\theta) A_{t}\left(s_{t}, a_{t}\right), \operatorname{clip}\left(r_t(\theta), 1-\epsilon, 1+\epsilon\right) A_t\left(s_{t}, a_{t}\right)\right)\right].
% 	\end{aligned}
% \end{equation}
\begin{equation}
	\begin{aligned}
		L^{\text{CLIP}}(\theta) = \mathbb{E}_{t}\left[\min \left( \frac{\pi_{\theta}\left(s_{t}, a_{t}\right)}{\pi_{\theta_{\text {old}}}\left(s_{t}, a_{t}\right)} A_{\pi_{\theta_{\text{old}}}}\left(s_{t}, a_{t}\right), \operatorname{clip}\left(\frac{\pi_{\theta}\left(s_{t}, a_{t}\right)}{\pi_{\theta_{\text{old}}}\left(s_{t}, a_{t}\right)}, 1-\epsilon, 1+\epsilon\right) A_{\pi_{\theta_{\text{old}}}}\left(s_{t}, a_{t}\right)\right)\right],
	\end{aligned}
\end{equation}
where $\epsilon$ is a hyperparameter. 
% Transitions are sampled by multiple actors in parallel~\citep{henderson2018deep}. 
%Although the clipping parameter $\epsilon$ has to be pre-determined, PPO has promising performance. 
We use PPO as the algorithm testbed to examine the efficacy of our anomaly detection framework. However, our detection methods are general and can be easily applied to other RL algorithms~\citep{zhang2020taxonomy} such as DQN~\citep{mnih2015human,hessel2018rainbow}, A3C~\citep{mnih2016asynchronous}, and DDPG~\citep{lillicrap2015continuous,haarnoja2018soft,fujimoto2018addressing,Bai_Zhang_Tao_Wu_Wang_Xu_2023}.

\noindent \textbf{Conformal Prediction.} Conformal anomaly detection~\citep{laxhammar2011sequential,ishimtsev2017conformal} is grounded in conformal prediction~\citep{shafer2008tutorial,angelopoulos2020uncertainty}, which aims to construct a confidence band $\mathcal{C}_{1-\alpha}(X)$ for $Y$ given a random data pair $(X, Y)\sim\mathcal{P}$ and a confidence level $1-\alpha$. Suppose we have a pre-trained model $\widehat{\mu}$ and a calibration dataset $(X_1, Y_1), ..., (X_n, Y_n)$ for conformal prediction.
% not used for training. 
We can then compute a predictive interval for the new sample $X_{n+1}$ to cover the unseen response $Y_{n+1}$ by leveraging the empirical quantiles of the residuals $|Y_i - \widehat{\mu}(X_i)|$ on the calibration dataset. This further leads to valid prediction intervals such that:
\begin{equation}\begin{aligned}\label{eq:conformal}
    \mathbb{P}(Y_{n+1} \in \mathcal{C}_{1-\alpha}(X_{n+1})) \geq 1 - \alpha,
\end{aligned}\end{equation}
where the confidence band is expected to be as small as possible while maintaining the desired coverage. A fundamental quantity in conformal prediction is the \textit{non-conformity measure}, e.g., the residual $|Y_i - \widehat{\mu}(X_i)|$, which measures how “different” an example is relative to a set of examples~\citep{vovk2005algorithmic}.

{\hongming{
\section{Mahalanobis Distance-based~(MDX) Detection Framework}

For a deep RL agent, acting on anomalous inputs could result in hazardous situations. Therefore, developing suitable anomaly detectors for deep RL agents is particularly important in safety-critical scenarios. \cref{fig:Detection Pipeline} illustrates the operational flow of our \algname \ framework.

\noindent \textbf{Description of Detection Framework.} Our detection framework is structured around two core components: feature extraction and detector estimation. The process begins by assessing whether a state is anomalous, which is crucially dependent on the associated policy. A state that prompts the policy to initiate a potentially unsafe action is labeled as an outlier. Specifically, we input the state into the policy network and extract the feature vector from the penultimate layer of this network. We categorize states according to the actions determined by the policy, based on the intuition that states associated with the same action share similar features. For each action class, we estimate the mean value ($\mu$) and covariance matrix ($\Sigma$) of the feature vectors as the class centroid. A threshold is set as the class boundary that partitions the feature space into inliers and outliers. 

\begin{figure*}[t]
\centering
% \vskip 0.1 in
\subfigure{\includegraphics[width=0.8\textwidth]{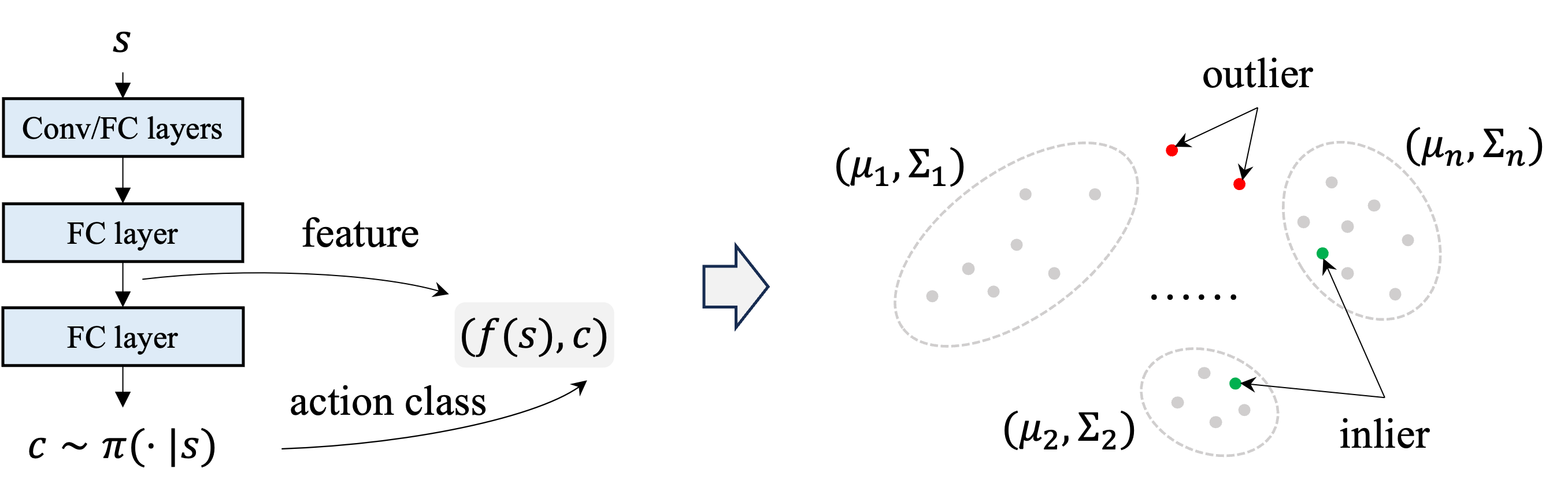}}
% \vskip -0.15in
\caption{\textbf{The detection pipeline of \algname}. We feed the state into the policy network to extract the feature vector and identify its class. For each class, we estimate $(\mu,\Sigma)$ and establish a detection threshold depicted as a dashed ellipse. To determine whether a new state is an outlier, we evaluate its features and compute the distance to the class centroids. If the distance falls below the set threshold, the state is classified as an inlier (green points). Conversely, the state is marked as an outlier (red points).}
\label{fig:Detection Pipeline}
    % \vskip -0.1in
\end{figure*}

After the class centroids are estimated, we can determine whether a new state is an outlier by computing its distance from the established class centroids using the Mahalanobis distance based on its feature vector. The Mahalanobis distance is a measure of the distance between a point and a distribution, which in this case is represented by the class centroids. A state is deemed an outlier if the distance surpasses the predefined threshold. An appropriate threshold can balance the trade-off between false positives and false negatives, ensuring that the detection system is both accurate and reliable. By ensuring that only states within the policy's capability are considered valid, \algname \ enhances the safety and reliability of the RL system.

Our detection framework is generic and can be applied to any agents that operates based on a learned neural network. Based on the proposed detection framework, we instantiate two detection methods in \cref{sec:offline setting}: distribution-based detection under the assumption of Gaussian distribution, and distribution-free detection, which is based on conformal prediction. The former method relies on Chi-square distribution to determine the threshold, while the latter method employs the conformity score to establish the threshold. We then extend the detection framework to the online setting in \cref{sec:online setting}, where the detectors are updated continuously as the agent interacts with the environment.
}}

\section{Anomaly Detection in the Offline RL Setting}
\label{sec:offline setting}

{\ke{Our MDX detection framework mainly induces two kinds of detection algorithms, including distribution-based detection under Gaussian assumption in~\cref{subsec:distributionbased} and distribution-free detection by leveraging conformal prediction in~\cref{subsec:distributionfree}. Finally, an integrated detection algorithm pipeline is provided, incorporating all variants of detection strategies in the RL context.}}

\ke{\subsection{Distribution-based Detection under Gaussian Assumption}
\label{subsec:distributionbased}}

\noindent \textbf{Gaussian Assumption.} The given pre-trained parameterized RL policy $\pi_{\theta}$ is a discriminative softmax classifier, $\pi(a_t=c|s_t)=\exp \left(\mathbf{w}_{c}^{\top} f(s_t)+b_{c}\right) / \sum_{c^{\prime}} \exp \left(\mathbf{w}_{c^{\prime}}^{\top} f(s_t)+b_{c^{\prime}}\right)$, where $\mathbf{w}_{c}$ and $b_{c}$ are the weight and bias of the policy classifier for action class $c$. The function $f(\cdot)$ represents the output of the penultimate layer of the policy network $\pi_{\theta}$, serving as the state feature vector. Here, $C=|A|$ is the size of the action space, and $\mu_c$ is the mean vector of $f(s)$ corresponding to the action class $c$~\footnote{\hongming{Our MDX detection framework currently focuses on environments with the discrete action spaces. For continuous action spaces, a natural solution is to discretize the actions into several bins and then follow the same detection pipeline, which deserves further validation.}}. If we assume that the class-conditional distribution follows a multivariate Gaussian distribution sharing a single covariance $\Sigma$ (tied covariance) in a generative classifier, i.e., $\pi(f(s) \mid a=c)=\mathcal{N}\left(f(s) \mid \mu_{c}, \Sigma \right)$, then the posterior distribution of $f(s)$ matches the form of a discriminative softmax classifier~\citep{lee2018simple}. This equivalence implies that $f(s)$ fit a Gaussian distribution under $\pi_{\theta}$. We approximate state feature vectors with a class-conditional Gaussian distribution with $\mu_c$ and $\Sigma_c$ \textit{for each action class}, rather than using a single "tied" covariance $\Sigma$ across all action classes~\citep{kamoi2020mahalanobis}. 

% \subsubsection{Vanilla MD-based Detection}

\noindent \textbf{Vanilla MD-based Detection.} An MD-based detection based on Gaussian assumption can be immediately developed based on the mean vectors $\mu_c$ and the covariance matrix $\Sigma_c$ calculated from $f(s)$ for each action class $c$. We first collect  $N_c$ state action pairs $\{ (s_i, a_i)\}$, separately for each action class $c$, and compute the empirical class mean and covariance of $c$:
\begin{equation}\begin{aligned}\label{eq:MD}
		\widehat{\mu}_{c}&=\frac{1}{N_{c}} \sum_{i: a_{i}=c} f\left(s_{i}\right), \quad \widehat{\Sigma}_c&=\frac{1}{N_c} \sum_{i: a_{i}=c}\left(f\left(s_{i}\right)-\widehat{\mu}_{c}\right)\left(f\left(s_{i}\right)-\widehat{\mu}_{c}\right)^{\top}.
\end{aligned}\end{equation}
In distance-based detection, a straightforward metric is Euclidean distance~(ED). However, MD generally outperforms ED in many tasks~\citep{lee2018simple,kamoi2020mahalanobis,ren2021simple}, as it incorporates the additional data covariance information to normalize the distance scales. Following the estimation in~\cref{eq:MD}, we derive the class-conditional Gaussian distribution to characterize the data structure within the state representation space for each action class. For each state $s$ observed by the agent, we compute its \textit{Detection Mahalanobis Distance} $M(s)$ between $s$ and the nearest class-conditional Gaussian distribution by:
\begin{equation}\begin{aligned}\label{eq:DetectionMD}
		M(s)=\min_{c} \left(f(s)-\widehat{\mu}_{c}\right)^{\top} \widehat{\Sigma}_c^{-1}\left(f(s)-\widehat{\mu}_{c}\right).
\end{aligned}\end{equation}
Unlike the previous work~\cite {lee2018simple}, which defined a Mahalanobis confidence score based on a binary classifier in a validation dataset, we utilize $M(s)$ as the detection metric within a statistical hypothesis test framework. Proposition~\ref{prop} demonstrates that $M(s)$ follows a Chi-squared distribution under the Gaussian assumption.

\begin{prop}(Test Distribution of Detection Mahalanobis distance $M(s)$) Let $f(\mathbf{s})$ be the $p$-dimensional state random vector for action class $c$. Under the Gaussian assumption $P(f(\mathbf{s}) | a=c)=\mathcal{N}\left(f(\mathbf{s}) \mid \mu_{c}, \Sigma_c \right)$, the Detection Mahalanobis Distance $M(\mathbf{s})$ in~\cref{eq:DetectionMD} is Chi-Square distributed: $M(\mathbf{s})\sim \chi^2_p$.
\label{prop}
\end{prop}

Please refer to~\cref{appendix:chi_square} for the proof. Based on Proposition~\ref{prop}, we can define a threshold $\Theta = \chi^2_p(1-\alpha)$ by selecting a $\alpha$ value from the specified Chi-Squared distribution to distinguish normal states from outliers.
Given a new state observation $s$ and a confidence level $1-\alpha$, if $M(s)>\Theta$, $s$ is detected as an outlier.

% \subsubsection{Robust MD-based Detection}
% \noindent \textbf{Motivation.} 
\noindent \textbf{Robust MD-based Detection.} The estimation of $\mu_{c}$ and $\Sigma_c$ in~\cref{eq:MD} relies on Maximum Likelihood Estimate~(MLE), which is sensitive to the presence of outliers in the dataset~\citep{rousseeuw1990unmasking}. As the offline data collected from the environment tends to be noisy, directly introducing MD for outlier detection in RL easily results in a less statistically effective estimation of $\mu_{c}$ and $\Sigma_c$, thus undermining the detection accuracy for outliers. This vulnerability of the MD-based detector against noisy states prompts us to instantiate \algname \ with a more robust estimator~\citep{huber2004robust}.

\begin{figure*}[b!]
	\centering
	\subfigure[Random Outliers.]{\includegraphics[width=0.32\textwidth]{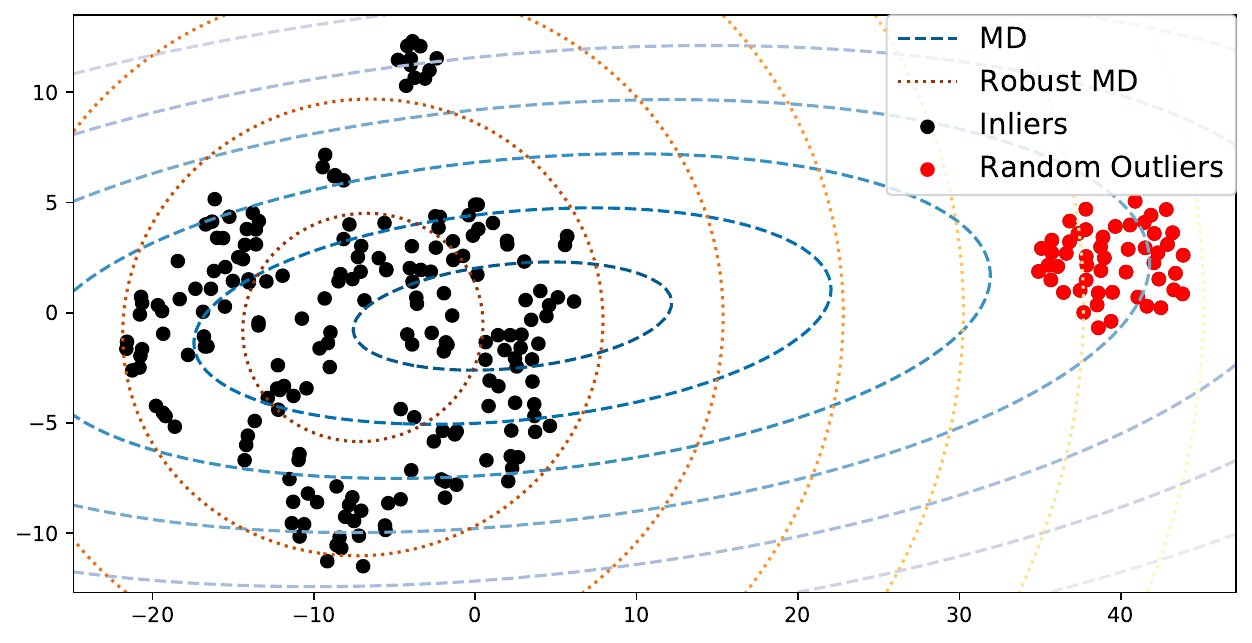}}
	\subfigure[Adversarial Outliers.]{\includegraphics[width=0.32\textwidth]{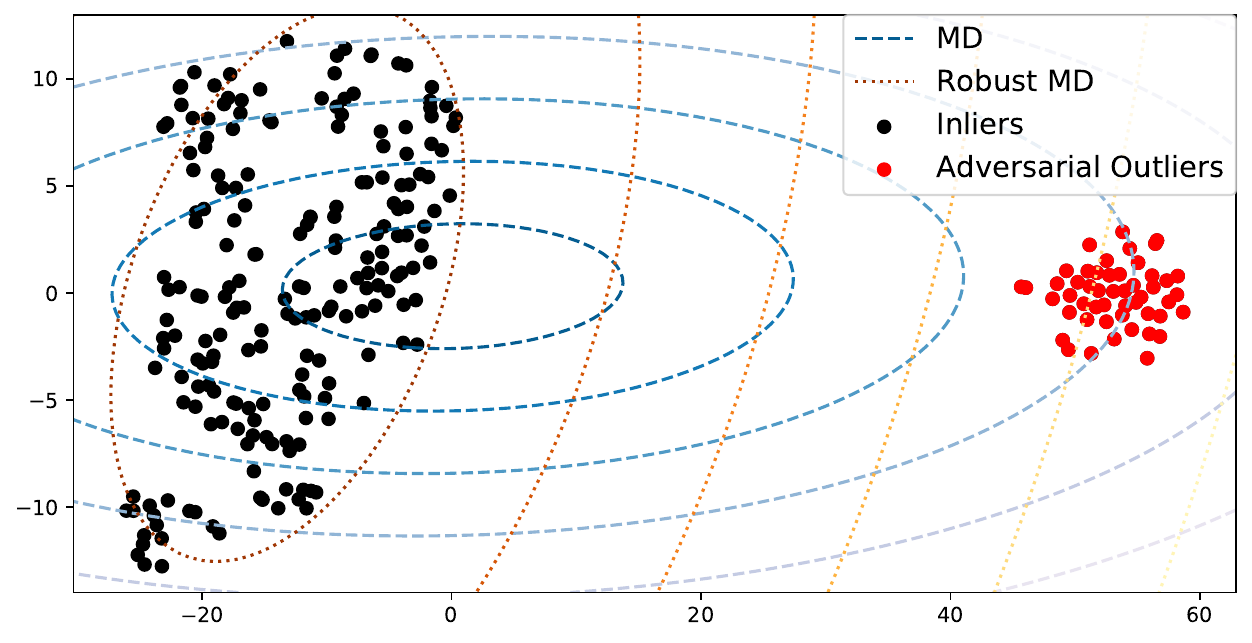}}
	\subfigure[OOD Outliers.]{\includegraphics[width=0.32\textwidth]{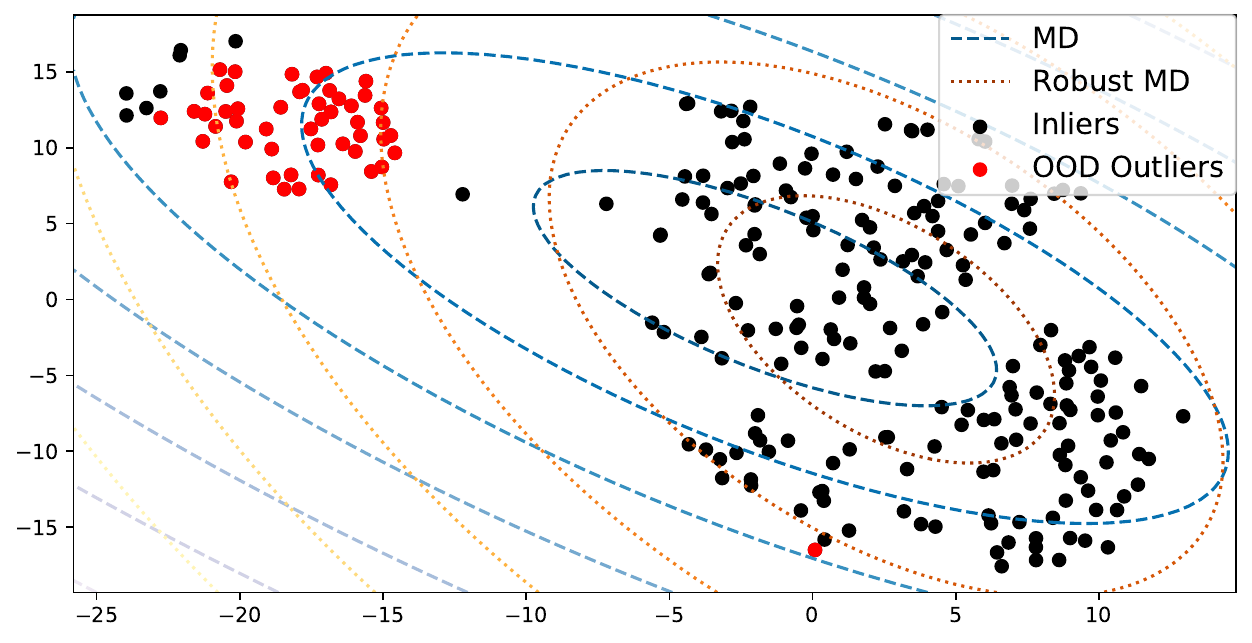}}
	\vskip -0.1in
	\caption{Contours under the estimation based on MD and Robust MD across different outlier types on Breakout. Black and red points denote inliers and outliers, respectively. The dimension of state feature vectors after a pre-trained PPO policy is reduced by t-SNE~\citep{van2008visualizing}.}
	\label{fig:contours}
\end{figure*}

To this end, we apply the Minimum Covariance Determinant (MCD) estimator~\citep{hubert2010minimum} to estimate $\mu_{c}$ and $\Sigma_c$ by only using a subset of all collected samples. It only uses the observations where the determinant of the covariance matrix is as small as possible. Concretely, MCD determines the subset $J$ of observations with a size $h$, while minimizing the determinant of the sample covariance matrix calculated solely from these $h$ points. The choice of $h$ determines the trade-off between the robustness and efficiency of the estimator. The robust MCD mean vector $\widehat{\mu}_c^{\text{rob}}$ and covariance matrix $\widehat{\Sigma}^{\text{rob}}_c$ in the action class $c$ are computed as
\begin{equation}\begin{aligned}\label{eq:RMD}
		\widehat{\mu}_c^{\text{rob}}=\frac{1}{h} \sum_{i: i\in J, a_{i}=c} f\left(s_{i}\right), \ \ J=\left\{\text { set of } h \text { points }:\left|\widehat{\Sigma}_{J}\right| \leq\left|\widehat{\Sigma}_{K}\right| \text { for all subsets } \mathrm{K} \right\},
%		\widehat{\Sigma}_c^{\text{rob}}&=\frac{1}{h} \sum_{i: i\in J,  a_{i}=c}\left(f\left(s_{i}\right)-\widehat{\mu}_c^{\text{rob}}\right)\left(f\left(s_{i}\right)-\widehat{\mu}_c^{\text{rob}}\right)^{\top}，
	\end{aligned}\end{equation}
where we set $h$ as $(\texttt{number\_of\_samples} + \texttt{number\_of\_features} + 1) / 2$~\citep{rousseeuw1984least}. 
% which is implemented in popular Python packages. 
$K$ represents the total number of subsets that contain $h$ points. In practice, the MCD estimator can be efficiently solved by the FAST-MCD algorithm~\citep{hubert2010minimum} instead of performing a brute-force search over all possible subsets. Akin to Mahalanobis Distance, we define the \textit{Detection Robust Mahalanobis Distance} $M_{\text{rob}}(s)$ as robust detection metric:
\begin{equation}\begin{aligned}\label{eq:DetectionMCD}
		M_{\text{rob}}(s)=\min_{c} \left(f(s)-\widehat{\mu}_{c}^{\text{rob}}\right)^{\top} \widehat{\Sigma}^{\text{rob}-1}_c\left(f(s)-\widehat{\mu}_{c}^{\text{rob}}\right).
	\end{aligned}\end{equation}
Since the robust Mahalanobis distance can still approximate the true Chi-squared distribution~\citep{hardin2005distribution}, we still employ the threshold value $\Theta = \chi^2_p(1-\alpha)$ for detecting outliers as in the MD case.

% \noindent \textbf{Potential Advantages of Robust MD on Real Data.} 
As a motivating example, \cref{fig:contours} displays contours computed by both MD and Robust MD detection methods for state feature vectors in the Breakout game from the popular Atari benchmark~\citep{bellemare13arcade,brockman2016openai} with different types of outliers. These results demonstrate that estimation based on Robust MD is less vulnerable to outlying states~(red points) and better fits inliers~(black points) than MD. This robust parameter estimation highlights the potential advantage of Robust MD for RL outlier detection, where the data used for estimation tends to be noisy.  

\ke{\subsection{Distribution-free Detection by Conformal Inference}
\label{subsec:distributionfree}}

% \noindent \textbf{Motivation.} 
Although robust MD-based detection is less vulnerable to noise in RL environments, both  MD and robust MD strategies heavily rely on the Gaussian assumption to construct the detection thresholds based on Proposition~\ref{prop}. This distribution assumption is often violated in practice, diminishing the effectiveness of MD and robust MD. In contrast, conformal prediction offers a mathematical framework that provides valid and rigorous prediction distribution without assuming a specific underlying data distribution. The resulting conformal anomaly detection circumvents the limitation of the distribution assumption, potentially improving the detection efficacy.

In the context of RL, conformal anomaly detection evaluates how a state conforms to a model's current prediction distribution, thereby discriminating abnormal states.  As a distribution-free detection approach, conformal anomaly detection can enhance the distance-based detectors by additionally tuning the anomaly threshold in the calibration dataset. To design the conformal anomaly detection method, we leverage the Detection Mahalanoibis Distance $M(s)$ as the \textit{non-conformity score}, which measures how dissimilar a state is from the instances in the calibration set. Following split conformal inference~\citep{papadopoulos2002inductive,shafer2008tutorial}, we split the the previously collected offline dataset into the the calibration set $\mathcal{D}_{\text{cal}}$ and the evaluation set. A simple way is to evaluate the quantiles of the resulting empirical distribution to create the corresponding confidence band. Using the calibration set $\mathcal{D}_{\text{cal}}$, we define the fitted quantiles $\widehat{Q}_{1-\alpha}^c$ of the conformity scores for the action class $c$ as follows:
\begin{equation}\begin{aligned}
\label{eq:conformity scores}
    \widehat{Q}_{1-\alpha}^c = \inf \left\{ q: \left( \frac{1}{N_c} \sum_{s_i \in \mathcal{D}_{\text{cal}},a_i=c} \mathbf{1}_{\{M^c(s_i) \leq q \}} \right) \geq 1-\alpha \right\},
\end{aligned} \end{equation}
where each $(s_i, a_i)$ is drawn from the calibration set $\mathcal{D}_{\text{cal}}$ and $c$ is calculated by $c = \arg\min M^c(s_i)$ in $M^c(s_i)$ among all action classes. Finally, we use the class-dependent and well-calibrated detection thresholding $\Theta=\widehat{Q}_{1-\alpha}^c$ in conformal MD-based detection instead of $\chi_p^2(1-\alpha)$ used in MD and Robust MD strategies.

\ke{\subsection{An Integrated MD-based Detection Algorithm in the Offline Setting}}

\ke{In practice, it is unclear whether the collected data is noisy or to what extent the Gaussian assumption is violated. Therefore, we provide an integrated algorithm pipeline.} \cref{alg:test} summarizes all the variant detection strategies of \algname \ in the offline setting. We compute the (robust) mean vector and covariance matrix among the
% (the dimension-reduced) 
% $p$-dimensional 
state feature vectors in the penultimate layer of $\pi_{\theta}$ for each action class. Next, given a state observation $s$, we compute the detection Mahalanobis distance
% evaluate the anomaly thresholding 
$d=M(s)$ or $d=M_{\text{rob}}(s)$ and compare it with the threshold $\Theta = \chi^2_p(1-\alpha)$ under the Gaussian assumption or $\Theta = \widehat{Q}^c_{1-\alpha}$ from distribution-free conformal quantiles. If $d > \Theta$, $s$ is detected as an outlier. Conversely, if $d \leq \Theta$, $s$ is identified as an inlier.
% A detailed practical algorithm is given in Algorithm~\ref{alg:test}.

\begin{algorithm}[t!]
	\caption{\algname \ Detection Framework in the Offline Setting}
	\begin{algorithmic}[1] %[1] enables line numbers
		\STATE \textbf{Input}: The given policy $\pi_{\theta}$, the dimension of state feature vectors $p$, and a confidence level $1-\alpha$.
		\STATE \textbf{Output}: Detection labels $\{y_s\}$ for each $s$ in the evaluation trajectory. 
		\STATE {\color{blue} / *  Step 1: Detection Design by Estimating Mean and Covariance * /}
		\STATE Given state action pairs $\{ (s_i, a_i)\}$ where $a_i \sim \pi_{\theta}(\cdot | s_i)$.
		\FOR{each action class $c$}	
		\IF{we choose MD detection}  
		\STATE Estimate $\widehat{\mu}_c$ and $\widehat{\Sigma}_c$ via~\cref{eq:MD}. {\color{blue} / * Approach 1: MD Detection * /} 
		\ELSIF{we choose Robust MD detection}
		\STATE Estimate $\widehat{\mu}^{\text{rob}}_c$ and $\widehat{\Sigma}^{\text{rob}}_c$ via~\cref{eq:RMD,eq:DetectionMCD}. {\color{blue} / * Approach 2: Robust MD Detection * / } \\
        % \ELSE \STATE Estimate $\widehat{\mu}_c$, $\widehat{\Sigma}_c$,  $M^c(s_i)$ $\forall{s_i}$ and calibrate $\widehat{Q}^c_{1-\alpha}$ via \cref{eq:conformity scores}  {\color{blue} / * Approach 3: Conformal MD-based Detection * /}  \\
        \ELSE \STATE Estimate $\widehat{\mu}_c, \widehat{\Sigma}_c$ via~\cref{eq:MD}, calibrate $\widehat{Q}^c_{1-\alpha}$ via~\cref{eq:conformity scores}  {\color{blue} / * Approach 3: Conformal MD-based Detection * /}  \\
		\ENDIF
		\ENDFOR
		\STATE {\color{blue} / *  Step 2: Detection Deployment * /}
		\FOR{$s$ in the noisy environment}	
  % \STATE Compute distance $d:\{M(s),M_{\text{rob}}(s)\}$ via~\cref{eq:DetectionMD} or~\cref{eq:DetectionMCD}.\\
    \STATE Compute distance $d=M(s)$ or $d=M_{\text{rob}}(s)$, and threshold $\Theta=\chi^2_p(1-\alpha)$ or $\Theta=\widehat{Q}^c_{1-\alpha}$. \\
  % \IF{we choose MD and Conformal MD-based detection}  
		% \STATE Evaluate $M(s)$ via Eq.~\ref{eq:DetectionMD} 
  % \ELSE
  % \STATE Evaluate $M_{\text{rob}}(s)$ via Eq.~\ref{eq:DetectionMCD}.
  % \ENDIF
    % \STATE   Set Detection label $y_s=1$ if $M(s) > \chi^2_p(1-\alpha)$ ($M_{\text{rob}}(s) > \chi^2_p(1-\alpha)$ or $M(s) > \widehat{Q}^c_{1-\alpha}$) else $y_s=-1$.
    \STATE   Set Detection label $y_s=1$ if $d > \Theta$ else $y_s=-1$.
		\ENDFOR
	\end{algorithmic}
	\label{alg:test}
\end{algorithm}

\section{Anomaly Detection in the Online RL Setting}
\label{sec:online setting}

In the online RL setting~\citep{sutton2018reinforcement,dong2020deep}, a policy is updated continuously, unlike the fixed pre-trained policy used in our offline setting. Robust policy training with noisy states is crucial in safe RL, as the agents are more likely to encounter state outliers during training. In this section, we extend \algname \ to the online RL training scenario. Unlike the offline setting, the challenge here stems from the dynamic nature of policy updates, requiring our detector to adapt to the evolving distribution of feature vector outputs. The complexity increases when the improved policy starts gathering new samples through exploration, posing a fundamental challenge in an online RL framework. An effective detection system must differentiate between actual noisy observations and newly collected data through exploration. Training the RL agent and estimating the detector are interleaved in a noisy online environment. Various options for managing detected outliers during training include removing or denoising the outlier states. In our detection framework, we focus on direct removal and assess the resulting learning curves in the presence of noisy states during the training process. To address the challenges in detecting abnormal states in the online training setting, we propose \textit{Moving Window Estimation} and \textit{Double Self-supervised Detectors}, both of which are pivotal for the empirical success of our anomaly detection approach.

\begin{algorithm}[t!]
	\caption{\algname \ Detection Framework in the Online Setting, PPO Style}
	\begin{algorithmic}[1] %[1] enables line numbers
		\STATE Initialize policy network $\pi_{\theta}$ and estimator $\widehat{\mu}_c$ and $\widehat{\Sigma}_c$. 
		\STATE Initialize confidence level $1-\alpha$, the moving window size $m$, inlier and outlier buffers $\mathcal{B}_I$, $\mathcal{B}_O$.
		\FOR{iteration $=1,2,...,K$}
		\FOR{actor $=1,2,...,N$}	
		\STATE Run policy $\pi_{\theta}$ in environment for $T$ timesteps.
		% \STATE Compute advantage estimates $\hat{A}_1,...,\hat{A}_T$.
		% \IF{iteration $\le K/2 $}
		% \STATE Add inlier trajectories to $\mathcal{B}_I$ automatically and optimize policy $\pi_{\theta}$ using inlier trajectories.
		% \ELSE
		% \STATE Reduce feature dimension from $p$ to $k$ by PCA. \\
		% \STATE Evaluate trajectories via (robust) MD metrics.
      \STATE Compute distance $d=M(s)$, and threshold $\Theta=\chi^2_p(1-\alpha)$ or $\Theta=\widehat{Q}^c_{1-\alpha}$. \\
		% \IF{$M(s) \le \chi^2_p(1-\alpha), \text{or } M_{\text{rob}}(s))  \le \chi^2_p(1-\alpha)$ ,  \text{or } $M(s)  \leq \widehat{Q}^c_{1-\alpha}$}
            \IF{$d \le \Theta $}
		\STATE Add it to $\mathcal{B}_I$.
		\ELSE
		\STATE Add it to $\mathcal{B}_O$.
		\ENDIF
            \ENDFOR
		\STATE Optimize policy $\pi_{\theta}$ using inlier trajectories.
            \STATE \hongming{Update $\widehat{\mu}_c$ and $\widehat{\Sigma}_c$ based on data in the moving window every $N_c$ new samples come.}
		% \STATE Update $\widehat{\mu}_c$ and $\widehat{\Sigma}_c$ of the two detectors based on $\mathcal{B}_I$ and $\mathcal{B}_O$ respectively every $N_c$ samples.
		% \ENDIF 
		\ENDFOR
	\end{algorithmic}
	\label{alg:train}
\end{algorithm}

\noindent \textbf{Moving Window Estimation.} 
In the online setting, improving the policy $\pi_{\theta}$ causes a shift in the data distribution within the replay buffer as the agent interacts with the environment~\citep{NEURIPS2019_fa7cdfad,xiao2019online}. To effectively utilize information from the updated data distribution, we maintain a moving window to store experiences throughout the interaction steps. 
\hongming{The moving window operates like a first-in-first-out buffer, storing the most recent samples and discarding the oldest ones. Data falling outside the window is cast away. The moving window can be adjusted to either prioritize a long historical context with a larger window size or consider only more recent experiences with a smaller size. In our experiments, since the environments we consider have finite horizons with restarts, catastrophic forgetting is not a concern. We set a small window size of 5120 to balance past and recent data used for detection estimation.}
Based on the constantly updated state feature vectors, $\mu_c$ and $\Sigma_c$ are continually estimated. This continuous updating allows us to accurately track the state feature distribution, ensuring that our detector remains sensitive to recent and historical data shifts.

\noindent \textbf{Double Self-Supervised Detectors.} 
Our current detector is continually refined using self-detected inliers, while any detected outliers are promptly discarded. However, a more practical approach is to leverage these outliers to create a complementary detector for outliers. This secondary self-supervised detector validates the detection results from the primary detector. 
For example, if the primary detector classifies a state as an inlier and the secondary detector agrees that it is not an outlier, the state is confidently classified as such. Conversely, if there is a difference between the discrimination of the two detectors, the state is randomly classified as either an outlier or an inlier. 
In the event of disagreement, this random classification is motivated by the need to avoid systematic bias that could arise from consistently favoring one detector's output over the other.
By introducing randomness, we ensure the system remains fair and does not overly rely on potentially flawed outputs from either detector. This approach also preserves the system's ability to learn and adapt over time, preventing the reinforcement of incorrect classifications. The double-detector system thus enhances the robustness and reliability of the detection process, ensuring more accurate and consistent identification of abnormal states.

% We demonstrate the effectiveness of our double self-supervised detectors through our extensive experiments and provide the ablation study in Appendix~\ref{appendix:double}.
%
\noindent \textbf{MD-based Detection Algorithm in the Online Setting.} \cref{alg:train} outlines our MD-based detection procedure for online RL, incorporating both moving window estimation and double self-supervised detectors. To update our double detectors, inliers and outliers are stored in buffers $\mathcal{B}_I$ and $\mathcal{B}_O$, respectively. For each class, a window size $m$ is specified. Within each class, the state-action pairs in the window are used to estimate $\widehat{\mu}_c$ and $\widehat{\Sigma}_c$.
% ($\widehat{\mu}_c^{\text{rob}}$ and $\widehat{\Sigma}_c^{\text{rob}}$). 
These parameters are updated after every $N_c$ newly collected data points in the window for action class $c$. This adaptive updating mechanism ensures that the detectors remain responsive to evolving data distributions.
% Given window size $m$, $m N_c$ state action pairs are used to estimate $\widehat{\mu}_c$ and $\widehat{\Sigma}_c$ ($\widehat{\mu}_c^{\text{rob}}$ and $\widehat{\Sigma}_c^{\text{rob}}$) for each class. Within each moving window update, for each action class $c$, $N_c$ pairs are newly collected while maintaining the previous $(m-1) N_c$ state-action pairs. 

\noindent \textbf{Online Anomaly Detection Procedure.} \hongming{Since our detection method relies on features extracted from the penultimate layer of the policy network, instead of training the policy from scratch, we pretrain the policy to ensure these features capture meaningful information about the environment. This pre-trained policy results in more meaningful state features and enhances the detection procedure, contributing to a rapid assessment among distinct detection algorithms through their learning curves. Moreover, deploying a randomly initialized policy in a real-world scenario is unreliable. Instead, it is common practice to use a pre-trained policy as a warm start and then further improve it. For example, in recommendation systems, a pre-trained policy is deployed initially to provide recommendations, and user feedback, such as click-through rates (CTR), is used to iteratively update the online policy. Similarly, within our online detection algorithm, we pre-train a policy using inlier data as a warm start.}
% In real-time scenarios, such as recommendation systems, we typically first deploy a pre-trained policy as a warm start in the online system to provide initial recommendations for each user. Feedback from users, such as the click-through rate (CTR), is then observed to update the online policy iteratively. Similarly, within our online detection algorithm, we pre-train a policy as a warm start. {\ke{This pre-trained policy results in more meaningful state features and enhances the detection procedure, contributing to a rapid assessment among distinct detection algorithms through their learning curves.}} 
After pre-training, the policy is introduced to the noisy environment for further online learning. Throughout this process, our \algname \ framework is used to identify outliers in the subsequent training phases. We then evaluate the training performance of algorithms equipped with these detection mechanisms. This systematic approach facilitates the gradual refinement of the policy while concurrently integrating outlier detection to enhance robustness in real-world settings.

\section{Experiments}
\label{sec:experiments}

\hongming{We first conduct experiments on both feature-input and image-input tasks to verify the effectiveness of our \algname \ framework in both offline and online settings. For feature-input tasks, we choose two classical control environments in OpenAI gym~\citep{brockman2016openai}, including Mountain Car~\citep{barto1983neuronlike} and Cart Pole~\citep{Moore90efficientmemory-based}. For image-input tasks, we choose six Atari games~\citep{bellemare13arcade}.} We divide the six Atari games into two different groups.
% We first conduct experiments on six typical Atari games~\citep{bellemare13arcade} to verify the effectiveness of our \algname \ framework in both offline and online settings. The Atari games are divided into two different groups. 
The first group includes Breakout, Asterix, and SpaceInvaders, which feature nearly static backgrounds. Enduro, FishingDerby, and Tutankham in the second group have time-changing or dramatically different backgrounds, presenting more challenging scenarios. 
We further conduct experiments on autonomous driving environments~\citep{dosovitskiy2017carla} as one potential application.
We select Proximal Policy Optimization~(PPO)~\citep{schulman2017proximal} as our baseline RL algorithm. 
\hongming{For feature-input classical control tasks, we use a policy network with two fully connected layers, each containing 128 units with ReLU activation functions. For image-input tasks, we use the same network architecture as described in the PPO paper~\citep{schulman2017proximal}.}

\noindent \textbf{Three Types of Outliers.} \textbf{(1) Random Outliers.} We generate random outliers by adding Gaussian noise with zero mean and different standard deviations on state observations, simulating natural measurement errors. \textbf{(2) Adversarial Outliers.} We perform white-box adversarial perturbations~\citep{szegedy2013intriguing,goodfellow2014explaining,cao2020adversarial} on state observations for the current policy, following the strategy proposed in~\citep{huang2017adversarial,pattanaik2017robust}. Particularly, we denote $a_w^{t}$ as the "worst" action, with the lowest probability from the current policy $\pi_{t}(a|s)$. The optimal adversarial perturbation $\eta_t$, constrained in an $\epsilon$-ball, can be derived by minimizing the objective function $J$: $\min_{\eta} J(s_t+\eta, \pi_t)= -\sum_{i=1}^{n} p_i^t \log \pi_t(a_i|s_t+\eta), s.t.  \Vert \eta \Vert \le \epsilon$, where $p_w^t=1$ and $p_i^t=0$ for $i \not= w$. We solve this minimization problem with the Fast Gradient Sign Method~(FGSM)~\citep{goodfellow2014explaining}, a typical adversarial attack method in the deep learning literature. The resulting adversarial outliers $s_t + \eta^*_t$ force the policy to choose $a_w^t$. \textbf{(3) Out-of-Distribution~(OOD) outliers.} OOD outliers arise from the disparity in data distribution across different environments. To simulate them, we randomly select states from other environments and introduce them to the current environment. 
In our experiments, we select images from other Atari games to serve as Out-of-Distribution (OOD) outliers within the considered environment. In the autonomous driving scenario, we designate rainy and nighttime observations as OOD outliers for the primary daytime setting on a sunny day. This deliberate selection of diverse outlier examples enables comprehensive testing of our method's robustness across varied environments.

\noindent \textbf{Baseline Methods.} A fundamental obstacle in assessing the anomaly detection strategies in RL lies in the scarcity of suitable baselines in deep RL settings as introduced in~\cref{sec:related_work}. 
% For instance, the detection proposed in \cite{haider2023out} is confined to model-based RL that requires access to transition and reward functions.
To rigorously substantiate the effectiveness of \algname, we initiate our evaluation by comparing them with the foundational baselines we have developed ourselves and implement two non-distance-based methods. (1) \textbf{Euclidean distance}~\textbf{(ED)} assumes that all features are independent under the Gaussian assumption with one standard deviation, which can be considered as a simplified version of our MD method with an identity covariance matrix.  (2) \textbf{MD with Tied covariance}~\textbf{(TMD)} follows the tied covariance assumption in~\citep{lee2018simple}, where features among all action classes share a single covariance matrix estimation. {\ke{(3) \textbf{PEOC} is a policy entropy-based detection method proposed in \cite{sedlmeier2020policy}. The authors assume that a successful training process reduces entropy for states encountered during training, which can then be used as a classification score to detect OOD states. (4) \textbf{EnvModel} follows the model-based detection algorithm utilizing
learned dynamics models and bootstrapped ensembles~\citep{haider2023out}. We train five autoencoders as environment transition models in each environment. For the offline setting, the autoencoders are trained based on the dataset, while for the online setting, they are continuously updated. Each autoencoder predicts the next state given the current state and action, and the minimum prediction error among the five models serves as the anomaly detection signal.}} (5) \textbf{MD} is our first proposed method with class-conditional Gaussian assumption. (6) \textbf{Robust MD~(RMD)} is the robust variant of MD under the Gaussian assumption. (7) \textbf{MD+C} uses well-calibrated conformality scores to construct a valid empirical distance distribution instead of relying on the Chi-Squared distribution established upon the Gaussian assumption. 

\begin{table}[!t]
% \begin{wraptable}{r}{0.55\textwidth}
% \vskip -0.15in
	\centering
	\scalebox{0.9}{
 {\ke{
		\begin{tabular}{cc|cccc|ccc}
			% \toprule[1pt]
           \bottomrule[1pt]
			 \textbf{Detection Accuracy $(\%)$} & Outliers &\textbf{ED} &\textbf{TMD} &{\ke{\textbf{PEOC}}}&{\ke{\textbf{EnvModel}}}&\textbf{MD} &\textbf{RMD} &\textbf{MD+C}   \\
           \hline
             \multirow{3}*{Cartpole} &  Random & 68.0 & 93.9 &50.0 &50.0&95.4 & 78.9 & \bf{96.2}  \\
           ~& Adversarial & 51.1  & 93.2 &50.0&50.0& 94.5 & 78.7 & \textbf{94.8} \\
            ~& OOD & 87.3 & 94.3 &50.0&93.8  & 96.5 & 79.1 & \bf{97.5} \\
            \hline
            \multirow{3}*{MountainCar} &  Random & 89.5 & 86.4 &50.0& 49.9 & 90.6 & 78.1 & \bf{93.6}  \\
           ~& Adversarial &64.0  & 81.0 &50.0&49.7 & 85.4 & 74.3 & \textbf{87.1} \\
            ~& OOD & 90.9 & 86.5 &50.0&48.7  & 90.5 & 77.2& \bf{91.7} \\
            \hline
            \hline
           \multirow{4}*{\textbf{Average}} &  Random & 78.8 & 90.1 &50.0&50.0 & 93.0 & 78.5 & \bf{94.9}  \\
           ~& Adversarial & 57.6  & 87.1 &50.0&49.8& 89.9 & 76.5 & \textbf{90.9} \\
            ~& OOD & 89.1 & 90.4 &50.0&71.3  & 93.5 & 78.2 & \bf{94.6} \\
           ~& \textbf{Average} & 75.1 & 89.2 &50.0&57.0 & 92.1 & 77.7 & \bf{93.5} \\
			\bottomrule[1pt]
           % \toprule[1pt]
		\end{tabular}
  }}
	}
	\caption{\textbf{Average detection accuracy} of MD, RMD, and MD+C compared with baselines across different outlier types in {\ke{two feature-input classical control}} environments in the \textbf{offline} setting. The averages are computed across environments and outlier types.  Accuracy is determined by applying detection techniques to the balanced data composed equally of clean and noisy states.}
	\label{table_offline_mainresults_conformal_classical}
 \vskip -0.1in
% \end{wraptable}
\end{table}

\begin{table}[b!]
% \begin{wraptable}{r}{0.55\textwidth}
% \vskip -0.15in
	\centering
	\scalebox{0.9}{
		\begin{tabular}{cc|cccc|ccc}
			% \toprule[1pt]
           \bottomrule[1pt]
			 \textbf{Detection Accuracy $(\%)$} & Outliers &\textbf{ED} &\textbf{TMD} &{\ke{\textbf{PEOC}}}&{\ke{\textbf{EnvModel}}}&\textbf{MD} &\textbf{RMD} &\textbf{MD+C}   \\
           \hline
            \multirow{3}*{Breakout} &  Random & 53.2 & 59.1 & 50.0& 50.0 & 61.4 & \bf{71.2} & 62.8  \\
           ~& Adversarial & 84.3  & 89.1 & 50.0&50.0 & 90.8 & 80.2 & \bf{91.7} \\
            ~& OOD & 56.9 & 47.8  &50.0 &\textbf{97.5}&49.6 & \bf{79.5} & 50.8 \\
            \hline
             \multirow{3}*{Asterix} &  Random & 43.9 & 45.1 &50.0&50.0 & 60.3 & \bf{69.5} & 54.7  \\
           ~& Adversarial & 83.7  & 85.6 & 50.0&50.0& 91.7 & 75.2 & \bf{94.0} \\
            ~& OOD & 39.6 & 40.8  &50.0&53.8& 46.1 & \bf{57.7} & 49.7 \\
            \hline
             \multirow{3}*{SpaceInvader} &  Random & 51.4 & 63.9 &50.0&50.0 & 70.2 & \bf{79.4} & 68.6  \\
           ~& Adversarial & 70.9  & 90.3 &50.0&50.0& 96.1 & 81.1 & \bf{96.5} \\
            ~& OOD & 45.3 & 45.9 &50.0&48.8  & 57.0 & \bf{81.0} & 53.6 \\
            \hline
             \multirow{3}*{Enduro} &  Random & 49.0 & 59.0 &50.0&50.0 & 72.7 & \bf{82.4} & 70.2  \\
           ~& Adversarial & 92.9  & 91.3  &50.0&50.0& 96.2 & 83.3 & \bf{97.5} \\
            ~& OOD & 57.1 & 74.9 &50.0&47.6  & 80.0 & \bf{83.4} & 63.3 \\
            \hline
             \multirow{3}*{FishingDerby} &  Random & 48.9 & 66.4 &50.0 &50.0&69.8 & \bf{85.7} & 66.5  \\
           ~& Adversarial & 86.3  & 92.5 &50.0&50.0& \bf{97.4} & 87.2 & \textbf{97.4} \\
            ~& OOD & 51.3 & 56.2 &50.0&61.7  & 59.1 & \bf{81.5} & 58.5 \\
            \hline
            \multirow{3}*{Tutankham} &  Random & 50.0 & 47.5 &50.0& 50.0 & 49.1 & \bf{74.0} & 49.8  \\
           ~& Adversarial &66.2  & 89.5 &50.0&50.0 & 95.3 & 77.1 & \textbf{96.5} \\
            ~& OOD & 55.0 & 83.3 &50.0&\textbf{97.5}  & 89.6 & 77.2& 79.7 \\
            \hline
            \hline
           \multirow{4}*{\textbf{Average}} &  Random & 49.4 & 56.8 &50.0&50.0 & 63.9 & \bf{77.0} & 61.7  \\
           ~& Adversarial & 80.7  & 89.7 &50.0&50.0& 94.6 & 80.7 & \textbf{95.6} \\
            ~& OOD & 50.8 & 58.1 &50.0&67.8  & 63.6 & \bf{76.7} & 59.3 \\
           ~& \textbf{Average} & 60.3 & 68.2 &50.0&55.7 & 74.0 & \textbf{78.1} & 72.2 \\
			\bottomrule[1pt]
           % \toprule[1pt]
		\end{tabular}
	}
	\caption{\textbf{Average detection accuracy} of MD, RMD, and MD+C compared with baselines across different outlier types in six Atari games in the \textbf{offline} setting. The averages are computed across environments and outlier types.  Accuracy is determined by applying detection techniques to the balanced data composed equally of clean and noisy states.}
	\label{table_offline_mainresults_conformal}
 \vskip -0.1in
% \end{wraptable}
\end{table}

\subsection{Anomaly Detection in the Offline Setting}
\label{sec:eva}

In the offline setting, we randomly split the states from the given dataset into calibration and evaluation sets, each containing 50\% of the data.
The calibration set is used to construct our detectors, and the evaluation set is for testing. We first use PCA to reduce the state feature vectors into a 50-dimensional space.  We then apply (robust) MD to estimate mean vectors and covariances and calibrate the conformality score based on the calibration dataset. Finally, we incorporate the three types of noises into the originally clean evaluation dataset. We assess the performance of our detection methods on the entire evaluation dataset. 
% Given a state $s$, we compute the detection metric $M(s)$ and $M_{\text{rob}}(s)$, and utilize test statistic $\chi^2_{p}(1-\alpha)$ or $Q^c_{1-\alpha}$ with significance level $\alpha$ for the detection. 

\noindent \textbf{Main results.} \ke{\cref{table_offline_mainresults_conformal_classical,table_offline_mainresults_conformal} show the detection accuracy of \algname \ instantiated with vanilla MD, robust MD, and conformal MD with $\alpha=0.05$ across a wide range of outlier types on each task. A higher accuracy indicates a more successful identification of anomalies for the evaluated detection method. We conclude that: (1) All MD-based methods, i.e., TMD, MD, RMD, and MD+C, outperform ED, confirming the usefulness of covariance matrix information in RL outlier detection. (2) MD+C performs consistently best on classic control tasks and excels in identifying adversarial outliers on Atari games. Robust MD generally performs the best on Atari games, significantly surpassing MD and other methods in detecting random and OOD outliers. Nonetheless, robust MD is not effective enough to detect adversarial outliers. We hypothesize that the robustness advantage resulting from RMD in detector estimation is more applicable in image input or the high-dimensional state space. 
% This helps to explain why RMD is less preferable when employed in feature-input tasks with low-dimensional states. 
(3) PEOC is almost ineffective across all considered state outliers, suggesting that the entropy difference is useless for identifying outliers. By contrast, EnvModel is only superior to the other approaches against some OOD outliers, which is not generally preferable. More detailed results are provided in \cref{table_offline_mainresults_conformal_details_classical,table_offline_mainresults_conformal_details} of \cref{appendix:evaluation_moreresults}.}

\noindent \textbf{Sensitivity Analysis on Feature Dimension Reduction.} We provide a sensitivity analysis on Atari games regarding the number of feature dimensions reduced by PCA, showing that the detection accuracy for all considered outliers tends to improve as the number of principal components increases. This indicates that better detection performance can be achieved with higher feature dimensions. The detailed results are presented in~\cref{appendix:sensitivity}.

\noindent \textbf{Effectiveness of Robust MD.}
In robust MD analysis, it is typically concluded that outlier states are more distinctly separated from inlier states. By comparing the Mahalanobis distance distributions between inliers and outliers under both MD and Robust MD, we show that this conclusion also applies to the RL anomaly detection scenario. This effect explains the detection advantage of robust MD in RL. Detailed results are provided in~\cref{appendix:boxplot}.
% \vskip -0.05in

\subsection{Anomaly Detection in the Online Setting}

\begin{figure*}[b!]
% \vskip -0.1in	
 \centering
	\subfigure{\includegraphics[width=0.2\textwidth]{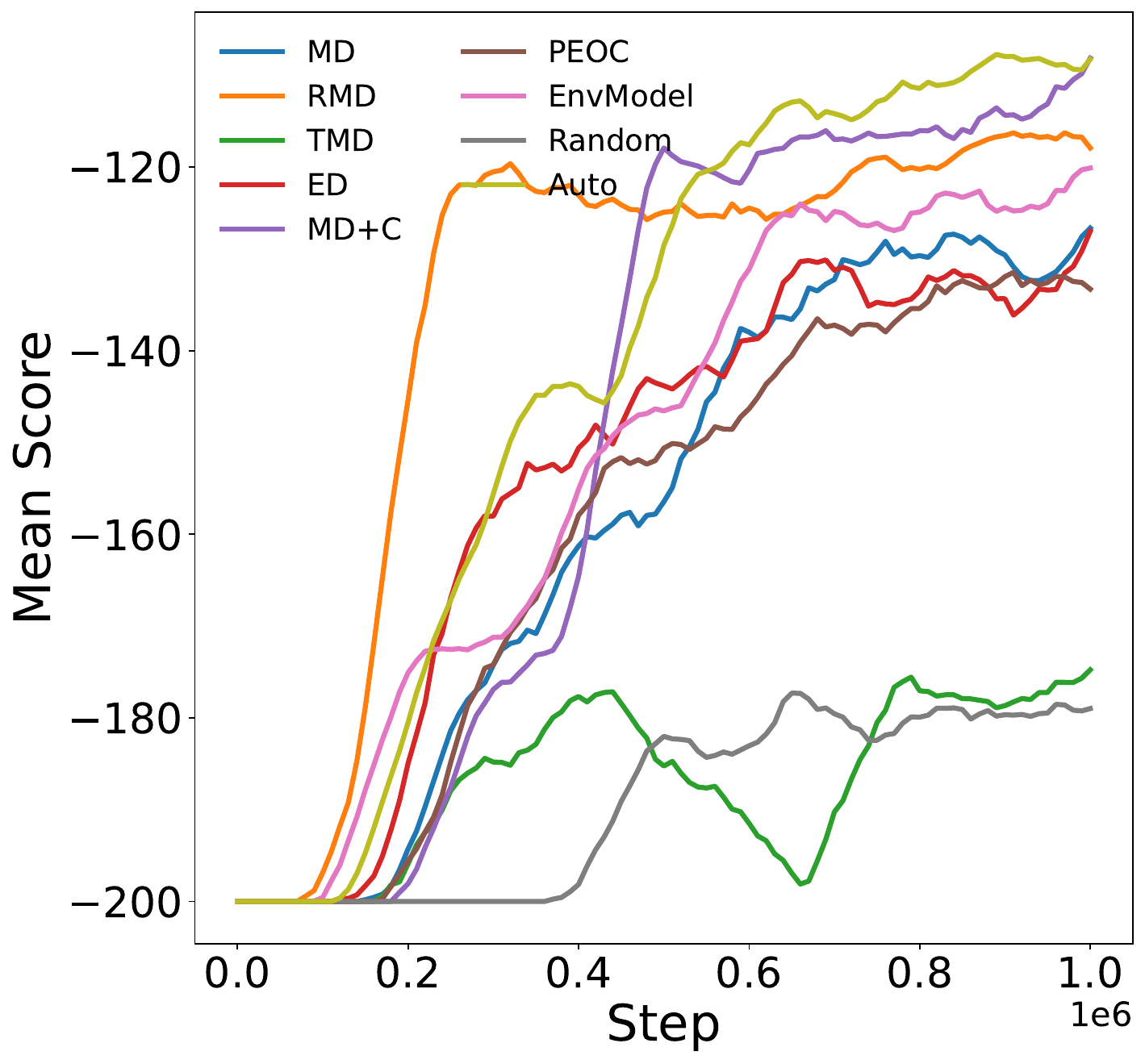}}
	\subfigure{\includegraphics[width=0.2\textwidth]{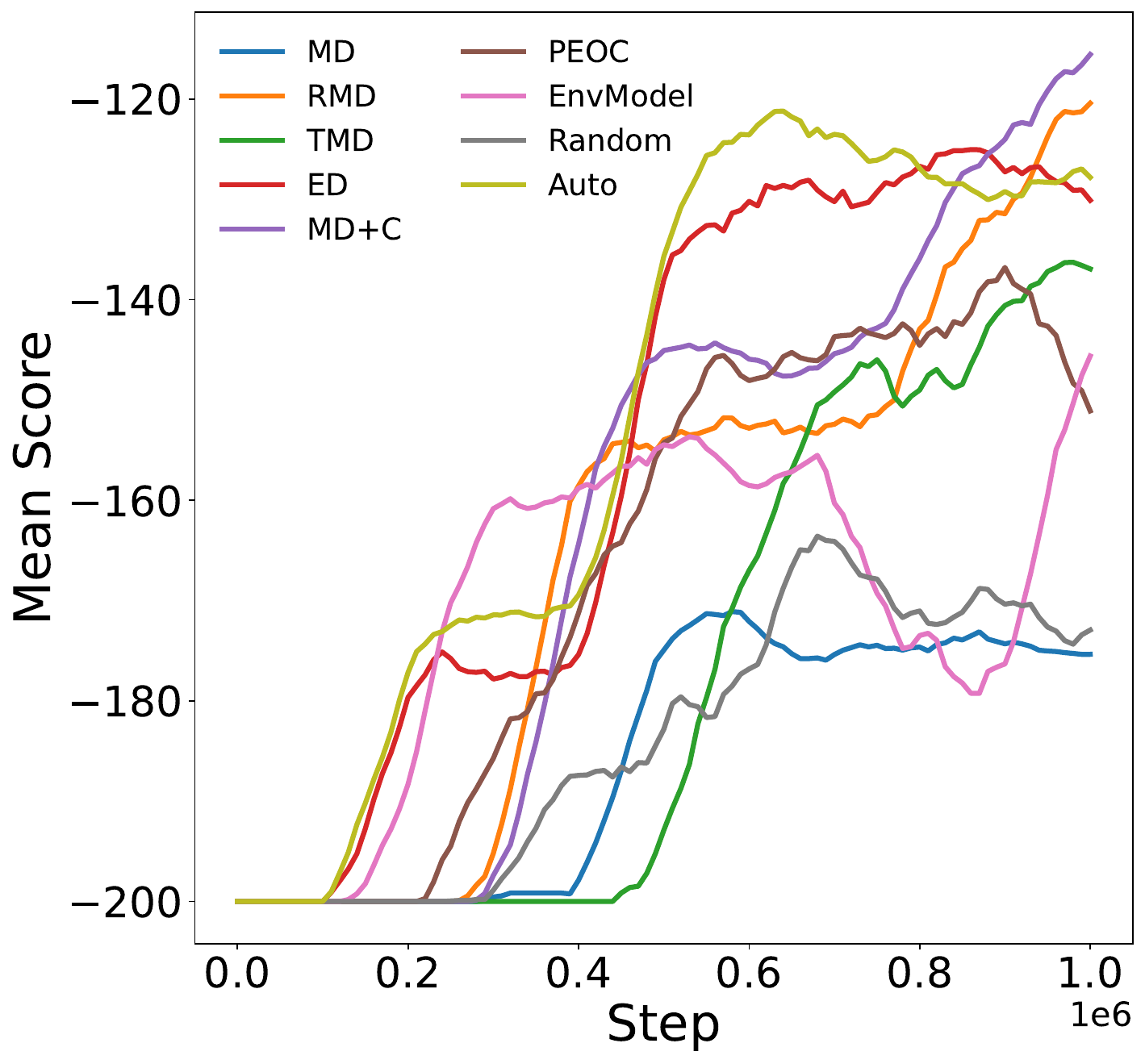}}
	\subfigure{\includegraphics[width=0.2\textwidth]{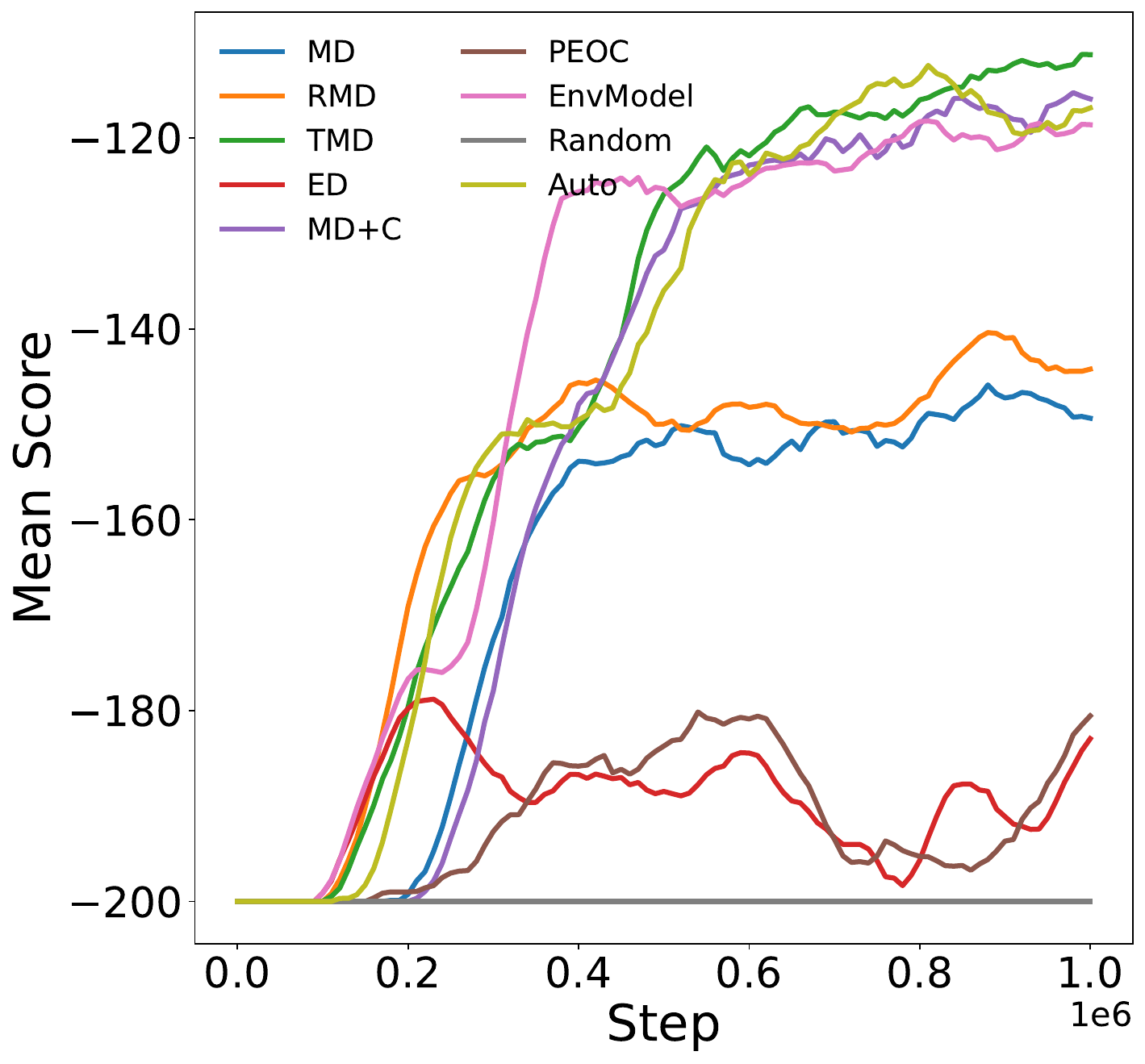}}
	\subfigure{\includegraphics[width=0.2\textwidth]{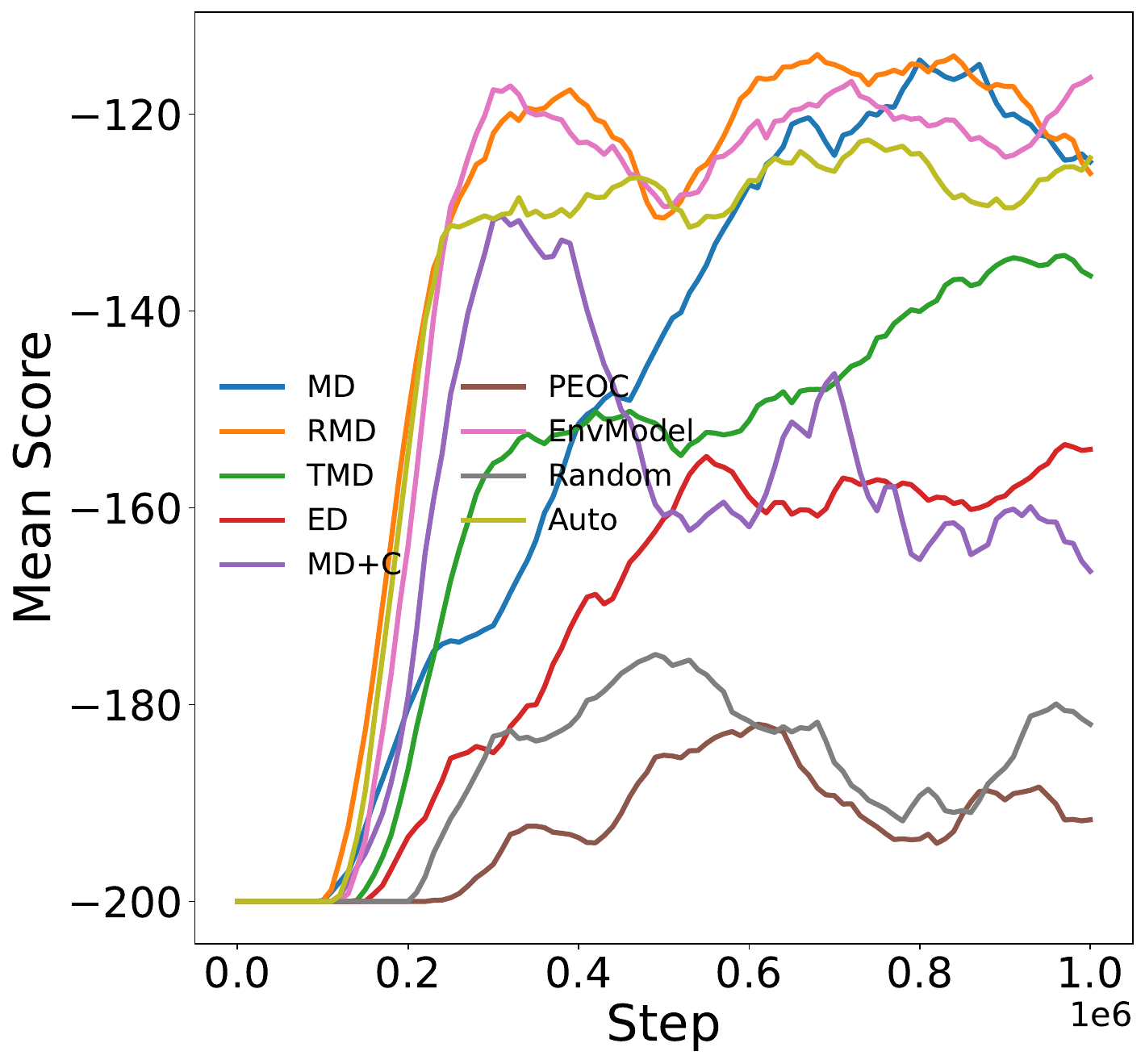}}
 \vskip -0.1in
 %    \setcounter{subfigure}{0}
	% \subfigure{\includegraphics[width=0.2\textwidth]{pic/woci_MountainCar_5_f1.pdf}}
	% \subfigure{\includegraphics[width=0.2\textwidth]{pic/woci_MountainCar_6_f1.pdf}}
	% \subfigure{\includegraphics[width=0.2\textwidth]{pic/woci_MountainCar_7_f1.pdf}}
	% \subfigure{\includegraphics[width=0.2\textwidth]{pic/woci_MountainCar_8_f1.pdf}}
 %     \vskip -0.1in
     \setcounter{subfigure}{0}
	\subfigure[Gaussian std=1]{\includegraphics[width=0.2\textwidth]{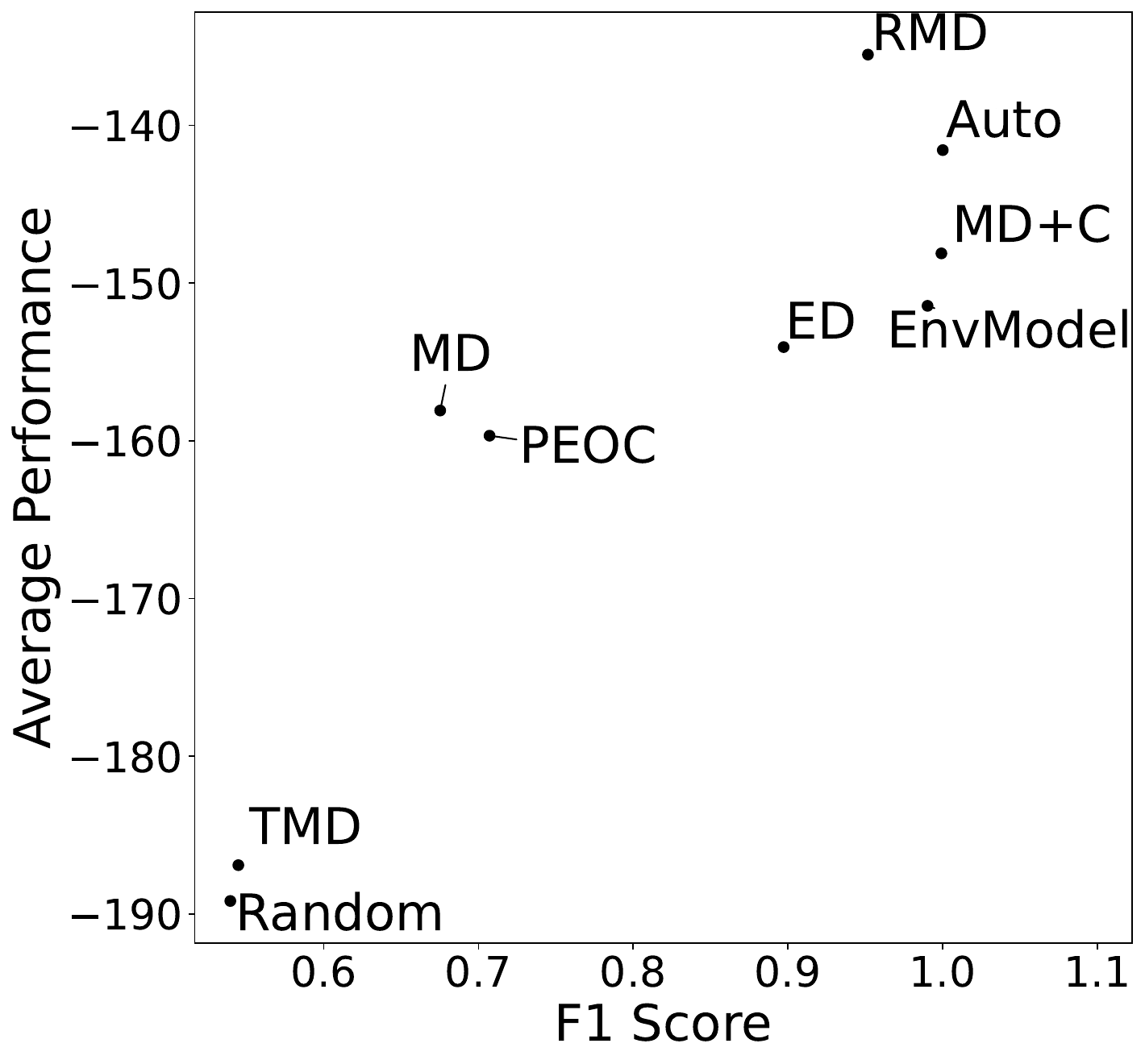}}
	\subfigure[Gaussian std=0.3]{\includegraphics[width=0.2\textwidth]{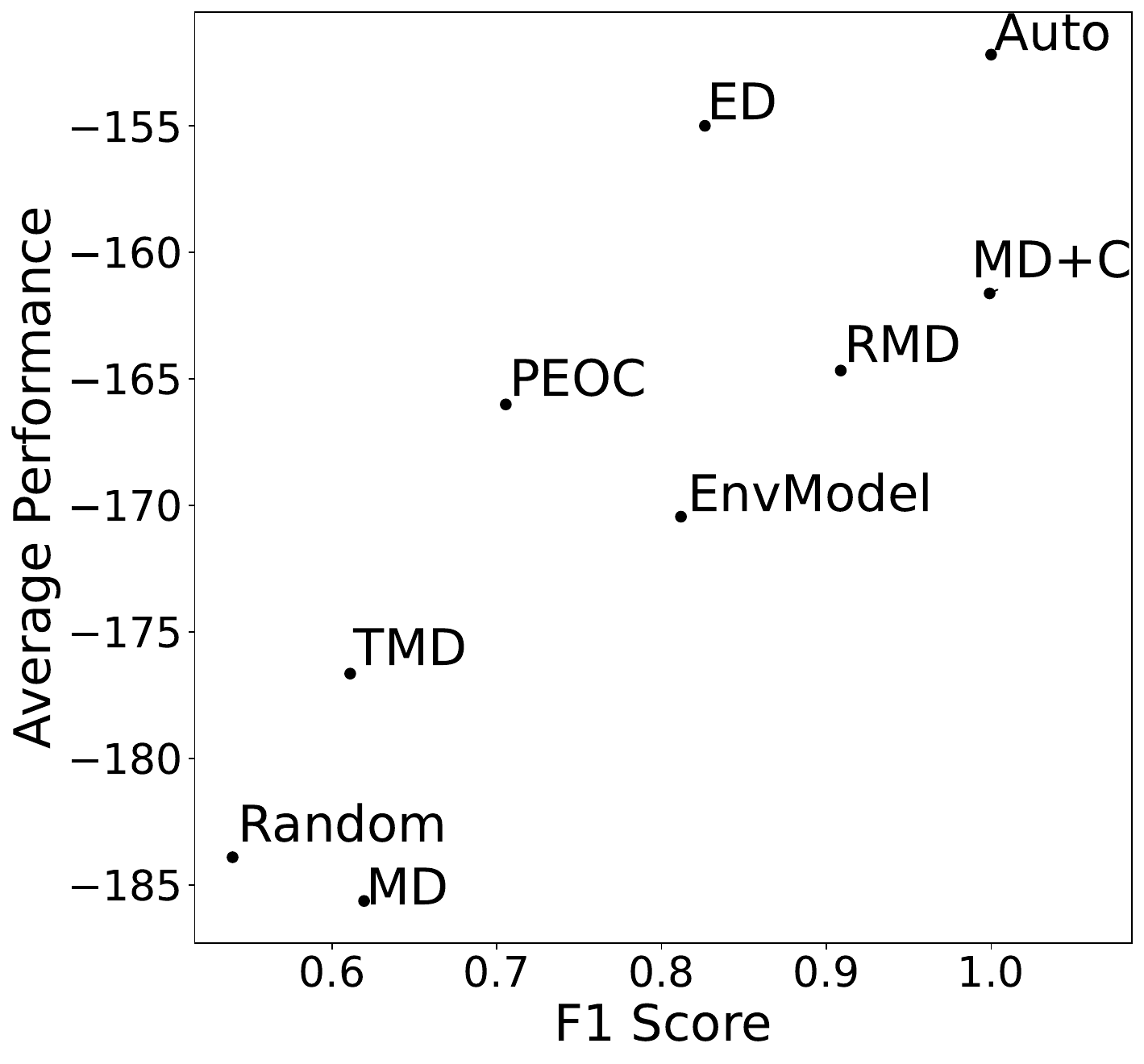}}
	\subfigure[OOD CartPole]{\includegraphics[width=0.2\textwidth]{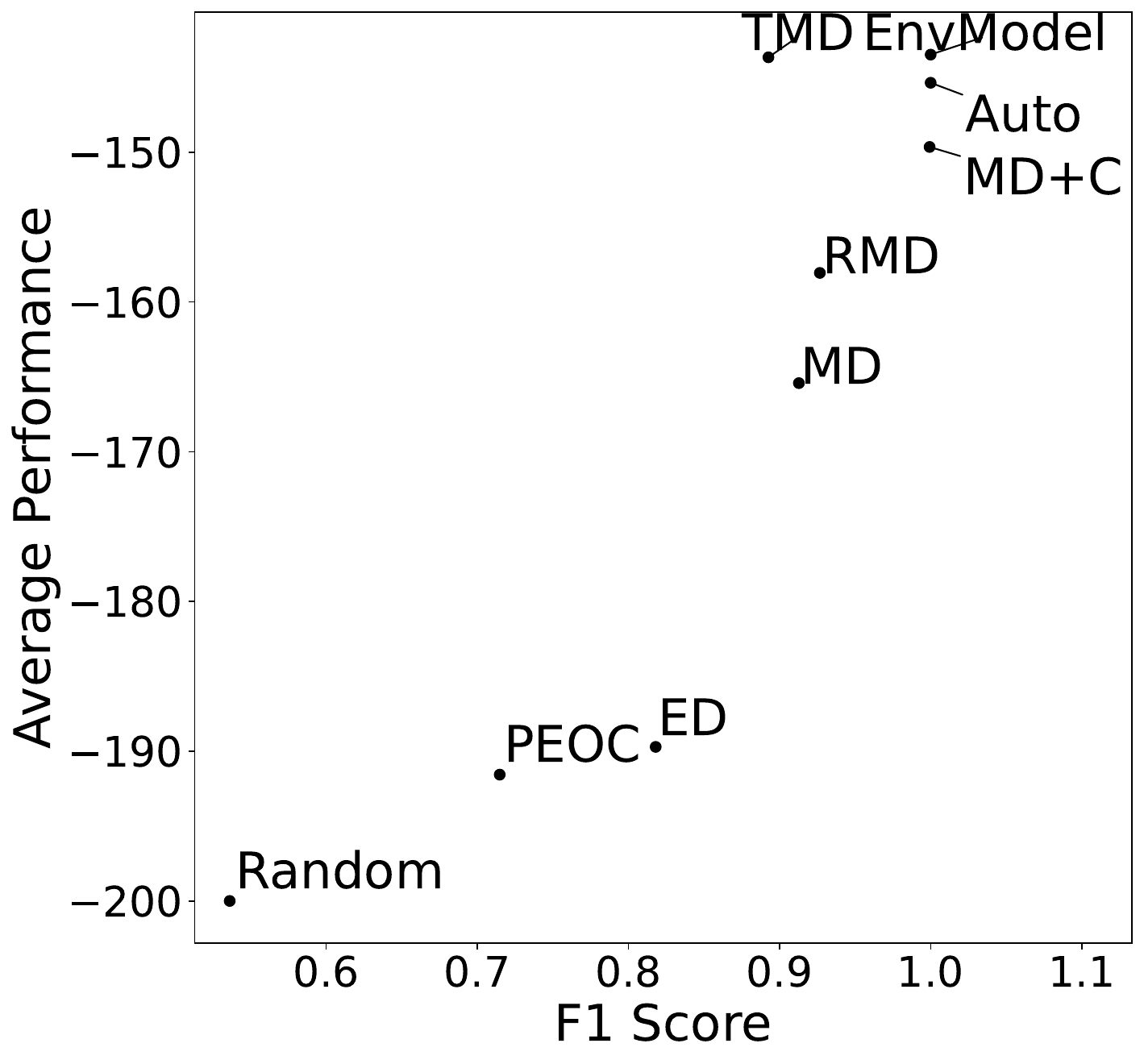}}
	\subfigure[Adversarial]{\includegraphics[width=0.2\textwidth]{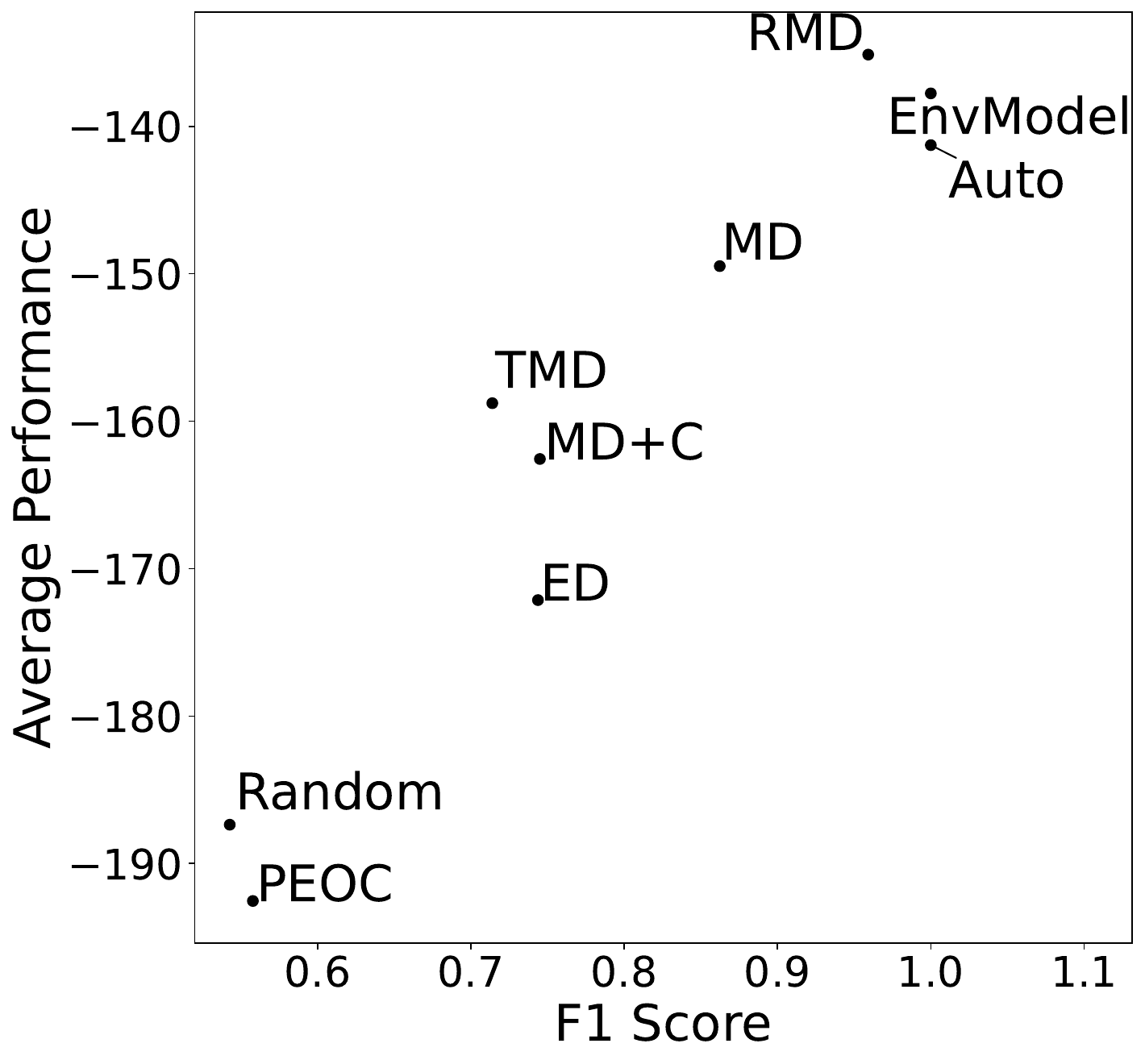}}
\vskip -0.1in	
 \caption{\hongming{Performance on MountainCar across various state outliers in online learning. The first row shows the policy performance during learning. The second row shows the relationship between the averaged detection accuracy and performance.}}
 % \caption{\hongming{\textbf{Learning curves and detection performance} across various state outliers in online learning on Tutankham. "Mean Score" in the first row indicates the cumulative rewards, and "F1 Score" in the second row evaluates the detection performance during training. The third row shows the relationship between the detection accuracy during learning and achieved final performance.}}
	\label{fig:MountainCar_online_full_main_page}
\end{figure*}

\begin{figure*}[t!]
	\centering
	\subfigure{\includegraphics[width=0.19\textwidth]{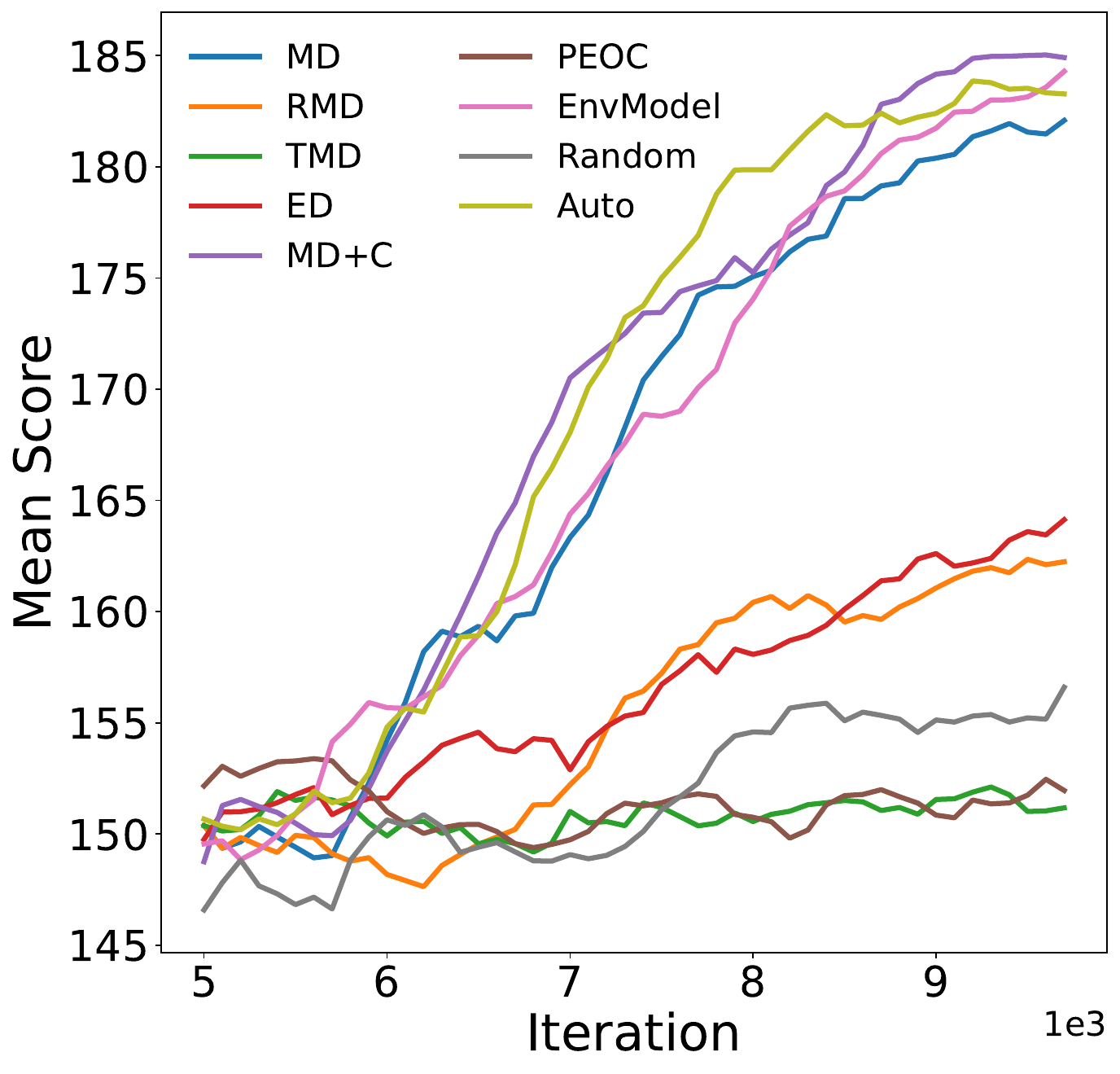}}
	\subfigure{\includegraphics[width=0.19\textwidth]{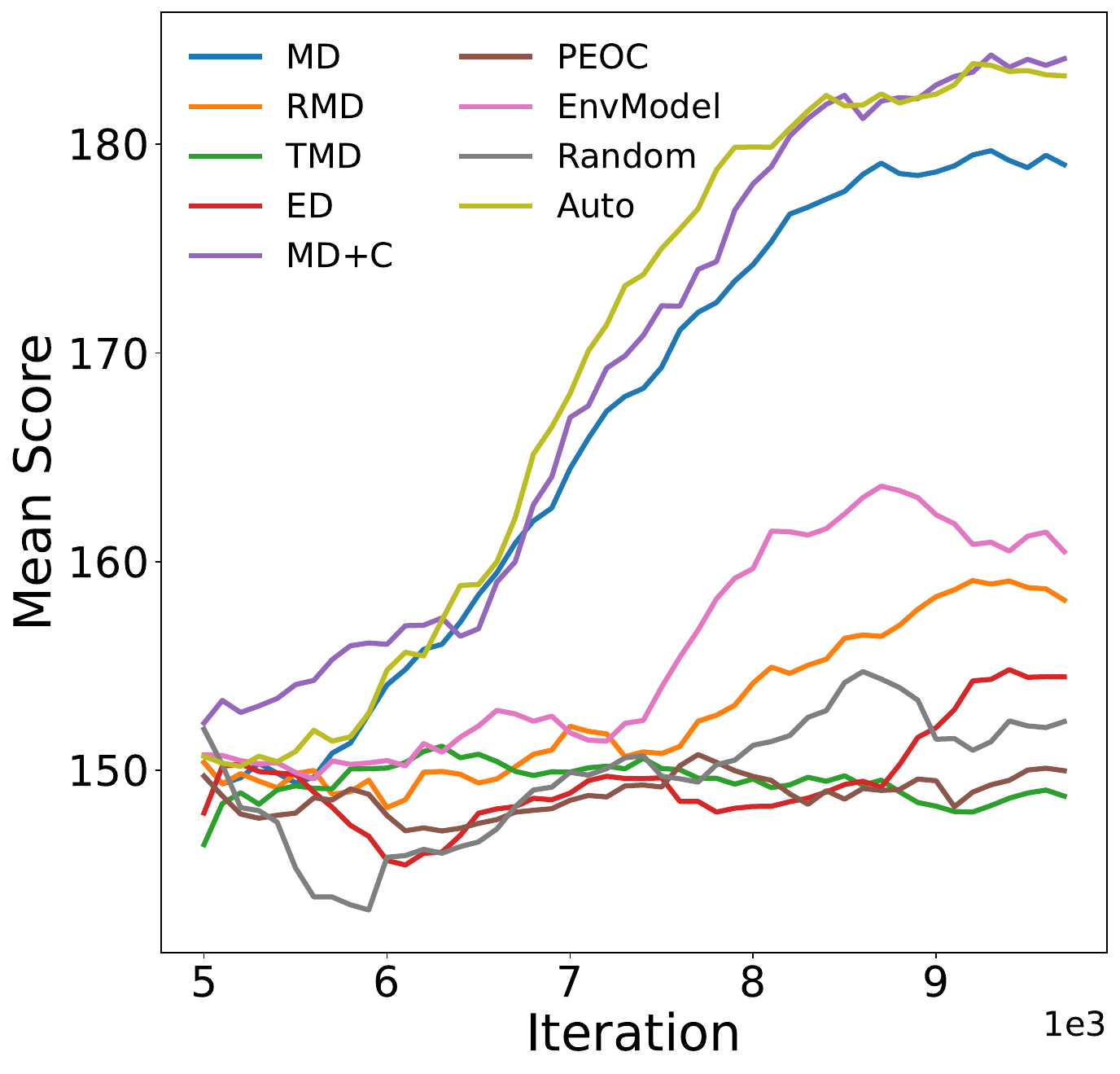}}
	\subfigure{\includegraphics[width=0.19\textwidth]{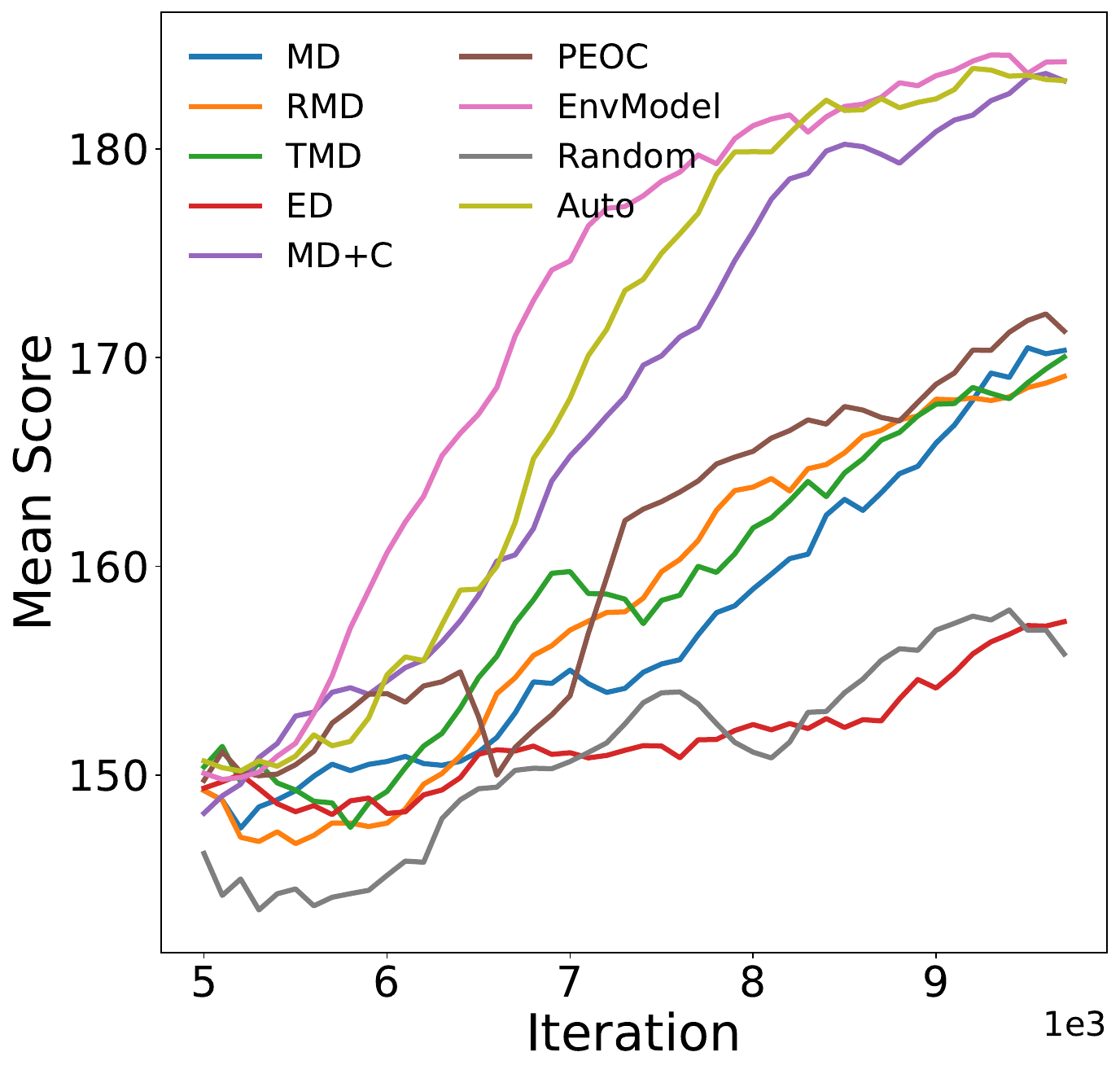}}
	\subfigure{\includegraphics[width=0.19\textwidth]{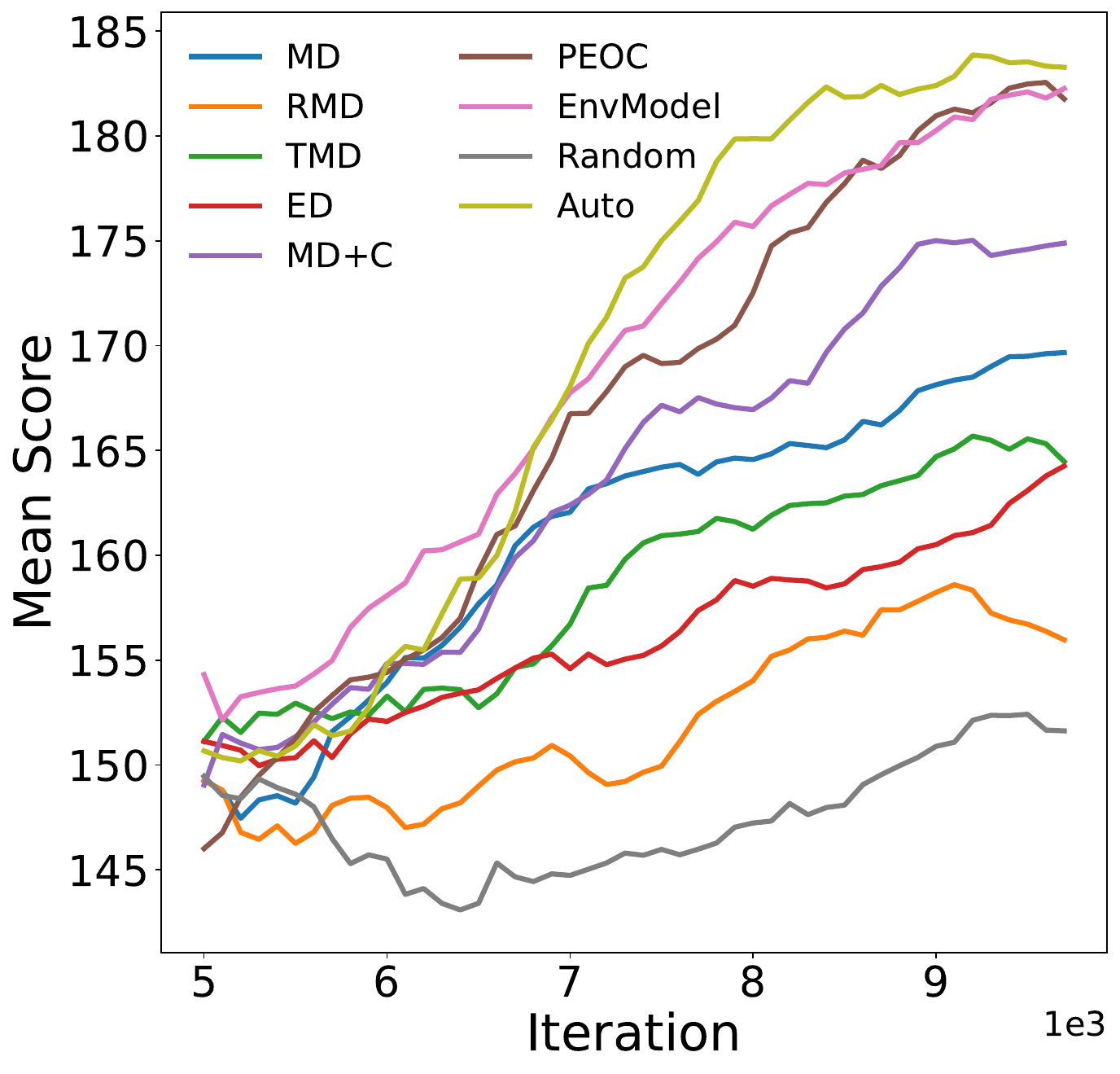}}
	\subfigure{\includegraphics[width=0.19\textwidth]{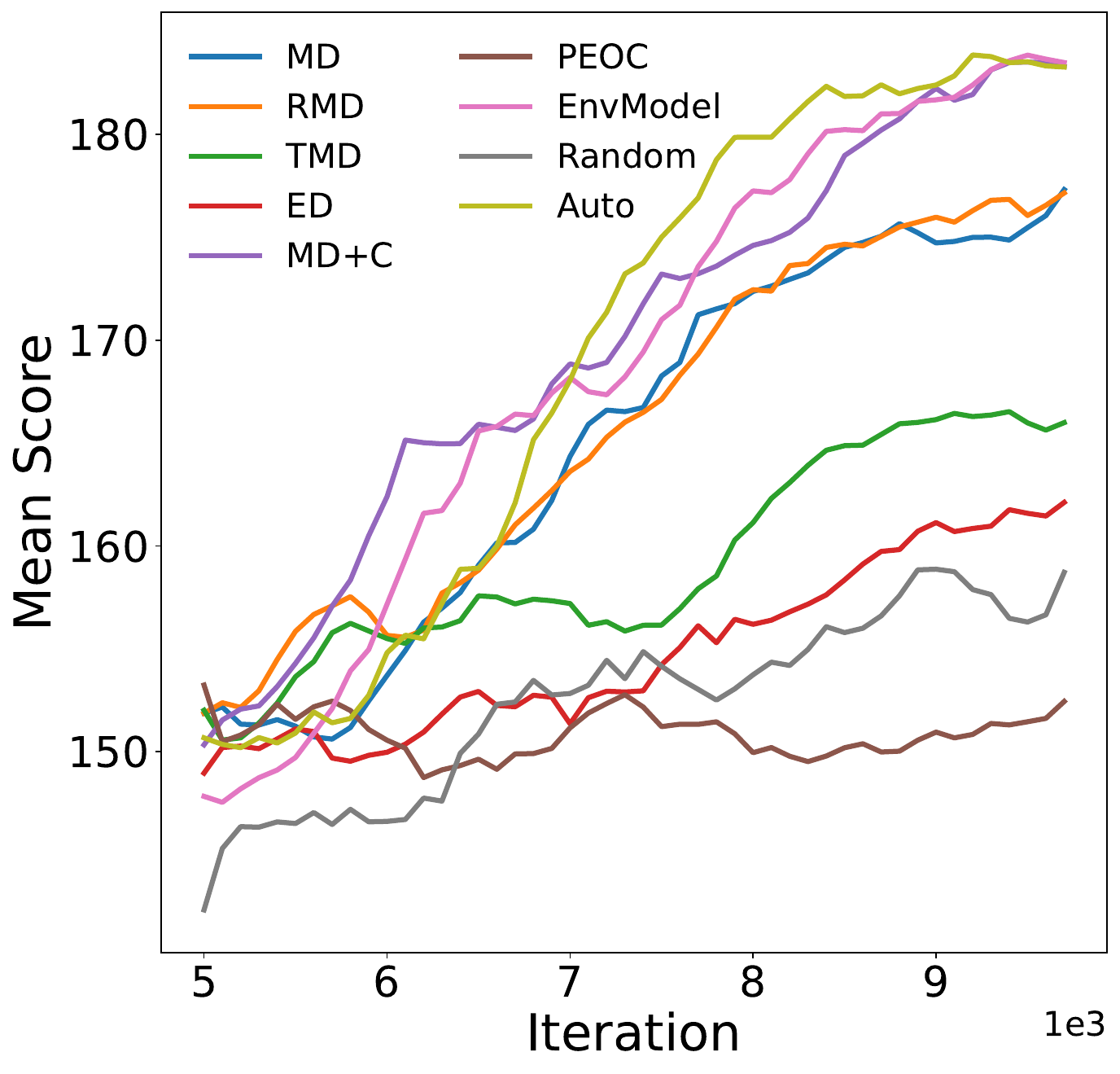}}
 \vskip -0.1in
 %    \setcounter{subfigure}{0}
	% \subfigure{\includegraphics[width=0.19\textwidth]{pic/woci_Tutankham_21_f1.pdf}}
	% \subfigure{\includegraphics[width=0.19\textwidth]{pic/woci_Tutankham_22_f1.pdf}}
	% \subfigure{\includegraphics[width=0.19\textwidth]{pic/woci_Tutankham_23_f1.pdf}}
	% \subfigure{\includegraphics[width=0.19\textwidth]{pic/woci_Tutankham_24_f1.pdf}}
	% \subfigure{\includegraphics[width=0.19\textwidth]{pic/woci_Tutankham_25_f1.pdf}}
	% \vskip -0.1in
     \setcounter{subfigure}{0}
 	\subfigure[Gaussian (std=1)]{\includegraphics[width=0.19\textwidth]{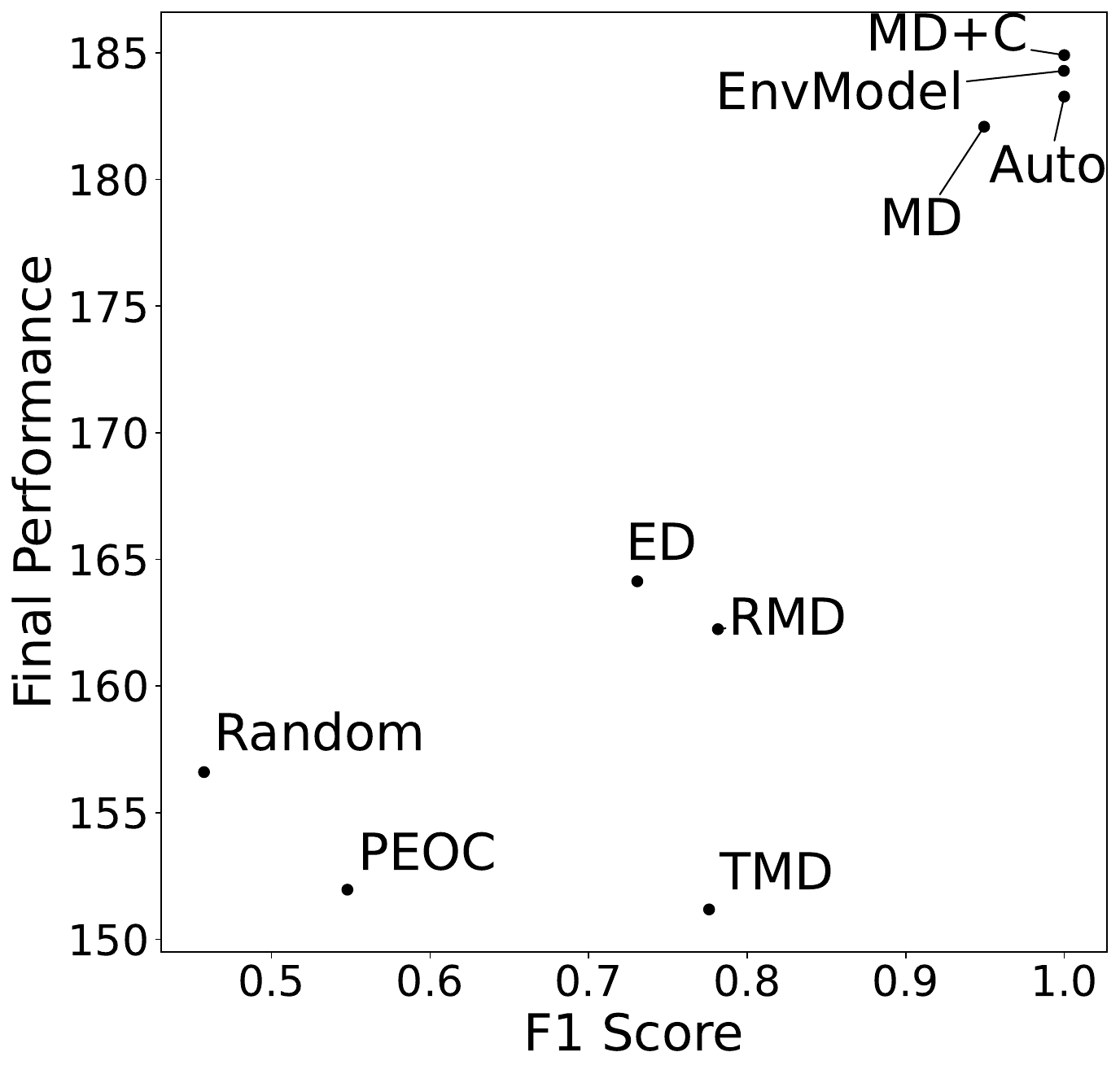}}
	\subfigure[Gaussian (std=0.3)]{\includegraphics[width=0.19\textwidth]{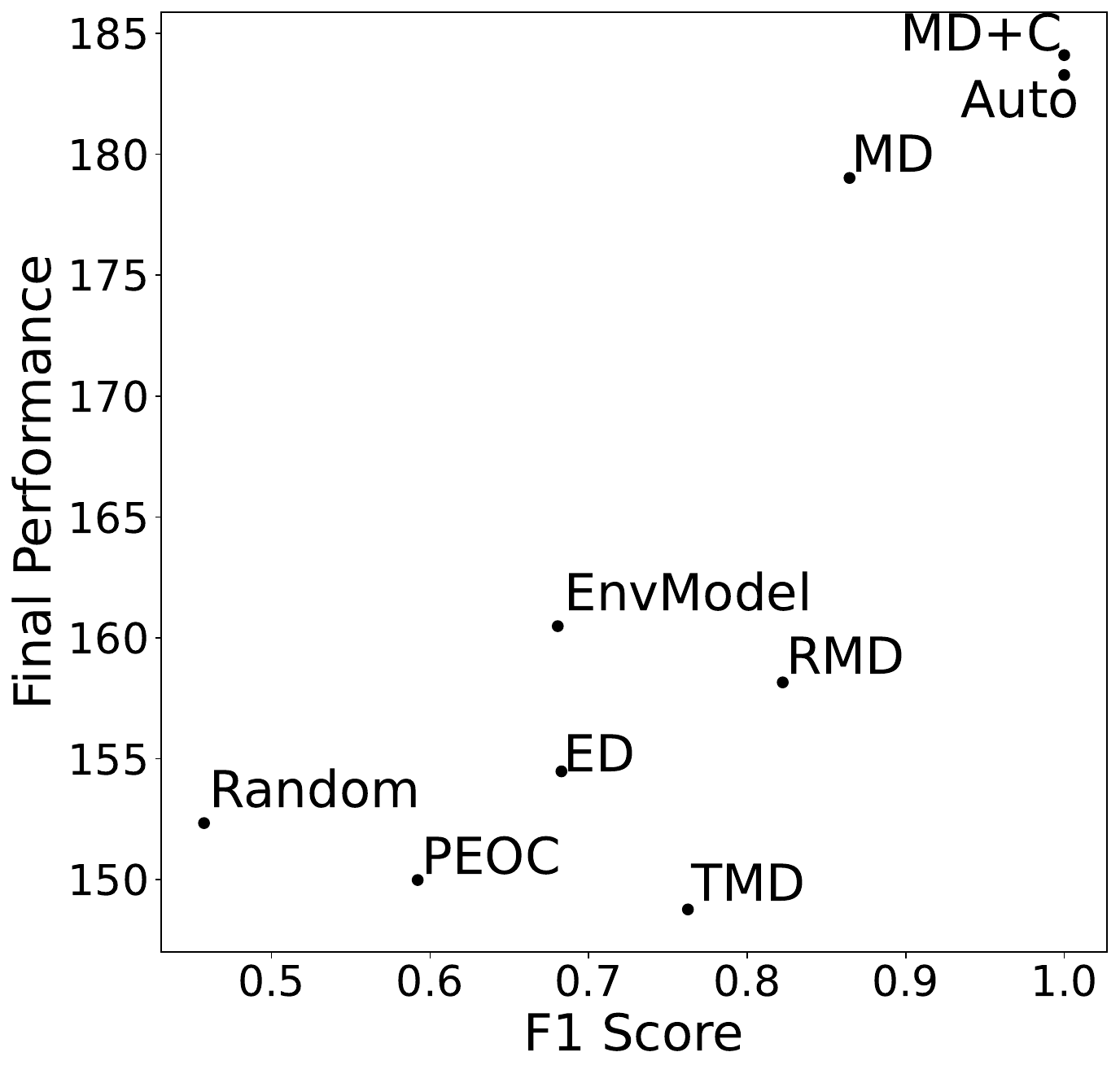}}
	\subfigure[OOD Enduro]{\includegraphics[width=0.19\textwidth]{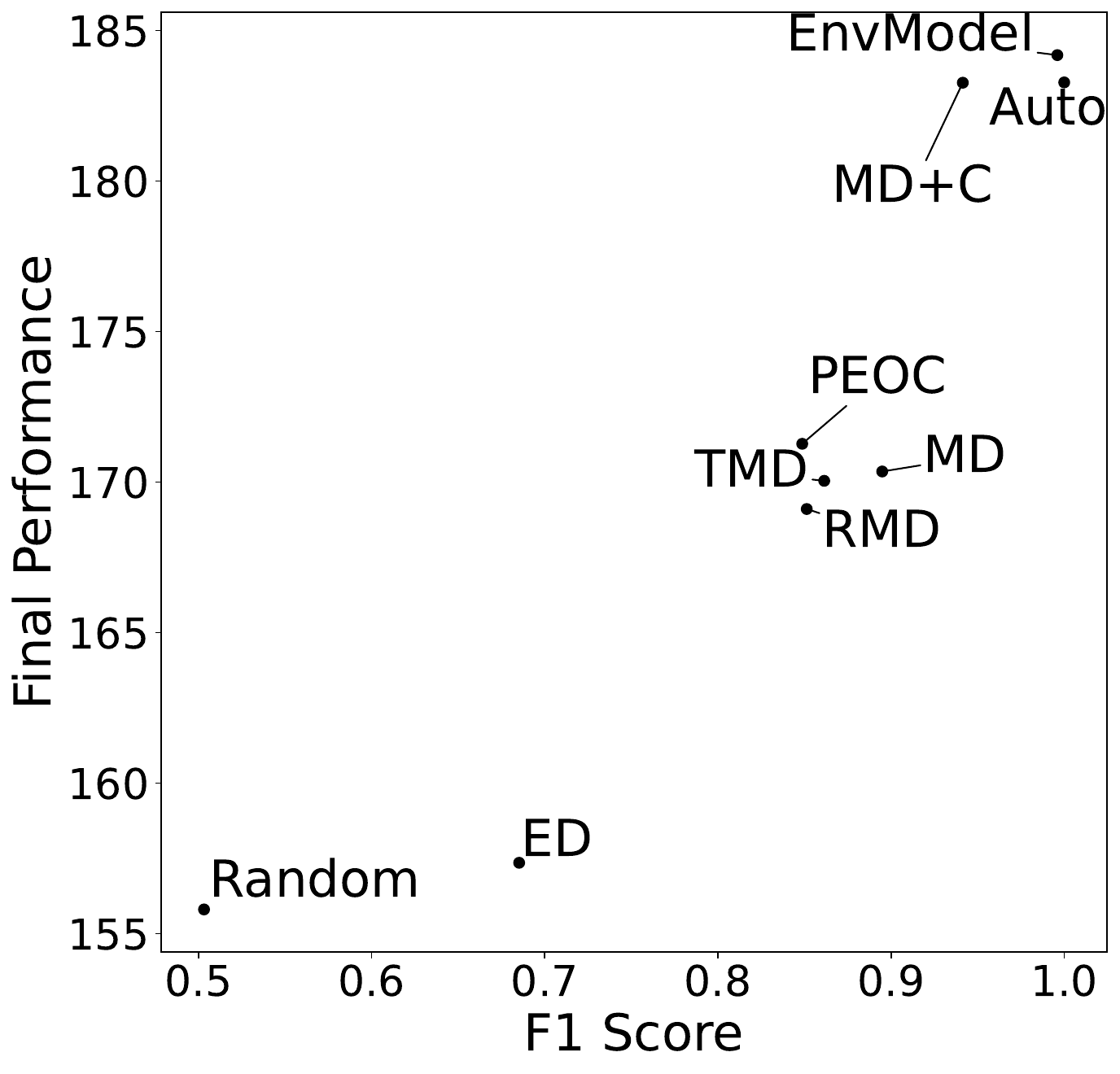}}
	\subfigure[OOD FishingDerby]{\includegraphics[width=0.19\textwidth]{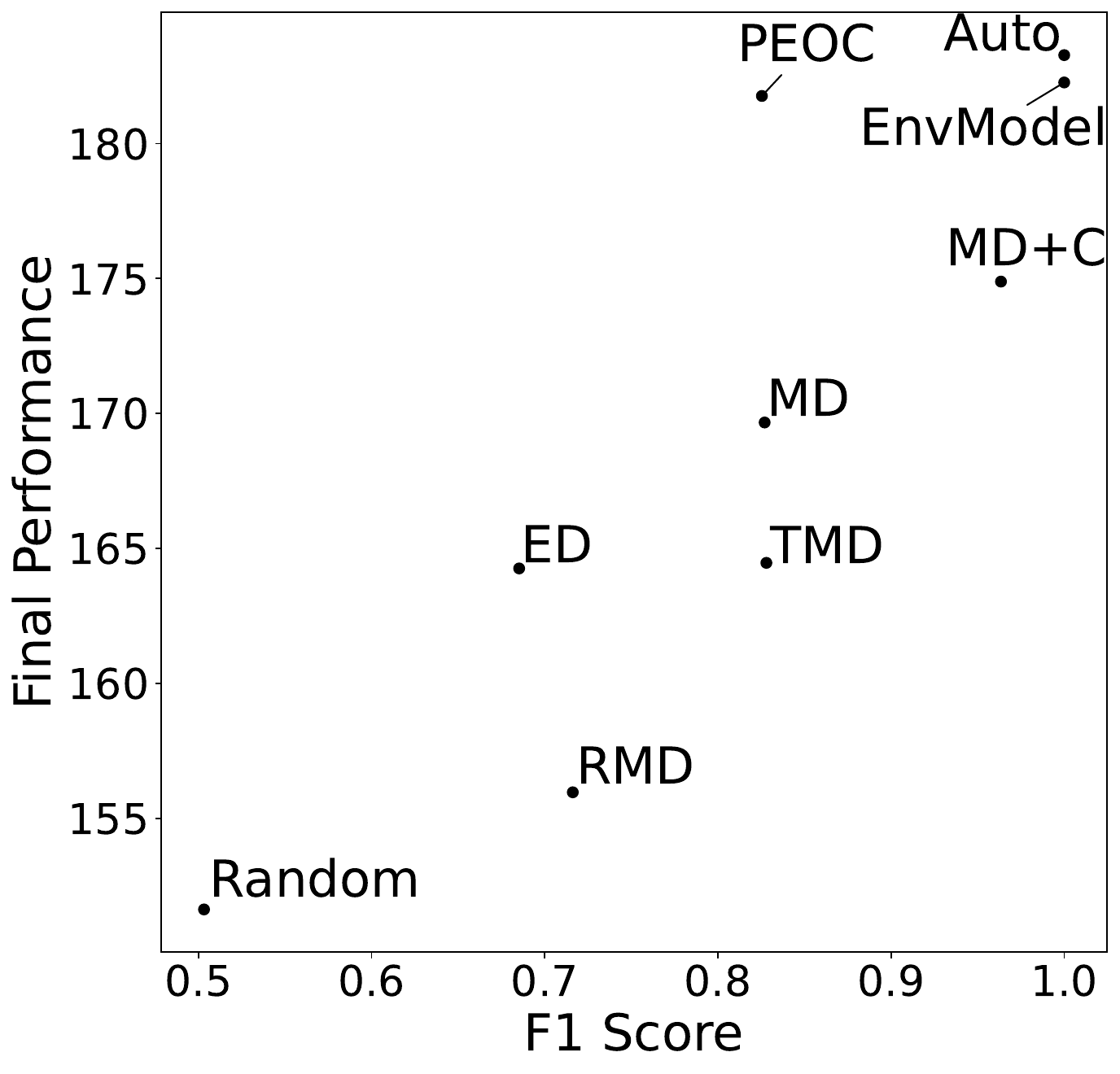}}
	\subfigure[Adversarial]{\includegraphics[width=0.19\textwidth]{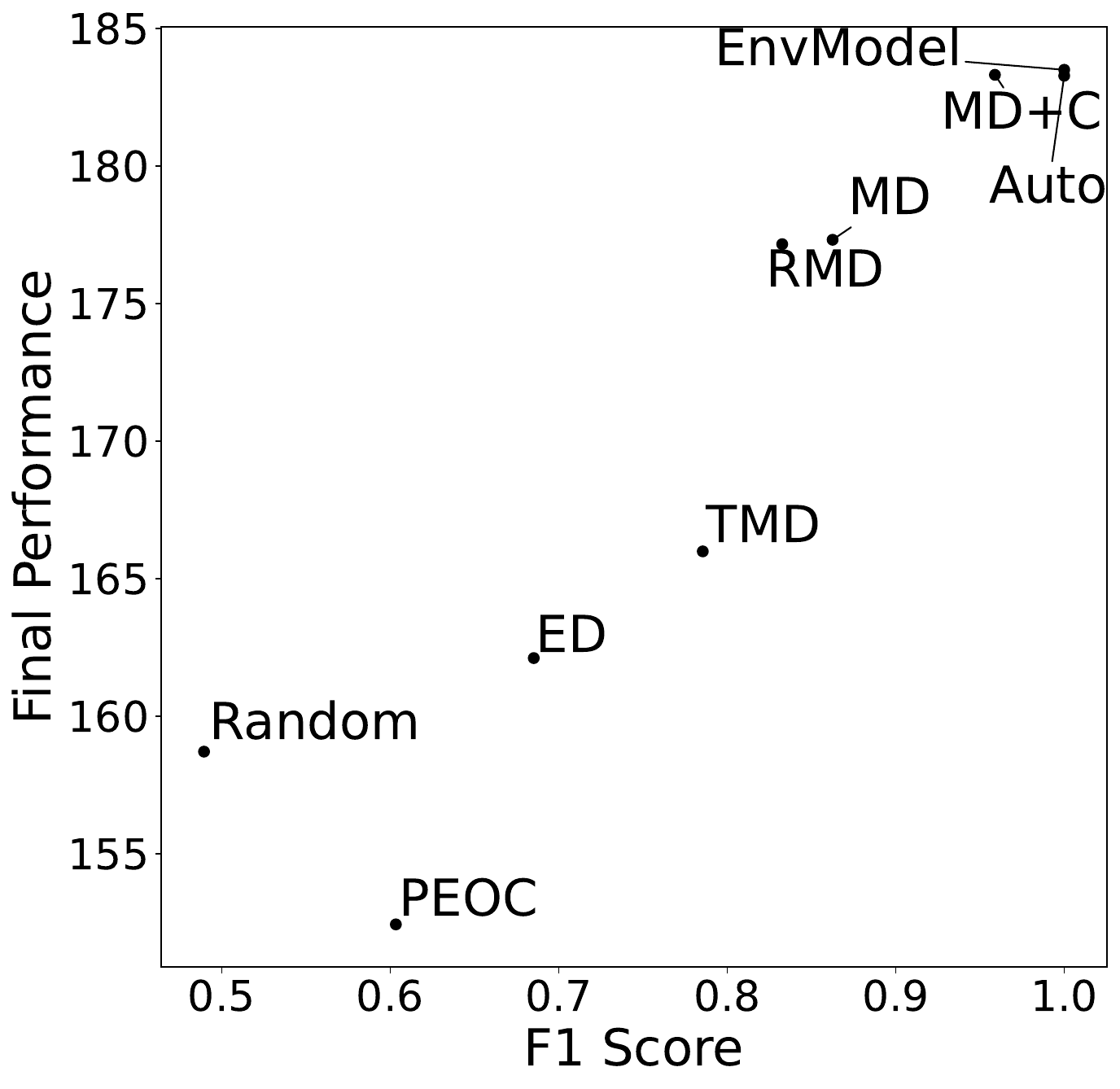}}
\vskip -0.1in	
 \caption{\hongming{Performance on Tutankham across various state outliers in online learning. The first row shows the policy performance during learning. The second row shows the relationship between the averaged detection accuracy and achieved final performance.}}
 % We present the average results over three random seeds while omitting confidence bands for a clearer comparison.}
	\label{fig:Tutankham_online_full_main_page}
\end{figure*}
% Owing to the parallel training of PPO and the estimation of GAE~\citep{schulman2015high}, we consider discriminating outliers according to a trajectory.
% rather than a single state. 
% We regard a trajectory as an outlier if half of its states are detected as outliers. 

The PPO agent, utilizing multi-processes as detailed in the original PPO algorithm~\citep{schulman2017proximal}, runs eight independent environments in parallel, and we introduce state outliers into four of these environments.
% We run eight environments in parallel as in the original PPO paper, and introduce state outliers into four environments.
For random and adversarial outliers, actions are determined based on the PPO policy network $\pi_{\theta}$. For OOD outliers, due to the potential differences in action spaces between the original environment and the OOD environment, we select OOD states from the OOD environment by taking random actions within its own action space. For the Robust MD method, we use PCA to reduce state feature vectors into a 50-dimensional space due to the expensive computation of the robust MD method. For the other methods, we use the original feature vectors output from the penultimate layer of $\pi_{\theta}$. 
Results are averaged over three seeds with hyperparameters given in~\cref{appendix:hyperparameter} of~\cref{appendix:Training Phase Main Results}. When our detectors identify an outlier, it is removed from training. We compare the resulting learning curves for different detection methods.

\noindent \textbf{Additional Baselines.} We add another two baselines as performance upper bound and lower bound. (1) For an ideal baseline, the method \textbf{Auto} automatically deletes true state outliers, showing the optimal training performance of algorithms \textit{without the interruption from outliers}. (2) At the other extreme, \textbf{Random} uses a totally random detector that detects a state as an inlier or outlier with a probability of 0.5. 
% Other baselines \textbf{E1}, \textbf{E2} and \textbf{TMD} are the same as introduced in evaluation phase in Section~\ref{sec:eva}.

\begin{table}[b!] % &~&~&~&~
	\centering
	\scalebox{0.85}{
		\begin{tabular}{cc|cccccccc}
			% \toprule[1pt]
            \bottomrule[1pt]
			\textbf{Superiority Rank}&\textbf{Outlier Type}&\textbf{Random}&\textbf{ED} &\textbf{TMD} &\hongming{\textbf{PECO}} &\hongming{\textbf{EnvModel}} &\textbf{MD} &\textbf{RMD} &\textbf{MD+C}   \\
            \hline
            \multirow{3}*{Performance} & Random & 6.75 & 3.75 & 7.25 & 6.00 & 3.25 & 4.50 & 3.25 & 1.25 \\
            ~& OOD & 7.75 & 6.25 & 4.25 & 5.00 & 1.50 & 3.75 & 4.50 & 3.00 \\
            ~& Adversarial & 7.00 & 6.00 & 5.00 & 8.00 & 1.50 & 3.50 & 2.50 & 2.50 \\
            \hline
            Average  & All & 7.20 & 5.20 & 5.60 & 6.00 & \textbf{2.20} & 4.00 & 3.60 & \textbf{2.20} \\
            % \toprule[1pt]
            \bottomrule[1pt]
            % \toprule[1pt]
            \multirow{3}*{F1 Score} & Random & 8.00 & 4.75 & 5.75 & 6.00 & 3.00 & 4.25 & 3.25 & 1.00 \\
            ~& OOD & 8.00 & 5.75 & 4.75 & 5.75 & 1.75 & 4.25 & 4.00 & 1.75 \\
            ~& Adversarial & 8.00 & 5.50 & 5.50 & 7.00 & 1.00 & 3.00 & 3.00 & 3.00 \\
            \hline
            Average  & All & 8.00 & 5.30 & 5.30 & 6.10 & 2.10 & 4.00 & 3.50 & \textbf{1.70} \\
			\bottomrule[1pt]
            % \toprule[1pt]
		\end{tabular}
	}
	\caption{The average superiority rank~(1 is best) of anomaly detection methods across all types of outliers. Numbers in bold represent the best results.}
	\label{table:training_results_rank}
\end{table}

\noindent \textbf{Main Results.}
\hongming{~\cref{fig:MountainCar_online_full_main_page,fig:Tutankham_online_full_main_page} present the online performance on feature-input task MountainCar and image-input Atari game Tutankham. The first row shows the learning curves of cumulative rewards based on the PPO algorithm.}
% corresponding detection F1 Score (second row) for all tested detection methods across three types of outliers in Tutankham games. 
To better highlight their differences, we omit the confidence bands in~\cref{fig:Tutankham_online_full_main_page}, while providing full results with confidence bands in~\cref{appendix:Training Phase Main Results}~\cref{fig:CartPole_online_full,fig:MountainCar_online_full,fig:Asterix_online_full,fig:Breakout_online_full,fig:SpaceInvaders_online_full,fig:Enduro_online_full,fig:Tutankham_online_full,fig:FishingDerby_online_full} for reference. \hongming{The second row illustrates the relationship between detection accuracy and policy performance. For Atari games, the x-axis represents the average F1 score during learning, while the y-axis represents the final performance. For feature-input tasks like MountainCar and CartPole, some methods can achieve the maximum score, leading to no significant difference in final performance. We exhibit the relationship between the average F1 score and the average policy performance during learning, similar to previous work~\citep{zhang2023replay}. We can find that higher detection accuracy is generally associated with better policy performance.} For each outlier type in~\cref{table:training_results_rank}, we evaluate the \textbf{superiority rank} of all detectors regarding the F1 score and policy performance, where rank 1 indicates the best performance. A smaller superiority rank implies a more effective detection. Our conclusions are as follows: (1) Conformal MD~(MD+C) generally achieves the best detection performance across all considered baselines~(except Auto). The superiority of MD+C over MD highlights the crucial role of accurately calibrated thresholds in the online RL detection setting. 
% (2) RMD is less effective than MD and MD+C, performing only on par with TMD and ED. This degradation is due to \textit{the information loss in the dimension reduction of the feature vectors for reducing the computational cost}. Thus, MD+C and MD are preferable to RMD in the computationally demanding online RL setting. 
\hongming{(2) PECO is ineffective in most of our experiments, which indicates that policy entropy is not a reliable indicator for detecting anomalies. (3) While EnvModel achieves competitive performance in most tasks, it is less practical because it requires executing the action in the environment to obtain the next state before detecting the current state. That being said, EnvModel is a hindsight method and is less applicable in real applications.}
% (3) The average superiority ranks of all considered detectors are similar in terms of performance and F1 score, verifying the consistency of our results.

 %    E2, MD, and RMD perform best or achieve competitive detection performance across different types of outliers, verifying the importance of using partial or complete covariance information in the detector design compared with E1. By contrast, TMD is slightly inferior to MD and RMD, which justifies that the "tied" covariance assumption may be implausible in some cases. 
	% \item The superiority of MD and RMD in the training phase may not be consistent with the evaluation phase. Specifically, MD and RMD do not outperform E2 under certain outlier circumstances. This performance degradation may be caused by the noisy estimation of the covariance matrix in the challenging online learning setting.
	% MD strategy tends to outperform RMD in some cases, which 

\noindent \textbf{Ablation Study on Double Self-Supervised Detectors.} We conduct an ablation study of double self-supervised detectors on Breakout with random and OOD outliers. Results in~\cref{fig:ablation_study_double} of~\cref{appendix:Training Phase_ablation_double} show that double self-supervised detectors reduce detection errors and improve detection accuracy.

\noindent \textbf{Ablation Study on Outlier Proportions.} We also demonstrate the robust detection performance across different proportions of outliers encountered by the agent during training. We conduct experiments on Breakout, and the results are provided in~\cref{fig:Ablation_study} of~\cref{appendix:Training Phase_ablation_noises}.

\subsection{Autonomous Driving Environment}

\begin{figure}[b!]
% \vskip -0.2in	
\centering
\subfigure[Original state]{\label{fig:carla original}\includegraphics[width=0.19\textwidth]{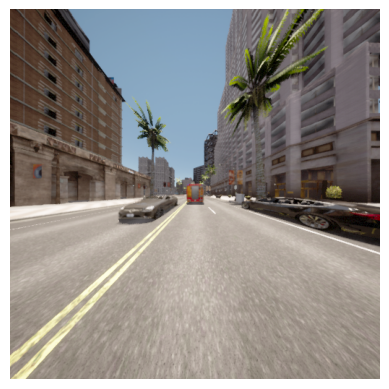}}
\subfigure[Random outlier]{\label{fig:carla random}\includegraphics[width=0.19\textwidth]{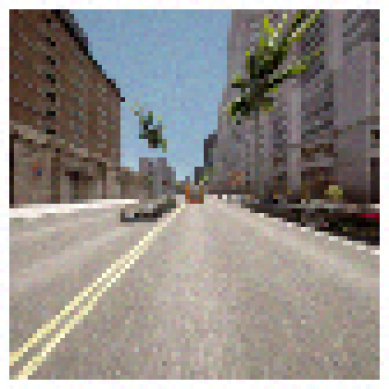}}
\subfigure[Adversarial outlier]{\label{fig:carla adv}\includegraphics[width=0.19\textwidth]{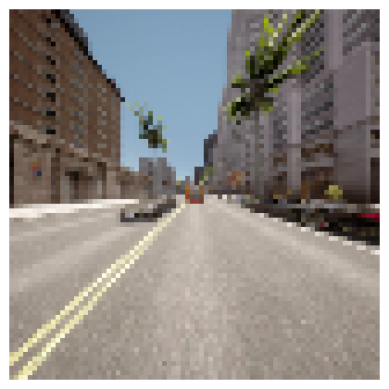}}
\subfigure[OOD outlier~(rainy)]{\label{fig:carla ood rainy}\includegraphics[width=0.19\textwidth]{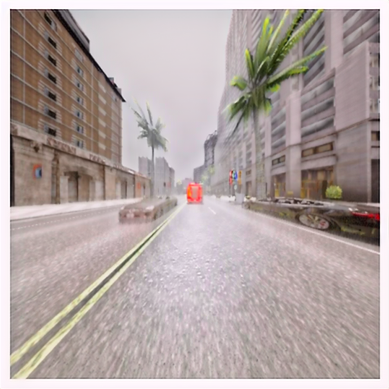}}
\subfigure[OOD outlier~(night)]{\label{fig:carla ood night}\includegraphics[width=0.19\textwidth]{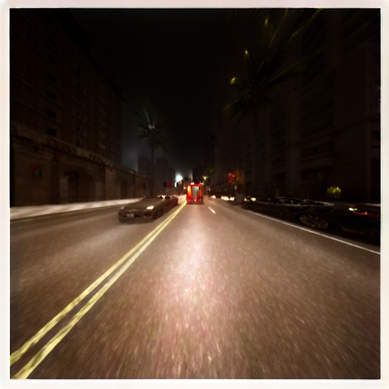}}
% \vskip -0.15in
\caption{The clean and noisy state observations in autonomous driving experiments.}
\label{fig: carla}
\end{figure}

To verify the broader applicability of our method, we perform experiments on autonomous driving environments \hongming{based on  CARLA~\citep{dosovitskiy2017carla}} and introduce practical scenarios in which all three types of anomalies commonly occur. \hongming{Since anomaly detection in autonomous driving is more practical in the offline setting and CARLA is a complex environment that exceeds our computational capacity for online training, we focus on the offline setting in this section.}

\textbf{Random Noise.} Malfunctioning sensors or cameras can introduce random noise into signal observations. For instance, a faulty camera lens may produce distorted images, while a malfunctioning LiDAR sensor might generate erroneous depth measurements. 
Such random noise can impair the reliability of perception systems in autonomous vehicles.

\textbf{Adversarial Attacks.} Adversarial attacks involve intentionally manipulating input signals to disrupt the functioning of RL systems~\citep{bai2024battle}. In the context of autonomous driving, an attacker might tamper with sensor data or traffic signs, resulting in misleading observations and potentially hazardous driving behavior. Adversarial states thus pose a significant threat to the robustness and safety of autonomous driving systems.

\textbf{Out-of-Distribution (OOD) States.} Consider a scenario where an RL policy is trained exclusively under sunny weather. Encountering rainy weather poses a challenge, as the observations captured under these conditions deviate from the training data distribution. Such observations are therefore considered Out-of-Distribution~(OOD) states. 

\noindent \textbf{Experimental Setup.} We conduct experiments using the CARLA environment~\citep{dosovitskiy2017carla}. CARLA is an open-source simulator for autonomous driving research known for its high-quality rendering and realistic physics. The environment includes 3D models of static objects, such as buildings, vegetation, traffic signs, and infrastructure, as well as dynamic objects, such as vehicles and pedestrians.
The task is to drive safely through the town. In each episode, the vehicle must reach a given goal without collision. The episode ends when the vehicle reaches the goal, collides with an obstacle, or exceeds the time limit.

\textbf{Noisy State Observations.} Following the approach used in Atari game settings, we introduce Gaussian noise to simulate random outliers and generate adversarial outliers using adversarial perturbations. For OOD outliers, we leverage CycleGAN-Turbo~\citep{zhu2017unpaired,parmar2024one}, a technique designed for adapting a single-step diffusion model~\citep{ho2020denoising} to new tasks and domains through adversarial learning~\citep{goodfellow2014generative}. This method can perform various image-to-image translation tasks and outperforms existing GAN-based and diffusion-based methods for various scene translation tasks, such as day-to-night conversion and adding/removing weather effects like fog, snow, and rain~\citep{parmar2024one}. Specifically, we use CycleGAN-Turbo to create \textbf{rainy} and \textbf{nighttime} outliers. Examples of different anomaly states are presented in \cref{fig: carla}.

\noindent \textbf{Main Results.} Given a fixed dataset and a pre-trained policy, we assess our detection methods across the three types of outliers. \cref{table:carla} shows the average accuracy, with MD+C achieving the highest performance in most scenarios, while RMD performs best in the presence of small random noises. {\ke{Similar to the results in classical control environments and Atari games, the entropy-based PEOC is still ineffective across all outlier settings. While EnvModel is competitive in identifying nighttime outliers, it is inferior to the other methods against the other considered outliers.}} These results suggest that our proposed method effectively detects outliers for realistic problems, such as autonomous driving.

% \begin{wraptable}{r}{0.55\textwidth}
\begin{table}[t!] % &~&~&~&~
% \vskip -0.1in
\centering
\scalebox{1.0}{
    \begin{tabular}{c|ccccc|cc}
        % \toprule[1pt]
        \bottomrule[1pt]
        \textbf{Detection Accuracy~$(\%)$} &\textbf{ED} &\textbf{TMD} &{\ke{\textbf{PEOC}}} & {\ke{\textbf{EnvModel}}}&\textbf{MD} &\textbf{RMD} &\textbf{MD+C}   \\
        \hline
         % Old $(\%)$ &  50.0 & 88.6 & 90.0 & 83.1 & \bf{94.8}  \\
         Random~(std $\in [0.005, 0.07]$) &  50.0 & 62.0 & 50.0&50.0& 68.7 & \bf{72.1} & 60.7  \\
         Random~(std $\in (0.07, 0.3]$)  &  50.0 & 95.4 &50.0&50.0& 95.2 & 73.8 & \bf{95.8}  \\
         Adversarial&  50.0 & 96.4 &50.0&50.0 & 91.0 & 73.8 & \bf{97.5}  \\
         OOD~(Rain)  &  50.0 & 96.4 &50.0&52.7 & 95.3 & 73.6 & \bf{97.5}  \\
         OOD~(Night)  &  50.0 & 96.4 & 50.0&\textbf{97.5}&95.3 & 73.8 & \bf{97.5}  \\
    \bottomrule[1pt]
        % \toprule[1pt]
    \end{tabular}
}
 \vskip -0.05in
\caption{ Detection accuracy on the CARLA \textit{town} environment over three types of outliers.}  % The number in bold represents the best performance.
\label{table:carla}
 % \vskip -0.15in
\end{table}
% \end{wraptable}
% % \begin{wraptable}{r}{0.55\textwidth}
% \begin{table}[t!] % &~&~&~&~
% \vskip -0.1in
% \centering
% \scalebox{0.9}{
%     \begin{tabular}{c|ccccc}
%         % \toprule[1pt]
%         \bottomrule[1pt]
%         \textbf{Detection Accuracy~$(\%)$} &\textbf{ED} &\textbf{TMD} &\textbf{MD} &\textbf{RMD} &\textbf{MD+C}   \\
%         \hline
%          % Old $(\%)$ &  50.0 & 88.6 & 90.0 & 83.1 & \bf{94.8}  \\
%          Random~(std $\in [0.005, 0.06]$) &  50.0 & 60.8 & 69.8 & \bf{72.1} & 61.7  \\
%          Random~(std $\in (0.06, 0.3]$)  &  50.0 & 95.0 & 95.2 & 73.6 & \bf{96.0}  \\
%          Adversarial&  50.0 & 96.7 & 95.3 & 73.8 & \bf{97.5}  \\
%          OOD~(Rain)  &  50.0 & 96.5 & 95.5 & 74.4 & \bf{97.5}  \\
%          OOD~(Night)  &  50.0 & 96.5 & 95.5 & 74.3 & \bf{97.5}  \\
%     \bottomrule[1pt]
%         % \toprule[1pt]
%     \end{tabular}
% }
%  \vskip -0.05in
% \caption{ Detection accuracy on the CARLA \textit{town} environment over three types of outliers.}  % The number in bold represents the best performance.
% \label{table:carla}
%  \vskip -0.15in
% \end{table}
% % \end{wraptable}

\section{Discussions and Conclusion}
In this paper, we present the first detailed study of a distance-based anomaly detection framework in deep RL, considering random, adversarial, and OOD state outliers in both offline and online settings. The primary detection backbone is based on Mahalanobis distance, and we extend it to robust and distribution-free versions by leveraging robust estimation and conformal prediction techniques. Experiments on \ke{classical control environments,} Atari games, and the autonomous driving environment, demonstrate the effectiveness of our proposed methods in detecting the three types of outliers. The conformal MD method achieves the best detection performance in most scenarios, especially in the online setting. Our research contributes to developing safe and trustworthy RL systems in real-world applications. 

\noindent \textbf{Limitations and Future Work.} In the online setting, especially with a high proportion of outliers, it may be preferable to denoise the detected state outliers via some neighboring smoothing techniques, e.g., \textit{mixup}~\citep{zhang2017mixup,wang2020improving}, rather than deleting them directly as performed in this paper. To relax the Gaussian assumption in the hypothesis test of our detection, we can consider other non-parametric methods, such as one-class support vector machines~\citep{choi2009least} or isolation forests~\citep{liu2008isolation}. A substantial challenge that remains for future work is to devise a more informed detector to distinguish between real ``bad'' outliers that can cause truly misleading actions and ``good'' \textit{new samples} collected through exploration, which can potentially benefit the policy learning, especially for image inputs~\citep{zhang2023robustness}.

\subsubsection*{Acknowledgements}
Ke Sun was supported by the State Scholarship Fund from China Scholarship Council (No:202006010082). Linglong Kong was partially supported by grants from the Canada CIFAR AI Chairs program, the Alberta Machine Intelligence Institute (AMII), and Natural Sciences and Engineering Council of Canada (NSERC), and the Canada Research Chair program from NSERC.
Hongming Zhang and Martin M\"uller were supported by UAHJIC, the Natural Sciences and Engineering Research Council of Canada (NSERC), and an Amii Canada CIFAR AI Chair.

\bibliography{main}

\begin{thebibliography}{98}
\providecommand{\natexlab}[1]{#1}
\providecommand{\url}[1]{\texttt{#1}}
\expandafter\ifx\csname urlstyle\endcsname\relax
  \providecommand{\doi}[1]{doi: #1}\else
  \providecommand{\doi}{doi: \begingroup \urlstyle{rm}\Url}\fi

\bibitem[Abel et~al.(2024)Abel, Barreto, Van~Roy, Precup, van Hasselt, and Singh]{abel2024definition}
David Abel, Andr{\'e} Barreto, Benjamin Van~Roy, Doina Precup, Hado~P van Hasselt, and Satinder Singh.
\newblock A definition of continual reinforcement learning.
\newblock \emph{Advances in Neural Information Processing Systems}, 36, 2024.

\bibitem[Ajay et~al.(2022)Ajay, Gupta, Ghosh, Levine, and Agrawal]{ajay2022distributionally}
Anurag Ajay, Abhishek Gupta, Dibya Ghosh, Sergey Levine, and Pulkit Agrawal.
\newblock Distributionally adaptive meta reinforcement learning.
\newblock \emph{Advances in Neural Information Processing Systems}, 35:\penalty0 25856--25869, 2022.

\bibitem[Anand \& Precup(2024)Anand and Precup]{anand2024prediction}
Nishanth Anand and Doina Precup.
\newblock Prediction and control in continual reinforcement learning.
\newblock \emph{Advances in Neural Information Processing Systems}, 36, 2024.

\bibitem[Angelopoulos et~al.(2021)Angelopoulos, Bates, Jordan, and Malik]{angelopoulos2020uncertainty}
Anastasios~Nikolas Angelopoulos, Stephen Bates, Michael Jordan, and Jitendra Malik.
\newblock Uncertainty sets for image classifiers using conformal prediction.
\newblock In \emph{International Conference on Learning Representations}, 2021.

\bibitem[Bai et~al.(2023)Bai, Zhang, Tao, Wu, Wang, and Xu]{Bai_Zhang_Tao_Wu_Wang_Xu_2023}
Fengshuo Bai, Hongming Zhang, Tianyang Tao, Zhiheng Wu, Yanna Wang, and Bo~Xu.
\newblock Picor: Multi-task deep reinforcement learning with policy correction.
\newblock \emph{Proceedings of the AAAI Conference on Artificial Intelligence}, 37\penalty0 (6):\penalty0 6728--6736, Jun. 2023.
\newblock \doi{10.1609/aaai.v37i6.25825}.
\newblock URL \url{https://ojs.aaai.org/index.php/AAAI/article/view/25825}.

\bibitem[Bai et~al.(2024{\natexlab{a}})Bai, Liu, Du, Wen, and Yang]{bai2024battle}
Fengshuo Bai, Runze Liu, Yali Du, Ying Wen, and Yaodong Yang.
\newblock {BATTLE}: Towards behavior-oriented adversarial attacks against deep reinforcement learning, 2024{\natexlab{a}}.
\newblock URL \url{https://openreview.net/forum?id=rp5vfyp5Np}.

\bibitem[Bai et~al.(2024{\natexlab{b}})Bai, Zhao, Zhang, Cui, Wen, Yang, Xu, and Han]{bai2024efficient}
Fengshuo Bai, Rui Zhao, Hongming Zhang, Sijia Cui, Ying Wen, Yaodong Yang, Bo~Xu, and Lei Han.
\newblock Efficient preference-based reinforcement learning via aligned experience estimation.
\newblock \emph{arXiv preprint arXiv:2405.18688}, 2024{\natexlab{b}}.

\bibitem[Barber et~al.(2023)Barber, Candes, Ramdas, and Tibshirani]{barber2023conformal}
Rina~Foygel Barber, Emmanuel~J Candes, Aaditya Ramdas, and Ryan~J Tibshirani.
\newblock Conformal prediction beyond exchangeability.
\newblock \emph{The Annals of Statistics}, 51\penalty0 (2):\penalty0 816--845, 2023.

\bibitem[Barto et~al.(1983)Barto, Sutton, and Anderson]{barto1983neuronlike}
Andrew~G Barto, Richard~S Sutton, and Charles~W Anderson.
\newblock Neuronlike adaptive elements that can solve difficult learning control problems.
\newblock \emph{IEEE transactions on systems, man, and cybernetics}, 5:\penalty0 834--846, 1983.

\bibitem[{Bellemare} et~al.(2013){Bellemare}, {Naddaf}, {Veness}, and {Bowling}]{bellemare13arcade}
M.~G. {Bellemare}, Y.~{Naddaf}, J.~{Veness}, and M.~{Bowling}.
\newblock The {Arcade} learning environment: An evaluation platform for general agents.
\newblock \emph{Journal of Artificial Intelligence Research}, 47:\penalty0 253--279, June 2013.

\bibitem[Boyan \& Moore(1994)Boyan and Moore]{boyan1994generalization}
Justin Boyan and Andrew Moore.
\newblock Generalization in reinforcement learning: Safely approximating the value function.
\newblock \emph{Advances in neural information processing systems}, 7, 1994.

\bibitem[Brockman et~al.(2016)Brockman, Cheung, Pettersson, Schneider, Schulman, Tang, and Zaremba]{brockman2016openai}
Greg Brockman, Vicki Cheung, Ludwig Pettersson, Jonas Schneider, John Schulman, Jie Tang, and Wojciech Zaremba.
\newblock {OpenAI Gym}, 2016.
\newblock URL \url{https://arxiv.org/abs/1606.01540}.

\bibitem[Butler et~al.(1993)Butler, Davies, and Jhun]{butler1993asymptotics}
RW~Butler, PL~Davies, and M~Jhun.
\newblock Asymptotics for the minimum covariance determinant estimator.
\newblock \emph{The Annals of Statistics}, pp.\  1385--1400, 1993.

\bibitem[Cao et~al.(2020)Cao, Chen, Yao, Wang, and Zhang]{cao2020adversarial}
Yuanjiang Cao, Xiaocong Chen, Lina Yao, Xianzhi Wang, and Wei~Emma Zhang.
\newblock Adversarial attacks and detection on reinforcement learning-based interactive recommender systems.
\newblock In \emph{Proceedings of the 43rd International ACM SIGIR Conference on Research and Development in Information Retrieval}, pp.\  1669--1672, 2020.

\bibitem[Chan et~al.(2020)Chan, Fishman, Korattikara, Canny, and Guadarrama]{chan2019measuring}
Stephanie~C.Y. Chan, Samuel Fishman, Anoop Korattikara, John Canny, and Sergio Guadarrama.
\newblock Measuring the reliability of reinforcement learning algorithms.
\newblock In \emph{International Conference on Learning Representations}, 2020.

\bibitem[Chen et~al.(2019)Chen, Cornelius, Martin, and Chau]{chen2019shapeshifter}
Shang-Tse Chen, Cory Cornelius, Jason Martin, and Duen~Horng Chau.
\newblock {ShapeShifter}: Robust physical adversarial attack on faster {R-CNN} object detector.
\newblock In \emph{Machine Learning and Knowledge Discovery in Databases: European Conference, ECML PKDD 2018, Dublin, Ireland, September 10--14, 2018, Proceedings, Part I 18}, pp.\  52--68. Springer, 2019.

\bibitem[Choi(2009)]{choi2009least}
Young-Sik Choi.
\newblock Least squares one-class support vector machine.
\newblock \emph{Pattern Recognition Letters}, 30\penalty0 (13):\penalty0 1236--1240, 2009.

\bibitem[Danesh \& Fern(2021)Danesh and Fern]{danesh2021out}
Mohamad~H Danesh and Alan Fern.
\newblock Out-of-distribution dynamics detection: {RL}-relevant benchmarks and results.
\newblock \emph{ICML 2021 Workshop on Uncertainty and Robustness in Deep Learning}, 2021.

\bibitem[De~Maesschalck et~al.(2000)De~Maesschalck, Jouan-Rimbaud, and Massart]{de2000mahalanobis}
Roy De~Maesschalck, Delphine Jouan-Rimbaud, and D{\'e}sir{\'e}~L Massart.
\newblock The {Mahalanobis} distance.
\newblock \emph{Chemometrics and Intelligent Laboratory Systems}, 50\penalty0 (1):\penalty0 1--18, 2000.

\bibitem[Dong et~al.(2020)Dong, Ding, and Zhang]{dong2020deep}
Hao Dong, Zihan Ding, and Shanghang Zhang.
\newblock \emph{Deep Reinforcement Learning: Fundamentals, Research and Applications}.
\newblock Springer Nature, 2020.

\bibitem[Dosovitskiy et~al.(2017)Dosovitskiy, Ros, Codevilla, Lopez, and Koltun]{dosovitskiy2017carla}
Alexey Dosovitskiy, German Ros, Felipe Codevilla, Antonio Lopez, and Vladlen Koltun.
\newblock {CARLA}: An open urban driving simulator.
\newblock In \emph{Conference on Robot Learning}, pp.\  1--16. PMLR, 2017.

\bibitem[Elmrabit et~al.(2020)Elmrabit, Zhou, Li, and Zhou]{elmrabit2020evaluation}
Nebrase Elmrabit, Feixiang Zhou, Fengyin Li, and Huiyu Zhou.
\newblock Evaluation of machine learning algorithms for anomaly detection.
\newblock In \emph{2020 international conference on cyber security and protection of digital services (cyber security)}, pp.\  1--8. IEEE, 2020.

\bibitem[Fujimoto et~al.(2018)Fujimoto, Hoof, and Meger]{fujimoto2018addressing}
Scott Fujimoto, Herke Hoof, and David Meger.
\newblock Addressing function approximation error in actor-critic methods.
\newblock In \emph{International conference on machine learning}, pp.\  1587--1596. PMLR, 2018.

\bibitem[Garc{\i}a \& Fern{\'a}ndez(2015)Garc{\i}a and Fern{\'a}ndez]{garcia2015comprehensive}
Javier Garc{\i}a and Fernando Fern{\'a}ndez.
\newblock A comprehensive survey on safe reinforcement learning.
\newblock \emph{Journal of Machine Learning Research}, 16\penalty0 (1):\penalty0 1437--1480, 2015.

\bibitem[Goodfellow et~al.(2014{\natexlab{a}})Goodfellow, Pouget-Abadie, Mirza, Xu, Warde-Farley, Ozair, Courville, and Bengio]{goodfellow2014generative}
Ian Goodfellow, Jean Pouget-Abadie, Mehdi Mirza, Bing Xu, David Warde-Farley, Sherjil Ozair, Aaron Courville, and Yoshua Bengio.
\newblock Generative adversarial nets.
\newblock \emph{Advances in neural information processing systems}, 27, 2014{\natexlab{a}}.

\bibitem[Goodfellow et~al.(2014{\natexlab{b}})Goodfellow, Shlens, and Szegedy]{goodfellow2014explaining}
Ian~J Goodfellow, Jonathon Shlens, and Christian Szegedy.
\newblock Explaining and harnessing adversarial examples.
\newblock \emph{International Conference on Learning Representations (ICLR)}, 2014{\natexlab{b}}.

\bibitem[Gr{\"u}bel(1988)]{grubel1988minimal}
R~Gr{\"u}bel.
\newblock A minimal characterization of the covariance matrix.
\newblock \emph{Metrika}, 35\penalty0 (1):\penalty0 49--52, 1988.

\bibitem[Gu et~al.(2024)Gu, Yang, Du, Chen, Walter, Wang, and Knoll]{gu2022review}
Shangding Gu, Long Yang, Yali Du, Guang Chen, Florian Walter, Jun Wang, and Alois Knoll.
\newblock A review of safe reinforcement learning: Methods, theory and applications, 2024.
\newblock URL \url{https://arxiv.org/abs/2205.10330}.

\bibitem[Haarnoja et~al.(2018)Haarnoja, Zhou, Abbeel, and Levine]{haarnoja2018soft}
Tuomas Haarnoja, Aurick Zhou, Pieter Abbeel, and Sergey Levine.
\newblock Soft actor-critic: Off-policy maximum entropy deep reinforcement learning with a stochastic actor.
\newblock In \emph{International conference on machine learning}, pp.\  1861--1870. PMLR, 2018.

\bibitem[Haider et~al.(2023)Haider, Roscher, Schmoeller~da Roza, and G{\"u}nnemann]{haider2023out}
Tom Haider, Karsten Roscher, Felippe Schmoeller~da Roza, and Stephan G{\"u}nnemann.
\newblock Out-of-distribution detection for reinforcement learning agents with probabilistic dynamics models.
\newblock In \emph{Proceedings of the 2023 International Conference on Autonomous Agents and Multiagent Systems}, pp.\  851--859, 2023.

\bibitem[Hardin \& Rocke(2005)Hardin and Rocke]{hardin2005distribution}
Johanna Hardin and David~M Rocke.
\newblock The distribution of robust distances.
\newblock \emph{Journal of Computational and Graphical Statistics}, 14\penalty0 (4):\penalty0 928--946, 2005.

\bibitem[Hastie et~al.(2009)Hastie, Tibshirani, Friedman, and Friedman]{hastie2009elements}
Trevor Hastie, Robert Tibshirani, Jerome~H Friedman, and Jerome~H Friedman.
\newblock \emph{The elements of statistical learning: data mining, inference, and prediction}, volume~2.
\newblock Springer, 2009.

\bibitem[Hendrycks et~al.(2021)Hendrycks, Carlini, Schulman, and Steinhardt]{hendrycks2021unsolved}
Dan Hendrycks, Nicholas Carlini, John Schulman, and Jacob Steinhardt.
\newblock Unsolved problems in ml safety.
\newblock \emph{arXiv preprint arXiv:2109.13916}, 2021.

\bibitem[Hessel et~al.(2018)Hessel, Modayil, Van~Hasselt, Schaul, Ostrovski, Dabney, Horgan, Piot, Azar, and Silver]{hessel2018rainbow}
Matteo Hessel, Joseph Modayil, Hado Van~Hasselt, Tom Schaul, Georg Ostrovski, Will Dabney, Dan Horgan, Bilal Piot, Mohammad Azar, and David Silver.
\newblock Rainbow: Combining improvements in deep reinforcement learning.
\newblock In \emph{Thirty-second AAAI conference on artificial intelligence}, 2018.

\bibitem[Hilal et~al.(2022)Hilal, Gadsden, and Yawney]{hilal2022financial}
Waleed Hilal, S~Andrew Gadsden, and John Yawney.
\newblock Financial fraud: a review of anomaly detection techniques and recent advances.
\newblock \emph{Expert systems With applications}, 193:\penalty0 116429, 2022.

\bibitem[Ho et~al.(2020)Ho, Jain, and Abbeel]{ho2020denoising}
Jonathan Ho, Ajay Jain, and Pieter Abbeel.
\newblock Denoising diffusion probabilistic models.
\newblock \emph{Advances in neural information processing systems}, 33:\penalty0 6840--6851, 2020.

\bibitem[Hu et~al.(2023)Hu, Li, Hou, Jiang, Liu, Chu, Huang, and Zhang]{hu2023potential}
Jincheng Hu, Jihao Li, Zhuoran Hou, Jingjing Jiang, Cunjia Liu, Liang Chu, Yanjun Huang, and Yuanjian Zhang.
\newblock Potential auto-driving threat: Universal rain-removal attack.
\newblock \emph{Iscience}, 26\penalty0 (9), 2023.

\bibitem[Hu et~al.(2022)Hu, Li, Shi, Wu, and Fryzlewicz]{hu2022doubly}
Liyuan Hu, Mengbing Li, Chengchun Shi, Zhenke Wu, and Piotr Fryzlewicz.
\newblock Doubly inhomogeneous reinforcement learning, 2022.
\newblock URL \url{https://arxiv.org/abs/2211.03983}.

\bibitem[Huang et~al.(2017)Huang, Papernot, Goodfellow, Duan, and Abbeel]{huang2017adversarial}
Sandy Huang, Nicolas Papernot, Ian Goodfellow, Yan Duan, and Pieter Abbeel.
\newblock Adversarial attacks on neural network policies.
\newblock \emph{International Conference on Learning Representations (ICLR) workshop}, 2017.

\bibitem[Huber(2004)]{huber2004robust}
Peter~J Huber.
\newblock \emph{Robust statistics}, volume 523.
\newblock John Wiley \& Sons, 2004.

\bibitem[Hubert \& Debruyne(2010)Hubert and Debruyne]{hubert2010minimum}
Mia Hubert and Michiel Debruyne.
\newblock Minimum covariance determinant.
\newblock \emph{Wiley interdisciplinary reviews: Computational statistics}, 2\penalty0 (1):\penalty0 36--43, 2010.

\bibitem[Ishimtsev et~al.(2017)Ishimtsev, Bernstein, Burnaev, and Nazarov]{ishimtsev2017conformal}
Vladislav Ishimtsev, Alexander Bernstein, Evgeny Burnaev, and Ivan Nazarov.
\newblock Conformal $ k $-nn anomaly detector for univariate data streams.
\newblock In \emph{Conformal and Probabilistic Prediction and Applications}, pp.\  213--227. PMLR, 2017.

\bibitem[Jordan et~al.(2020)Jordan, Chandak, Cohen, Zhang, and Thomas]{jordan2020evaluating}
Scott Jordan, Yash Chandak, Daniel Cohen, Mengxue Zhang, and Philip Thomas.
\newblock Evaluating the performance of reinforcement learning algorithms.
\newblock In \emph{International Conference on Machine Learning}, pp.\  4962--4973. PMLR, 2020.

\bibitem[Kamoi \& Kobayashi(2020)Kamoi and Kobayashi]{kamoi2020mahalanobis}
Ryo Kamoi and Kei Kobayashi.
\newblock Why is the mahalanobis distance effective for anomaly detection?, 2020.
\newblock URL \url{https://arxiv.org/abs/2003.00402}.

\bibitem[Kazari et~al.(2023)Kazari, Shereen, and D{\'a}n]{kazari2023decentralized}
Kiarash Kazari, Ezzeldin Shereen, and Gy{\"o}rgy D{\'a}n.
\newblock Decentralized anomaly detection in cooperative multi-agent reinforcement learning.
\newblock In \emph{IJCAI}, pp.\  162--170, 2023.

\bibitem[Khetarpal et~al.(2022)Khetarpal, Riemer, Rish, and Precup]{khetarpal2022towards}
Khimya Khetarpal, Matthew Riemer, Irina Rish, and Doina Precup.
\newblock Towards continual reinforcement learning: A review and perspectives.
\newblock \emph{Journal of Artificial Intelligence Research}, 75:\penalty0 1401--1476, 2022.

\bibitem[Kiran et~al.(2021)Kiran, Sobh, Talpaert, Mannion, Al~Sallab, Yogamani, and P{\'e}rez]{kiran2021deep}
B~Ravi Kiran, Ibrahim Sobh, Victor Talpaert, Patrick Mannion, Ahmad~A Al~Sallab, Senthil Yogamani, and Patrick P{\'e}rez.
\newblock Deep reinforcement learning for autonomous driving: A survey.
\newblock \emph{IEEE Transactions on Intelligent Transportation Systems}, 23\penalty0 (6):\penalty0 4909--4926, 2021.

\bibitem[Klecka et~al.(1980)Klecka, Iversen, and Klecka]{klecka1980discriminant}
William~R Klecka, Gudmund~R Iversen, and William~R Klecka.
\newblock \emph{Discriminant analysis}, volume~19.
\newblock Sage, 1980.

\bibitem[Laxhammar \& Falkman(2011)Laxhammar and Falkman]{laxhammar2011sequential}
Rikard Laxhammar and G{\"o}ran Falkman.
\newblock Sequential conformal anomaly detection in trajectories based on {Hausdorff} distance.
\newblock In \emph{14th international conference on information fusion}, pp.\  1--8. IEEE, 2011.

\bibitem[Lee et~al.(2018)Lee, Lee, Lee, and Shin]{lee2018simple}
Kimin Lee, Kibok Lee, Honglak Lee, and Jinwoo Shin.
\newblock A simple unified framework for detecting out-of-distribution samples and adversarial attacks.
\newblock \emph{Advances in Neural Information Processing Systems}, 31, 2018.

\bibitem[Levine et~al.(2020)Levine, Kumar, Tucker, and Fu]{levine2020offline}
Sergey Levine, Aviral Kumar, George Tucker, and Justin Fu.
\newblock Offline reinforcement learning: Tutorial, review, and perspectives on open problems, 2020.
\newblock URL \url{https://arxiv.org/abs/2005.01643}.

\bibitem[Lillicrap et~al.(2016)Lillicrap, Hunt, Pritzel, Heess, Erez, Tassa, Silver, and Wierstra]{lillicrap2015continuous}
Timothy~P. Lillicrap, Jonathan~J. Hunt, Alexander Pritzel, Nicolas Heess, Tom Erez, Yuval Tassa, David Silver, and Daan Wierstra.
\newblock Continuous control with deep reinforcement learning.
\newblock \emph{International Conference on Learning Representations (ICLR)}, 2016.

\bibitem[Liu et~al.(2008)Liu, Ting, and Zhou]{liu2008isolation}
Fei~Tony Liu, Kai~Ming Ting, and Zhi-Hua Zhou.
\newblock Isolation forest.
\newblock In \emph{2008 Eighth IEEE International Conference on Data Mining}, pp.\  413--422. IEEE, 2008.

\bibitem[Liu et~al.(2022)Liu, Bai, Du, and Yang]{NEURIPS2022_8be9c134}
Runze Liu, Fengshuo Bai, Yali Du, and Yaodong Yang.
\newblock Meta-reward-net: Implicitly differentiable reward learning for preference-based reinforcement learning.
\newblock In S.~Koyejo, S.~Mohamed, A.~Agarwal, D.~Belgrave, K.~Cho, and A.~Oh (eds.), \emph{Advances in Neural Information Processing Systems}, volume~35, pp.\  22270--22284. Curran Associates, Inc., 2022.
\newblock URL \url{https://proceedings.neurips.cc/paper_files/paper/2022/file/8be9c134bb193d8bd3827d4df8488228-Paper-Conference.pdf}.

\bibitem[Maronna \& Yohai(2014)Maronna and Yohai]{maronna2014robust}
Ricardo~A Maronna and V{\'\i}ctor~J Yohai.
\newblock Robust estimation of multivariate location and scatter.
\newblock \emph{Wiley StatsRef: Statistics Reference Online}, pp.\  1--12, 2014.

\bibitem[Mnih et~al.(2015)Mnih, Kavukcuoglu, Silver, Rusu, Veness, Bellemare, Graves, Riedmiller, Fidjeland, Ostrovski, et~al.]{mnih2015human}
Volodymyr Mnih, Koray Kavukcuoglu, David Silver, Andrei~A Rusu, Joel Veness, Marc~G Bellemare, Alex Graves, Martin Riedmiller, Andreas~K Fidjeland, Georg Ostrovski, et~al.
\newblock Human-level control through deep reinforcement learning.
\newblock \emph{Nature}, 518\penalty0 (7540):\penalty0 529--533, 2015.

\bibitem[Mnih et~al.(2016)Mnih, Badia, Mirza, Graves, Lillicrap, Harley, Silver, and Kavukcuoglu]{mnih2016asynchronous}
Volodymyr Mnih, Adria~Puigdomenech Badia, Mehdi Mirza, Alex Graves, Timothy Lillicrap, Tim Harley, David Silver, and Koray Kavukcuoglu.
\newblock Asynchronous methods for deep reinforcement learning.
\newblock In \emph{International conference on machine learning}, pp.\  1928--1937. PMLR, 2016.

\bibitem[Moore(1990)]{Moore90efficientmemory-based}
Andrew~William Moore.
\newblock Efficient memory-based learning for robot control.
\newblock Technical report, University of Cambridge, 1990.

\bibitem[M{\"u}ller et~al.(2022)M{\"u}ller, Illium, Phan, Haider, and Linnhoff-Popien]{muller2022towards}
Robert M{\"u}ller, Steffen Illium, Thomy Phan, Tom Haider, and Claudia Linnhoff-Popien.
\newblock Towards anomaly detection in reinforcement learning.
\newblock In \emph{Proceedings of the 21st International Conference on Autonomous Agents and Multiagent Systems}, pp.\  1799--1803, 2022.

\bibitem[Nagabandi et~al.(2018)Nagabandi, Clavera, Liu, Fearing, Abbeel, Levine, and Finn]{nagabandi2018learning}
Anusha Nagabandi, Ignasi Clavera, Simin Liu, Ronald~S Fearing, Pieter Abbeel, Sergey Levine, and Chelsea Finn.
\newblock Learning to adapt in dynamic, real-world environments through meta-reinforcement learning.
\newblock \emph{arXiv preprint arXiv:1803.11347}, 2018.

\bibitem[Pang et~al.(2021)Pang, Shen, Cao, and Hengel]{pang2021deep}
Guansong Pang, Chunhua Shen, Longbing Cao, and Anton Van~Den Hengel.
\newblock Deep learning for anomaly detection: A review.
\newblock \emph{ACM Computing Surveys (CSUR)}, 54\penalty0 (2):\penalty0 1--38, 2021.

\bibitem[Papadopoulos et~al.(2002)Papadopoulos, Proedrou, Vovk, and Gammerman]{papadopoulos2002inductive}
Harris Papadopoulos, Kostas Proedrou, Volodya Vovk, and Alex Gammerman.
\newblock Inductive confidence machines for regression.
\newblock In \emph{Machine learning: ECML 2002: 13th European conference on machine learning Helsinki, Finland, August 19--23, 2002 proceedings 13}, pp.\  345--356. Springer, 2002.

\bibitem[Parisotto et~al.(2015)Parisotto, Ba, and Salakhutdinov]{parisotto2015actor}
Emilio Parisotto, Jimmy~Lei Ba, and Ruslan Salakhutdinov.
\newblock Actor-mimic: Deep multitask and transfer reinforcement learning.
\newblock \emph{arXiv preprint arXiv:1511.06342}, 2015.

\bibitem[Parmar et~al.(2024)Parmar, Park, Narasimhan, and Zhu]{parmar2024one}
Gaurav Parmar, Taesung Park, Srinivasa Narasimhan, and Jun-Yan Zhu.
\newblock One-step image translation with text-to-image models, 2024.
\newblock URL \url{https://arxiv.org/abs/2403.12036}.

\bibitem[Pattanaik et~al.(2017)Pattanaik, Tang, Liu, Bommannan, and Chowdhary]{pattanaik2017robust}
Anay Pattanaik, Zhenyi Tang, Shuijing Liu, Gautham Bommannan, and Girish Chowdhary.
\newblock Robust deep reinforcement learning with adversarial attacks.
\newblock \emph{Advances in Neural Information Processing Systems}, 2017.

\bibitem[Patterson et~al.(2020)Patterson, Neumann, White, and White]{patterson2023empirical}
Andrew Patterson, Samuel Neumann, Martha White, and Adam White.
\newblock Draft: Empirical design in reinforcement learning.
\newblock \emph{Journal of Artificial Intelligence Research}, 1, 2020.

\bibitem[Pinto et~al.(2017)Pinto, Davidson, Sukthankar, and Gupta]{pinto2017robust}
Lerrel Pinto, James Davidson, Rahul Sukthankar, and Abhinav Gupta.
\newblock Robust adversarial reinforcement learning.
\newblock \emph{International Conference on Machine Learning (ICML)}, 2017.

\bibitem[Ren et~al.(2021)Ren, Fort, Liu, Roy, Padhy, and Lakshminarayanan]{ren2021simple}
Jie Ren, Stanislav Fort, Jeremiah Liu, Abhijit~Guha Roy, Shreyas Padhy, and Balaji Lakshminarayanan.
\newblock A simple fix to {Mahalanobis} distance for improving near-{OOD} detection.
\newblock \emph{International Conference on Machine Learning (ICML) workshop}, 2021.

\bibitem[Rolnick et~al.(2019)Rolnick, Ahuja, Schwarz, Lillicrap, and Wayne]{NEURIPS2019_fa7cdfad}
David Rolnick, Arun Ahuja, Jonathan Schwarz, Timothy Lillicrap, and Gregory Wayne.
\newblock Experience replay for continual learning.
\newblock In H.~Wallach, H.~Larochelle, A.~Beygelzimer, F.~d\textquotesingle Alch\'{e}-Buc, E.~Fox, and R.~Garnett (eds.), \emph{Advances in Neural Information Processing Systems}, volume~32. Curran Associates, Inc., 2019.

\bibitem[Rousseeuw(1984)]{rousseeuw1984least}
Peter~J Rousseeuw.
\newblock Least median of squares regression.
\newblock \emph{Journal of the American Statistical Association}, 79\penalty0 (388):\penalty0 871--880, 1984.

\bibitem[Rousseeuw \& Van~Zomeren(1990)Rousseeuw and Van~Zomeren]{rousseeuw1990unmasking}
Peter~J Rousseeuw and Bert~C Van~Zomeren.
\newblock Unmasking multivariate outliers and leverage points.
\newblock \emph{Journal of the American Statistical Association}, 85\penalty0 (411):\penalty0 633--639, 1990.

\bibitem[Schulman et~al.(2015)Schulman, Levine, Abbeel, Jordan, and Moritz]{schulman2015trust}
John Schulman, Sergey Levine, Pieter Abbeel, Michael Jordan, and Philipp Moritz.
\newblock Trust region policy optimization.
\newblock In \emph{Proceedings of The 32nd International Conference on Machine Learning}, pp.\  1889--1897. PMLR, 2015.

\bibitem[Schulman et~al.(2017)Schulman, Wolski, Dhariwal, Radford, and Klimov]{schulman2017proximal}
John Schulman, Filip Wolski, Prafulla Dhariwal, Alec Radford, and Oleg Klimov.
\newblock Proximal policy optimization algorithms, 2017.
\newblock URL \url{https://arxiv.org/abs/1707.06347}.

\bibitem[Schulman et~al.(2018)Schulman, Moritz, Levine, Jordan, and Abbeel]{schulman2015high}
John Schulman, Philipp Moritz, Sergey Levine, Michael Jordan, and Pieter Abbeel.
\newblock High-dimensional continuous control using generalized advantage estimation.
\newblock \emph{International Conference on Learning Representations}, 2018.

\bibitem[Sedlmeier et~al.(2020{\natexlab{a}})Sedlmeier, Gabor, Phan, Belzner, and Linnhoff-Popien]{sedlmeier2019uncertainty}
Andreas Sedlmeier, Thomas Gabor, Thomy Phan, Lenz Belzner, and Claudia Linnhoff-Popien.
\newblock Uncertainty-based out-of-distribution classification in deep reinforcement learning.
\newblock In \emph{Proceedings of the 12th International Conference on Agents and Artificial Intelligence}. SCITEPRESS - Science and Technology Publications, 2020{\natexlab{a}}.
\newblock \doi{10.5220/0008949905220529}.

\bibitem[Sedlmeier et~al.(2020{\natexlab{b}})Sedlmeier, M{\"u}ller, Illium, and Linnhoff-Popien]{sedlmeier2020policy}
Andreas Sedlmeier, Robert M{\"u}ller, Steffen Illium, and Claudia Linnhoff-Popien.
\newblock Policy entropy for out-of-distribution classification.
\newblock In \emph{Artificial Neural Networks and Machine Learning--ICANN 2020: 29th International Conference on Artificial Neural Networks, Bratislava, Slovakia, September 15--18, 2020, Proceedings, Part II 29}, pp.\  420--431. Springer, 2020{\natexlab{b}}.

\bibitem[Shafer \& Vovk(2008)Shafer and Vovk]{shafer2008tutorial}
Glenn Shafer and Vladimir Vovk.
\newblock A tutorial on conformal prediction.
\newblock \emph{Journal of Machine Learning Research}, 9\penalty0 (3), 2008.

\bibitem[Sutton \& Barto(2018)Sutton and Barto]{sutton2018reinforcement}
Richard~S Sutton and Andrew~G Barto.
\newblock \emph{Reinforcement learning: An introduction}.
\newblock MIT press, 2018.

\bibitem[Szegedy et~al.(2013)Szegedy, Zaremba, Sutskever, Bruna, Erhan, Goodfellow, and Fergus]{szegedy2013intriguing}
Christian Szegedy, Wojciech Zaremba, Ilya Sutskever, Joan Bruna, Dumitru Erhan, Ian Goodfellow, and Rob Fergus.
\newblock Intriguing properties of neural networks.
\newblock \emph{International Conference on Learning Representations (ICLR)}, 2013.

\bibitem[Taylor \& Stone(2009)Taylor and Stone]{taylor2009transfer}
Matthew~E Taylor and Peter Stone.
\newblock Transfer learning for reinforcement learning domains: A survey.
\newblock \emph{Journal of Machine Learning Research}, 10\penalty0 (7), 2009.

\bibitem[Teng et~al.(2023)Teng, Wen, Zhang, Bengio, Gao, and Yuan]{teng2022predictive}
Jiaye Teng, Chuan Wen, Dinghuai Zhang, Yoshua Bengio, Yang Gao, and Yang Yuan.
\newblock Predictive inference with feature conformal prediction.
\newblock In \emph{The Eleventh International Conference on Learning Representations}, 2023.

\bibitem[Tibshirani et~al.(2019)Tibshirani, Foygel~Barber, Candes, and Ramdas]{tibshirani2019conformal}
Ryan~J Tibshirani, Rina Foygel~Barber, Emmanuel Candes, and Aaditya Ramdas.
\newblock Conformal prediction under covariate shift.
\newblock In H.~Wallach, H.~Larochelle, A.~Beygelzimer, F.~d\textquotesingle Alch\'{e}-Buc, E.~Fox, and R.~Garnett (eds.), \emph{Advances in Neural Information Processing Systems}, volume~32. Curran Associates, Inc., 2019.

\bibitem[Todorov et~al.(2012)Todorov, Erez, and Tassa]{todorov2012mujoco}
Emanuel Todorov, Tom Erez, and Yuval Tassa.
\newblock Mujoco: A physics engine for model-based control.
\newblock In \emph{2012 IEEE/RSJ International Conference on Intelligent Robots and Systems}, pp.\  5026--5033, 2012.
\newblock \doi{10.1109/IROS.2012.6386109}.

\bibitem[Van~der Maaten \& Hinton(2008)Van~der Maaten and Hinton]{van2008visualizing}
Laurens Van~der Maaten and Geoffrey Hinton.
\newblock Visualizing data using t-sne.
\newblock \emph{Journal of machine learning research}, 9\penalty0 (11), 2008.

\bibitem[Vovk et~al.(2005)Vovk, Gammerman, and Shafer]{vovk2005algorithmic}
Vladimir Vovk, Alexander Gammerman, and Glenn Shafer.
\newblock \emph{Algorithmic learning in a random world}, volume~29.
\newblock Springer, 2005.

\bibitem[Wang et~al.(2024)Wang, Erfani, Alpcan, and Leckie]{wang2024oil}
Chen Wang, Sarah Erfani, Tansu Alpcan, and Christopher Leckie.
\newblock Oil-ad: An anomaly detection framework for sequential decision sequences, 2024.
\newblock URL \url{https://arxiv.org/abs/2402.04567}.

\bibitem[Wang et~al.(2020)Wang, Kang, Shao, and Feng]{wang2020improving}
Kaixin Wang, Bingyi Kang, Jie Shao, and Jiashi Feng.
\newblock Improving generalization in reinforcement learning with mixture regularization.
\newblock In H.~Larochelle, M.~Ranzato, R.~Hadsell, M.F. Balcan, and H.~Lin (eds.), \emph{Advances in Neural Information Processing Systems}, volume~33, pp.\  7968--7978. Curran Associates, Inc., 2020.

\bibitem[Wilson \& Hilferty(1931)Wilson and Hilferty]{wilson1931distribution}
Edwin~B Wilson and Margaret~M Hilferty.
\newblock The distribution of chi-square.
\newblock \emph{Proceedings of the National Academy of Sciences of the United States of America}, 17\penalty0 (12):\penalty0 684, 1931.

\bibitem[Xiao et~al.(2019)Xiao, Huang, He, Silva, Emrani, and Chaudhuri]{xiao2019online}
Wei Xiao, Xiaolin Huang, Fan He, Jorge Silva, Saba Emrani, and Arin Chaudhuri.
\newblock Online robust principal component analysis with change point detection.
\newblock \emph{IEEE Transactions on Multimedia}, 22\penalty0 (1):\penalty0 59--68, 2019.

\bibitem[Xu et~al.(2018)Xu, van Hasselt, and Silver]{xu2018meta}
Zhongwen Xu, Hado~P van Hasselt, and David Silver.
\newblock Meta-gradient reinforcement learning.
\newblock \emph{Advances in neural information processing systems}, 31, 2018.

\bibitem[Zhang \& Yu(2020)Zhang and Yu]{zhang2020taxonomy}
Hongming Zhang and Tianyang Yu.
\newblock Taxonomy of reinforcement learning algorithms.
\newblock \emph{Deep reinforcement learning: Fundamentals, research and applications}, pp.\  125--133, 2020.

\bibitem[Zhang et~al.(2023{\natexlab{a}})Zhang, Xiao, Wang, Jin, Bo, and M{\"u}ller]{zhang2023replay}
Hongming Zhang, Chenjun Xiao, Han Wang, Jun Jin, Xu~Bo, and Martin M{\"u}ller.
\newblock Replay memory as an empirical mdp: Combining conservative estimation with experience replay.
\newblock In \emph{The Eleventh International Conference on Learning Representations}, 2023{\natexlab{a}}.

\bibitem[Zhang et~al.(2018)Zhang, Cisse, Dauphin, and Lopez-Paz]{zhang2017mixup}
Hongyi Zhang, Moustapha Cisse, Yann~N. Dauphin, and David Lopez-Paz.
\newblock mixup: Beyond empirical risk minimization.
\newblock In \emph{International Conference on Learning Representations}, 2018.

\bibitem[Zhang et~al.(2020)Zhang, Chen, Xiao, Li, Boning, and Hsieh]{zhang2020robust}
Huan Zhang, Hongge Chen, Chaowei Xiao, Bo~Li, Duane Boning, and Cho-Jui Hsieh.
\newblock Robust deep reinforcement learning against adversarial perturbations on observations.
\newblock \emph{Advances in Neural Information Processing Systems}, 2020.

\bibitem[Zhang \& Ranganath(2023)Zhang and Ranganath]{zhang2023robustness}
Lily~H Zhang and Rajesh Ranganath.
\newblock Robustness to spurious correlations improves semantic out-of-distribution detection.
\newblock In \emph{Proceedings of the AAAI Conference on Artificial Intelligence}, volume~37, pp.\  15305--15312, 2023.

\bibitem[Zhang et~al.(2023{\natexlab{b}})Zhang, Shi, and Luo]{zhang2023conformal}
Yingying Zhang, Chengchun Shi, and Shikai Luo.
\newblock Conformal off-policy prediction.
\newblock In \emph{International Conference on Artificial Intelligence and Statistics}, pp.\  2751--2768. PMLR, 2023{\natexlab{b}}.

\bibitem[Zhu et~al.(2017)Zhu, Park, Isola, and Efros]{zhu2017unpaired}
Jun-Yan Zhu, Taesung Park, Phillip Isola, and Alexei~A Efros.
\newblock Unpaired image-to-image translation using cycle-consistent adversarial networks.
\newblock In \emph{Proceedings of the IEEE international conference on computer vision}, pp.\  2223--2232, 2017.

\bibitem[Zhu et~al.(2023)Zhu, Lin, Jain, and Zhou]{zhu2023transfer}
Zhuangdi Zhu, Kaixiang Lin, Anil~K Jain, and Jiayu Zhou.
\newblock Transfer learning in deep reinforcement learning: A survey.
\newblock \emph{IEEE Transactions on Pattern Analysis and Machine Intelligence}, 2023.

\end{thebibliography}
\bibliographystyle{tmlr}

\clearpage
\appendix
\section{Proof of Proposition~\ref{prop}}\label{appendix:chi_square}
\begin{proof}
	We show that for each action class $c$, the square of Mahalanobis distance $d$ is identically independent Chi-squared distributed under the Gaussian assumption. Without loss of generality, we denote $\mu$ and $\Sigma$ as the mean and variance matrix of the closest class-conditional Gaussian distribution. We need to show $d=(f(\mathbf{s})-\mu)^{\top} \Sigma^{-1}(f(\mathbf{s})-\mu)$ is Chi-squared distributed. Firstly, by eigenvalue decomposition, we have
	\begin{equation}\begin{aligned}
			\Sigma^{-1} = \sum_{k=1}^{p} \lambda_k^{-1} u_k u_k^{\top},
	\end{aligned}\end{equation}
	where $\lambda_k$ and $u_k$ are the $k$-th eigenvalue and eigenvector of $\Sigma$. Plugging it into the form of $d$, we immediately obtain
	\begin{equation}\begin{aligned}
			d=&(f(\mathbf{s})-\mu)^{\top} \Sigma^{-1}(f(\mathbf{s})-\mu)\\
			=&(f(\mathbf{s})-\mu)^{\top} (\sum_{k=1}^{p} \lambda_k^{-1} u_k u_k^{\top}) (f(\mathbf{s})-\mu)\\
			=&\sum_{k=1}^{p} \lambda_k^{-1} (f(\mathbf{s})-\mu)^{\top} u_k u_k^{\top} (f(\mathbf{s})-\mu)\\
			=&\sum_{k=1}^{p}  \left[\lambda_k^{-\frac{1}{2}} u_k^{\top} (f(\mathbf{s})-\mu)\right]^2\\
			=&\sum_{k=1}^{p} \mathbf{X}^2_k,
	\end{aligned}\end{equation}
	where $\mathbf{X}^2_k$ is a new Gaussian variable that results from the linear transform of a Gaussian distribution $f(\mathbf{s})$ where $f(\mathbf{s}) \sim \mathcal{N}(\mu, \Sigma)$. Therefore, the resulting variance $\sigma_k^2$ can be derived as
	\begin{equation}\begin{aligned}
			\sigma_k^2 &=\lambda_k^{-\frac{1}{2}} u_k^{\top} \Sigma \lambda_k^{-\frac{1}{2}} u_k =\lambda_k^{-1} u_k^{\top} (\sum_{j=1}^{p} \lambda_j u_j u_j^{\top}) u_k =\sum_{j=1}^{p} \lambda_k^{-1} \lambda_j u_k^{\top} u_j u_j^{\top} u_k
	\end{aligned}\end{equation}
	
	As the $\mu_j$ and $\mu_k$ are orthogonal if $j\neq k$, the variance $\sigma_k^2$ can be further reduced to
	\begin{equation}\begin{aligned}
			\sigma_k^2 &=\lambda_k^{-1} \lambda_k u_k^{\top} u_k u_k^{\top} u_k=\Vert u_k \vert^2  \Vert u_k \vert^2 =1.
	\end{aligned}\end{equation}
	
	Each $\mathbf{X}_k$ is a standard Gaussian distribution. Then we have $d$, the square of Mahalanobis distance, Chi-squared distributed, i.e., $d\sim \chi^2(p)$, independent of the action class $c$. Without loss of generality, the smallest $d$ among all action classes, i.e., $M(\mathbf{s})$, is also a Chi-squared distribution. That is to say, $M(\mathbf{s}) \sim \chi^2(p)$.
	
\end{proof}

\section{Results in Offline Setting}\label{appendix:main}

\subsection{Results across Different Noise Strengths}\label{appendix:evaluation_moreresults}

\begin{table*}[htbp] % &~&~&~&~
	\centering
	\caption{Detection accuracy~(\%) of our MD, Robust MD, and conformal MD strategies compared with other baseline methods on two classical control environments with $\alpha=0.05$.}
 	\vspace{1.5ex}
	\scalebox{0.9}{

 {\ke{

		\begin{tabular}{lcccccc|ccc}
			\toprule[1pt]
			\textbf{Environments} &\textbf{Outliers}&\textbf{Perturbation }&\textbf{ED}  &\textbf{TMD}  &\textbf{PEOC}  &\textbf{EnvModel} &\textbf{MD} &\textbf{RMD} &\textbf{MD+C} \\
			\hline  
			
			\multirow{5}*{Cartpole} &\multirow{2}*{Random}&std=0.3&60.54&93.24&50.00 & 50.01& 94.74&78.7& \textbf{95.33} \\
			~&~&std=0.5&75.54&94.49&50.00&49.98&96.14&79.15& \textbf{97.00}\\
		
			~&\multirow{2}*{Adversarial}&$\epsilon$=0.15&50.58&92.53&50.00&49.98&93.65&78.48 &\textbf{93.79} \\
			~&~&$\epsilon$=0.2&51.65&93.89&50.00&50.01&95.31& 78.97& \textbf{95.76} \\
			
			~&OOD&MountainCar&87.27&94.26&50.00&93.84&96.45&79.09&\textbf{97.46}  \\

			\hline

			\multirow{5}*{MountainCar} &\multirow{2}*{Random}&std=0.3 &86.91&85.82&50.00&50.11&89.81&77.84& \textbf{92.52}  \\
			~&~&std=0.5 &92.08&86.96&50.00&49.71&91.37&78.35&\textbf{94.73} \\
			~&\multirow{2}*{Adversarial}&$\epsilon$=0.001 &63.86&80.72&50.00&49.69&85.09&74.04& \textbf{86.83} \\
			~&~&$\epsilon$=0.01&64.13&81.30&50.00&49.693&85.72&74.63&\textbf{87.32} \\
			~&OOD&Cartpole&90.89&86.51&50.00&48.73&90.49& 77.21&\textbf{91.69} \\
			\hline

   \bottomrule[1pt]
   
		\end{tabular}
  }}
}
\label{table_offline_mainresults_conformal_details_classical}
\end{table*}

\begin{table*}[htbp] % &~&~&~&~
	\centering
	\caption{Detection accuracy~(\%) of our MD, Robust MD, and conformal MD strategies compared with other baseline methods on six Atari games with $\alpha=0.05$.}
 	\vspace{1.5ex}
	\scalebox{0.9}{

 {\ke{

		\begin{tabular}{lcccccc|ccc}
			\toprule[1pt]
			\textbf{Games} &\textbf{Outliers}&\textbf{Perturbation }&\textbf{ED}  &\textbf{TMD}  &\textbf{PEOC}  &\textbf{EnvModel} &\textbf{MD} &\textbf{RMD} &\textbf{MD+C} \\
			\hline  
			
			\multirow{6}*{Breakout} &\multirow{2}*{Random}&std=0.02&50.13&52.01&50.00 & 50.00& 54.89&\textbf{62.80}& 54.46 \\
			~&~&std=0.04&56.18&66.26&50.00&50.00&67.85&\textbf{79.64}& 66.76\\
		
			~&\multirow{2}*{Adversarial}&$\epsilon$=0.001&81.28&87.44&50.00&50.00&89.39&79.85 &\textbf{89.54} \\
			~&~&$\epsilon$=0.01&87.36&90.67&50.00&50.00&92.23& 80.57& \textbf{93.87} \\
			
			~&\multirow{2}*{OOD}&Asterix&66.80&47.91&49.99&\textbf{97.50}&50.32&80.73&51.15  \\
			~&~&SpaceInvaders&47.07&47.74&49.99&\textbf{97.50}&48.92&78.22& 50.37 \\
			\hline

			\multirow{6}*{Asterix} &\multirow{2}*{Random}&std=0.1 &42.56&42.87&50.01&49.99&48.28&\textbf{63.05}& 49.04  \\
			~&~&std=0.2 &45.29&47.37&50.01&49.99&72.31&\textbf{75.86}&60.31 \\
			~&\multirow{2}*{Adversarial}&$\epsilon$=0.001 &83.41&85.31&50.01&50.00&91.38&75.02& \textbf{93.62} \\
			~&~&$\epsilon$=0.01&83.94&85.90&50.01&50.00&91.95&75.40&\textbf{94.35} \\
			~&\multirow{2}*{OOD}&Breakout&41.73&42.94&50.01&48.95&48.24& \textbf{75.72}&51.85 \\
			~&~&SpaceInvaders&37.38&38.61&50.01&\textbf{58.55}&43.92&39.62& 47.50 \\
			\hline

			\multirow{6}*{SpaceInvaders} &\multirow{2}*{Random}&std=0.02&50.00&48.46&50.01&50.00&53.26 &\textbf{75.24}&52.53  \\
			~&~&std=0.04&52.88&79.38&50.01&50.00&\textbf{87.20}& 83.57&84.62  \\
			~&\multirow{2}*{Adversarial}&$\epsilon$=0.001&67.13&89.69&50.01&50.00&95.64&83.16&\textbf{95.87}  \\
			~&~&$\epsilon$=0.01&74.66&90.93 &50.01&50.00&96.65& 79.07& \textbf{97.05}\\
			~&\multirow{2}*{OOD}&Breakout&45.81&45.64&50.01&48.97&56.56& \textbf{78.88}&54.23 \\
			~&~&Asterix&44.71&46.22&50.01&48.72&57.45&\textbf{83.06}& 53.03\\
\hline			

\multirow{6}*{Enduro} &\multirow{2}*{Random}&std=0.1&49.42&45.27&50.00&50.00&54.03&\textbf{81.65}& 52.24\\
~&~&std=0.2&48.67&72.70&50.00&49.99&\textbf{91.40}&83.15& 88.17\\
~&\multirow{2}*{Adversarial}&$\epsilon$=0.001&91.86&91.26&50.00&49.99&96.24& 83.27  & \textbf{97.48}\\
~&~&$\epsilon$=0.01&93.93&91.26&50.00&49.99&96.24&83.37 & \textbf{97.48} \\
~&\multirow{2}*{OOD}&FishingDerby&63.95&83.15&50.00&47.54&\textbf{85.10}& 83.44 & 61.39 \\
~&~&Tutankham&50.19&66.60&50.00&47.67&74.98&\textbf{83.28 }& 65.13 \\
                \hline
\multirow{6}*{FishingDerby} &\multirow{2}*{Random}&std=0.2 &48.80&48.20&50.00&50.00&51.65& \textbf{83.77} &50.82 \\
~&~&std=0.3 &49.03&84.60&50.00&50.00&\textbf{87.86}&87.71&82.09 \\
~&\multirow{2}*{Adversarial}&$\epsilon$=0.001&83.44&92.48&50.00&50.01&\textbf{97.31}& 86.90&97.33 \\
~&~&$\epsilon$=0.01&89.12&92.54&50.00&50.01&\textbf{97.49}& 87.41& 97.46\\
~&\multirow{2}*{OOD}&Enduro&48.89&56.69&50.00&75.34&60.31&\textbf{86.45}& 59.78\\
~&~&Tutankham&53.64&55.65&49.96&47.97&57.93&\textbf{76.59} &57.25 \\

   \hline 
                
   \multirow{6}*{Tutankham} &\multirow{2}*{Random}&std=0.04&50.00&48.30&50.00&50.03&49.31&\textbf{71.49}& 50.00 \\
~&~&std=0.06&50.00&46.68&50.03&50.00&48.79&\textbf{76.51}& 49.57 \\
~&\multirow{2}*{Adversarial}&$\epsilon$=0.01&60.08&89.37&50.03&50.01&95.24&77.07& \textbf{95.56} \\
~&~&$\epsilon$=0.05&72.36&89.55&50.03&50.01&95.28&77.05& \textbf{97.49} \\
   ~&\multirow{2}*{OOD}&Enduro&49.96&89.17&50.03&\textbf{97.50}&95.12& 77.18& 91.59\\
   ~&~&FishingDerby&50.0&77.34&50.03&\textbf{97.50}&84.12&77.17& 67.77\\
   \bottomrule[1pt]
		\end{tabular}
  }}
	}

		% \vspace{1.5ex}
	\label{table_offline_mainresults_conformal_details}
\end{table*}

We provide detailed detection accuracy of various detection methods across different noise strengths. The results on two classical control environments and six Atari games are gived in~\cref{table_offline_mainresults_conformal_details_classical,table_offline_mainresults_conformal_details}.

\subsection{Visualization of Outlier States on Six Games}\label{appendix:visualization}

\begin{figure*}[b!]
	\centering
	% breakout 
	\subfigure[Breakout: Clean]{\includegraphics[width=0.19\textwidth,trim=0 0 0 0,clip]{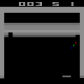}}
	\subfigure[Breakout: Random]{\includegraphics[width=0.19\textwidth,trim=0 0 0 0,clip]{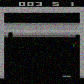}}
	\subfigure[Breakout: Adversarial]{\includegraphics[width=0.19\textwidth,trim=0 0 0 10,clip]{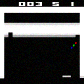}}
	\subfigure[Breakout: Asterix]{\includegraphics[width=0.19\textwidth,trim=0 0 0 0,clip]{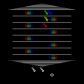}}
	\subfigure[Breakout: Space]{\includegraphics[width=0.19\textwidth,trim=0 0 0 0,clip]{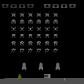}}
	\subfigure[Asterix: Clean]{\includegraphics[width=0.19\textwidth,trim=0 0 0 0,clip]{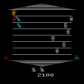}}
	\subfigure[Asterix: Random]{\includegraphics[width=0.19\textwidth,trim=0 0 0 0,clip]{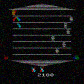}}
	\subfigure[Asterix: Adversarial]{\includegraphics[width=0.19\textwidth,trim=0 0 0 10,clip]{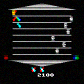}}
	\subfigure[Asterix: Breakout]{\includegraphics[width=0.19\textwidth,trim=0 0 0 0,clip]{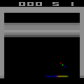}}
	\subfigure[Asterix: Space]{\includegraphics[width=0.19\textwidth,trim=0 0 0 0,clip]{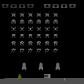}}
	\subfigure[Space: Clean]{\includegraphics[width=0.19\textwidth,trim=0 0 0 0,clip]{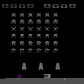}}
	\subfigure[Space: Random]{\includegraphics[width=0.19\textwidth,trim=0 0 0 0,clip]{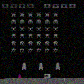}}
	\subfigure[Space: Adversarial]{\includegraphics[width=0.19\textwidth,trim=0 0 0 10,clip]{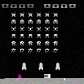}}
	\subfigure[Space: Breakout]{\includegraphics[width=0.19\textwidth,trim=0 0 0 0,clip]{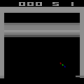}}
	\subfigure[Space: Asterix]{\includegraphics[width=0.19\textwidth,trim=0 0 0 0,clip]{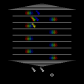}}
	
	\caption{Visualization of various state outliers on Breakout, Asterix, and SpaceInvaders games.}
	\label{fig:images_BAS}
\end{figure*}

\begin{figure*}[htbp]
	\centering
	% enduro 
	\subfigure[Enduro: Clean]{\includegraphics[width=0.19\textwidth,trim=0 0 0 0,clip]{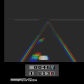}}
	\subfigure[Enduro: Random]{\includegraphics[width=0.19\textwidth,trim=0 0 0 0,clip]{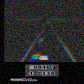}}
	\subfigure[Enduro: Adversarial]{\includegraphics[width=0.19\textwidth,trim=0 0 0 10,clip]{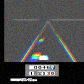}}
	\subfigure[Enduro: Fishing]{\includegraphics[width=0.19\textwidth,trim=0 0 0 0,clip]{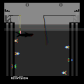}}
	\subfigure[Enduro: Tutankham]{\includegraphics[width=0.19\textwidth,trim=0 0 0 0,clip]{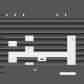}}
	\subfigure[Fishing: Clean]{\includegraphics[width=0.19\textwidth,trim=0 0 0 0,clip]{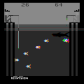}}
	\subfigure[Fishing: Random]{\includegraphics[width=0.19\textwidth,trim=0 0 0 0,clip]{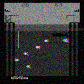}}
	\subfigure[Fishing: Adversarial]{\includegraphics[width=0.19\textwidth,trim=0 0 0 10,clip]{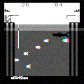}}
	\subfigure[Fishing: Enduro]{\includegraphics[width=0.19\textwidth,trim=0 0 0 0,clip]{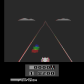}}
	\subfigure[Fishing: Tutankham]{\includegraphics[width=0.19\textwidth,trim=0 0 0 0,clip]{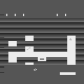}}
	\subfigure[Tutankham: Clean]{\includegraphics[width=0.19\textwidth,trim=0 0 0 0,clip]{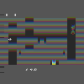}}
	\subfigure[Tutankham: Random]{\includegraphics[width=0.19\textwidth,trim=0 0 0 0,clip]{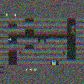}}
	\subfigure[Tutankham: Adv]{\includegraphics[width=0.19\textwidth,trim=0 0 0 10,clip]{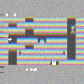}}
	\subfigure[Tutankham: Enduro]{\includegraphics[width=0.19\textwidth,trim=0 0 0 0,clip]{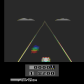}}
	\subfigure[Tutankham: Fishing]{\includegraphics[width=0.19\textwidth,trim=0 0 0 0,clip]{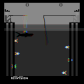}}
	
	\caption{Visualization of various state outliers on Enduro, FishingDerby, and Tutankham games.}
	\label{fig:images_EFT}
\end{figure*}

We plot the outlier states on Breakout, Asterix, and SpaceInvaders games in \cref{fig:images_BAS} and outliers states on Enduro, FishingDerby, and Tutankham in \cref{fig:images_EFT}.

\clearpage
\subsection{Effectiveness of Robust MD}\label{appendix:boxplot}

\begin{figure*}[b!]
	\centering
	\subfigure[Random Outliers.]{\includegraphics[width=0.32\textwidth]{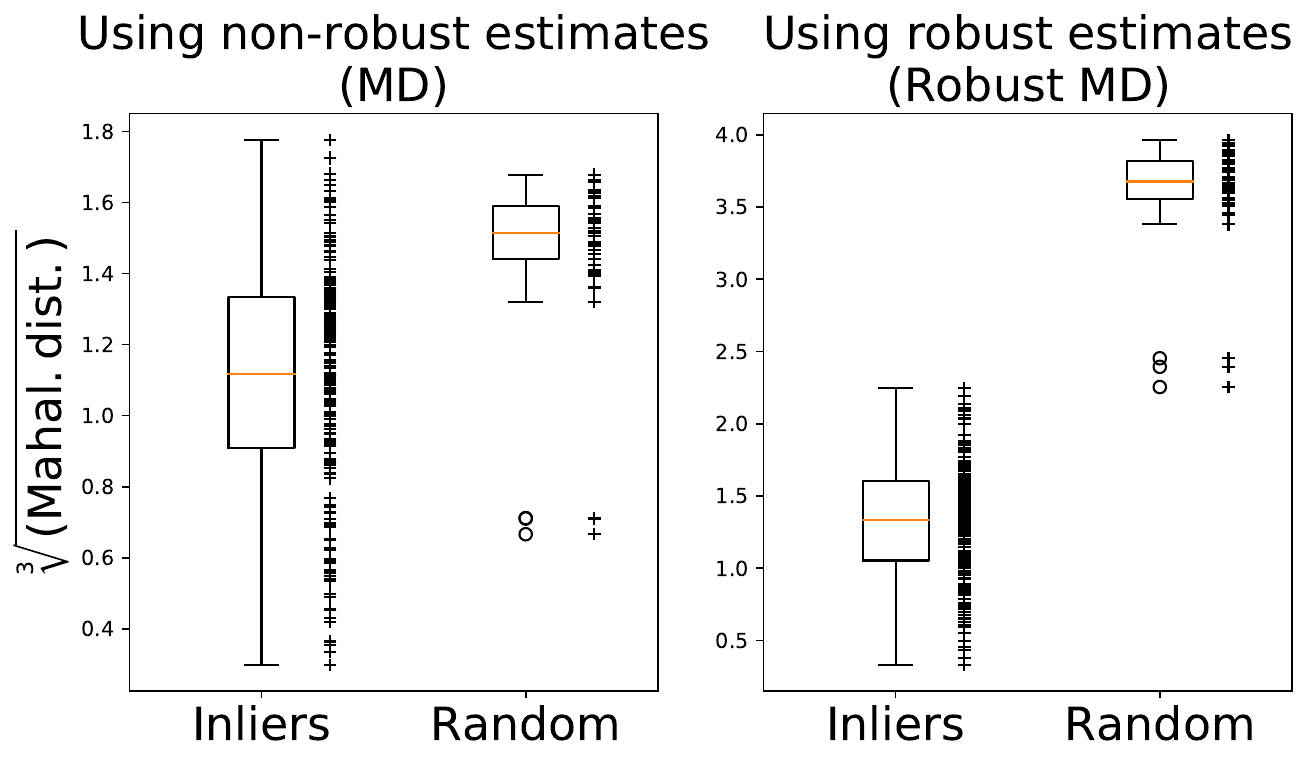}}
	\subfigure[Adversarial Outliers.]{\includegraphics[width=0.32\textwidth]{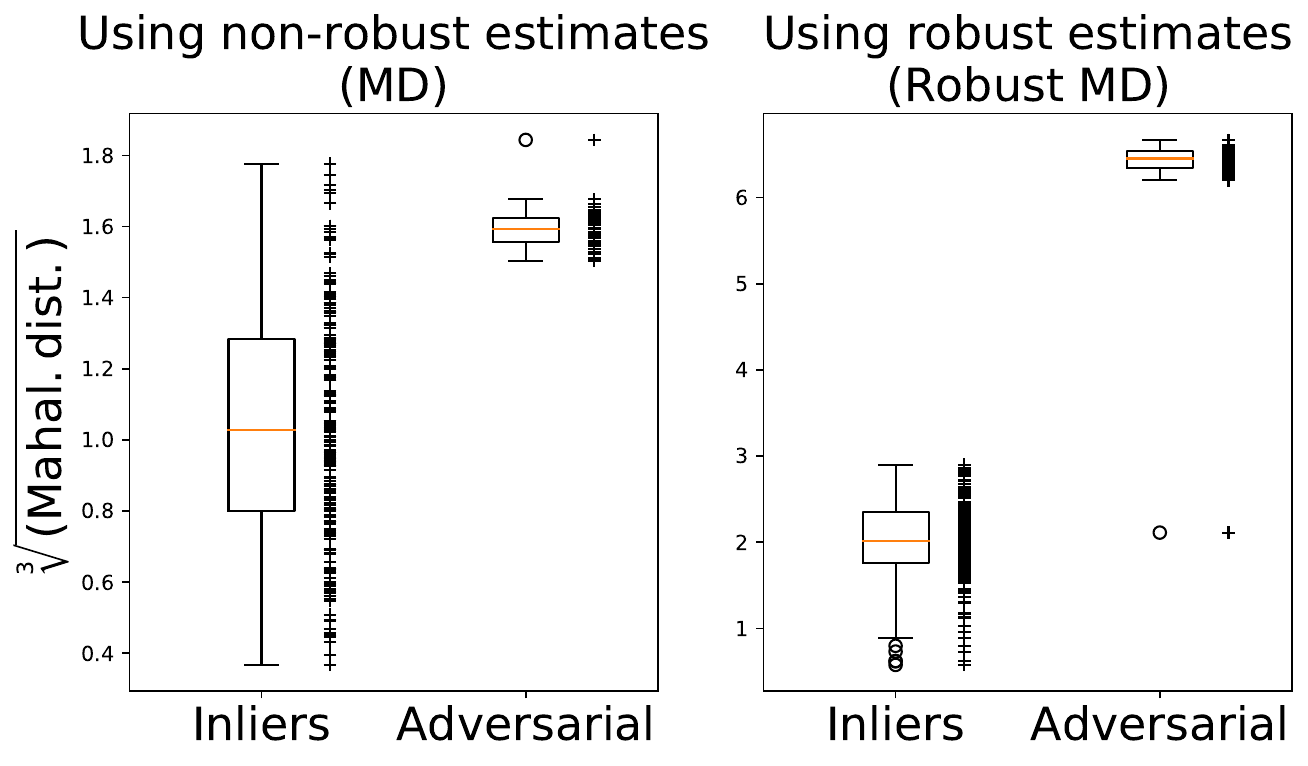}}
	\subfigure[OOD Outliers.]{\includegraphics[width=0.32\textwidth]{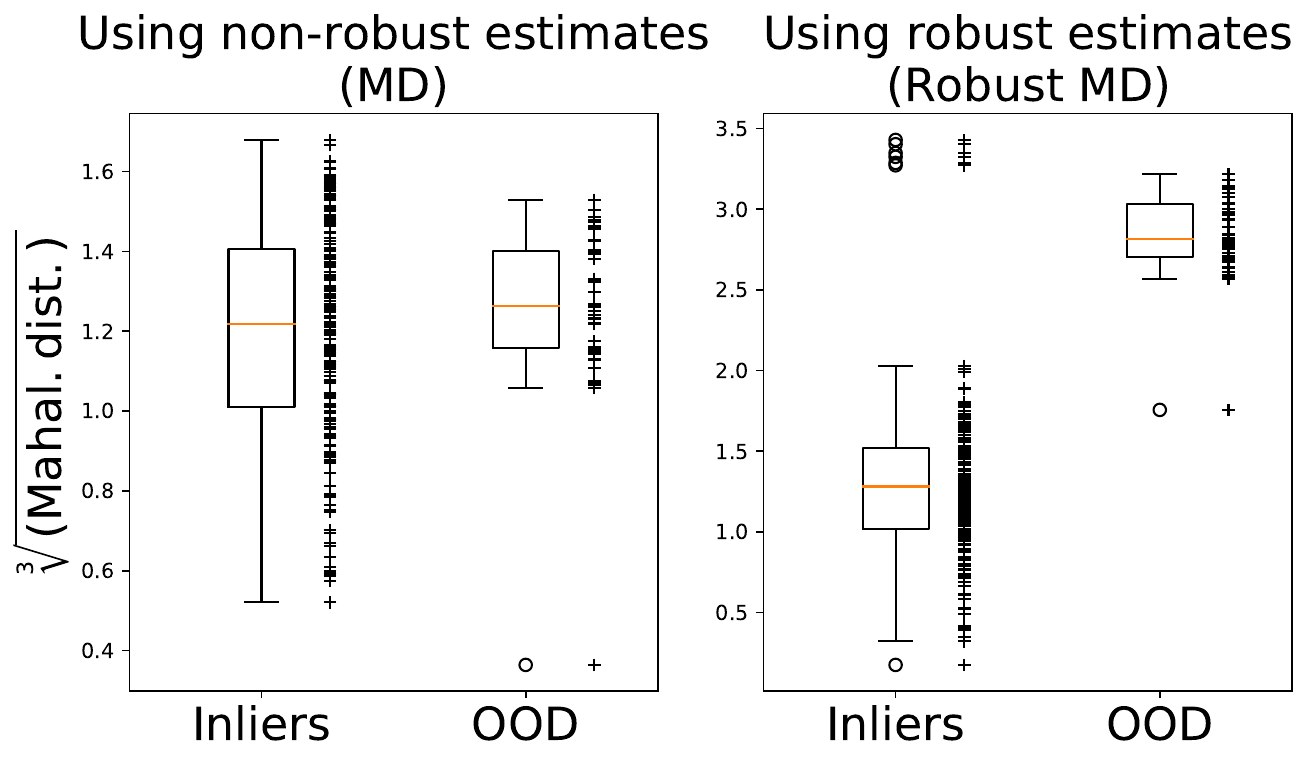}}
	\vskip -0.1in
	\caption{Boxplot of distributions between inliers and three types of outliers in an action class on Breakout game.}
	\label{fig:boxplot_breakout}
\end{figure*}

We take the cubic root of the Mahalanobis distances, yielding approximately normal distributions~\citep{wilson1931distribution}. In this experiment, 250 clean states are drawn from the replay buffer, and 50 abnormal states are drawn from each of the three types of outliers. We reduce the state feature dimension to 2 via t-SNE and compute Mahalanobis distances of these two kinds of states to their centrality within each action class under the estimation based on MD or Robust MD, respectively. \cref{fig:boxplot_breakout} suggests that Robust MD separates inliers and outliers better than MD on Breakout within a random action class, indicating its effectiveness in detecting RL evaluation. Similar results are also given in other games.

\begin{figure*}[b!]
	\centering
	\subfigure[Random Outliers.]{\includegraphics[width=0.32\textwidth]{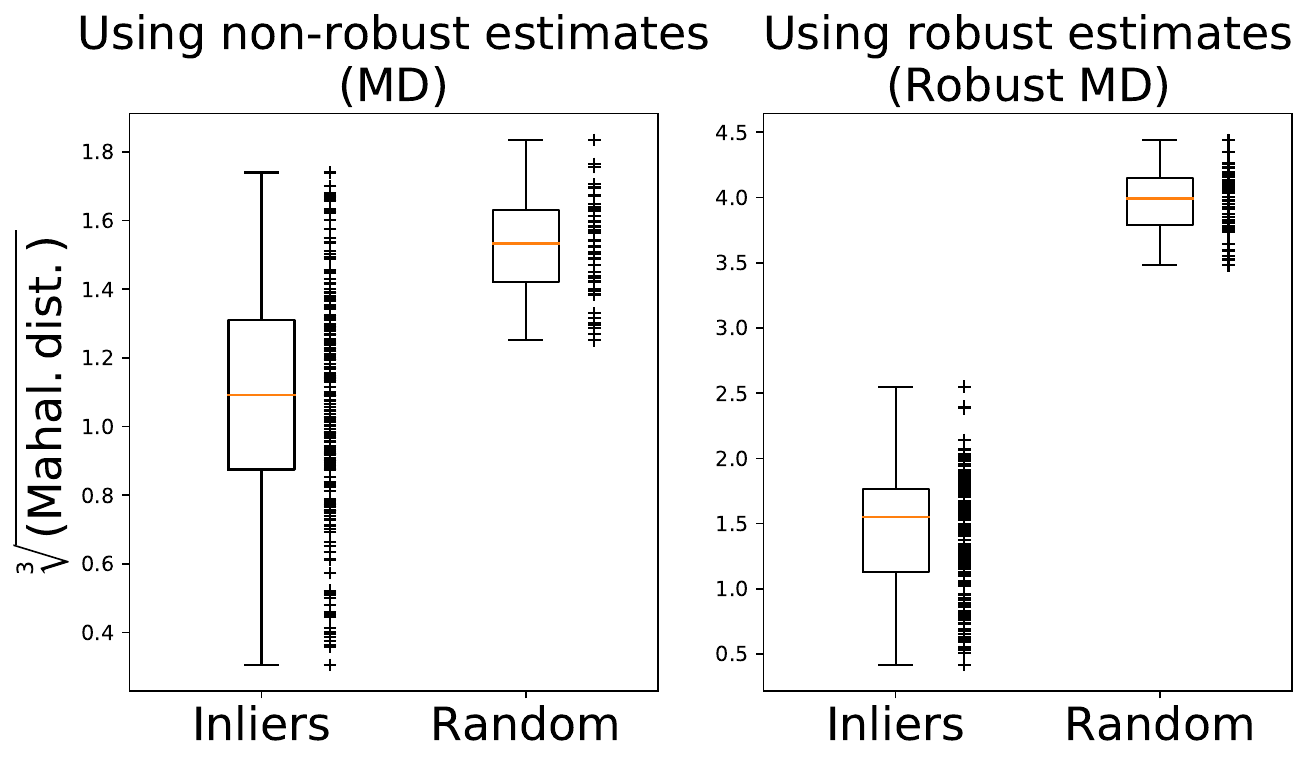}}
	\subfigure[Adversarial Outliers.]{\includegraphics[width=0.32\textwidth]{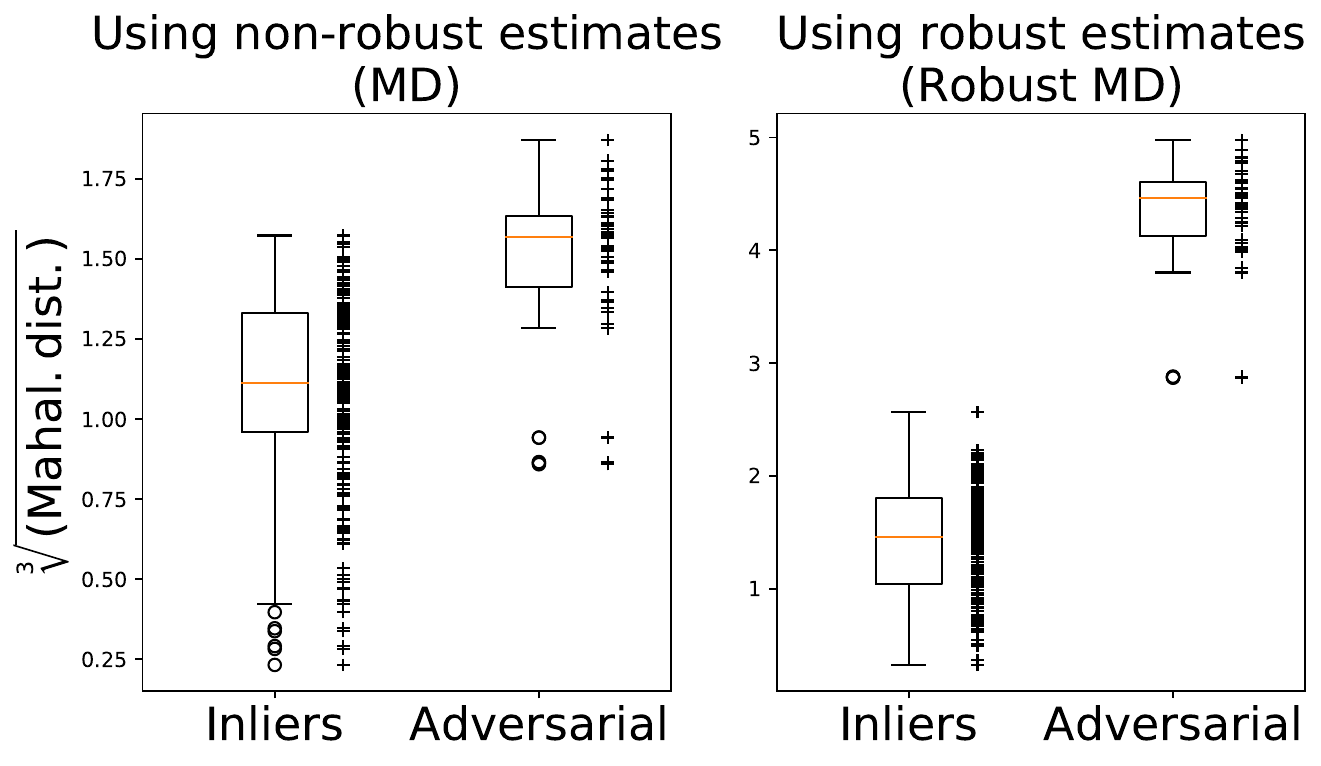}}
	\subfigure[OOD Outliers.]{\includegraphics[width=0.32\textwidth]{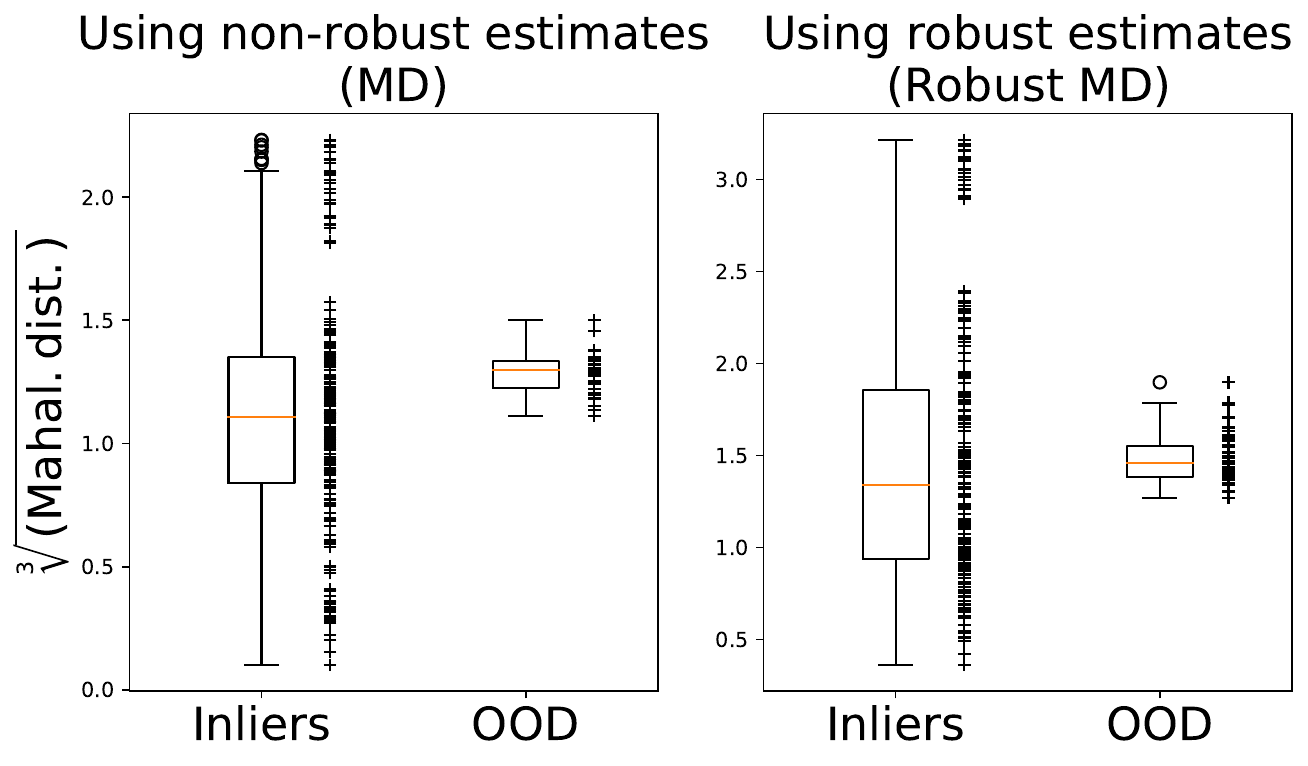}}
	\caption{Boxplot of distributions between inliers and three types of outliers in an action class on SpaceInvaders game.}
	\label{fig:boxplot_SpaceInvaders}
\end{figure*}

\begin{figure*}[b!]
	\centering
	\subfigure[Random Outliers.]{\includegraphics[width=0.32\textwidth]{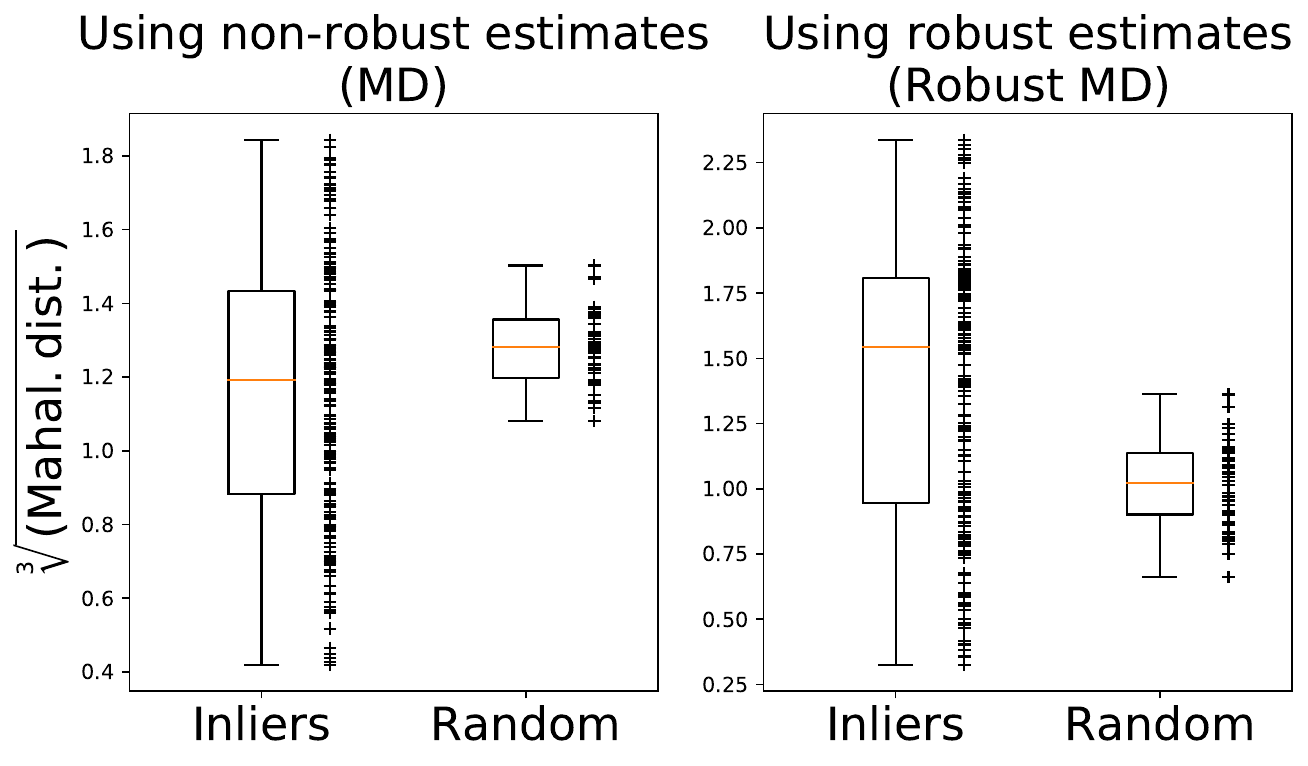}}
	\subfigure[Adversarial Outliers.]{\includegraphics[width=0.32\textwidth]{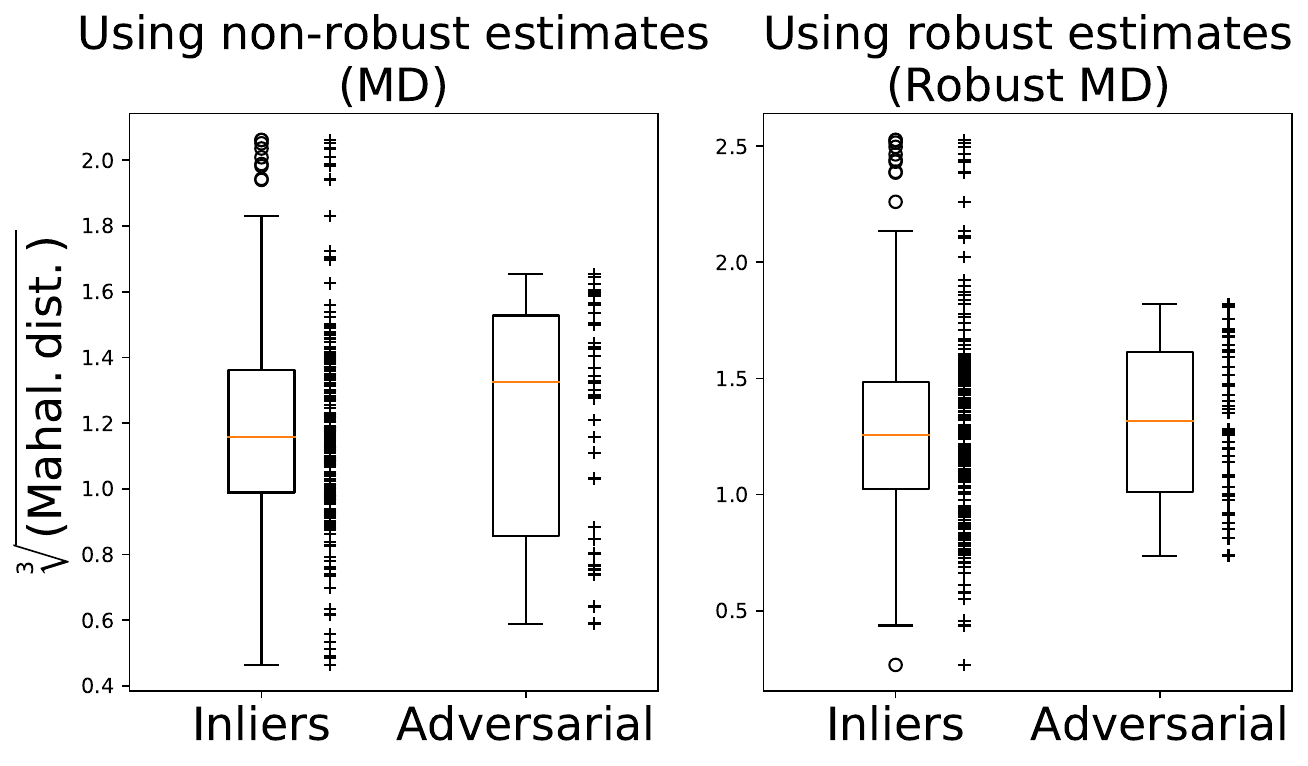}}
	\subfigure[OOD Outliers.]{\includegraphics[width=0.32\textwidth]{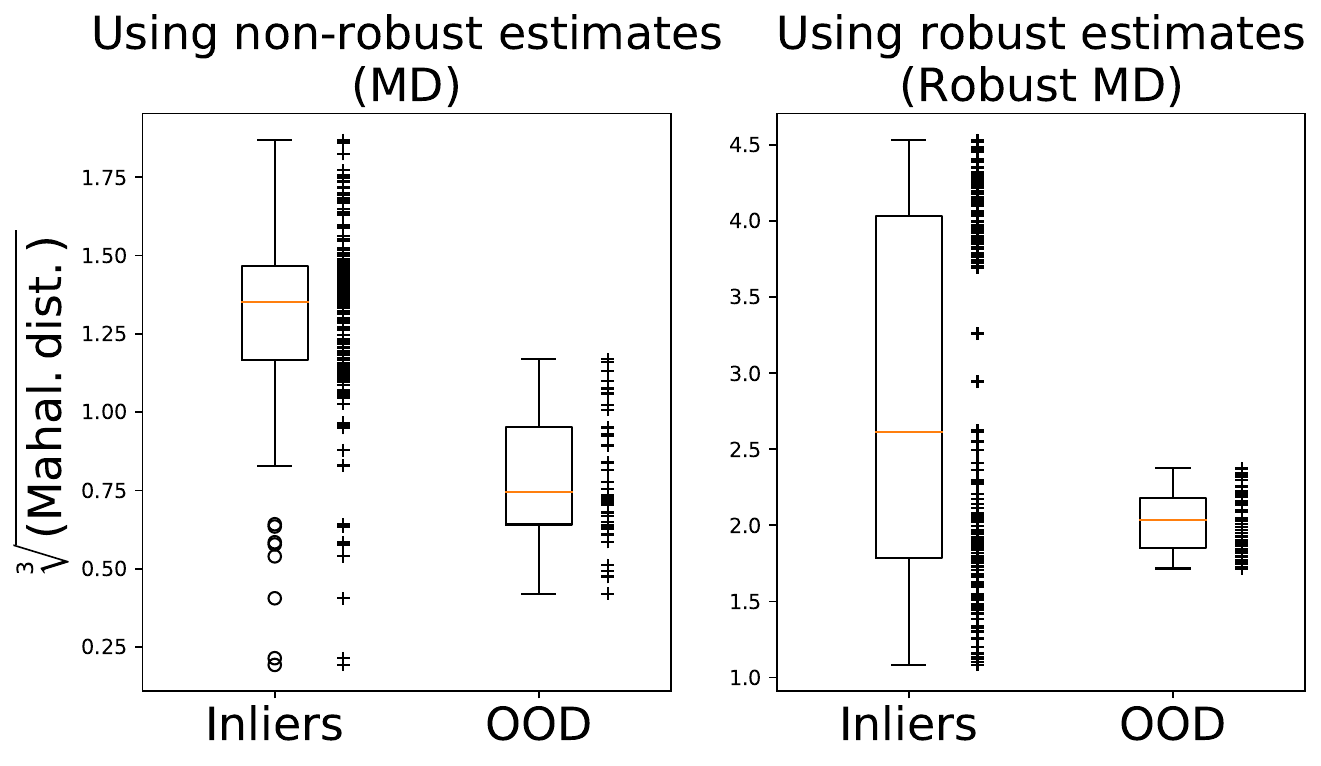}}
	\caption{Boxplot of distributions between inliers and three types of outliers in an action class on Asterix game.}
	\label{fig:boxplot_Asterix}
\end{figure*}

We plot the distributions of inliers and three types of outliers on SpaceInvaders and Asterix games in~\cref{fig:boxplot_SpaceInvaders,fig:boxplot_Asterix}, respectively. It is worth noting that Robust MD is also capable of enlarging the separation of distributions between inliers and both random and adversarial outliers on SpaceInvaders game, while its benefit seems to be negligible on OOD outliers~(Breakout) on SpaceInvaders games as well as in Asterix game. We speculate that it is determined by the game's difficulty. Specifically, the PPO algorithm can achieve desirable performance on the simple Breakout game, thus yielding informative feature space vectors. By contrast, there is room for the generalization of PPO on both SpaceInvaders and Asterix games, such that Robust MD might not help when handling the less meaningful state feature vectors in these two games.

\subsection{Sensitivity Analysis}
\label{appendix:sensitivity}

\begin{figure*}[htbp] \centering\includegraphics[width=0.7\textwidth,trim=10 10 10 10,clip]{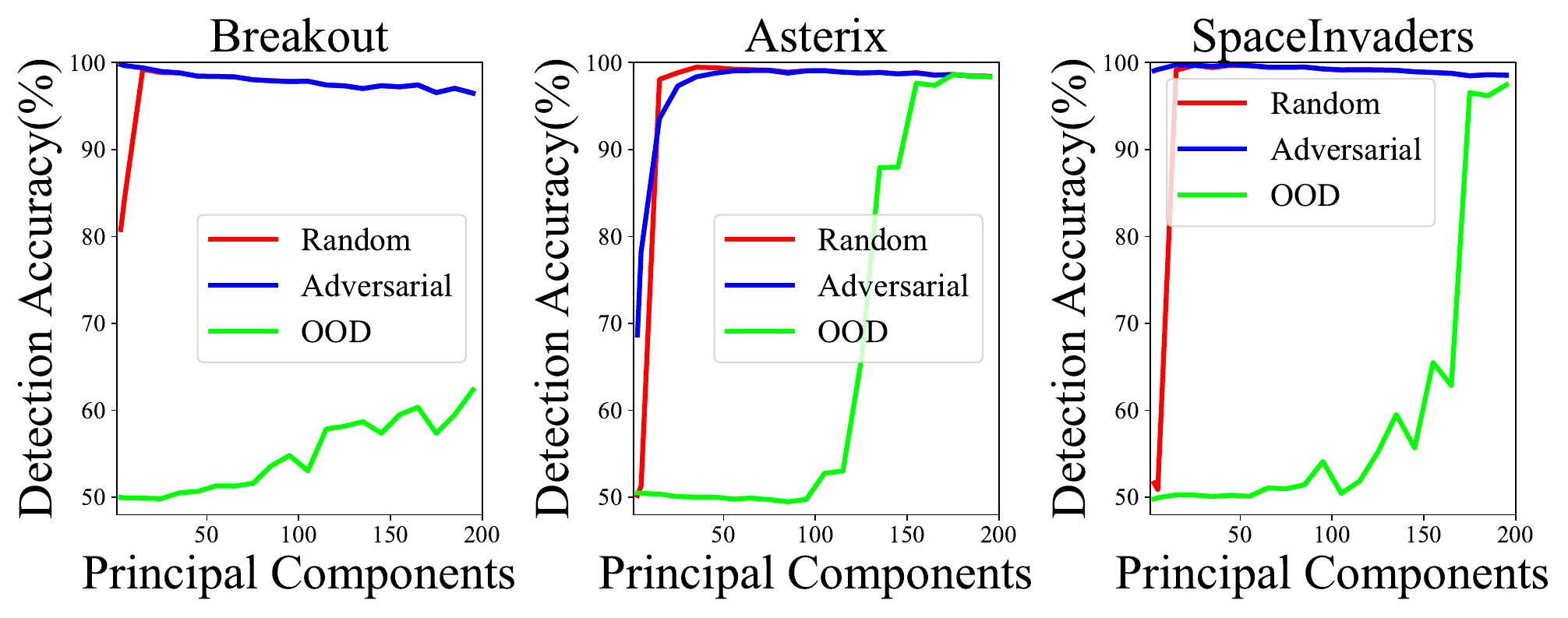}
	% \vskip -0.1in	
	\caption{Detection performance under \textbf{Robust MD} as the number of principal components increases.}
	\label{fig:sensitivity_RMD}
\end{figure*}

We provide the sensitivity analysis of Robust MD in terms of the PCA dimension in \cref{fig:sensitivity_RMD}. The impact of the number of principal components on the detection performance for robust MD detection is shown in \cref{fig:sensitivity_RMD}. The detection accuracy over all considered outliers improves as the number of principal components increases, except for a slight decline for random and adversarial outliers~(red and blue lines) on the Breakout game. The increase implies that the subspace spanned by principal components with small explained variance also contains valuable information for detecting anomalous states from in-distribution states, which coincides with the conclusion in~\citep{kamoi2020mahalanobis}. 

\begin{figure*}[htbp]
\centering\includegraphics[width=0.7\textwidth,trim=10 10 10 10,clip]{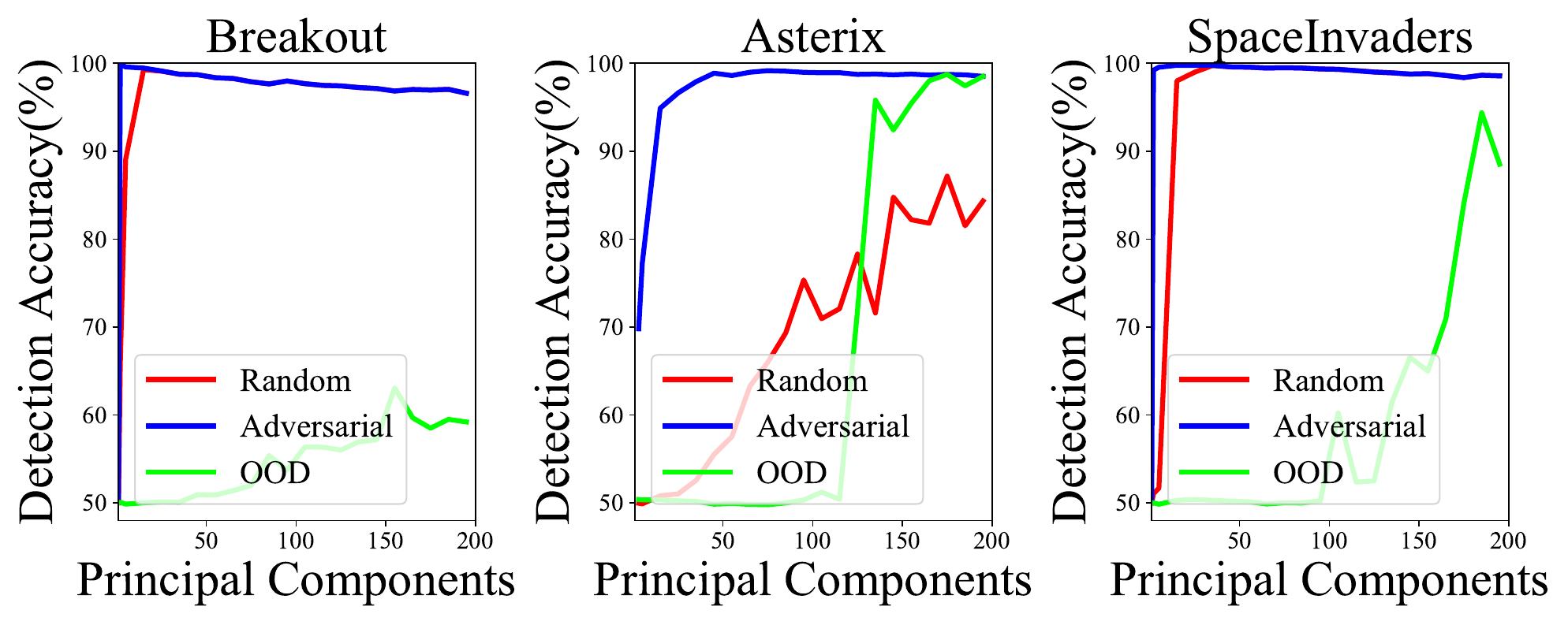}
	% \vskip -0.1in
	\caption{Detection performance under \textbf{MD} as the number of principal components increases.}
	\label{fig:sensitivity_MD}
\end{figure*}

The result of MD estimation manifests in \cref{fig:sensitivity_MD}. It suggests that there is still an ascending tendency of detection accuracy as the number of principal components increases.

\clearpage
\section{Results in Online Setting}

\subsection{Setup and Full Main Results}\label{appendix:Training Phase Main Results}

\begin{figure*}[htbp]
	\centering
	\subfigure{\includegraphics[width=0.2\textwidth]{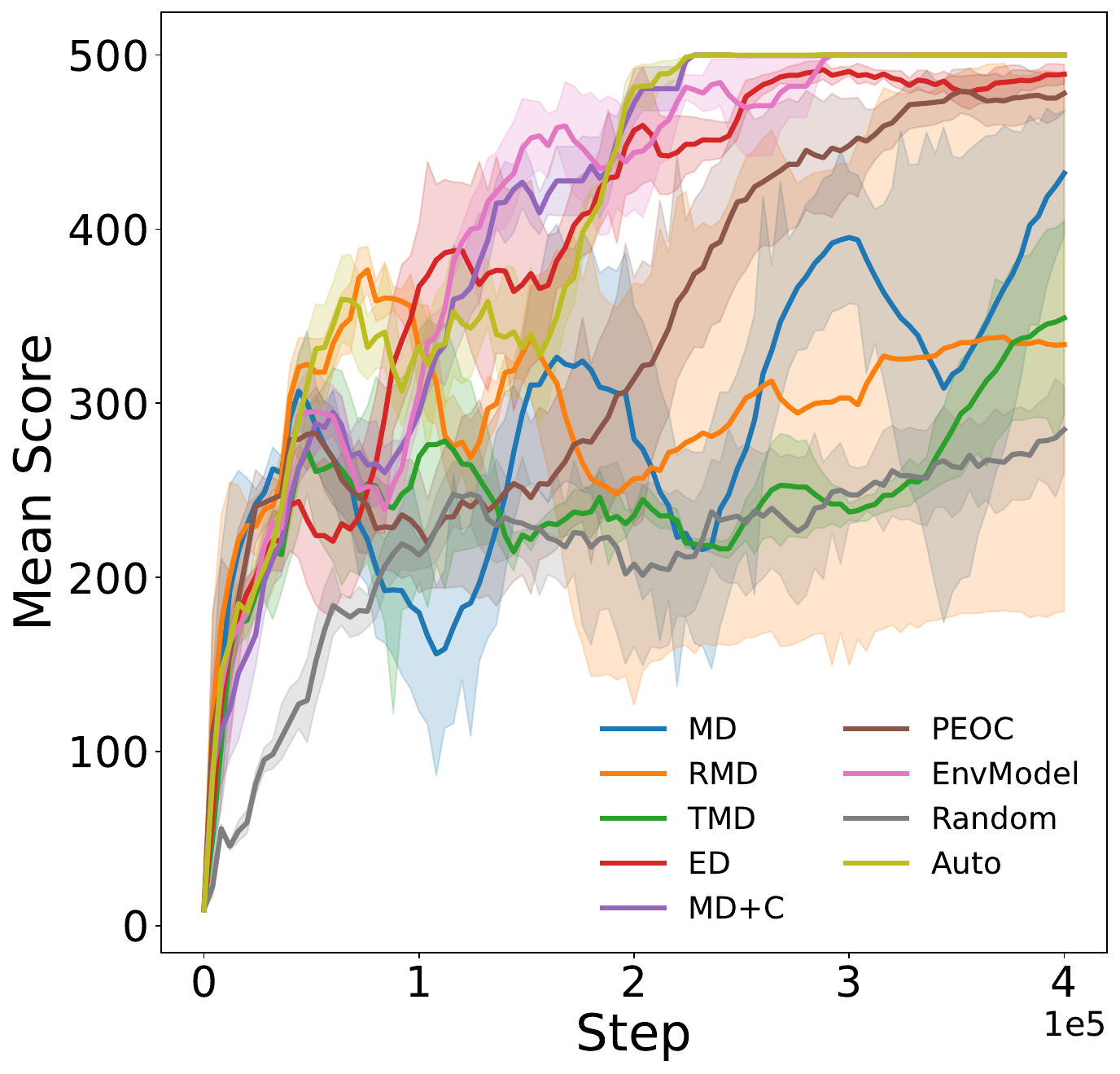}}
	\subfigure{\includegraphics[width=0.2\textwidth]{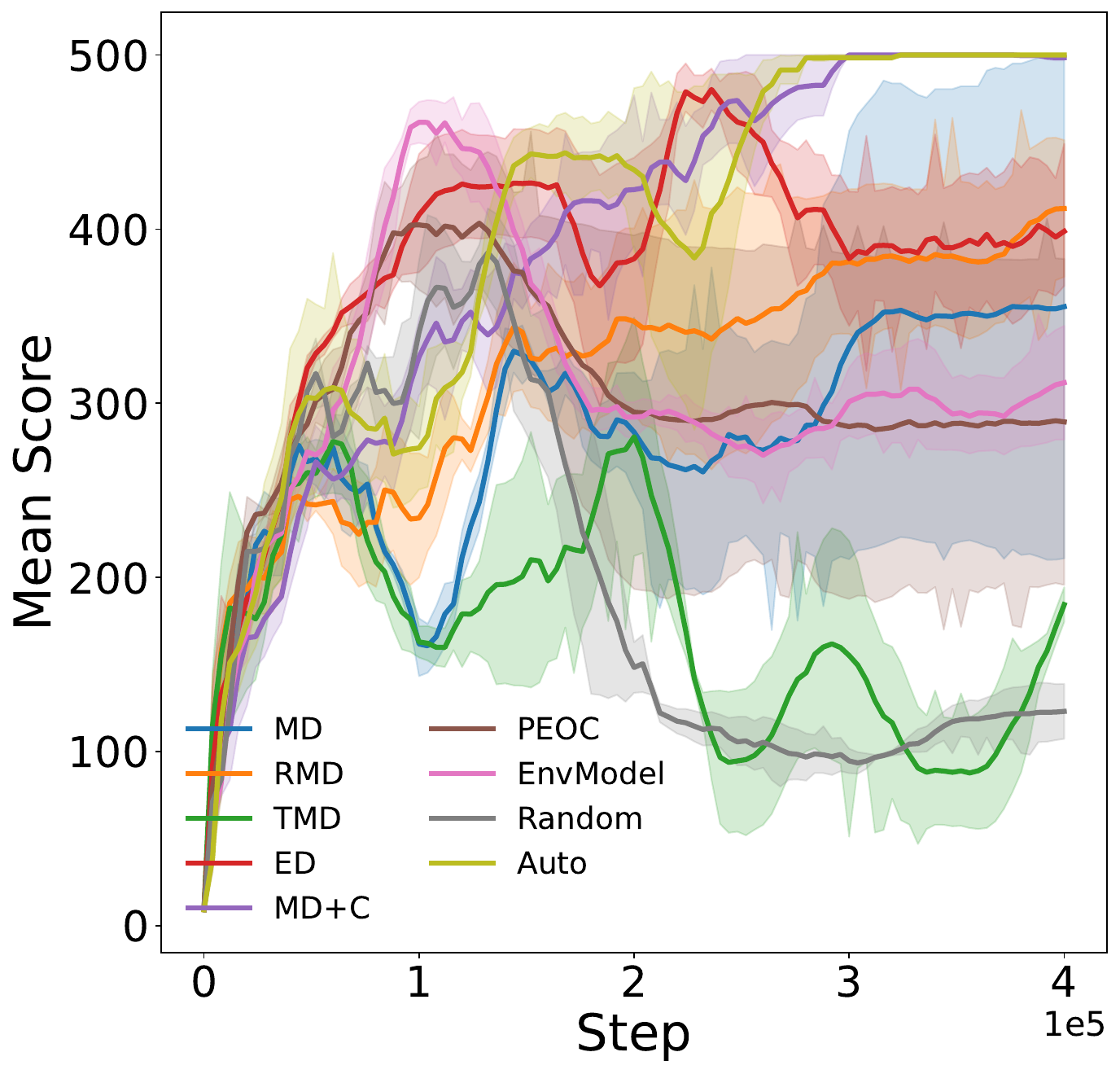}}
	\subfigure{\includegraphics[width=0.2\textwidth]{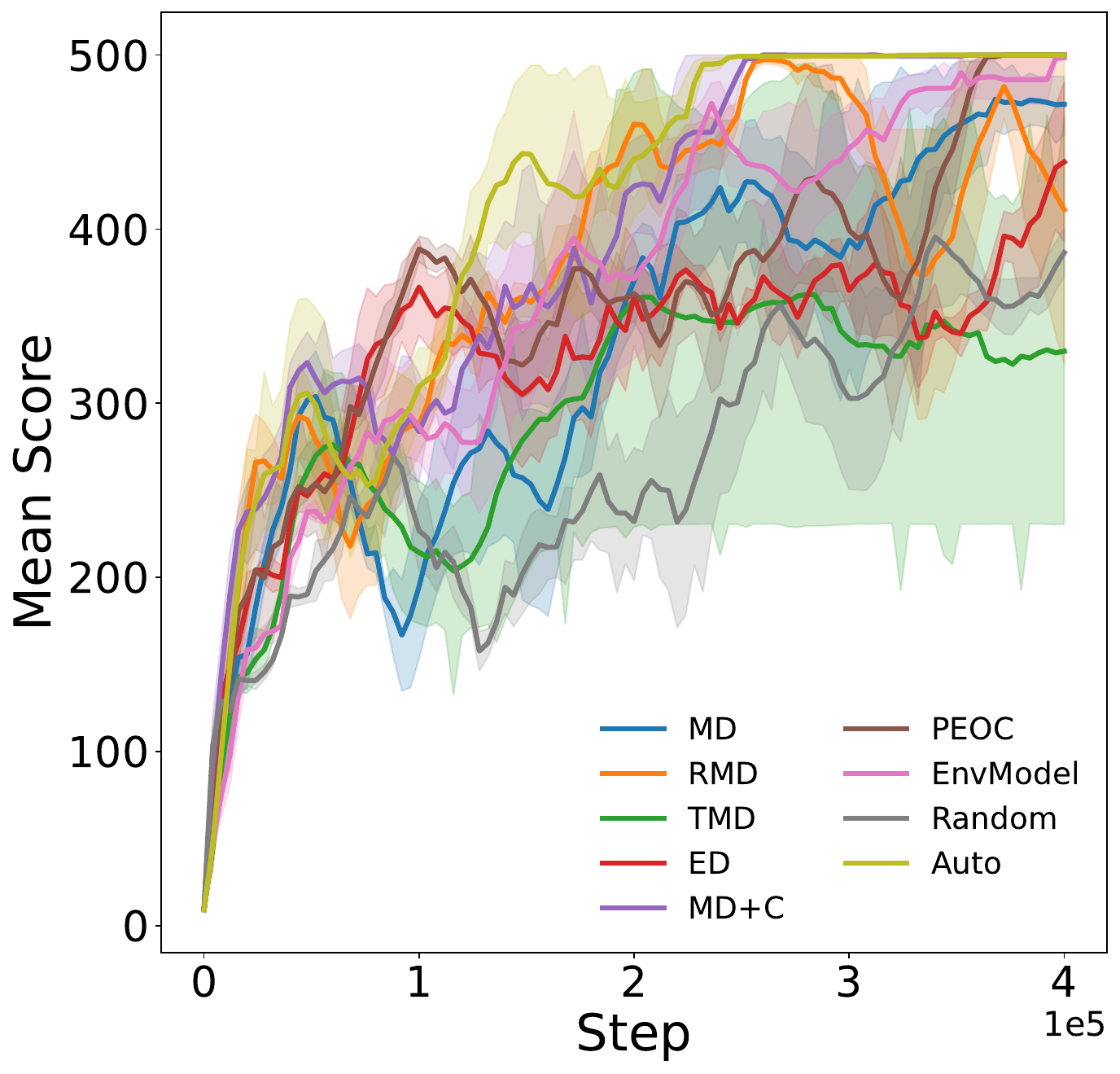}}
	\subfigure{\includegraphics[width=0.2\textwidth]{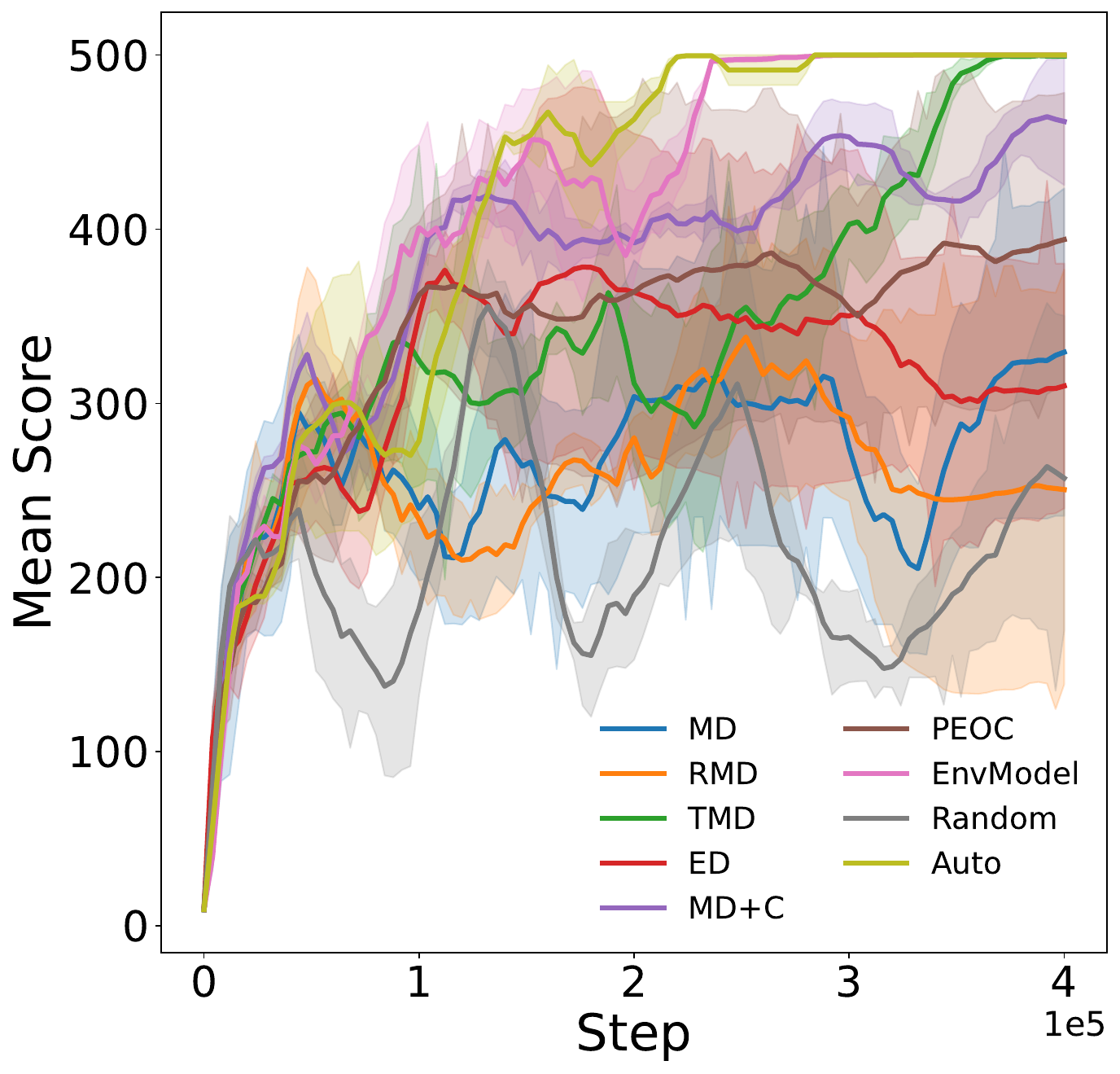}}
    \setcounter{subfigure}{0}
	\subfigure{\includegraphics[width=0.2\textwidth]{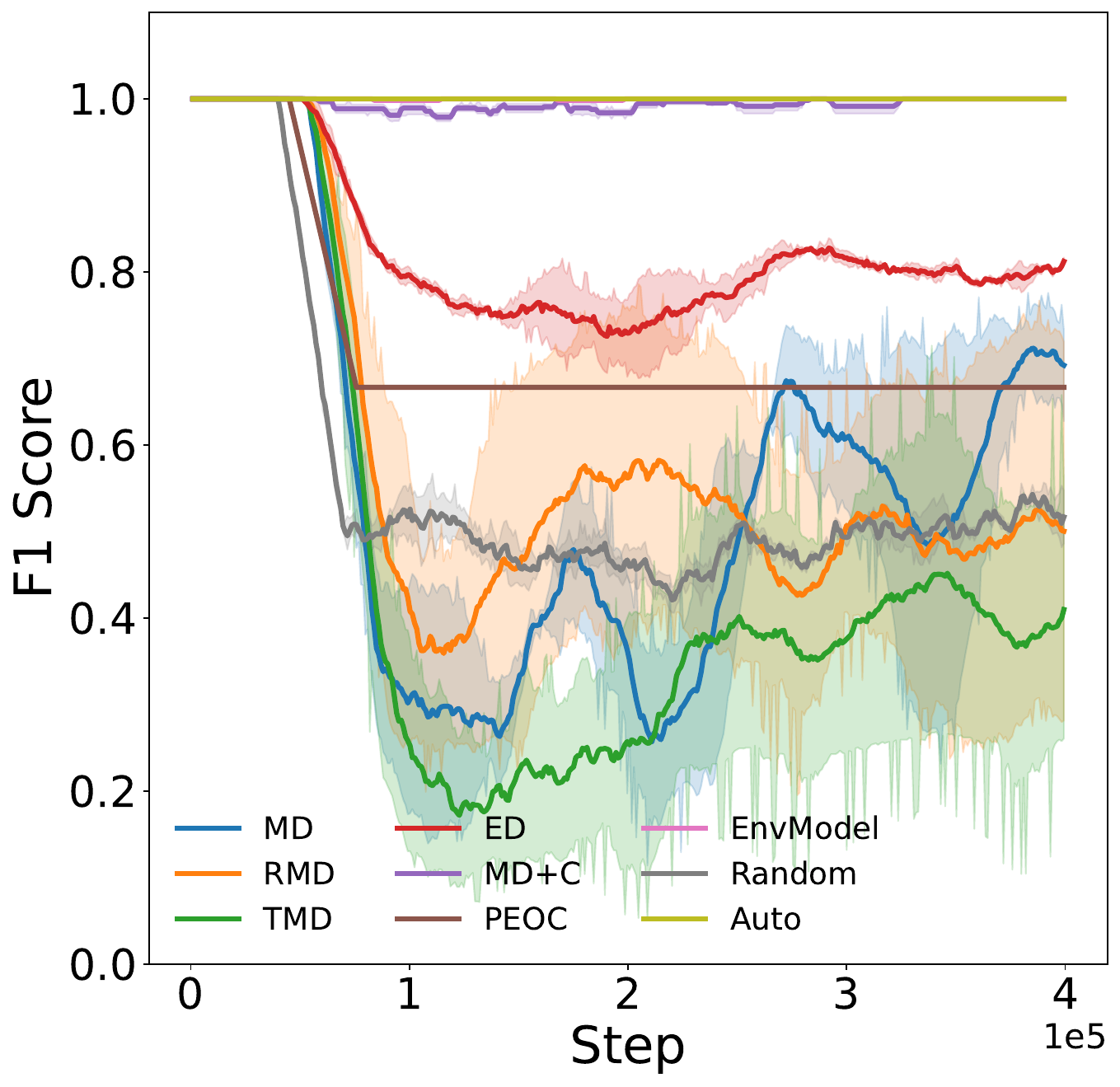}}
	\subfigure{\includegraphics[width=0.2\textwidth]{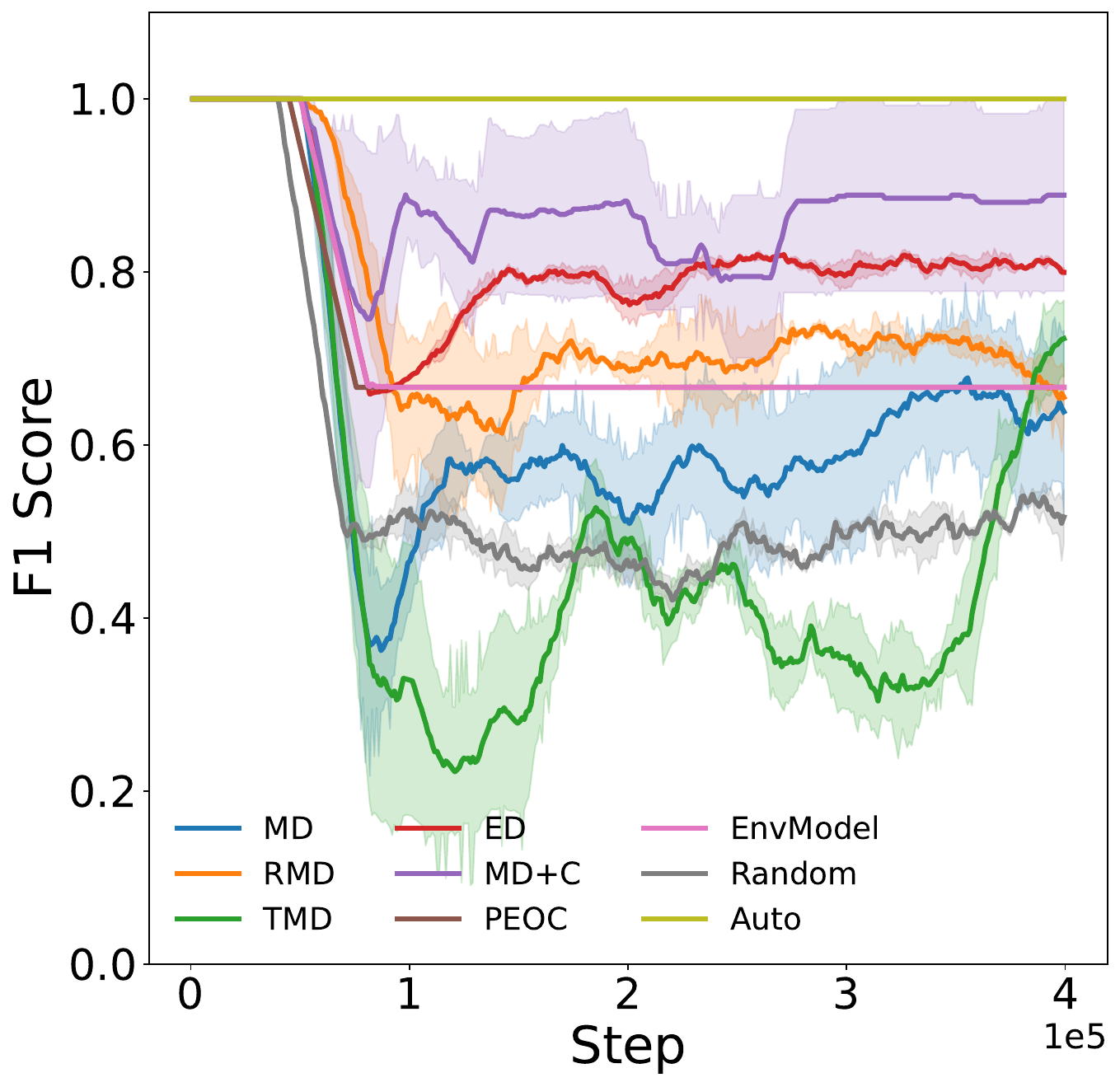}}
	\subfigure{\includegraphics[width=0.2\textwidth]{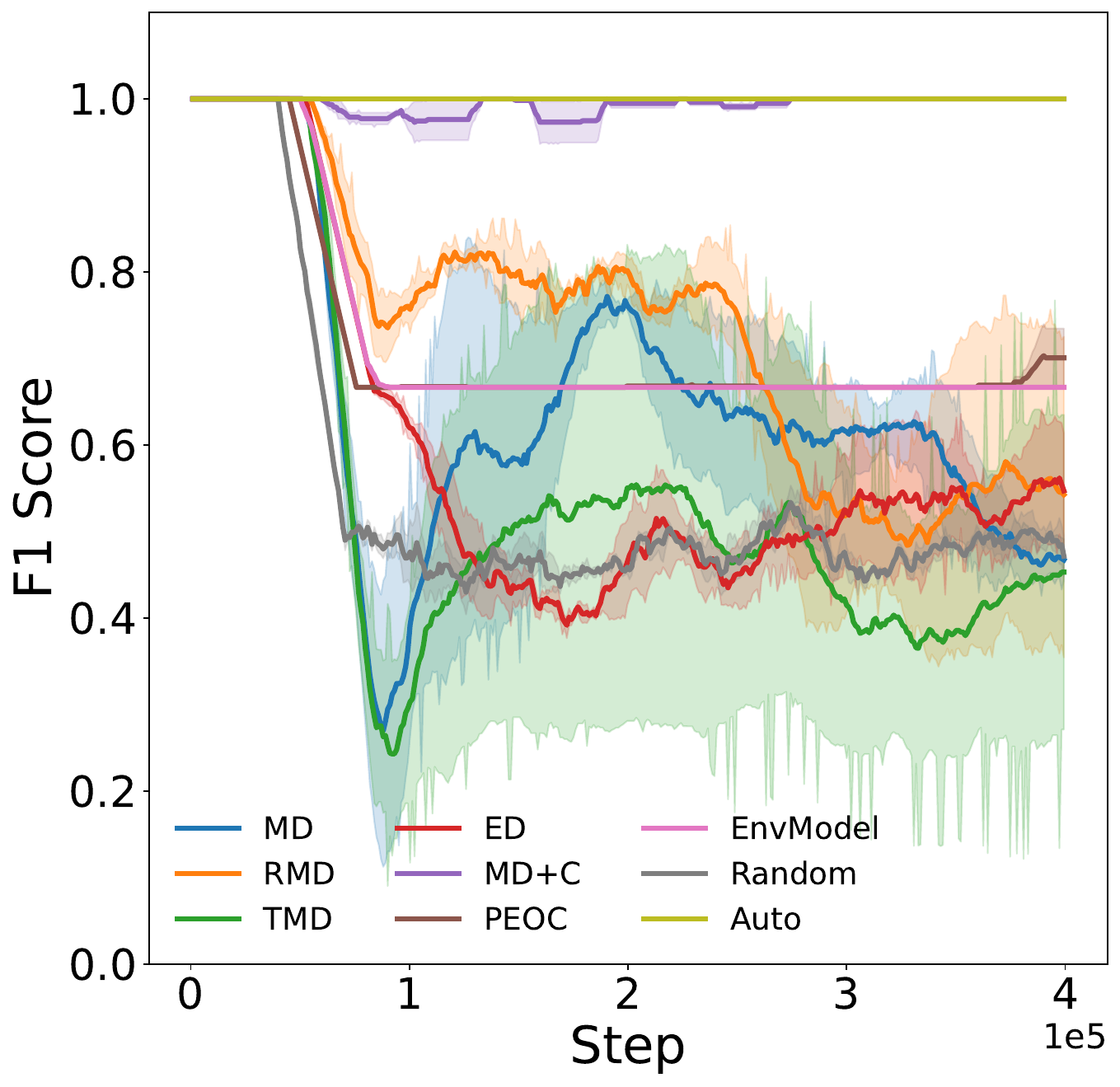}}
	\subfigure{\includegraphics[width=0.2\textwidth]{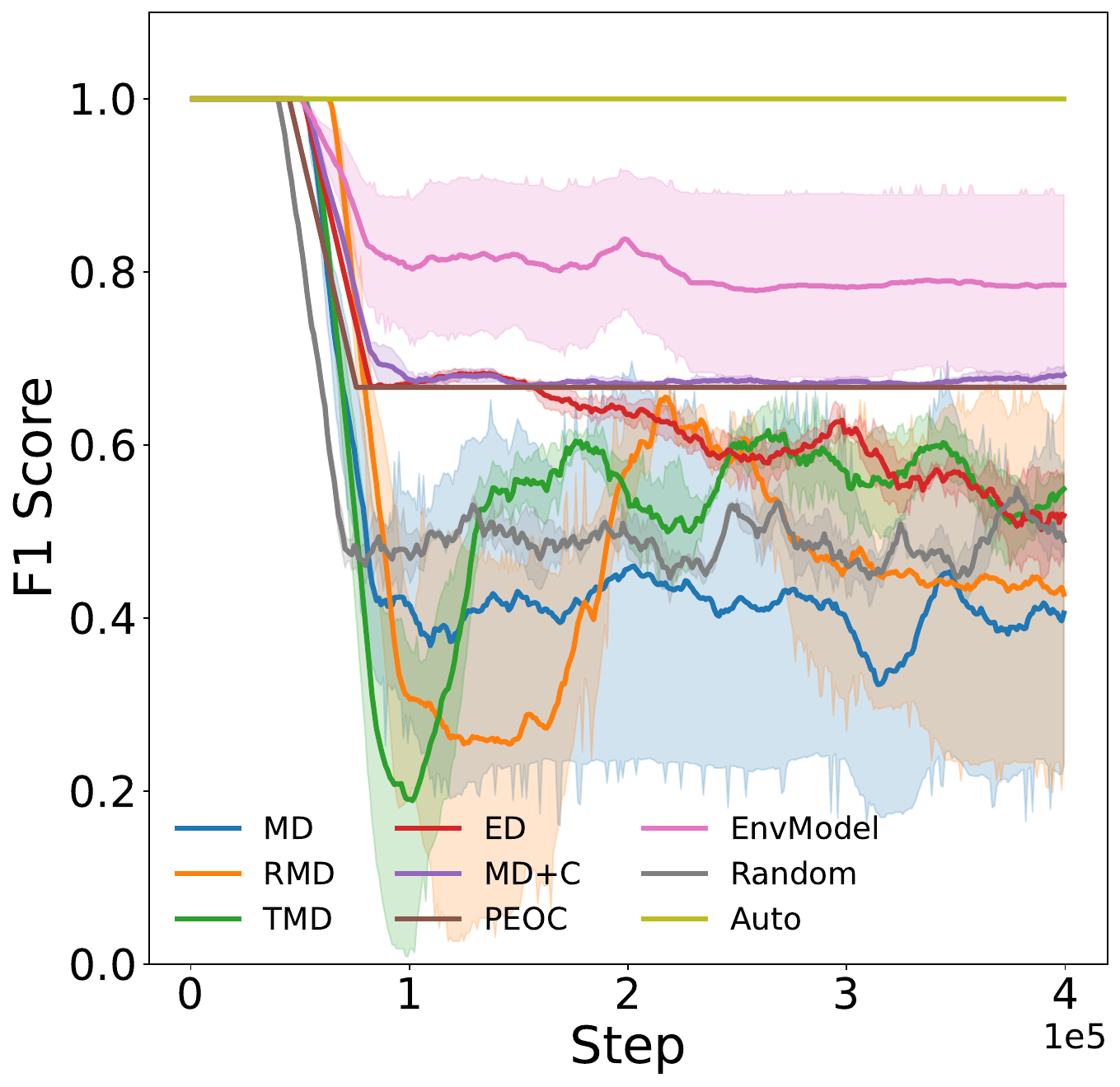}}
     \setcounter{subfigure}{0}
	\subfigure[Gaussian std=1]{\includegraphics[width=0.2\textwidth]{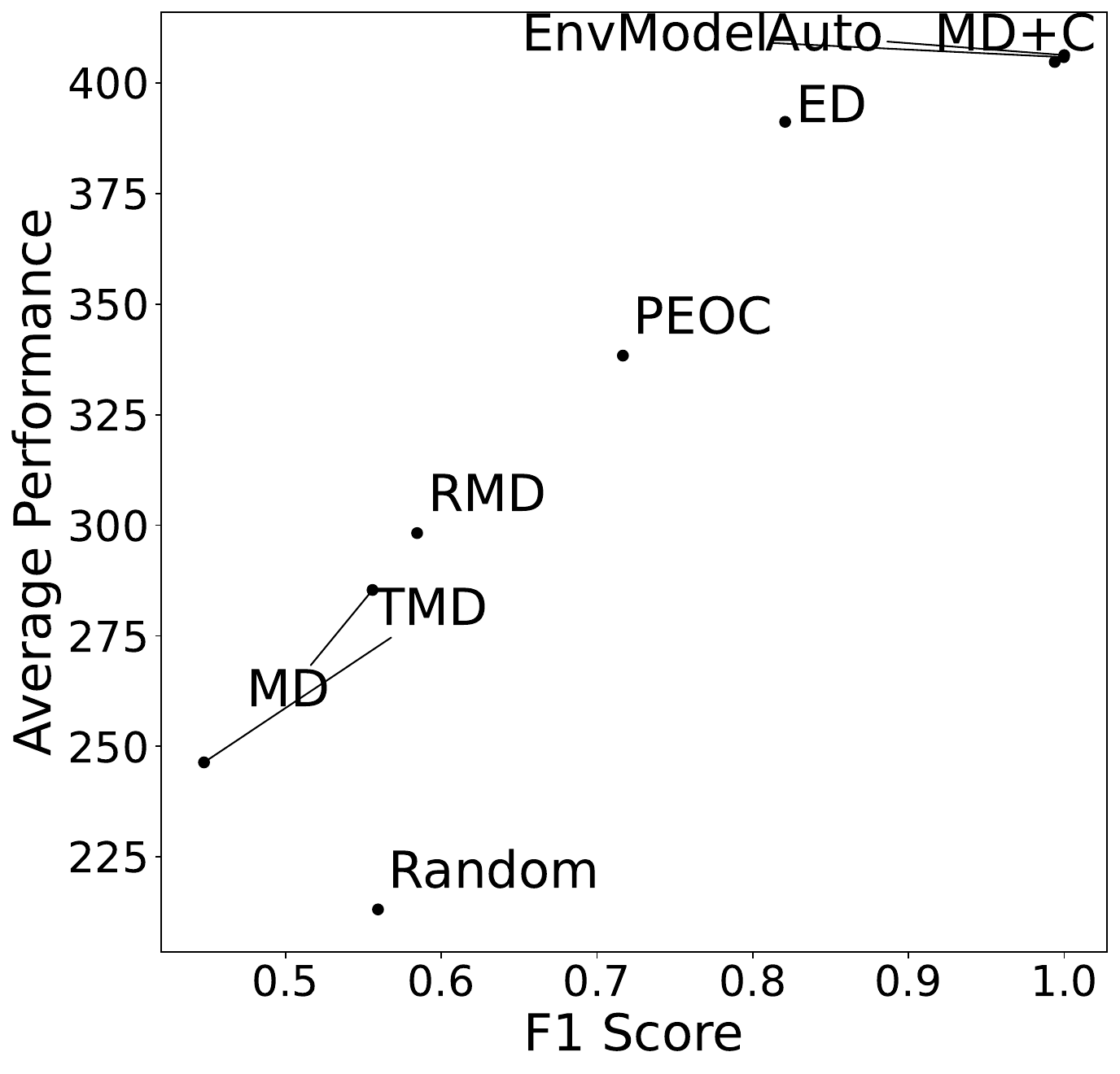}}
	\subfigure[Gaussian std=0.3]{\includegraphics[width=0.2\textwidth]{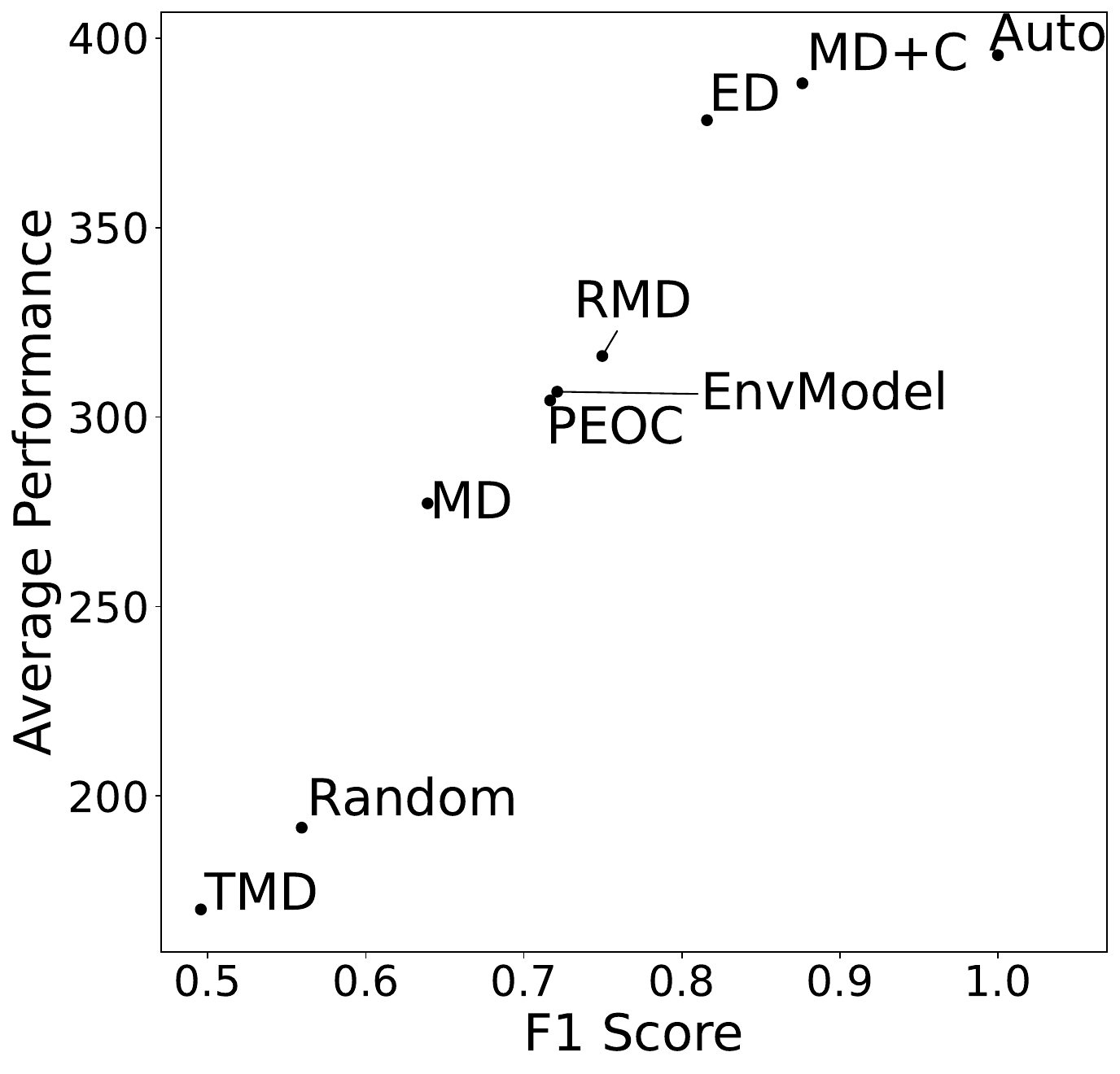}}
	\subfigure[OOD MountainCar]{\includegraphics[width=0.2\textwidth]{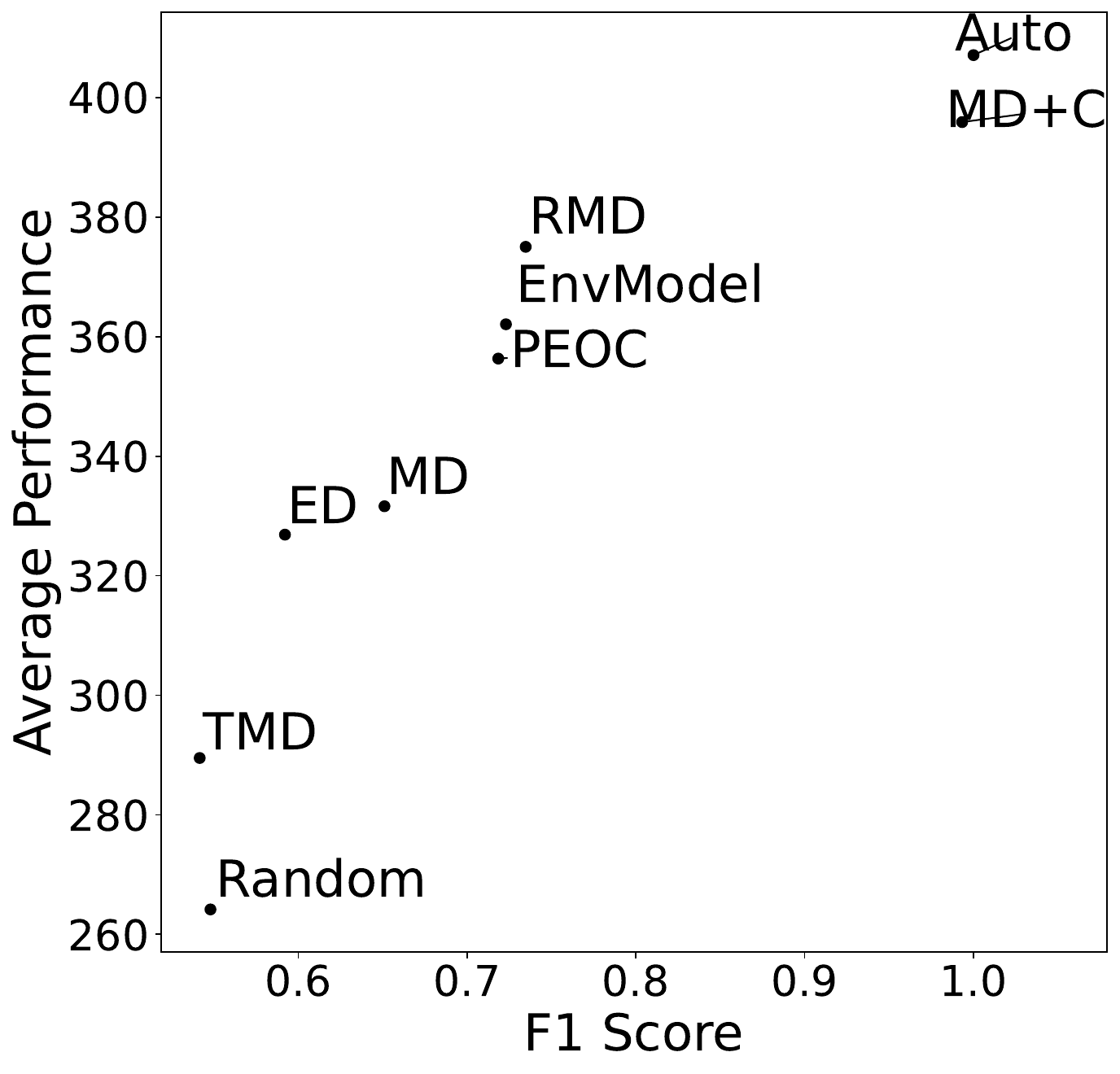}}
	\subfigure[Adversarial]{\includegraphics[width=0.2\textwidth]{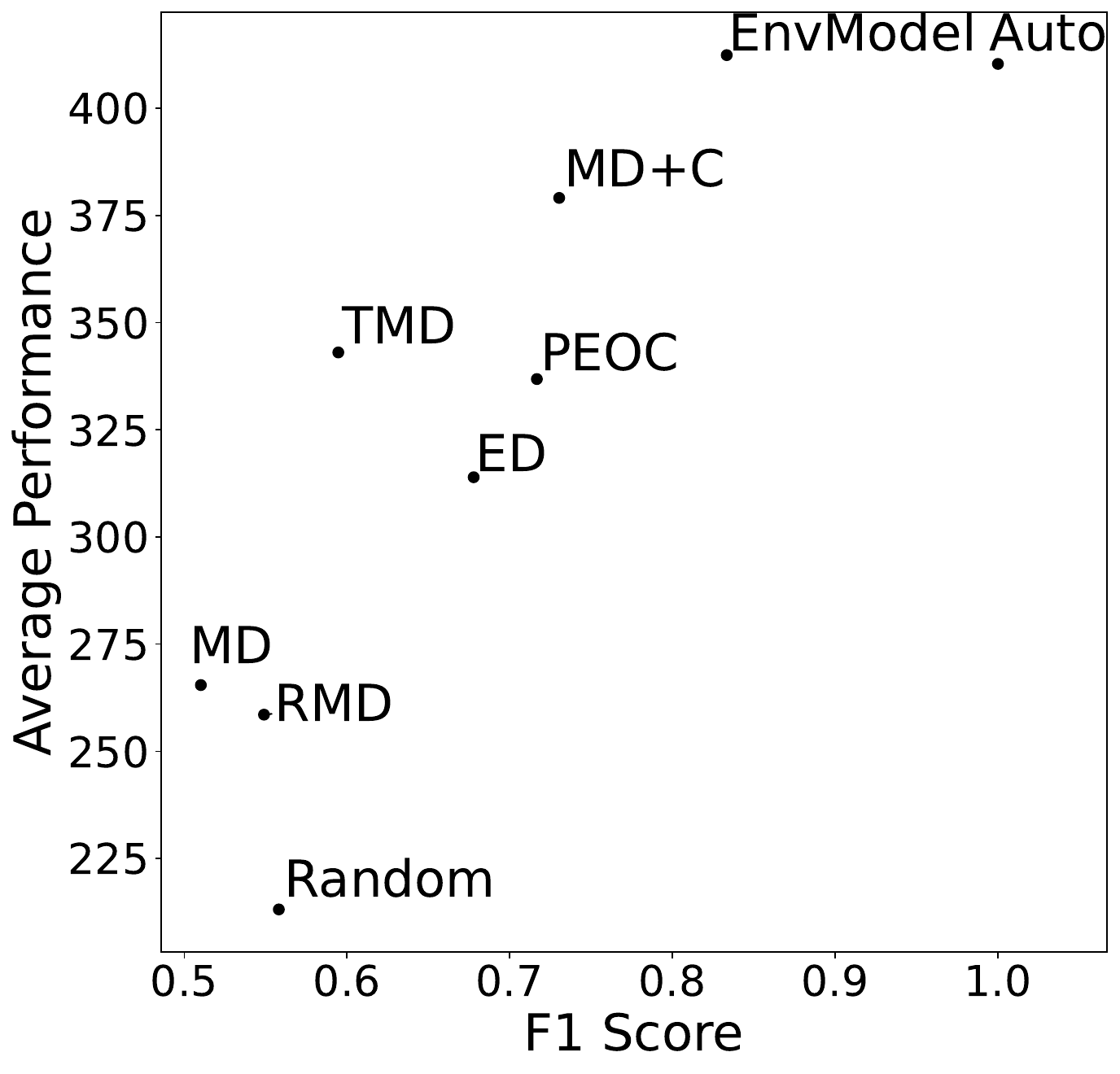}}
	\caption{\hongming{Detection performance across various state outliers in the online training on CartPole.}}
	\label{fig:CartPole_online_full}
\end{figure*}

\begin{figure*}[b]
	\centering
	\subfigure{\includegraphics[width=0.2\textwidth]{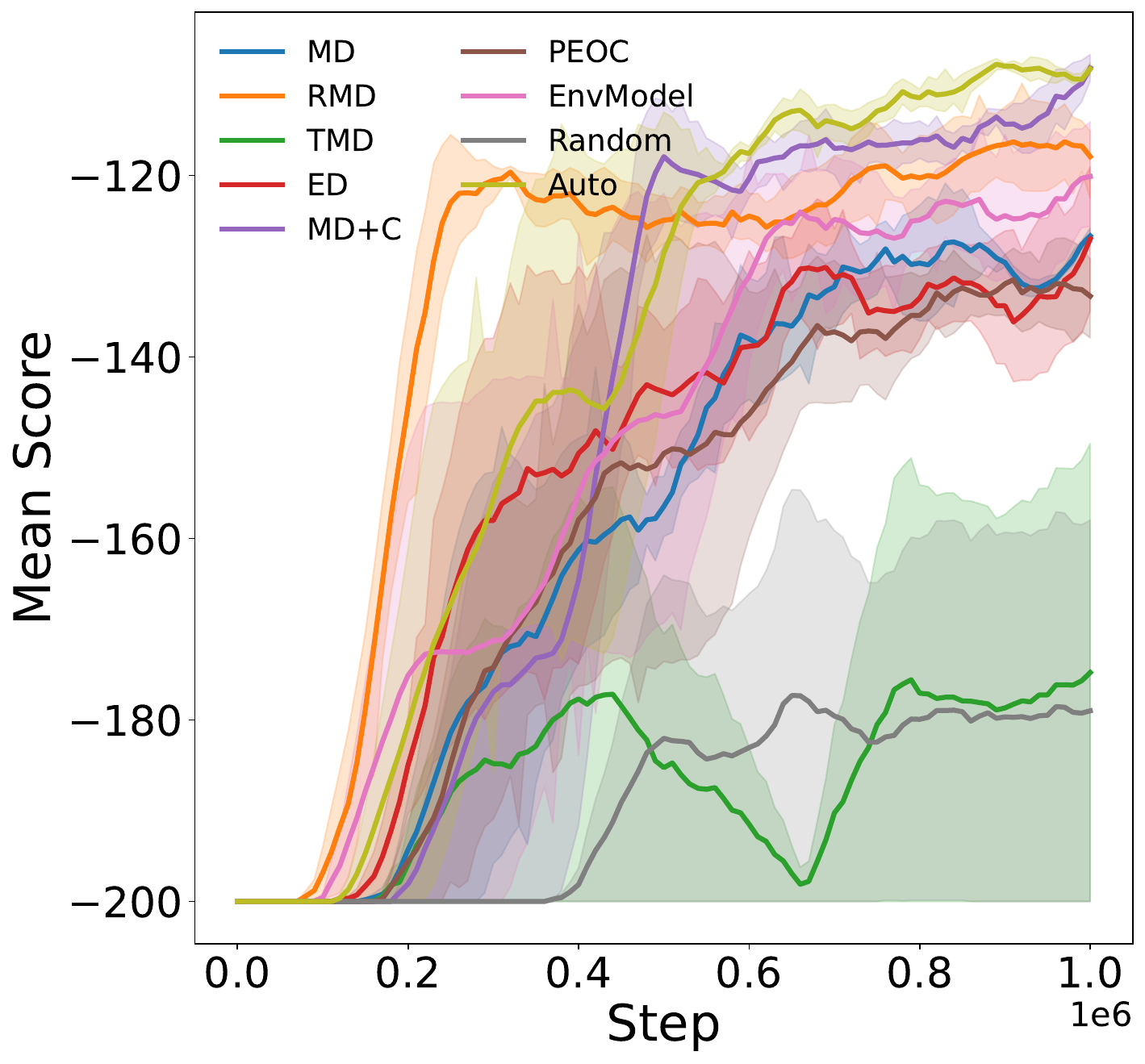}}
	\subfigure{\includegraphics[width=0.2\textwidth]{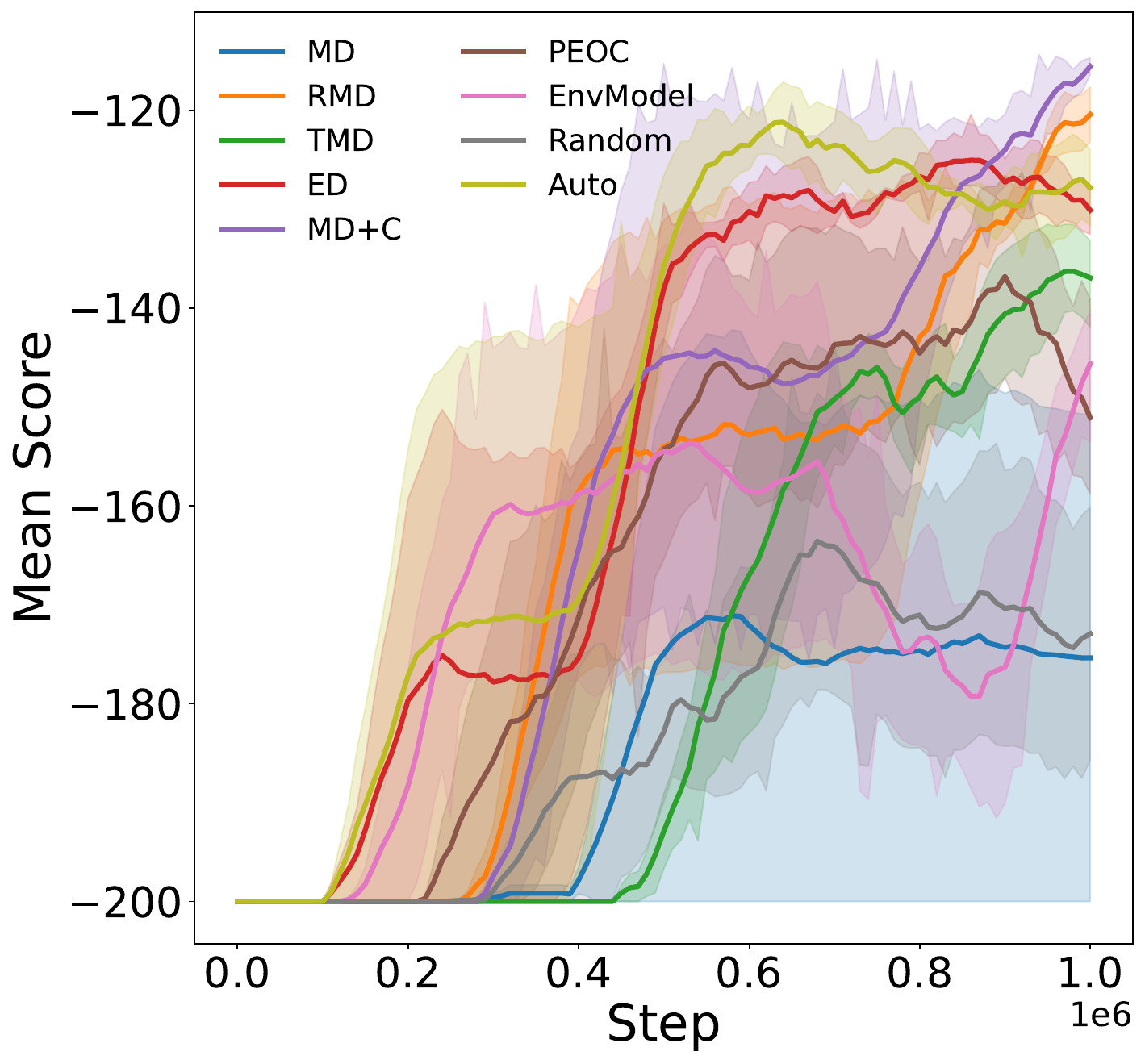}}
	\subfigure{\includegraphics[width=0.2\textwidth]{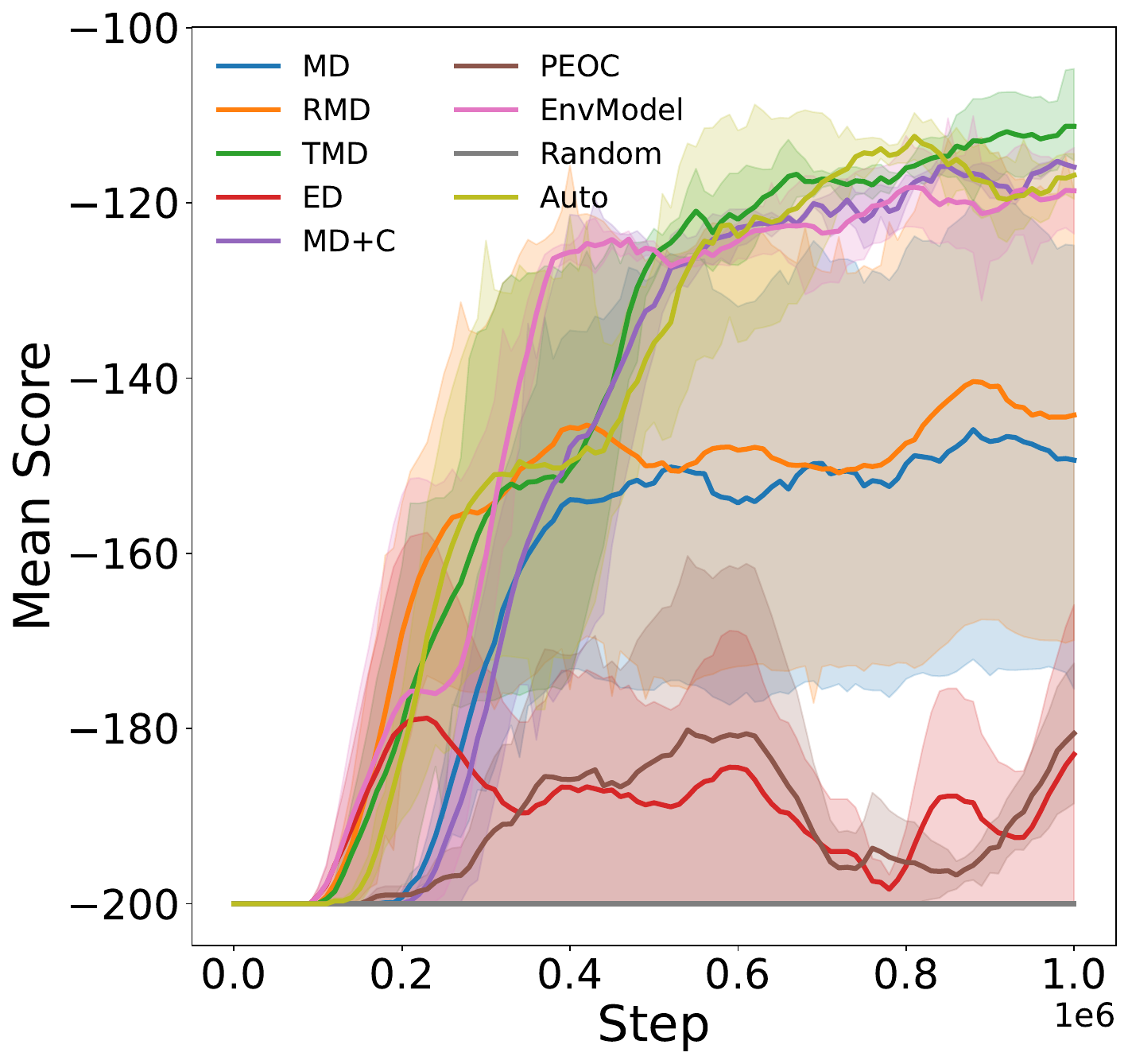}}
	\subfigure{\includegraphics[width=0.2\textwidth]{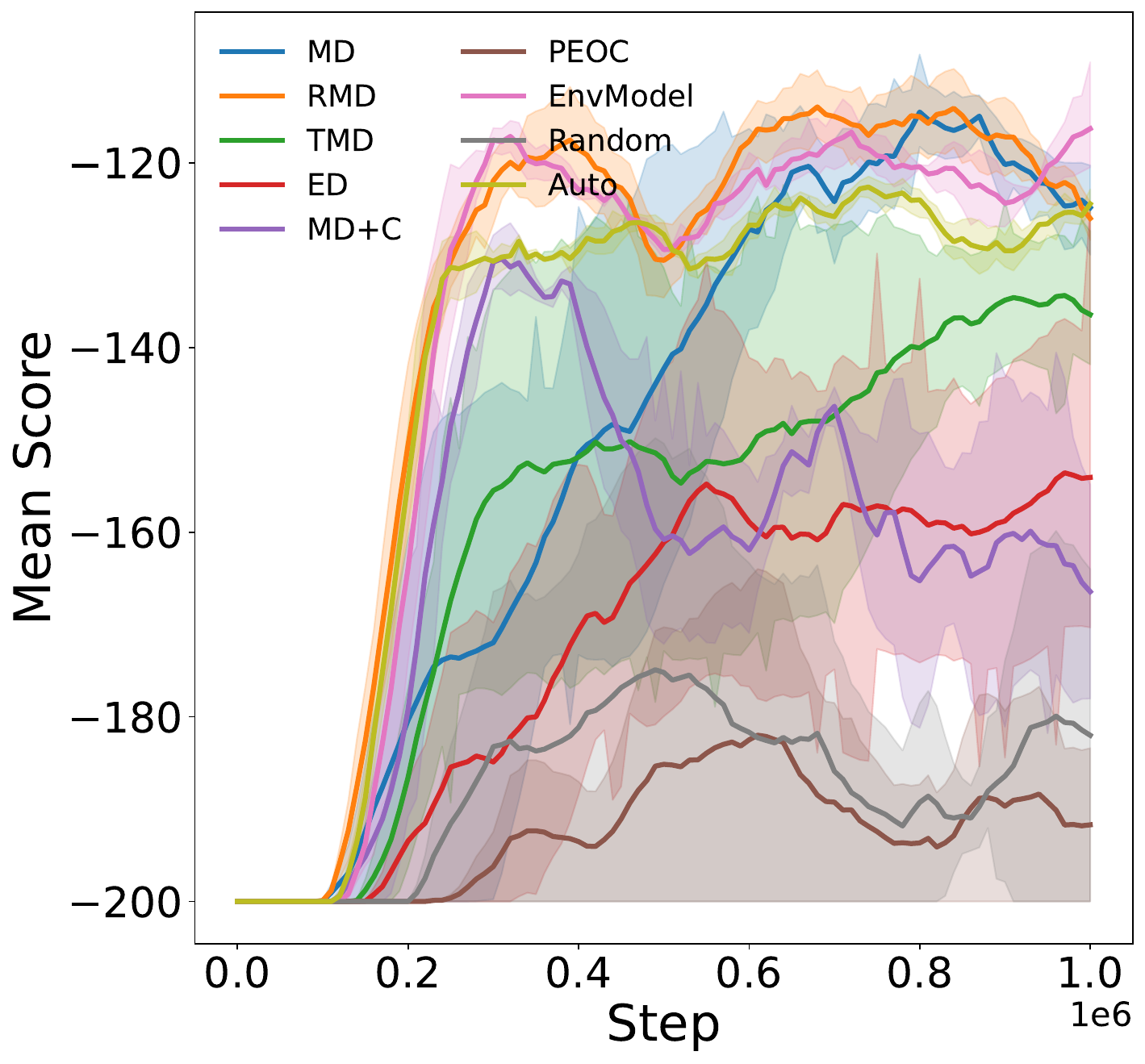}}
    \setcounter{subfigure}{0}
	\subfigure{\includegraphics[width=0.2\textwidth]{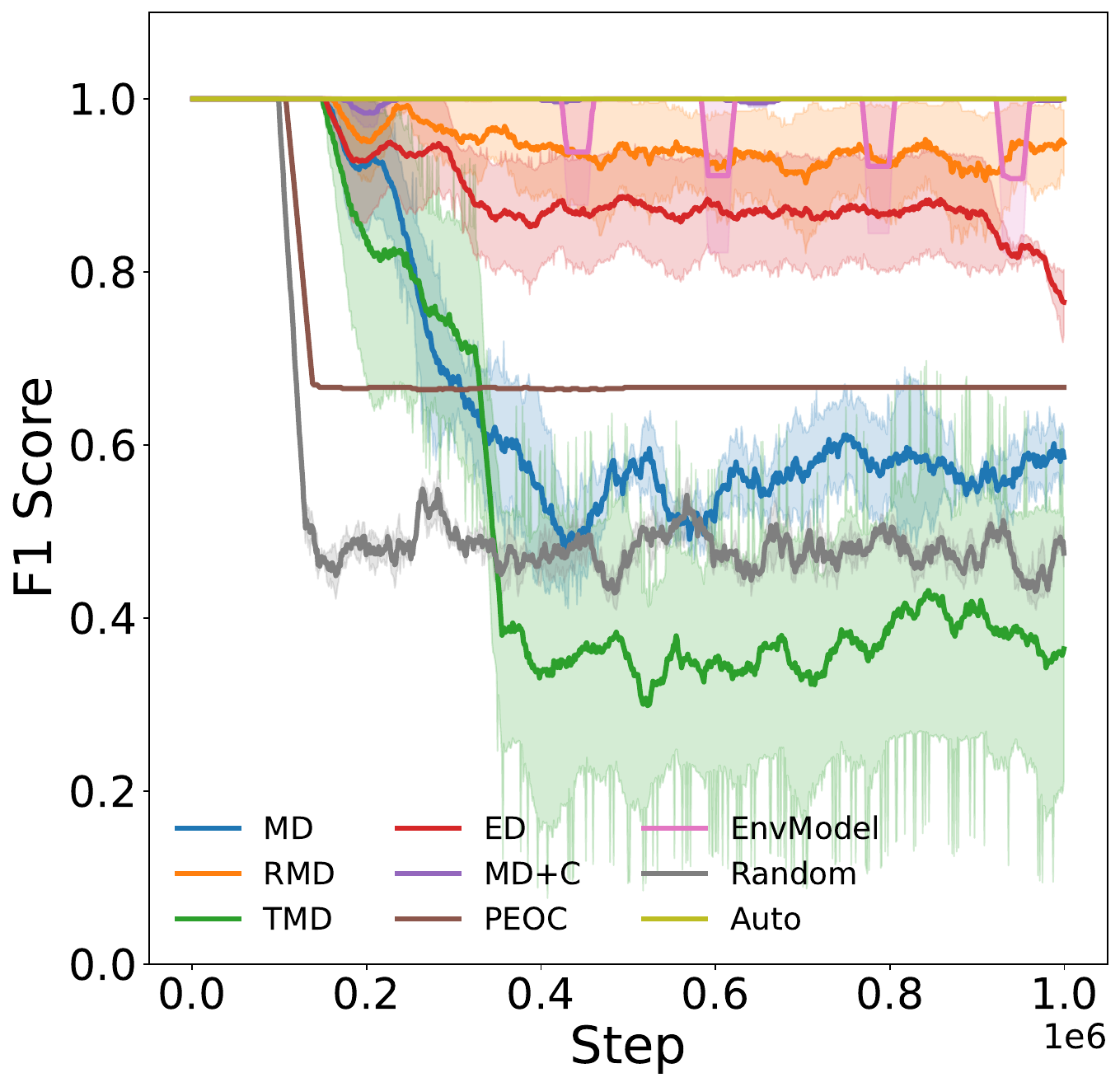}}
	\subfigure{\includegraphics[width=0.2\textwidth]{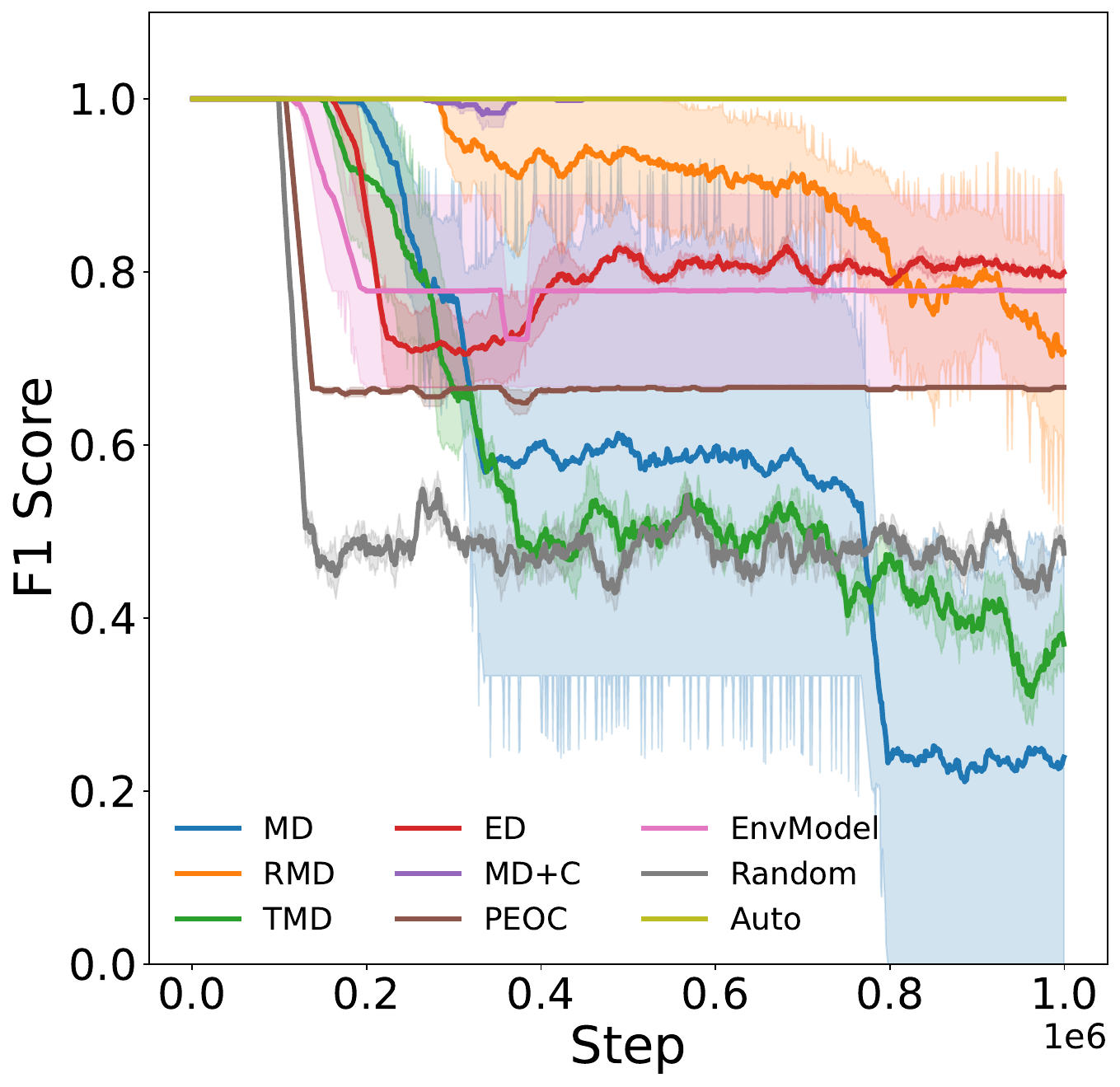}}
	\subfigure{\includegraphics[width=0.2\textwidth]{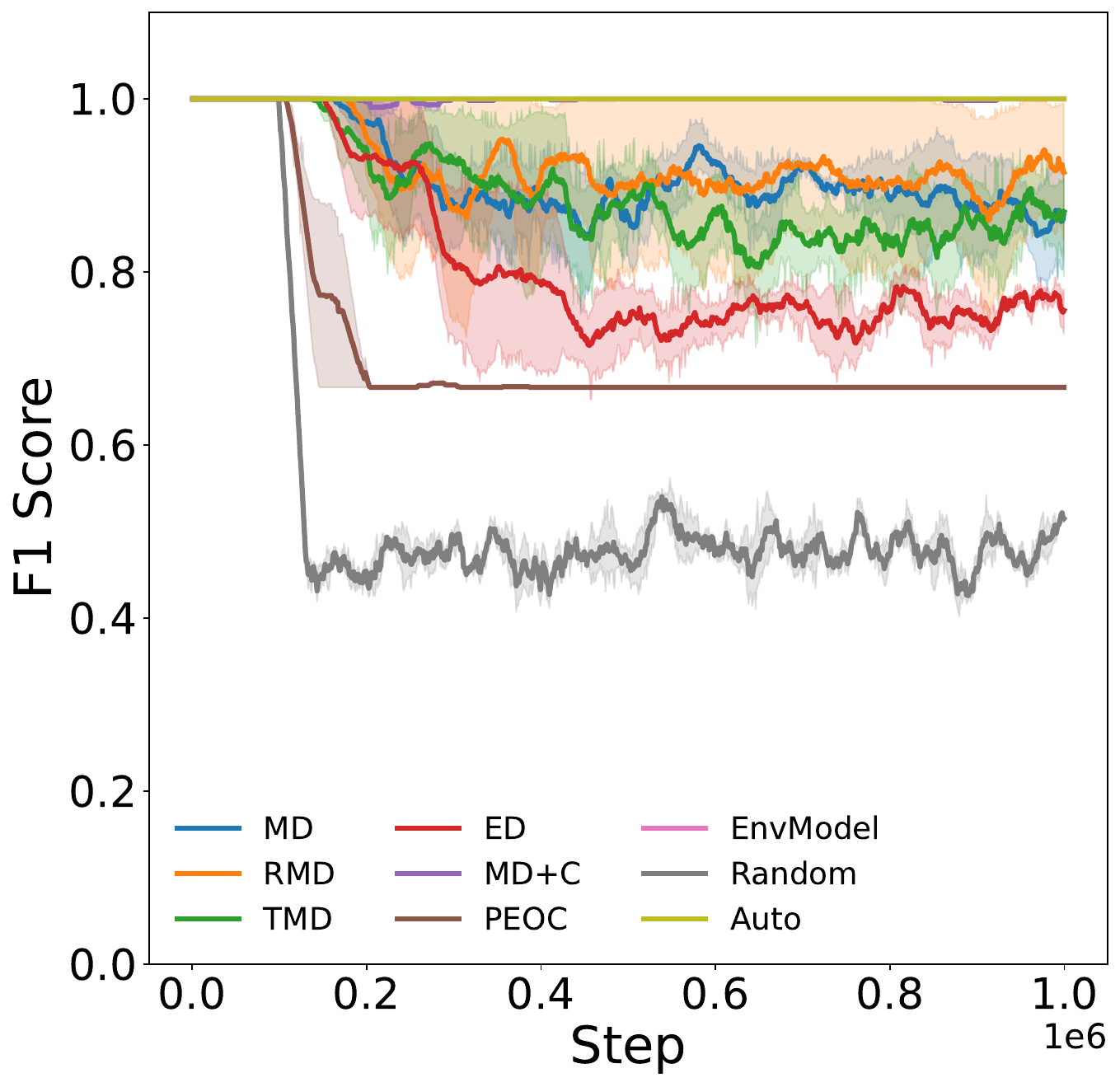}}
	\subfigure{\includegraphics[width=0.2\textwidth]{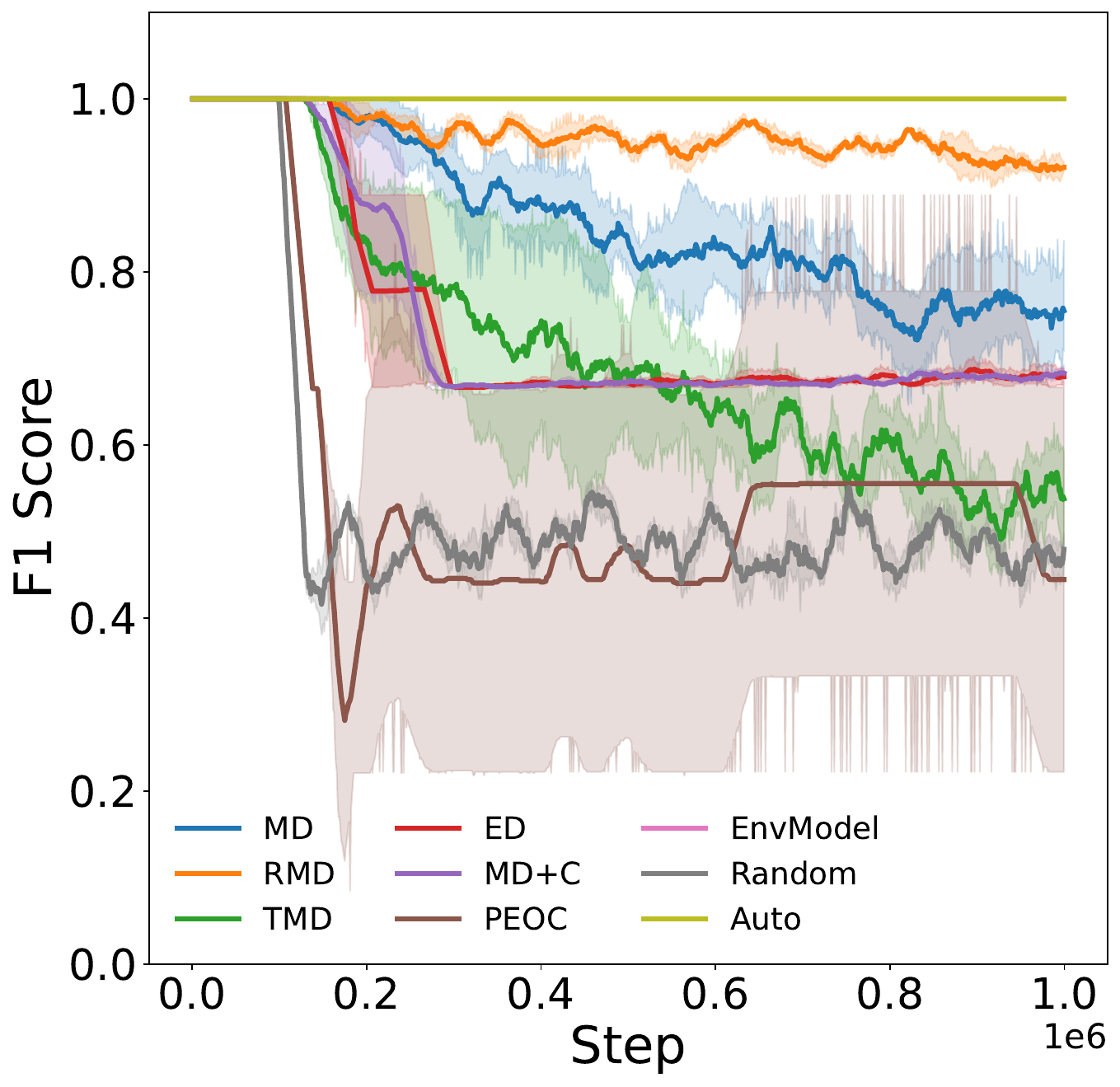}}
     \setcounter{subfigure}{0}
	\subfigure[Gaussian std=1]{\includegraphics[width=0.2\textwidth]{pic/relation_MountainCar_5.pdf}}
	\subfigure[Gaussian std=0.3]{\includegraphics[width=0.2\textwidth]{pic/relation_MountainCar_6.pdf}}
	\subfigure[OOD CartPole]{\includegraphics[width=0.2\textwidth]{pic/relation_MountainCar_7.pdf}}
	\subfigure[Adversarial]{\includegraphics[width=0.2\textwidth]{pic/relation_MountainCar_8.pdf}}
	\caption{\hongming{Detection performance across various state outliers in the online training on MountainCar.}}
	\label{fig:MountainCar_online_full}
\end{figure*}

\begin{figure*}[htbp]
	\centering
	\subfigure{\includegraphics[width=0.19\textwidth]{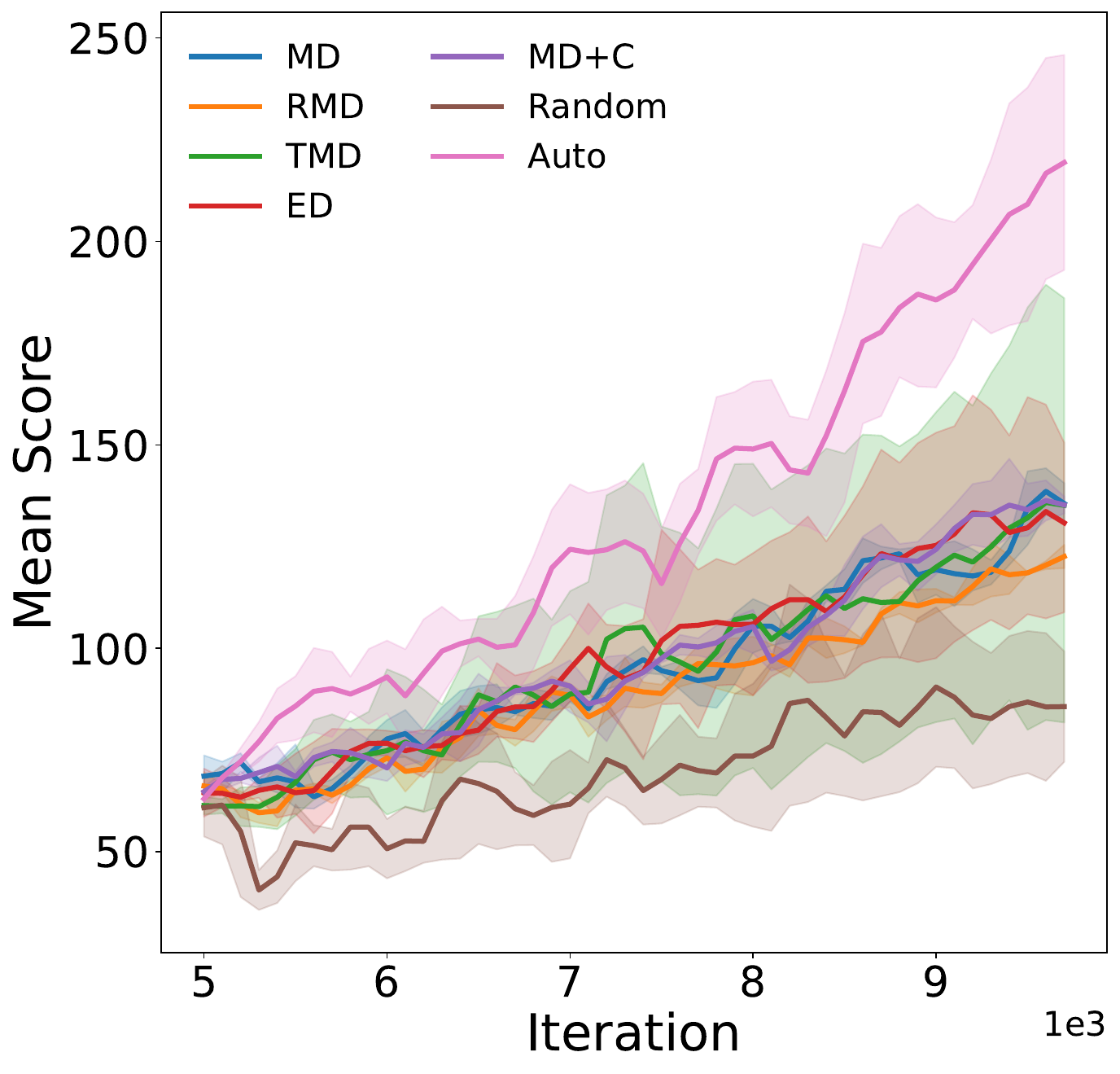}}
	\subfigure{\includegraphics[width=0.19\textwidth]{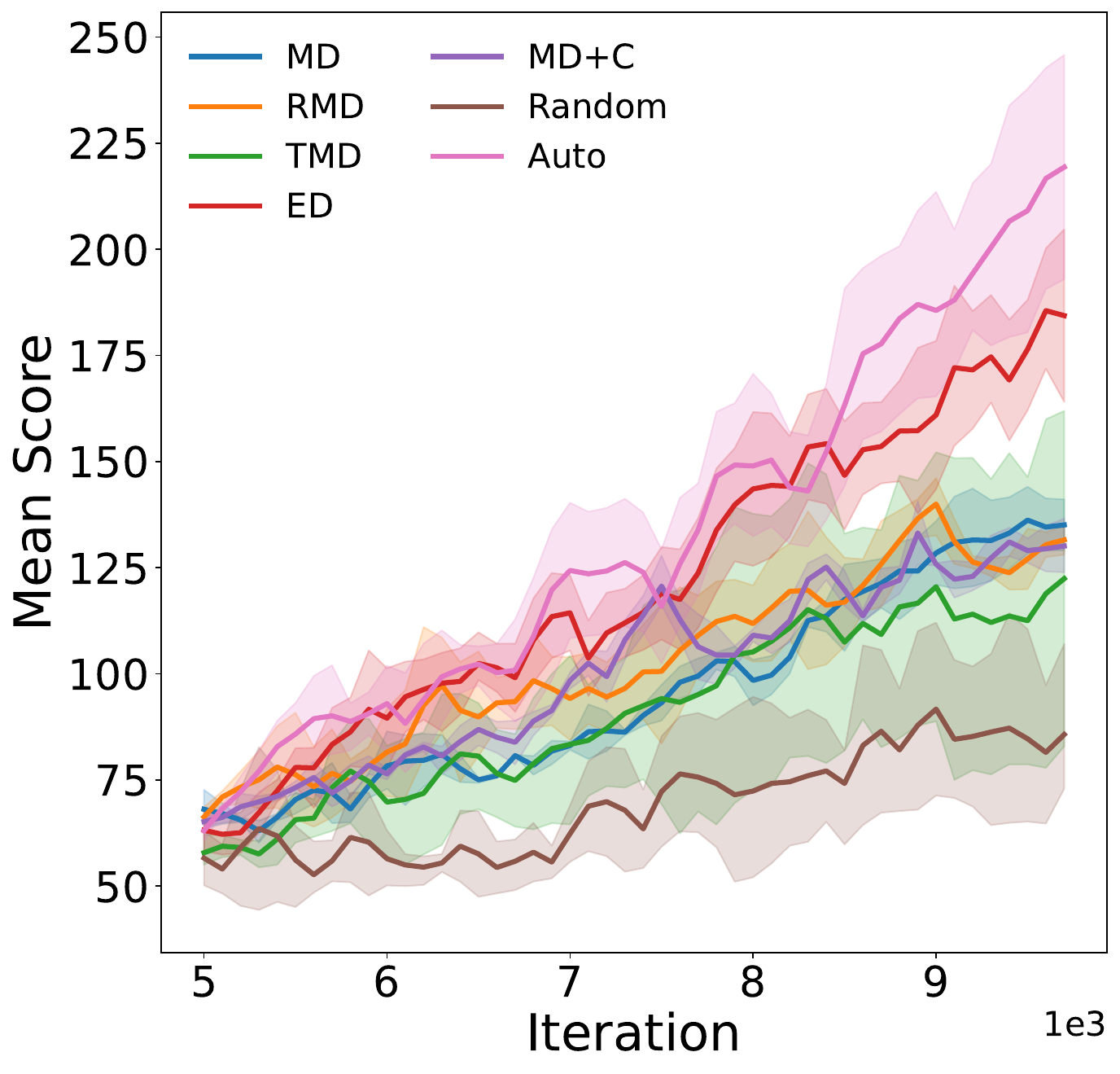}}
	\subfigure{\includegraphics[width=0.19\textwidth]{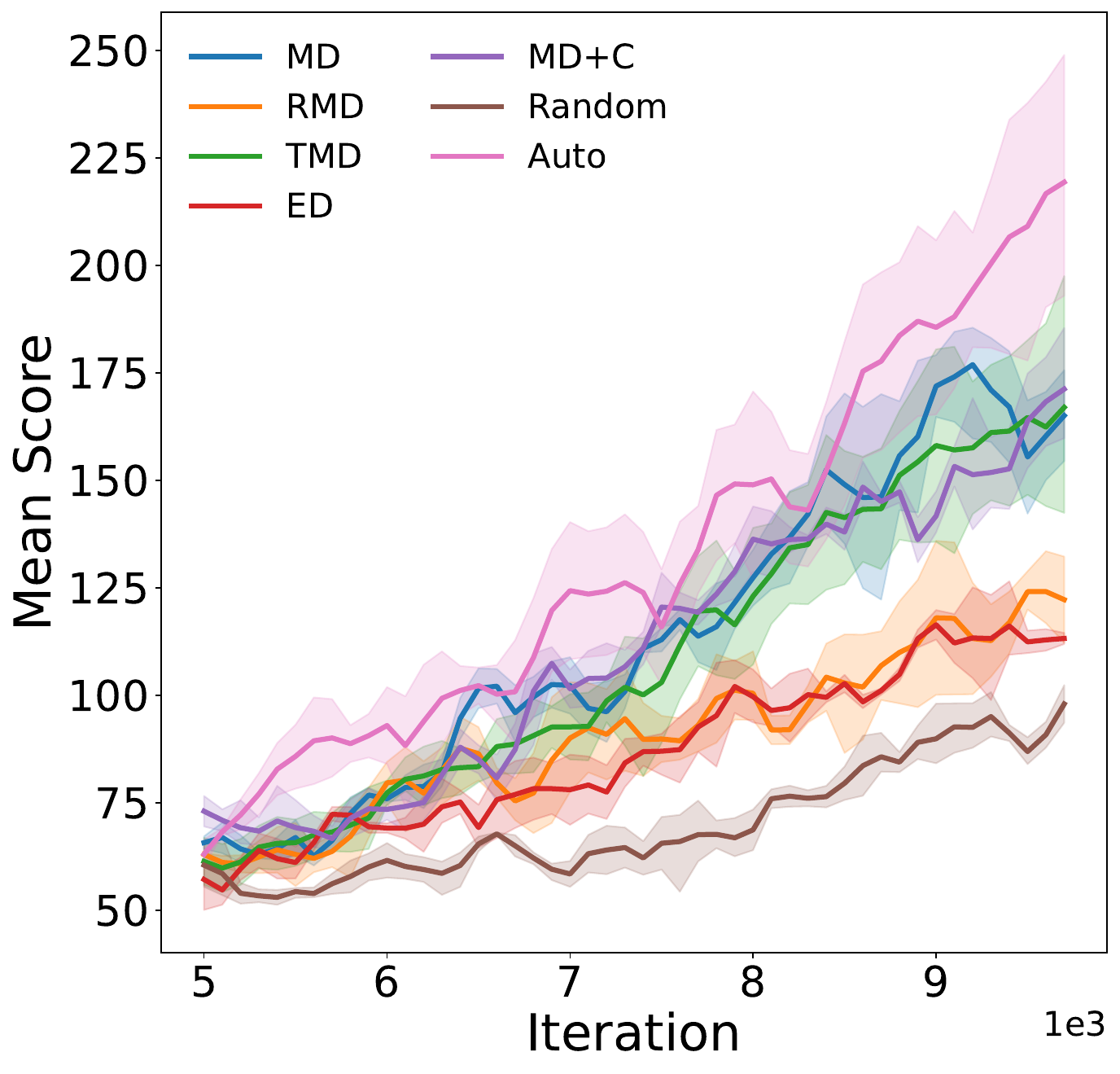}}
	\subfigure{\includegraphics[width=0.19\textwidth]{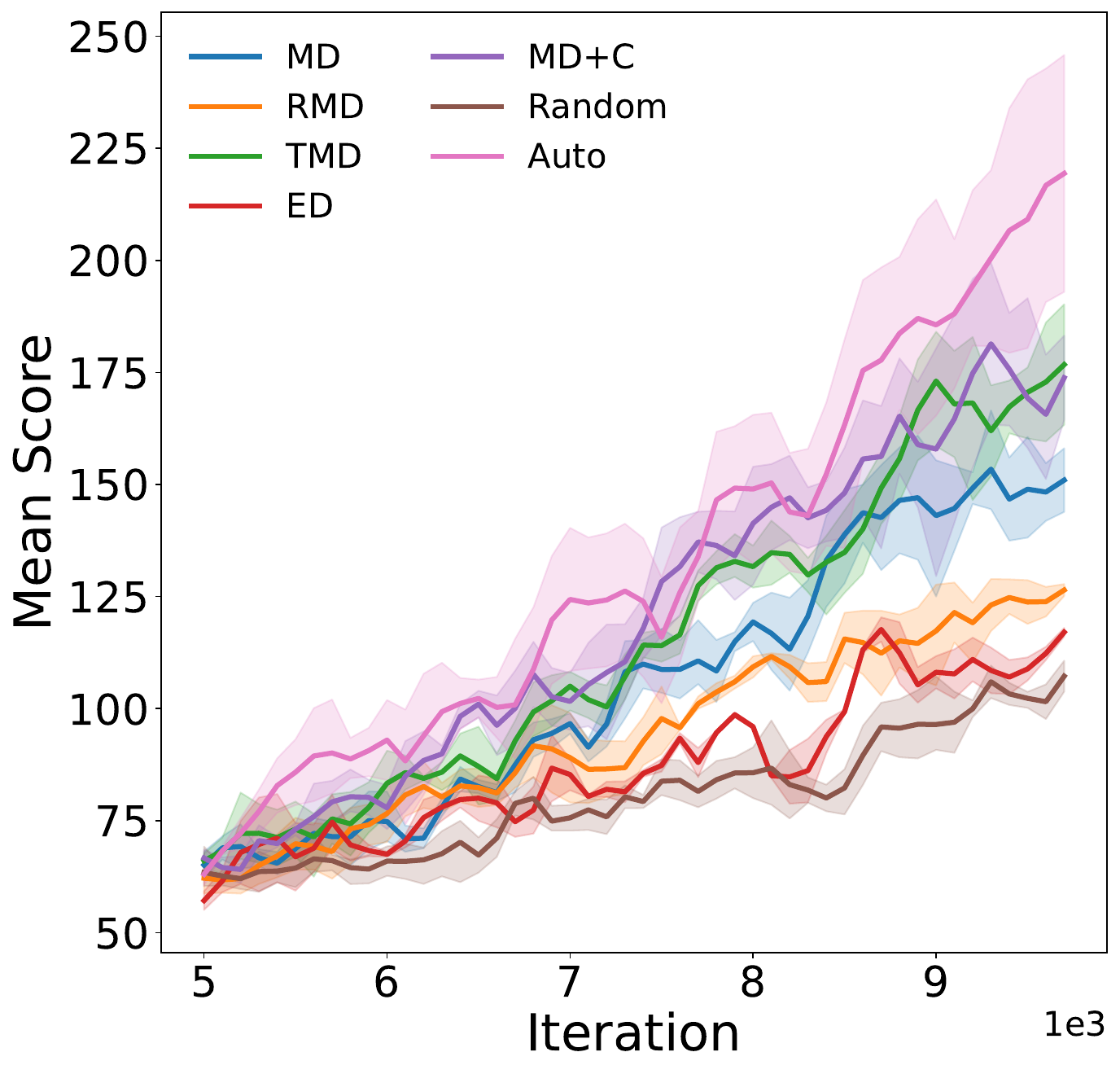}}
	\subfigure{\includegraphics[width=0.19\textwidth]{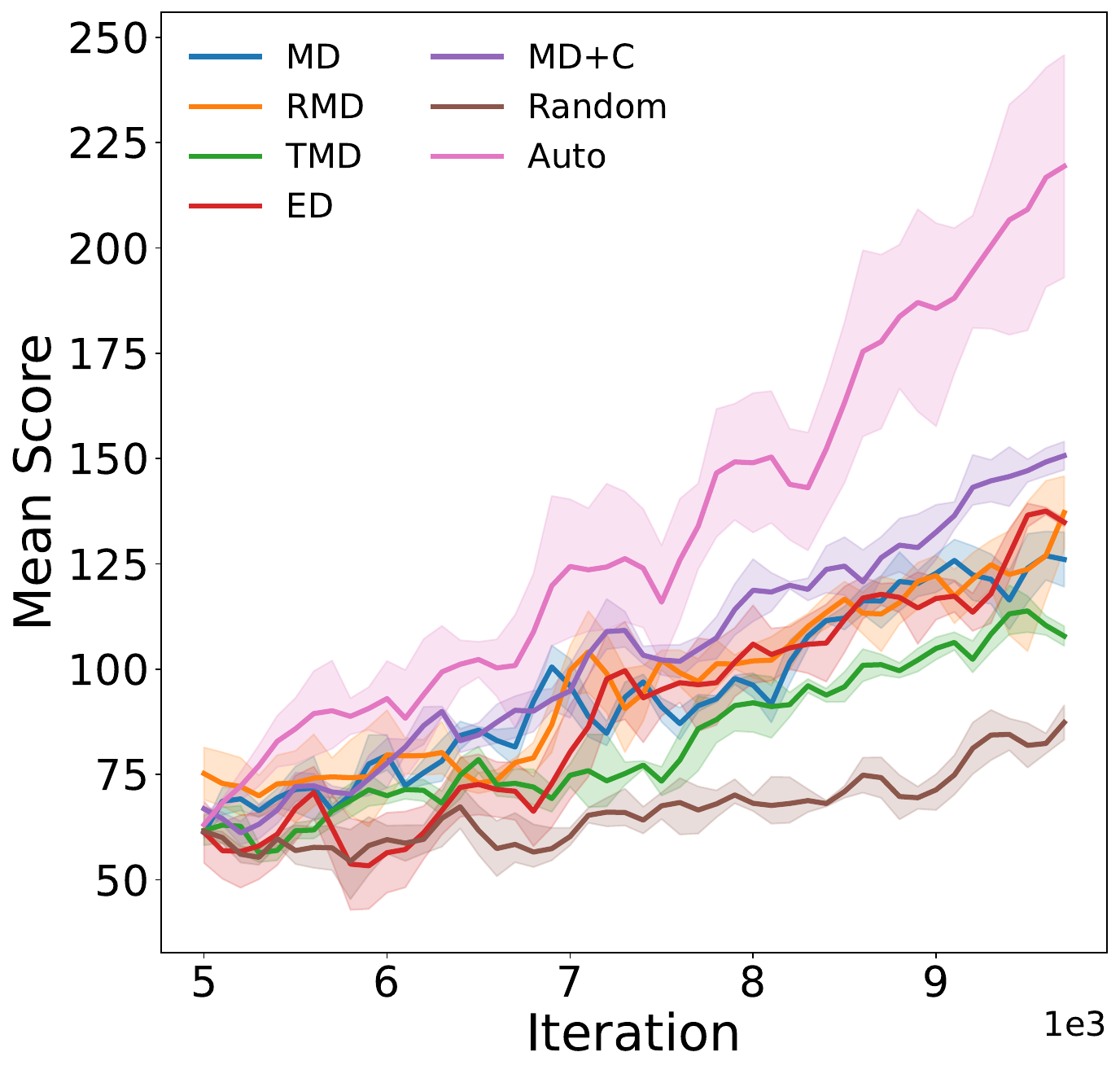}}
    \setcounter{subfigure}{0}
	\subfigure{\includegraphics[width=0.19\textwidth]{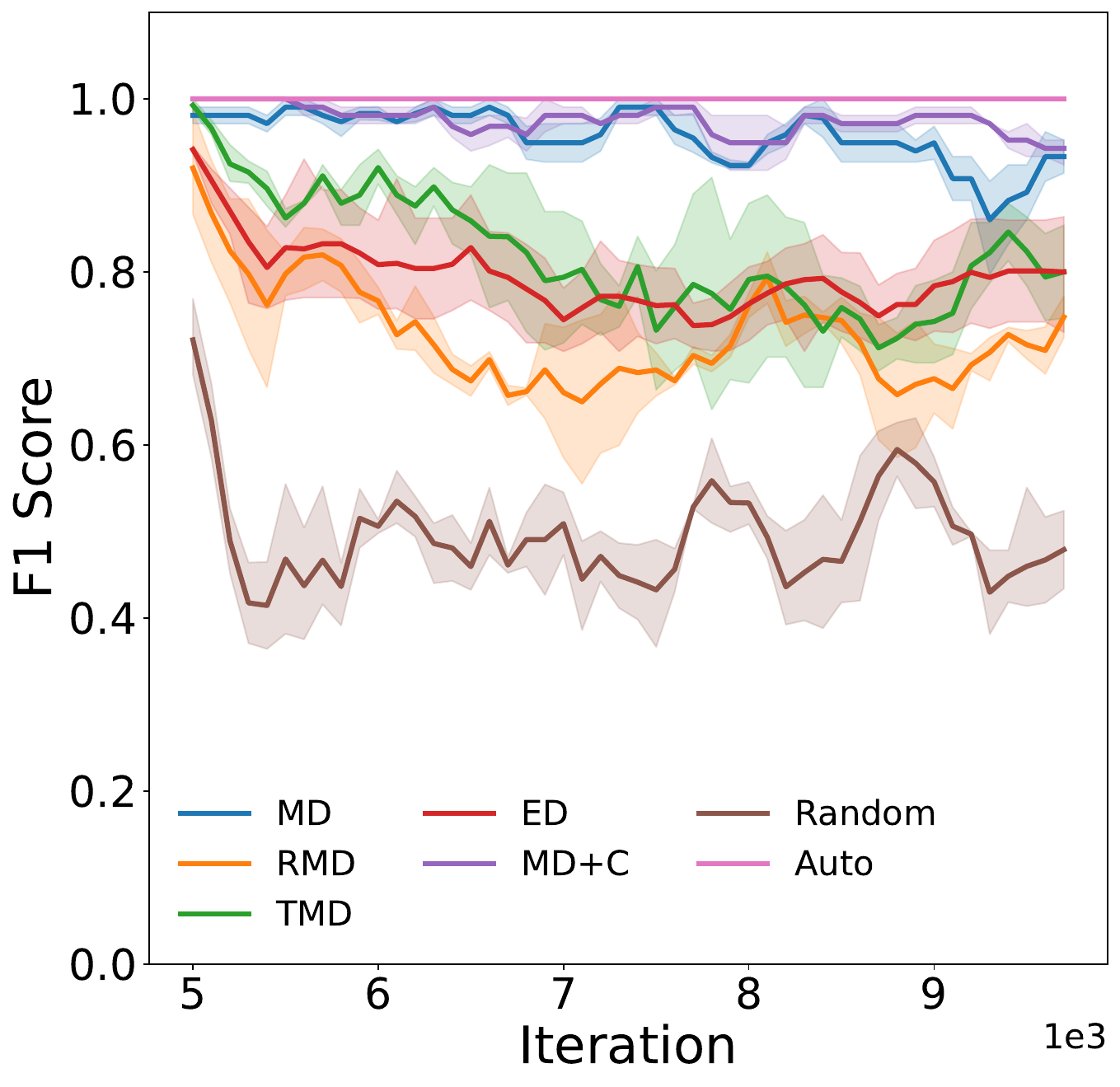}}
	\subfigure{\includegraphics[width=0.19\textwidth]{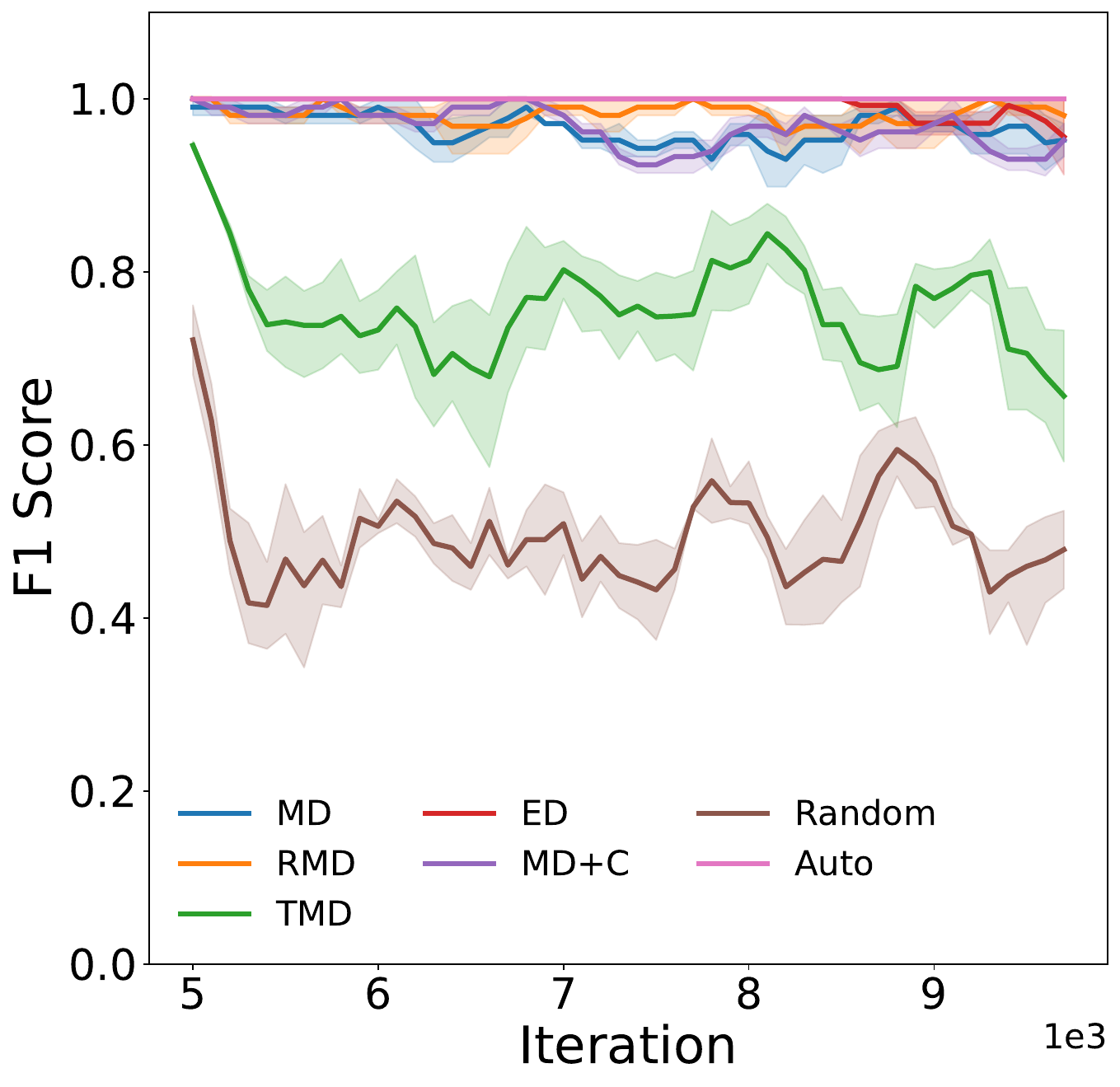}}
	\subfigure{\includegraphics[width=0.19\textwidth]{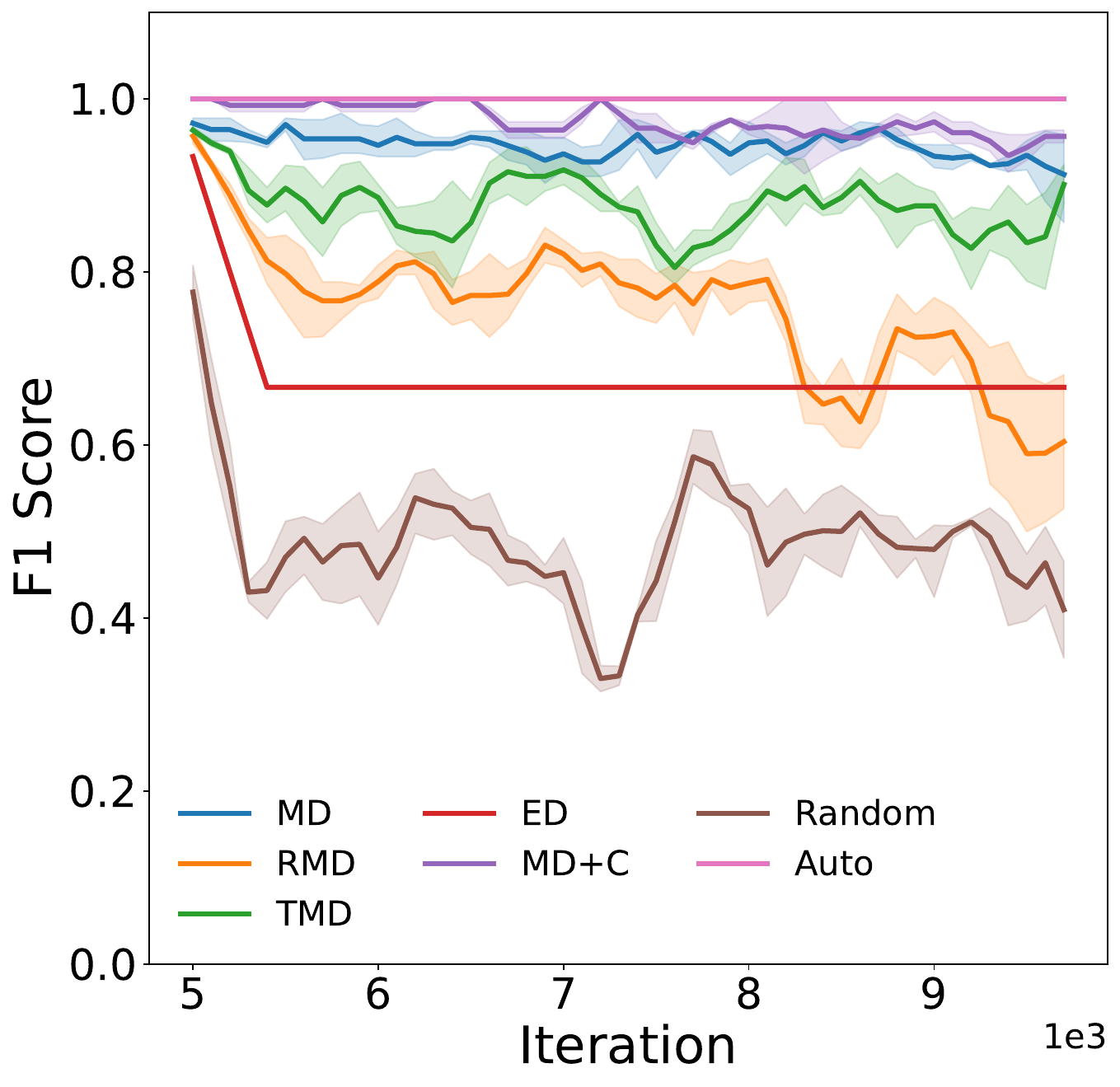}}
	\subfigure{\includegraphics[width=0.19\textwidth]{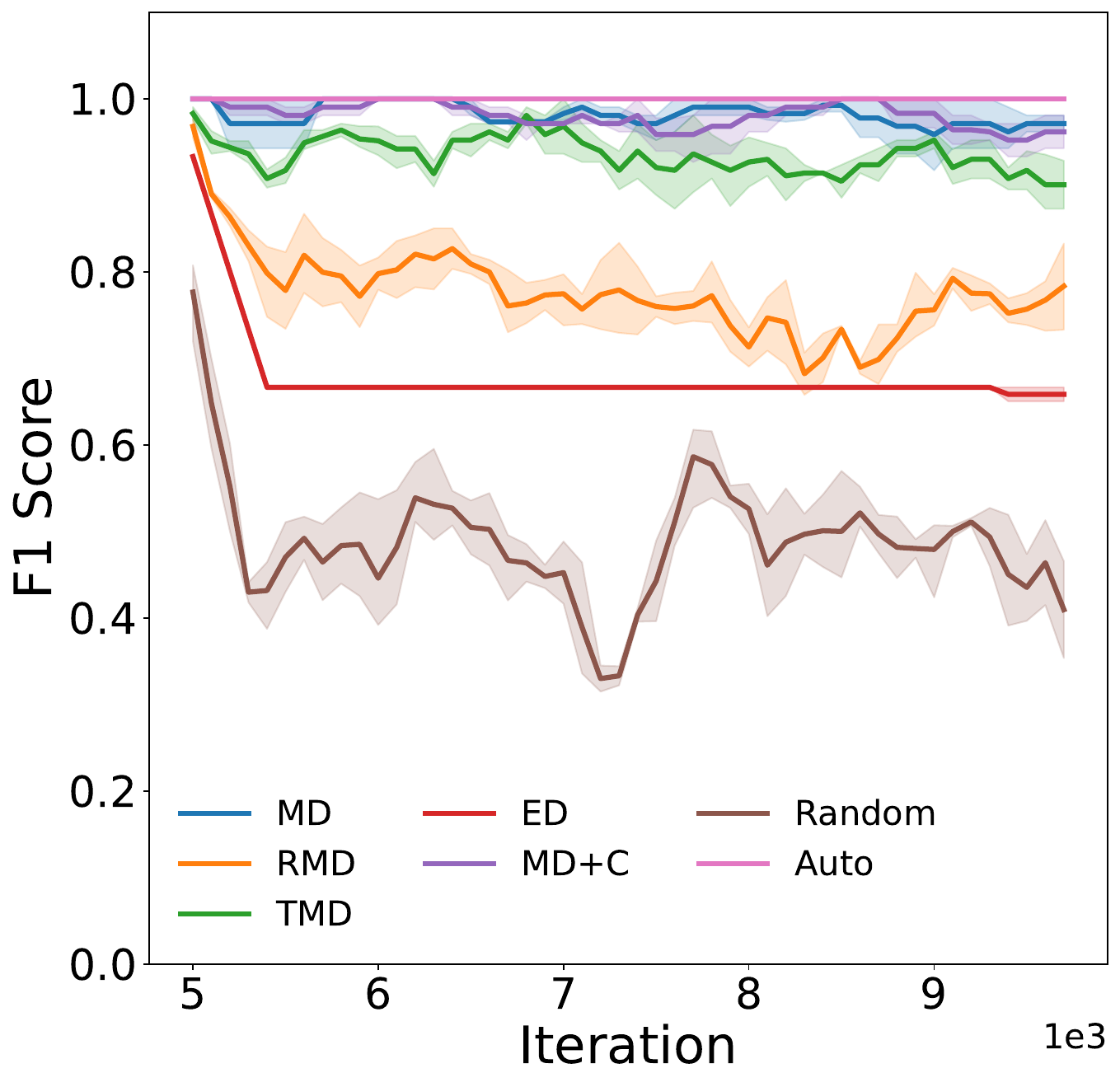}}
	\subfigure{\includegraphics[width=0.19\textwidth]{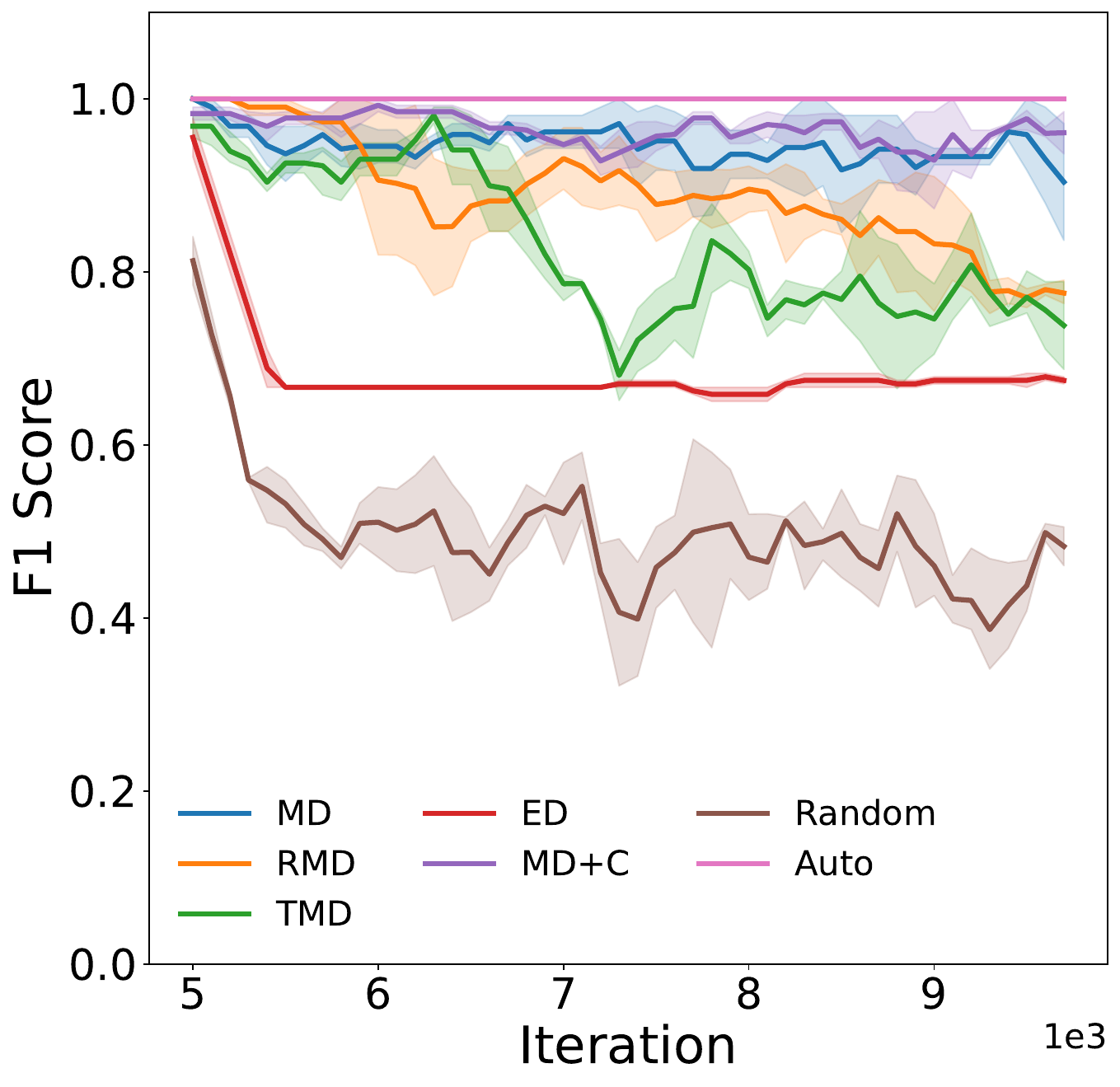}}
     \setcounter{subfigure}{0}
	\subfigure[Gaussian std=1]{\includegraphics[width=0.19\textwidth]{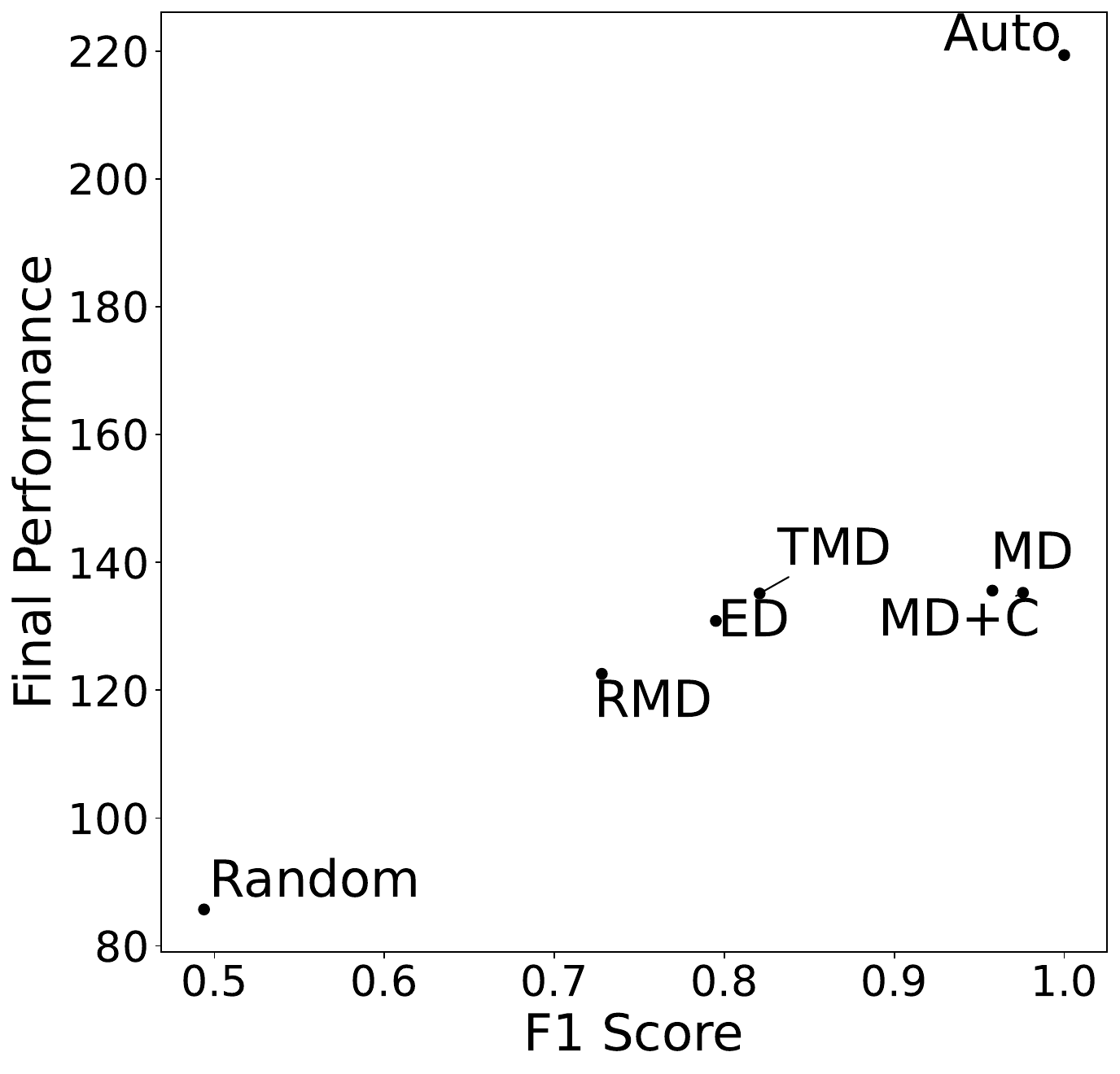}}
	\subfigure[Gaussian std=0.3]{\includegraphics[width=0.19\textwidth]{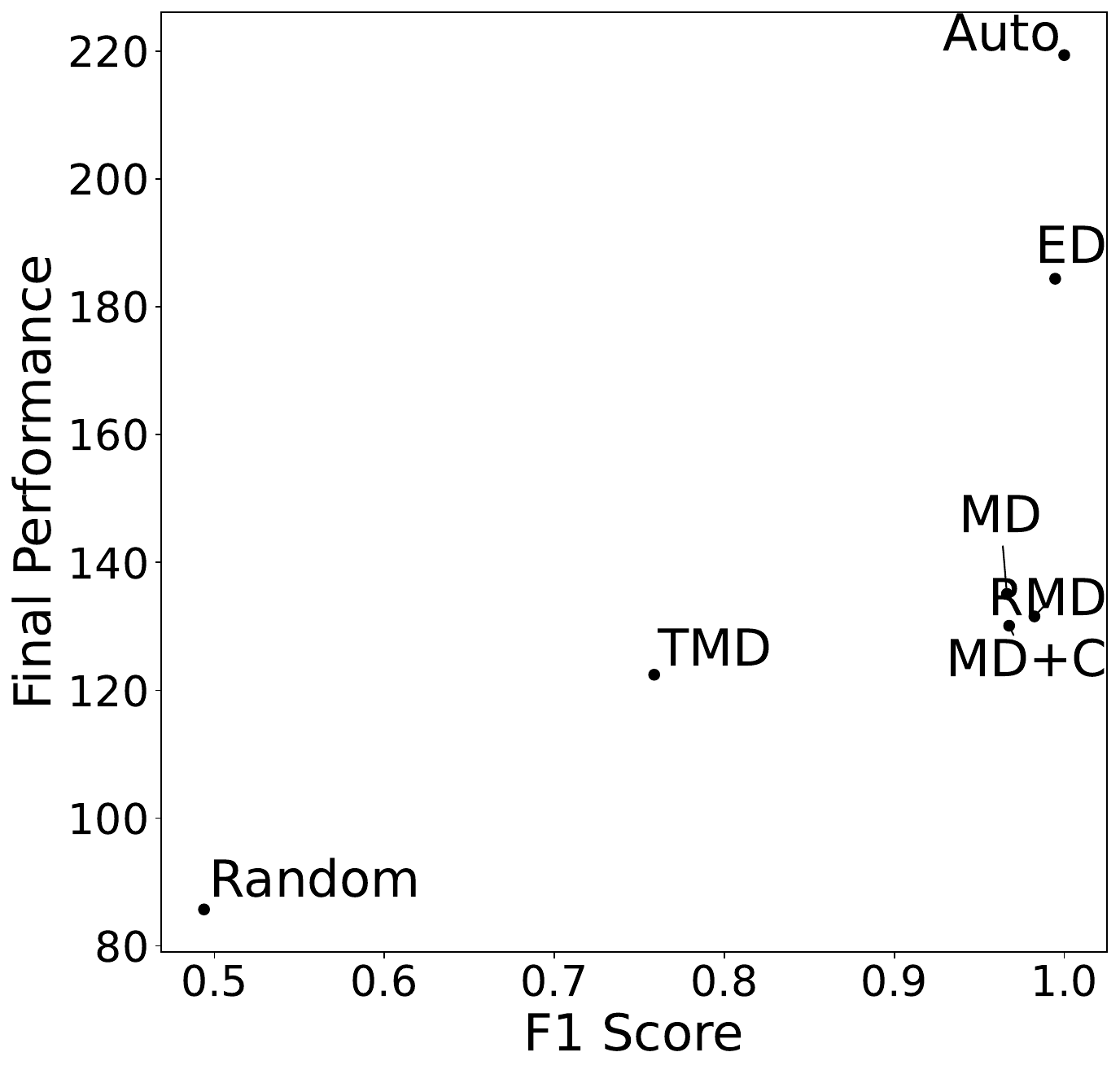}}
	\subfigure[OOD SpaceInvaders]{\includegraphics[width=0.19\textwidth]{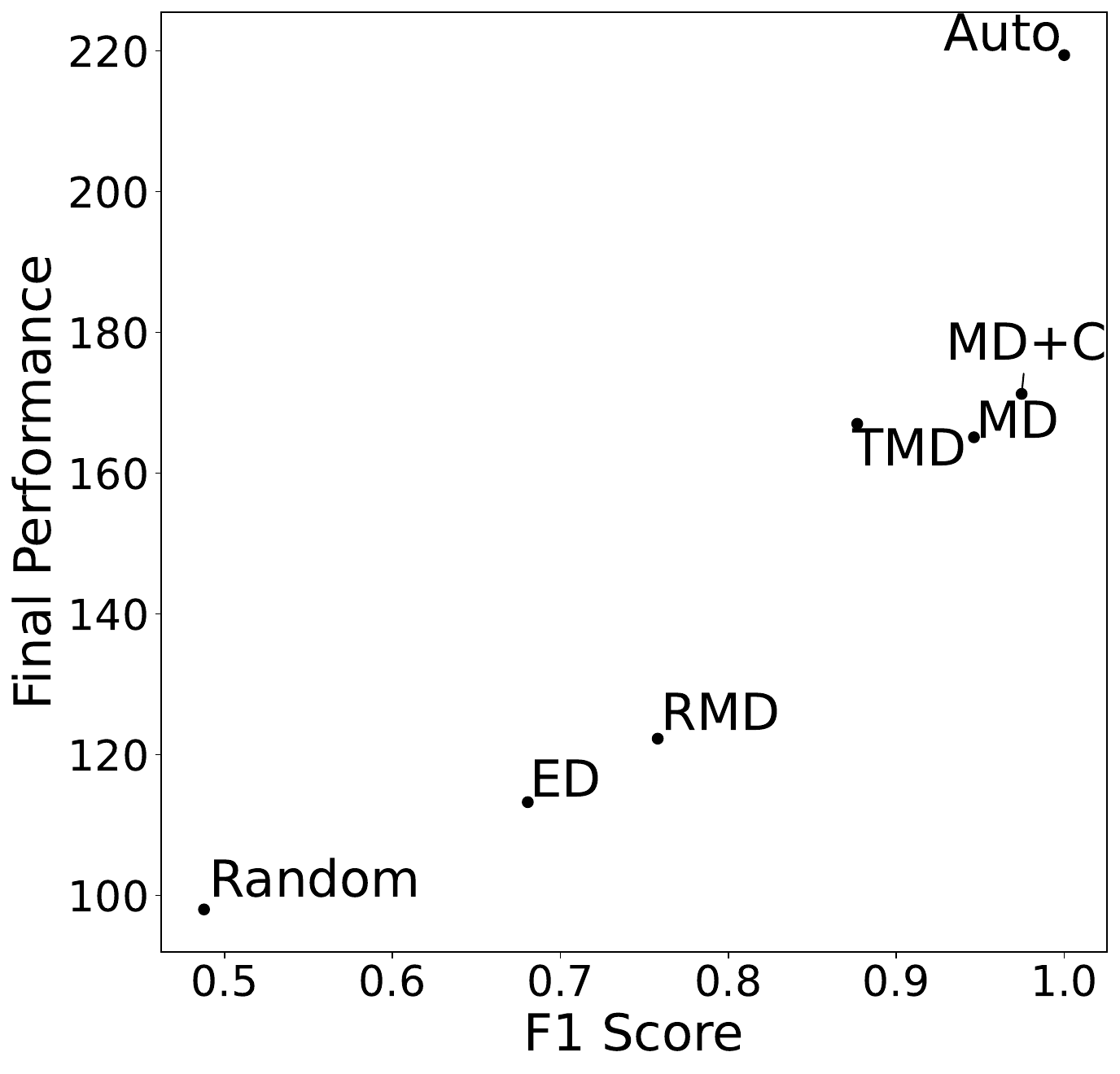}}
	\subfigure[OOD Asterix]{\includegraphics[width=0.19\textwidth]{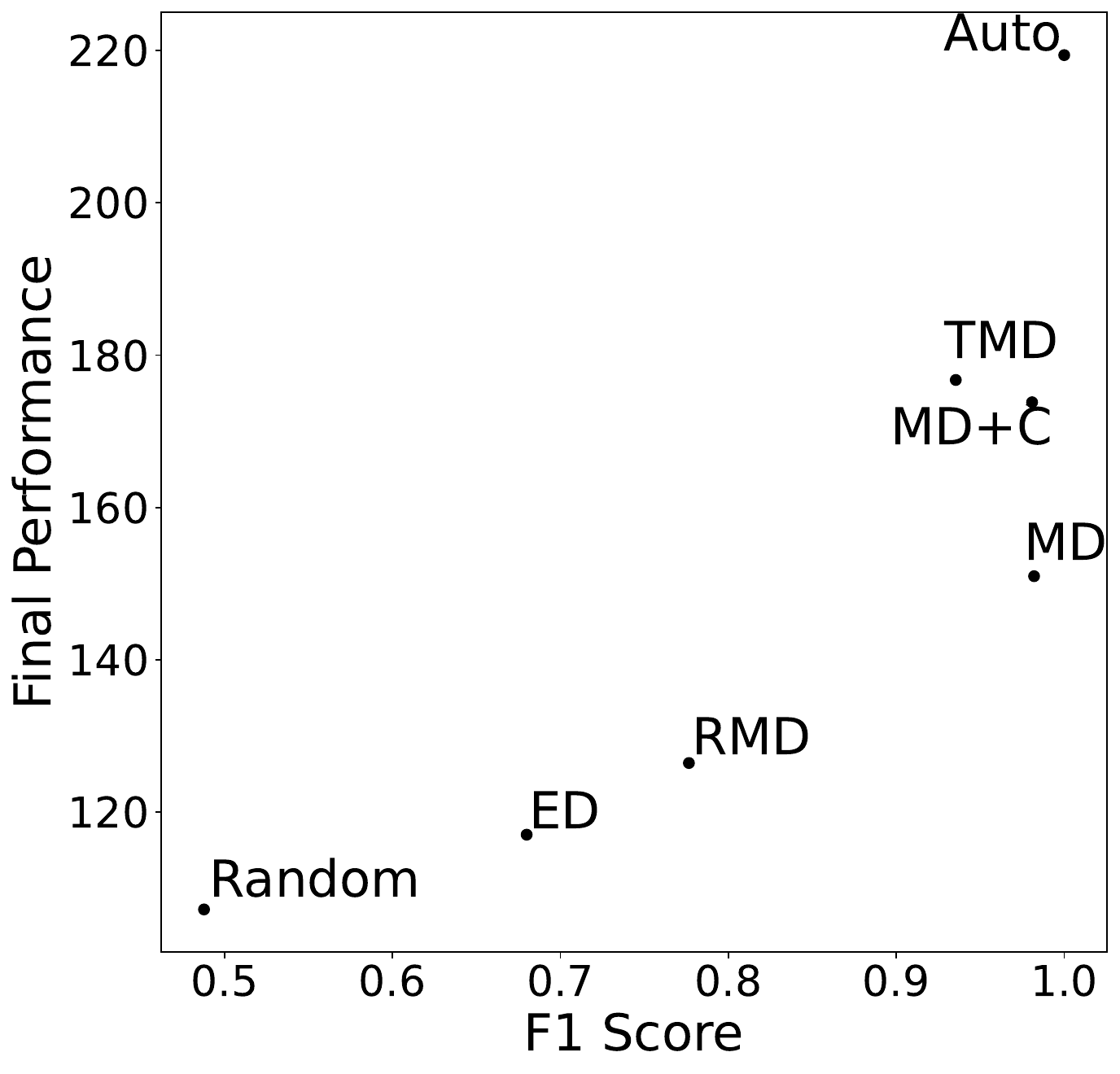}}
	\subfigure[Adversarial]{\includegraphics[width=0.19\textwidth]{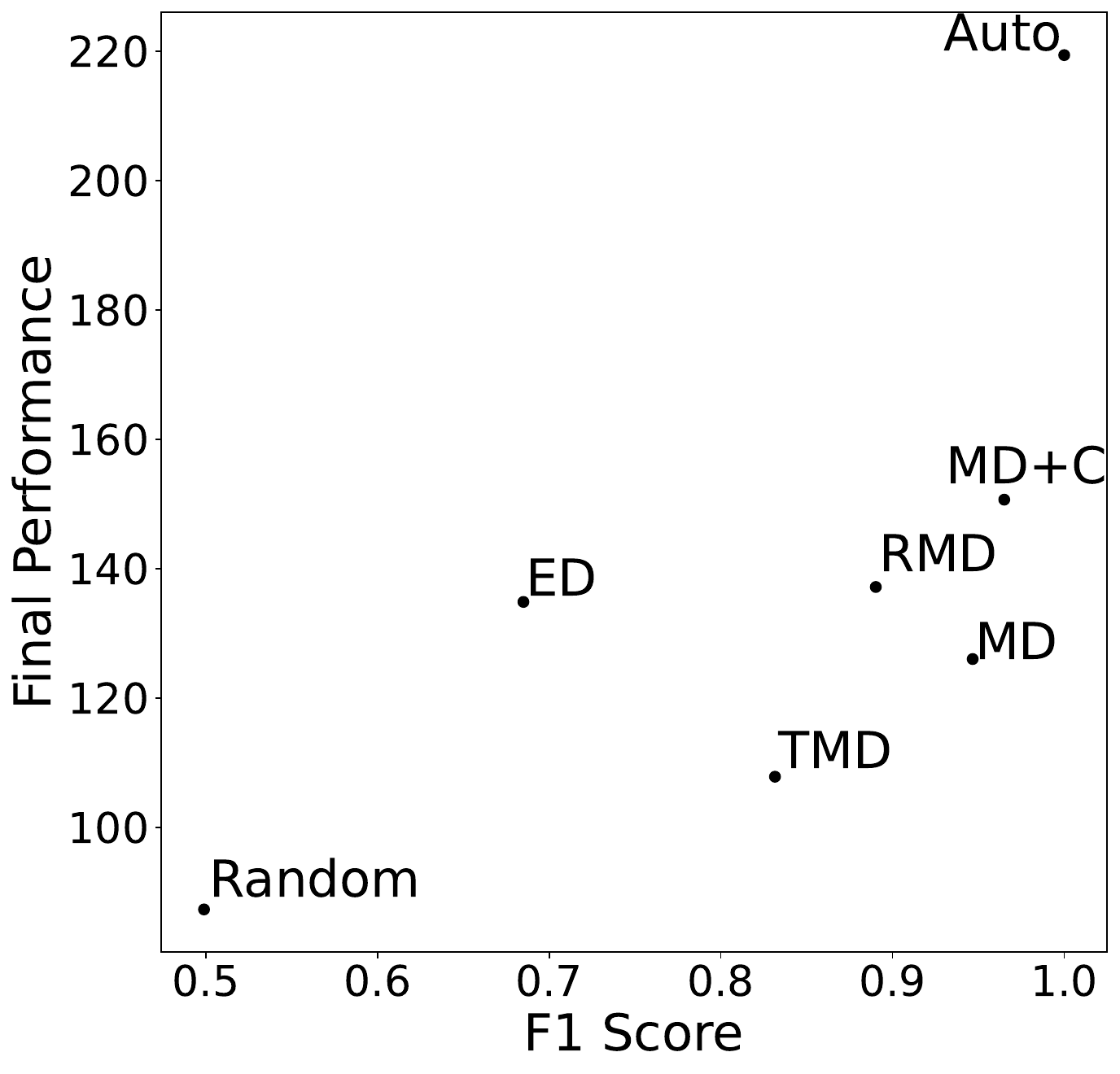}}
	\caption{Detection performance across various state outliers in the online training on Breakout.}
	\label{fig:Breakout_online_full}
\end{figure*}

\begin{figure*}[htbp]
	\centering
	\subfigure{\includegraphics[width=0.19\textwidth]{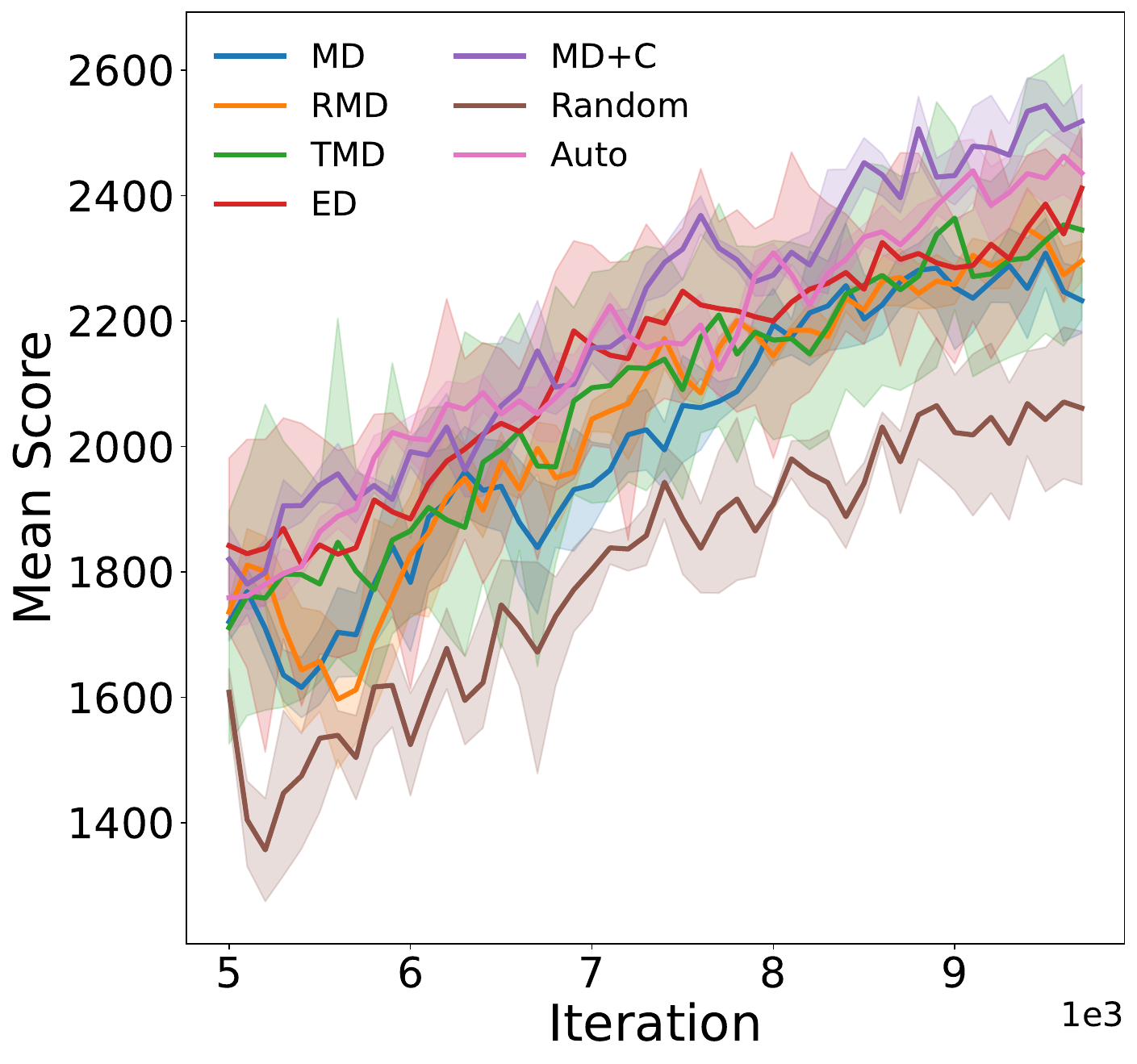}}
	\subfigure{\includegraphics[width=0.19\textwidth]{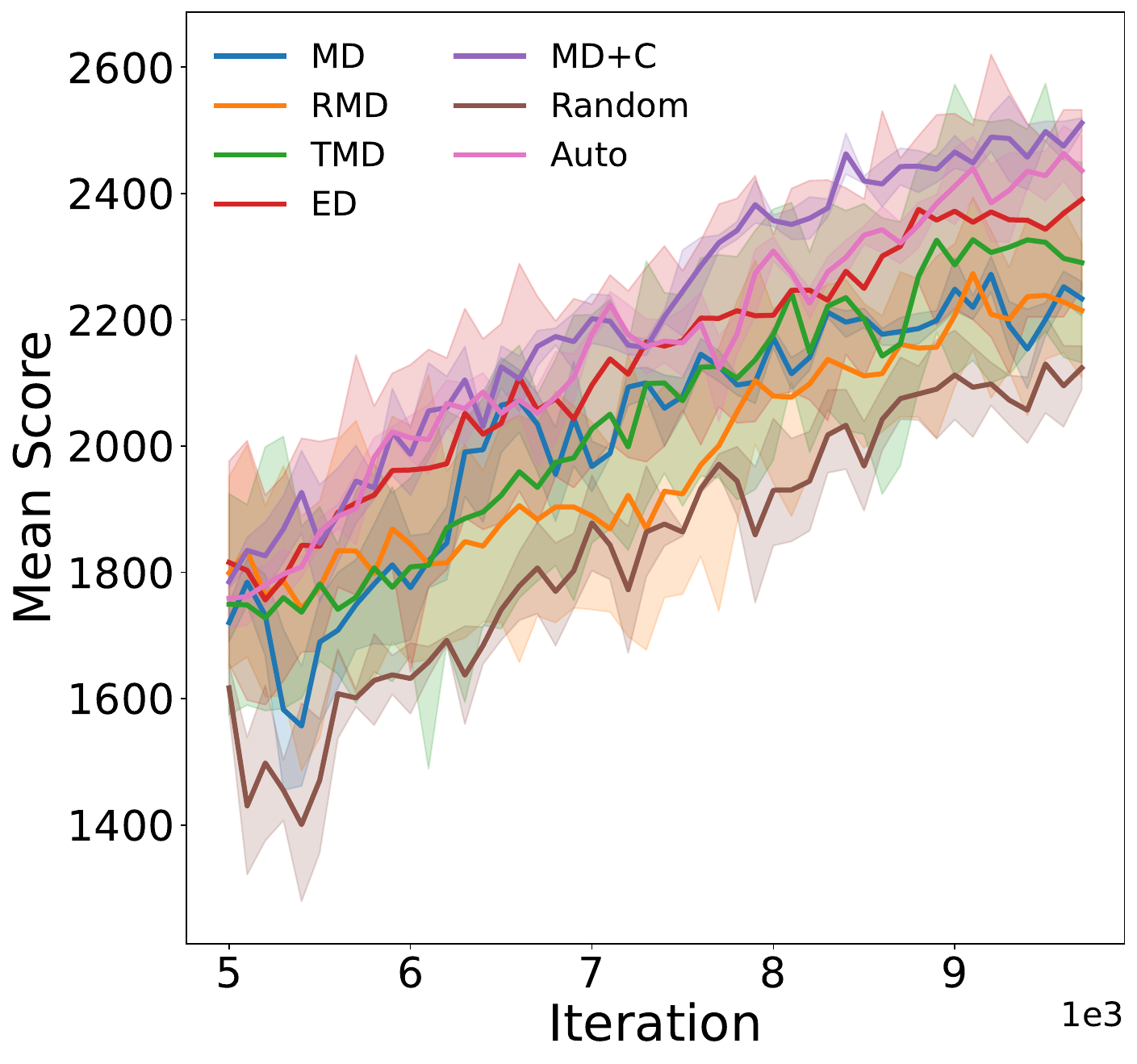}}
	\subfigure{\includegraphics[width=0.19\textwidth]{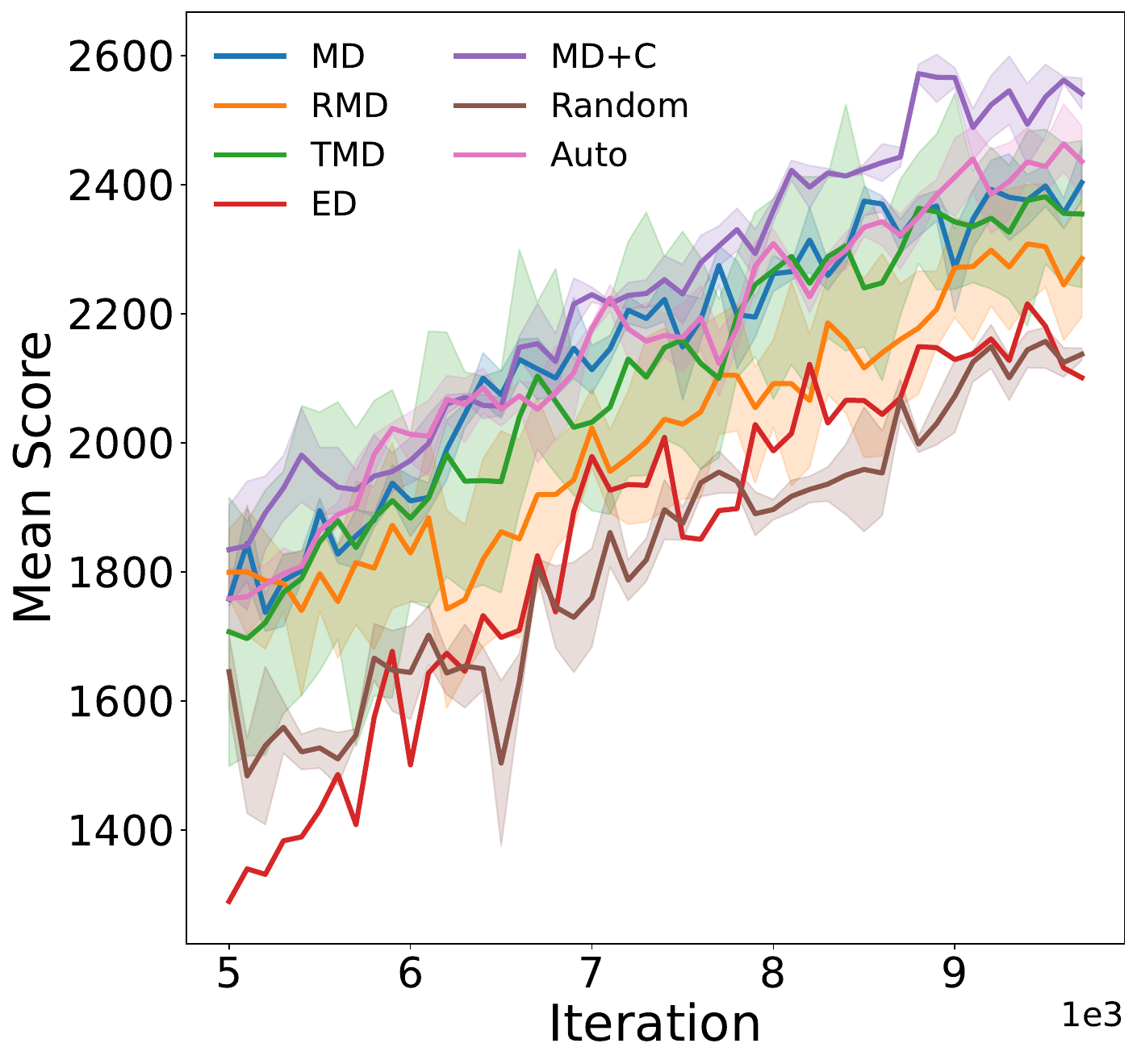}}
	\subfigure{\includegraphics[width=0.19\textwidth]{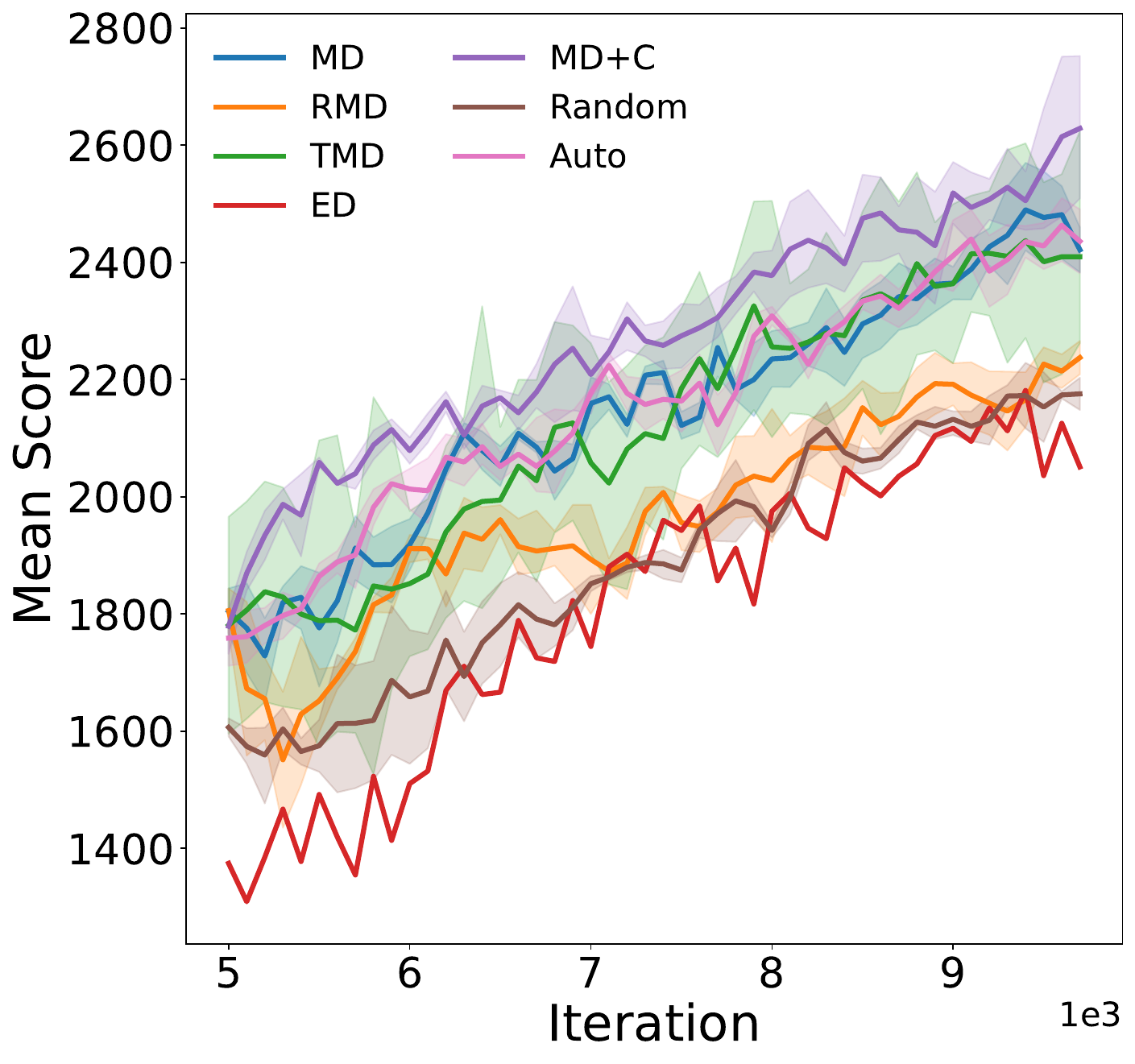}}
	\subfigure{\includegraphics[width=0.19\textwidth]{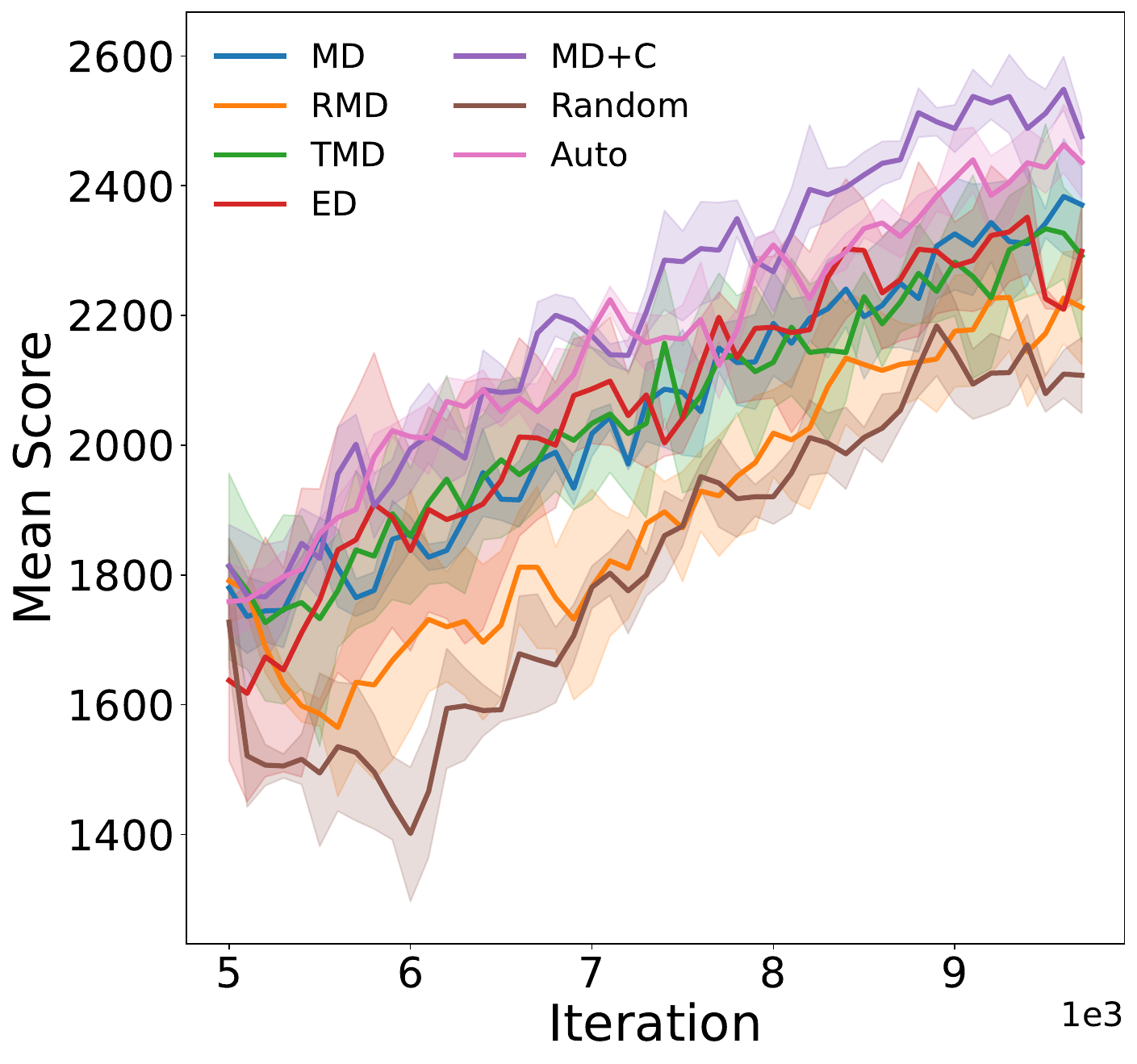}}
    \setcounter{subfigure}{0}
	\subfigure{\includegraphics[width=0.19\textwidth]{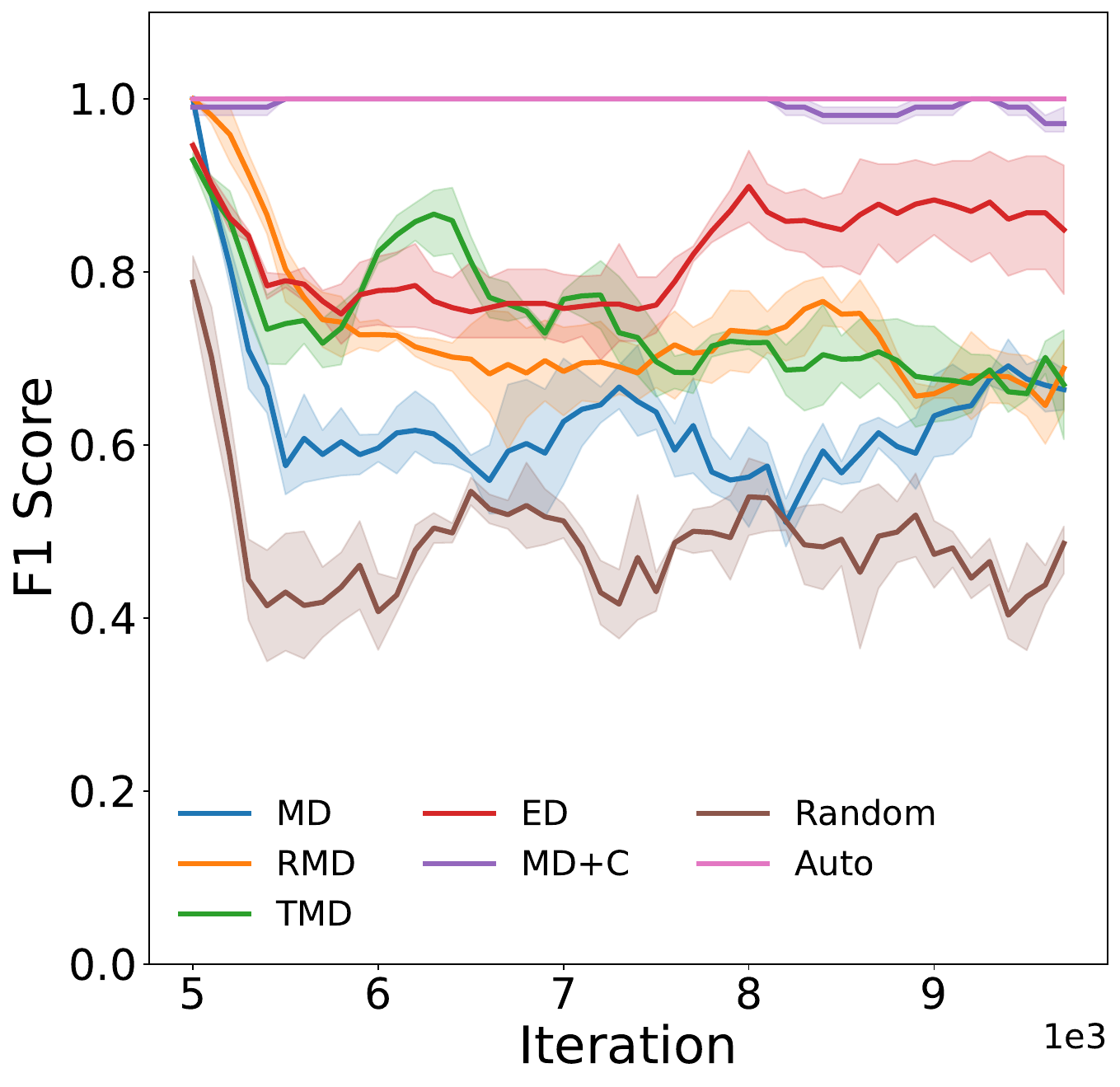}}
	\subfigure{\includegraphics[width=0.19\textwidth]{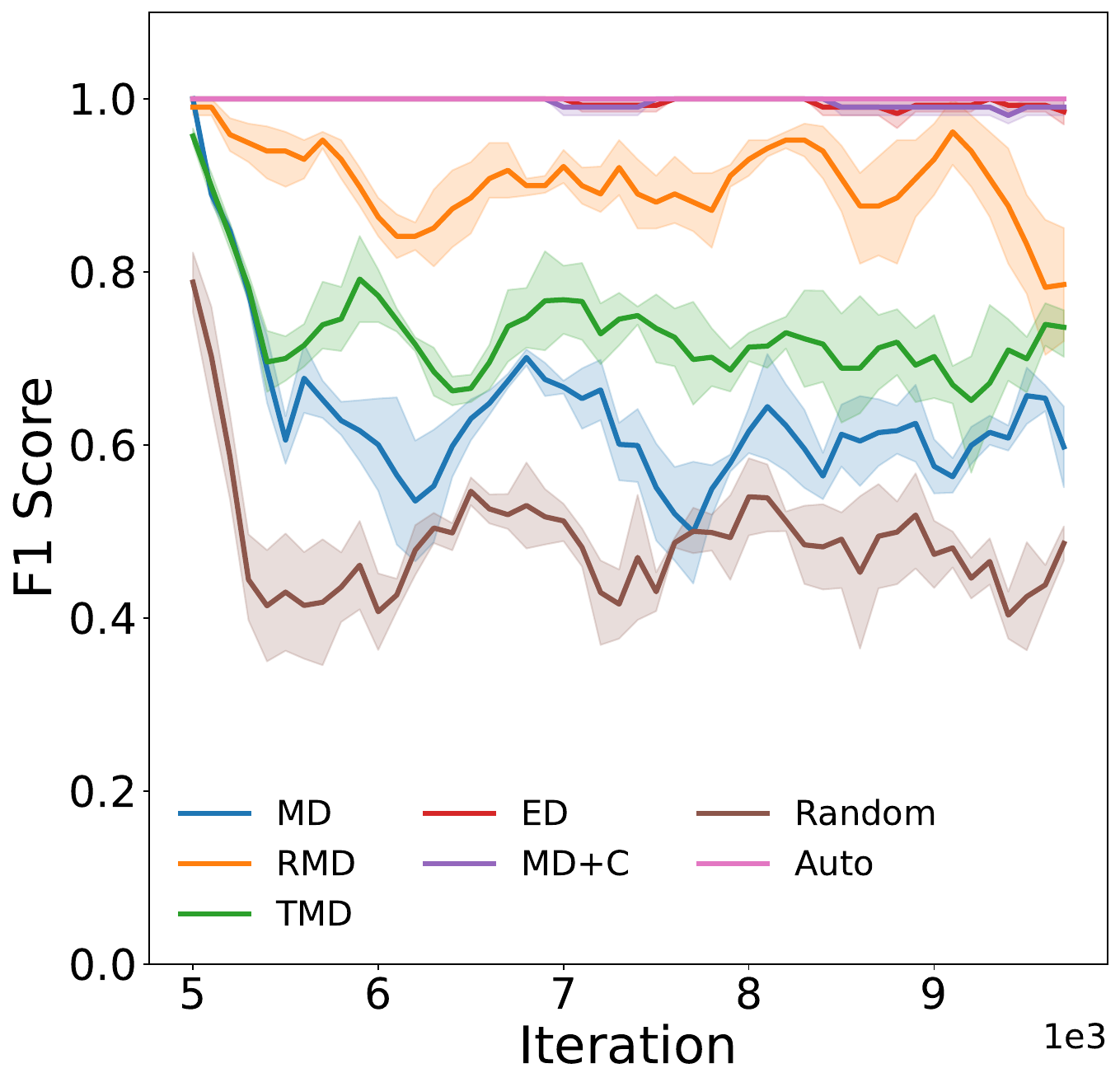}}
	\subfigure{\includegraphics[width=0.19\textwidth]{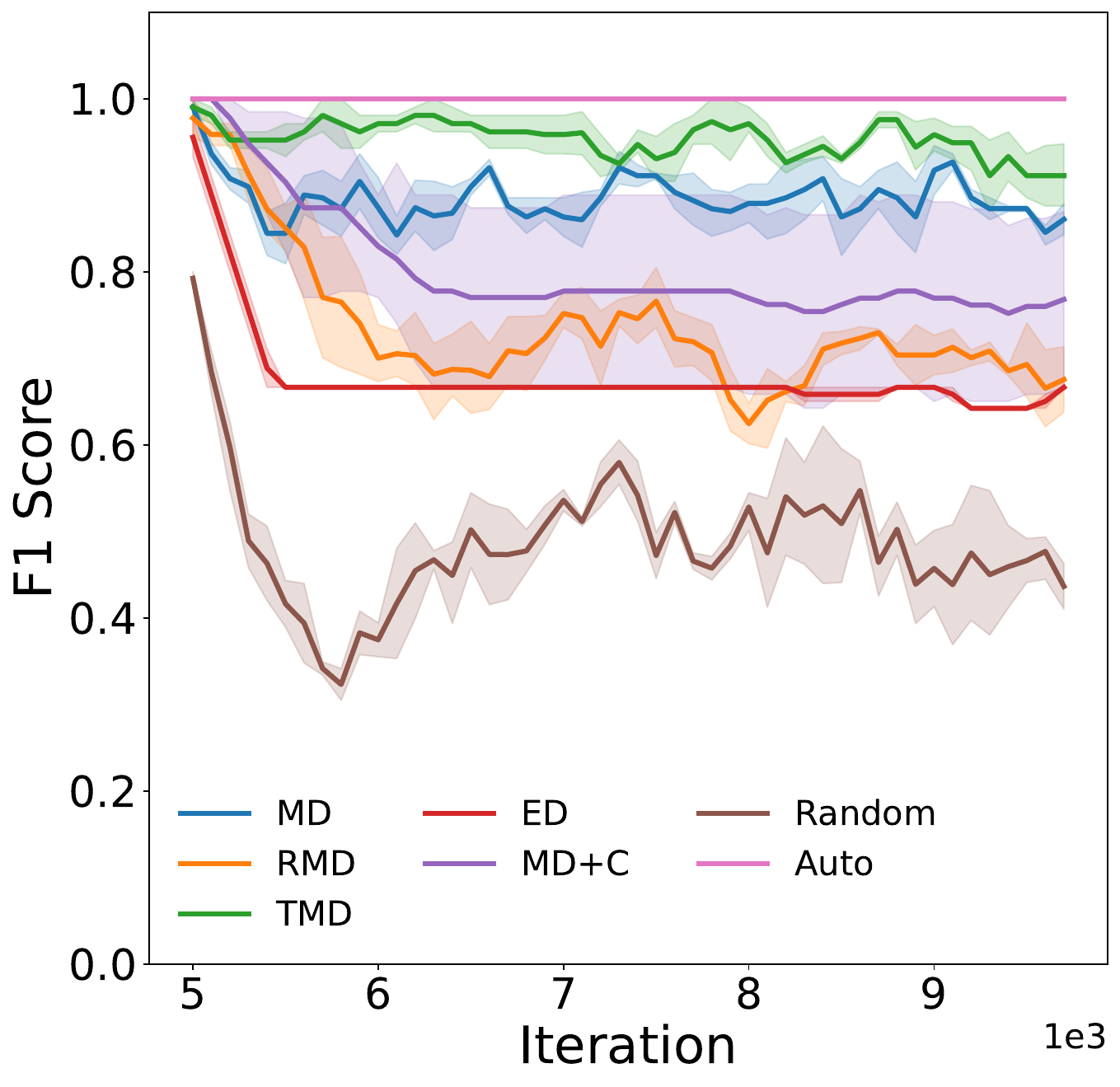}}
	\subfigure{\includegraphics[width=0.19\textwidth]{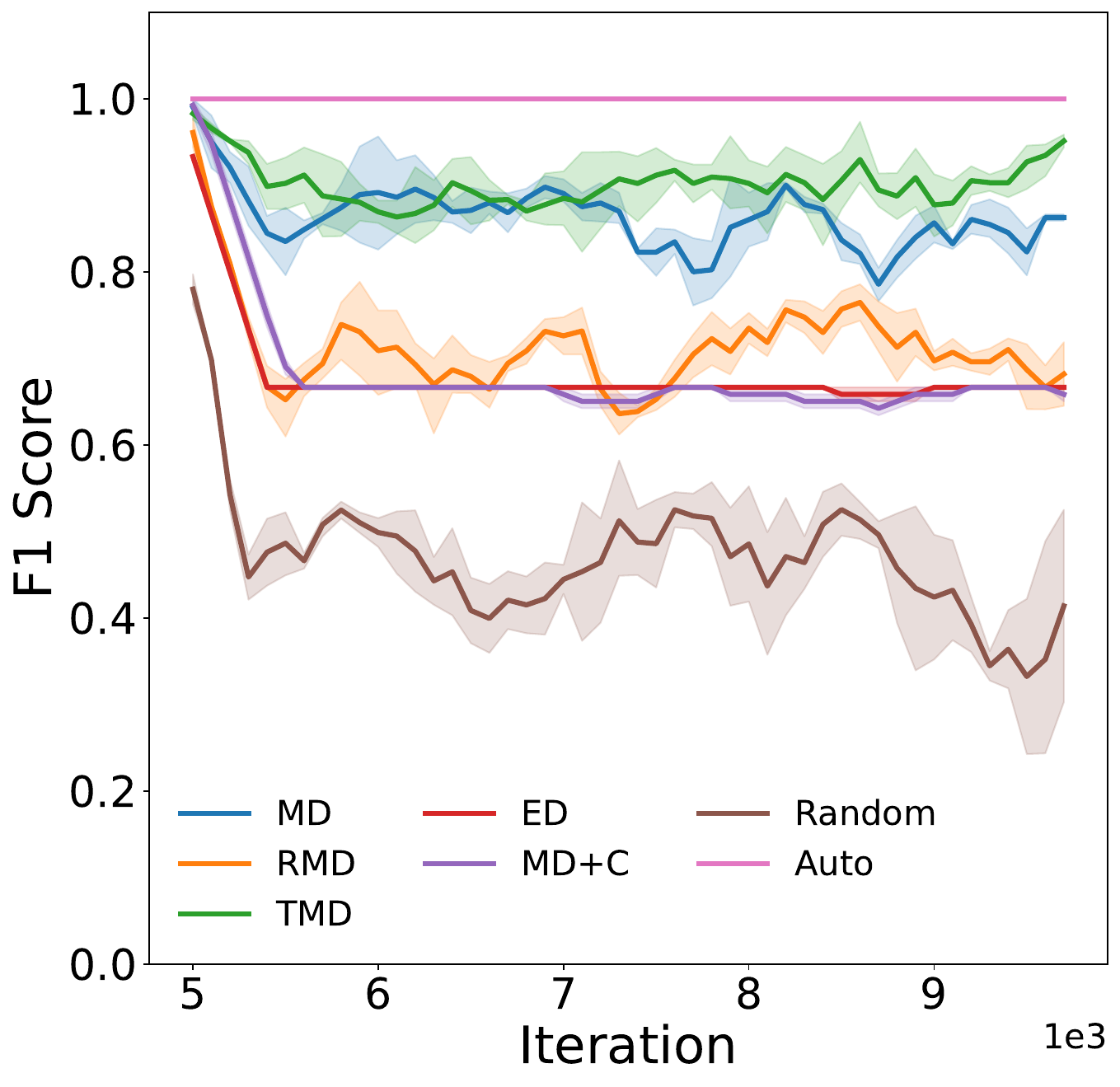}}
	\subfigure{\includegraphics[width=0.19\textwidth]{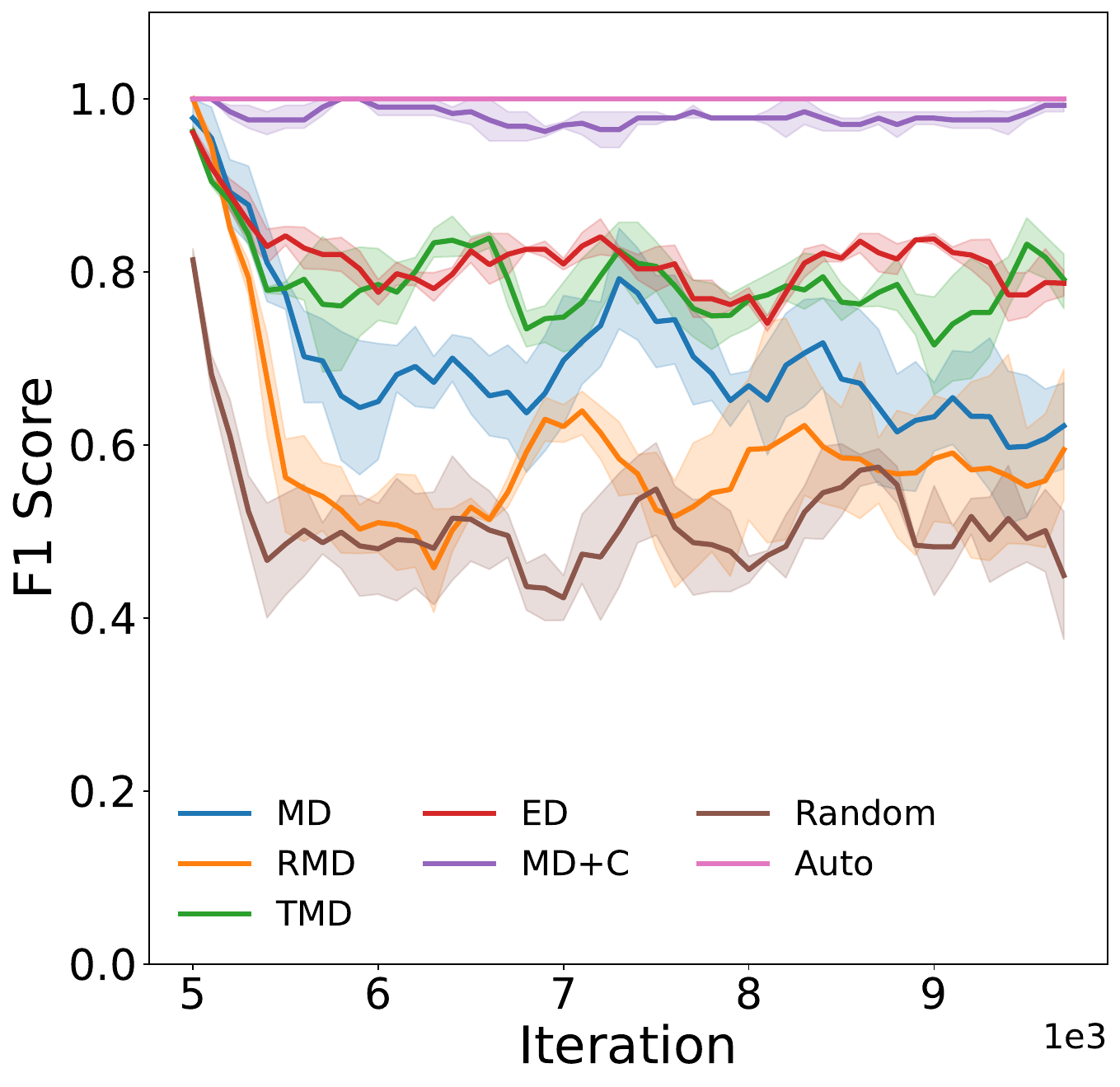}}    
 \setcounter{subfigure}{0} 	
        \subfigure[Gaussian std=1]{\includegraphics[width=0.19\textwidth]{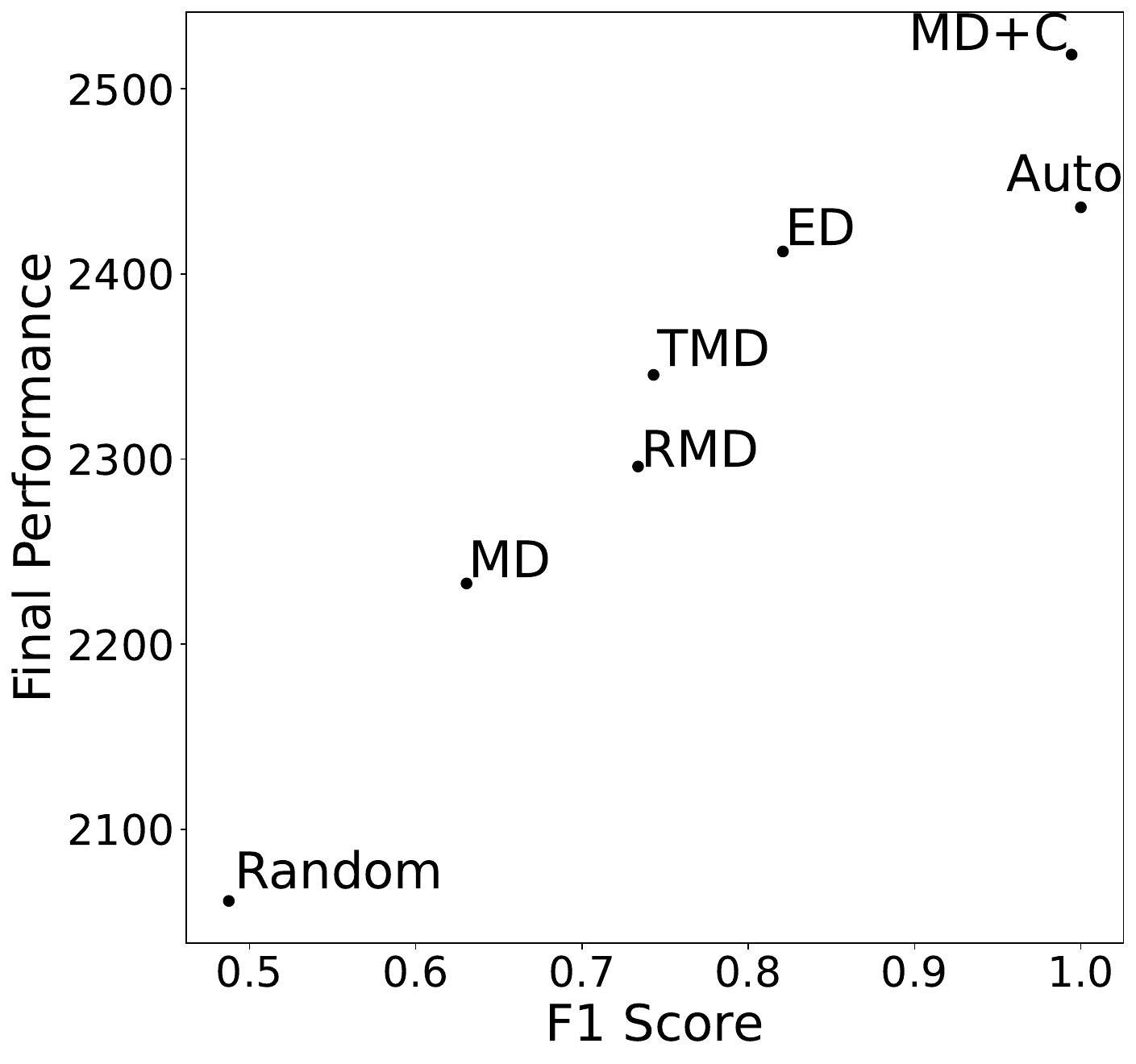}}
	\subfigure[Gaussian std=0.3]{\includegraphics[width=0.19\textwidth]{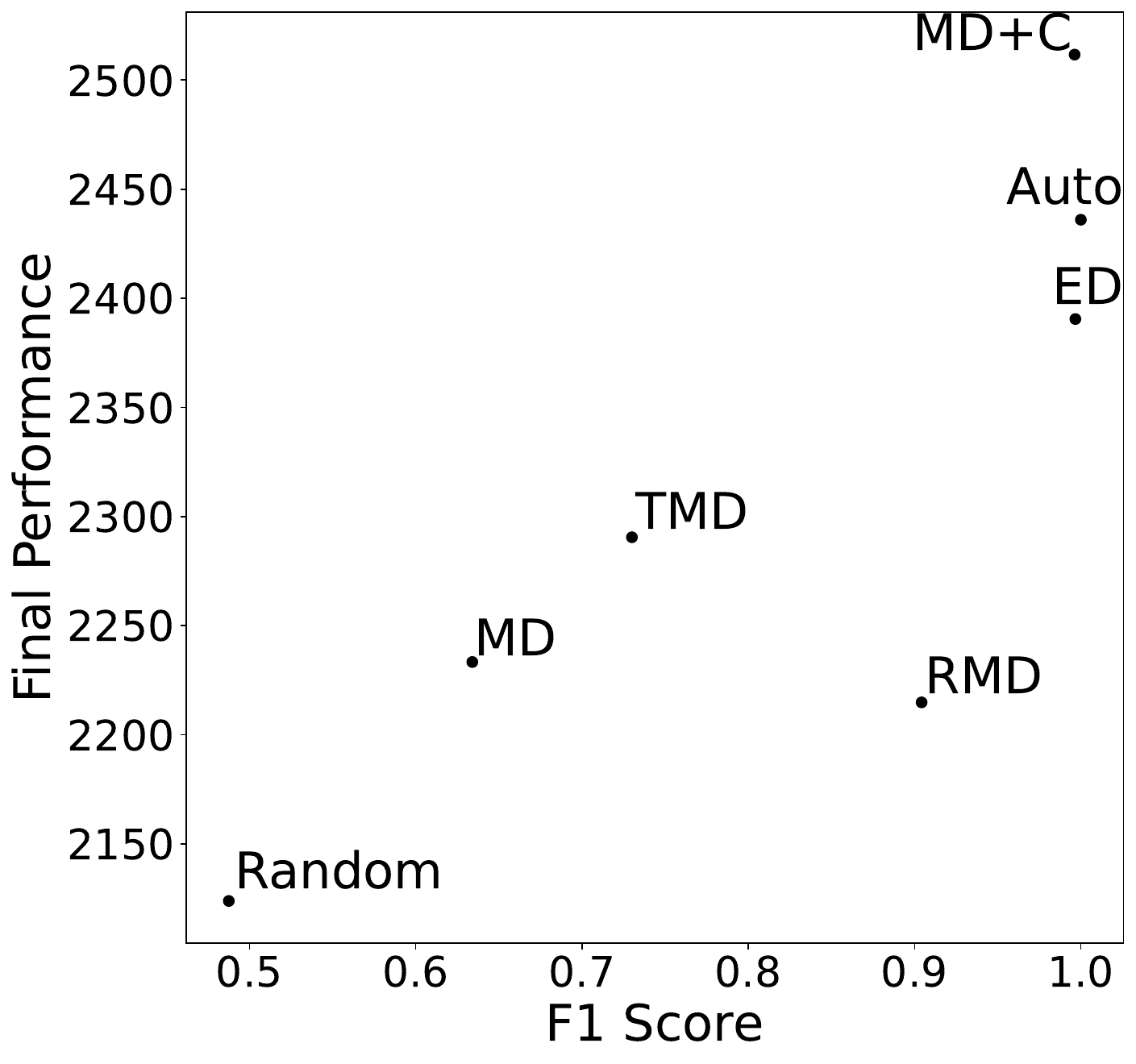}}
	\subfigure[OOD Breakout]{\includegraphics[width=0.19\textwidth]{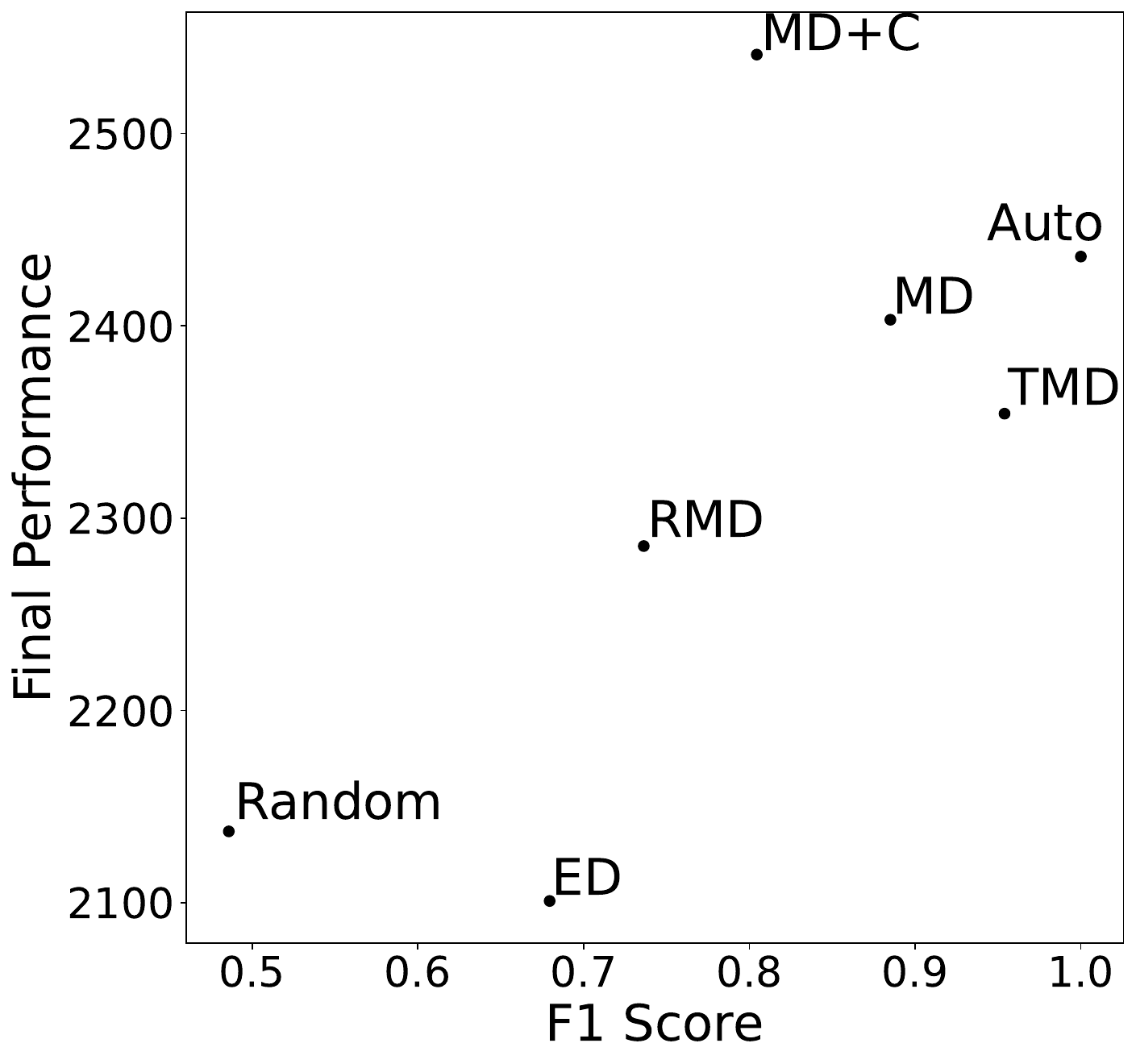}}
	\subfigure[OOD SpaceInvaders]{\includegraphics[width=0.19\textwidth]{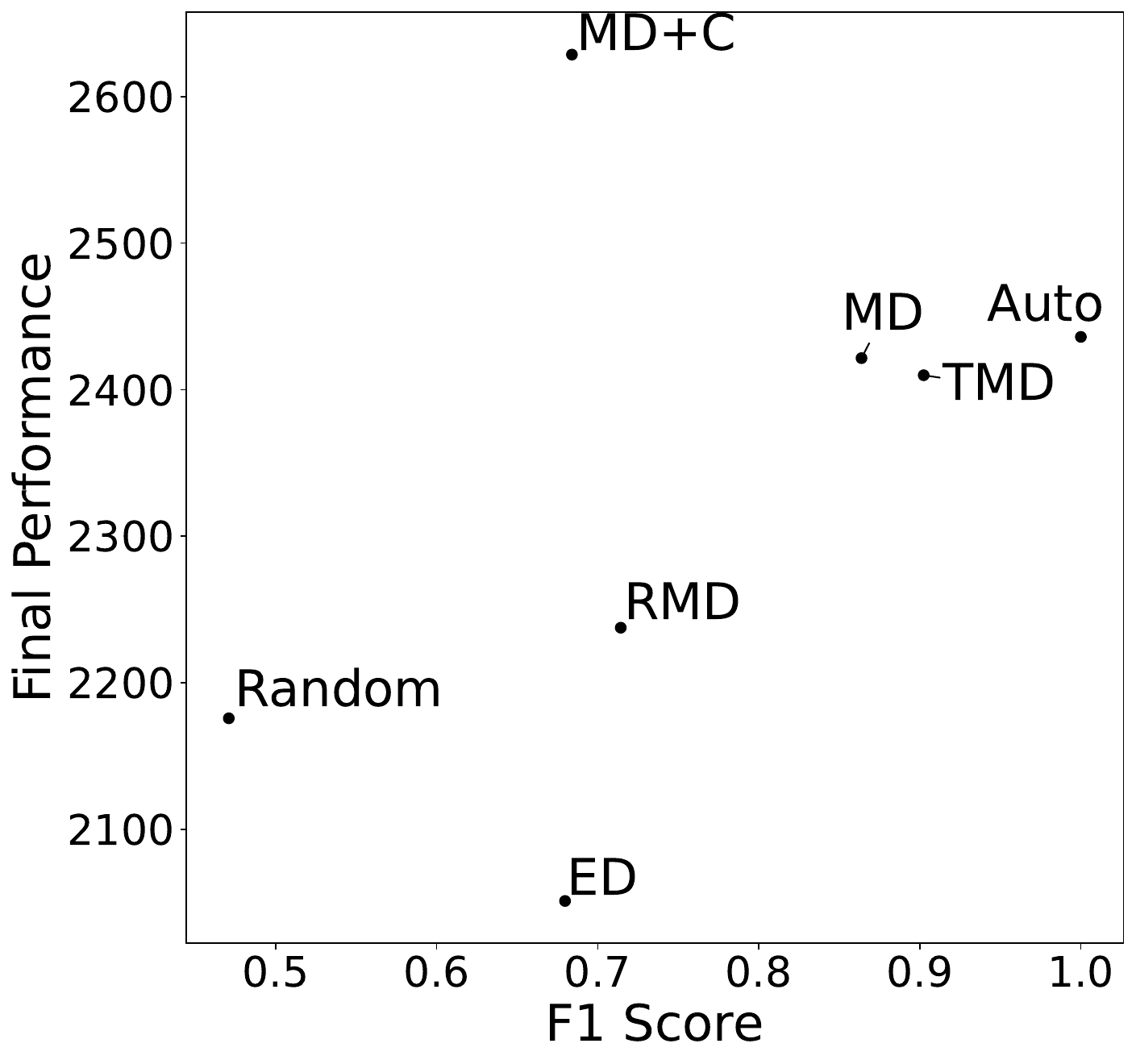}}
	\subfigure[Adversarial]{\includegraphics[width=0.19\textwidth]{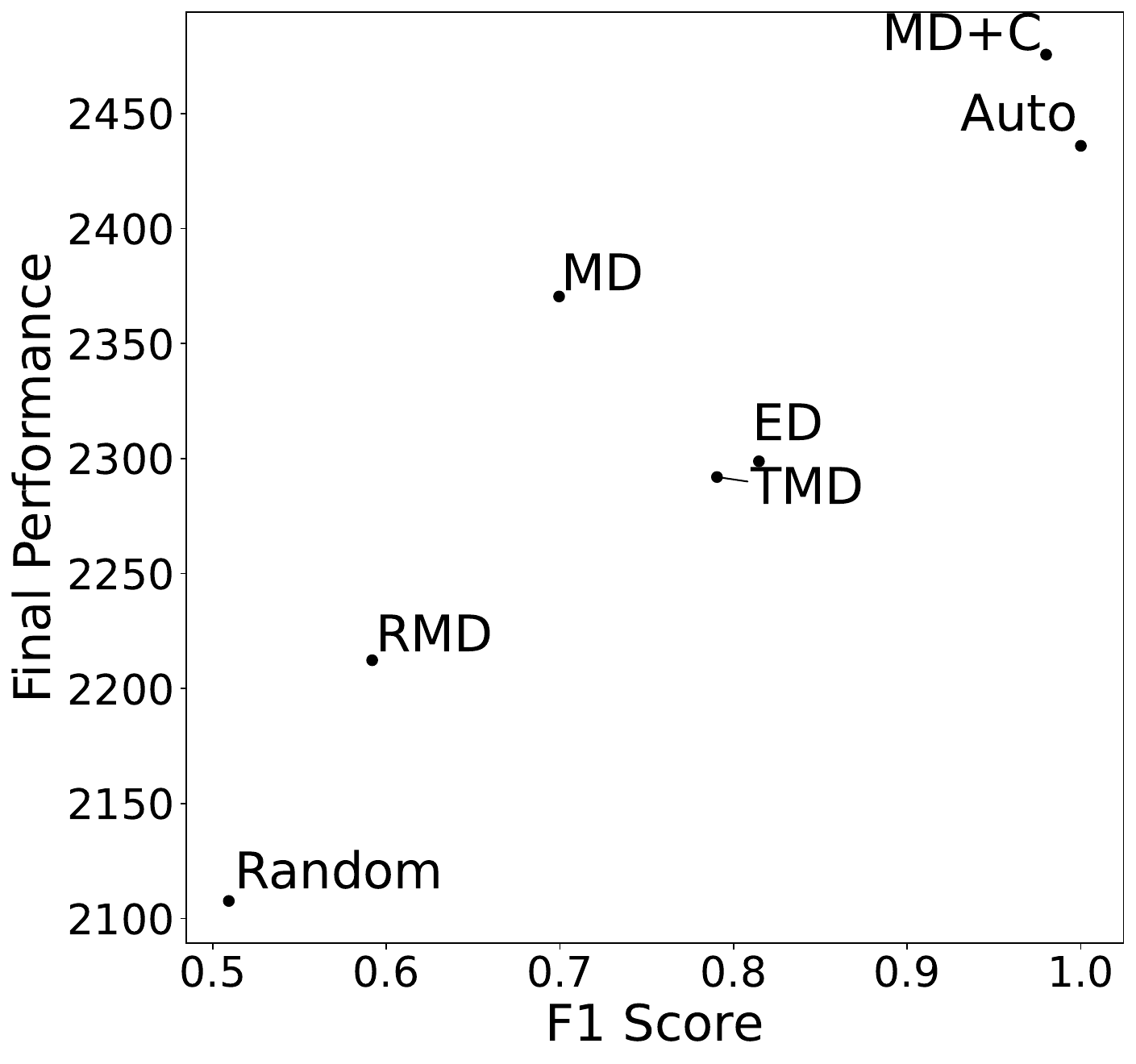}}
	\caption{Detection performance across various state outliers in the online training on Asterix.}
	\label{fig:Asterix_online_full}
\end{figure*}

\begin{figure*}[htbp]
	\centering
	\subfigure{\includegraphics[width=0.19\textwidth]{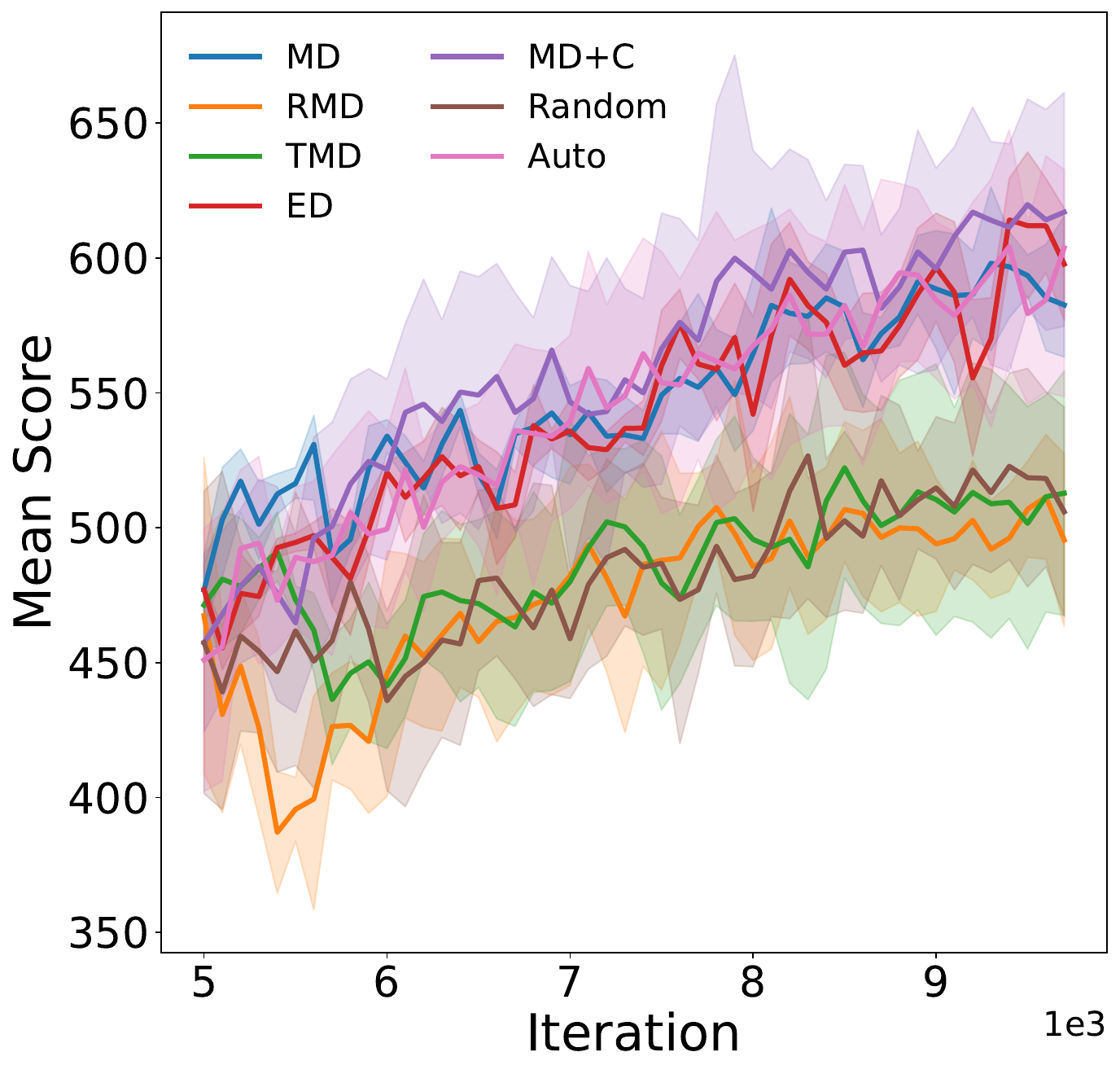}}
	\subfigure{\includegraphics[width=0.19\textwidth]{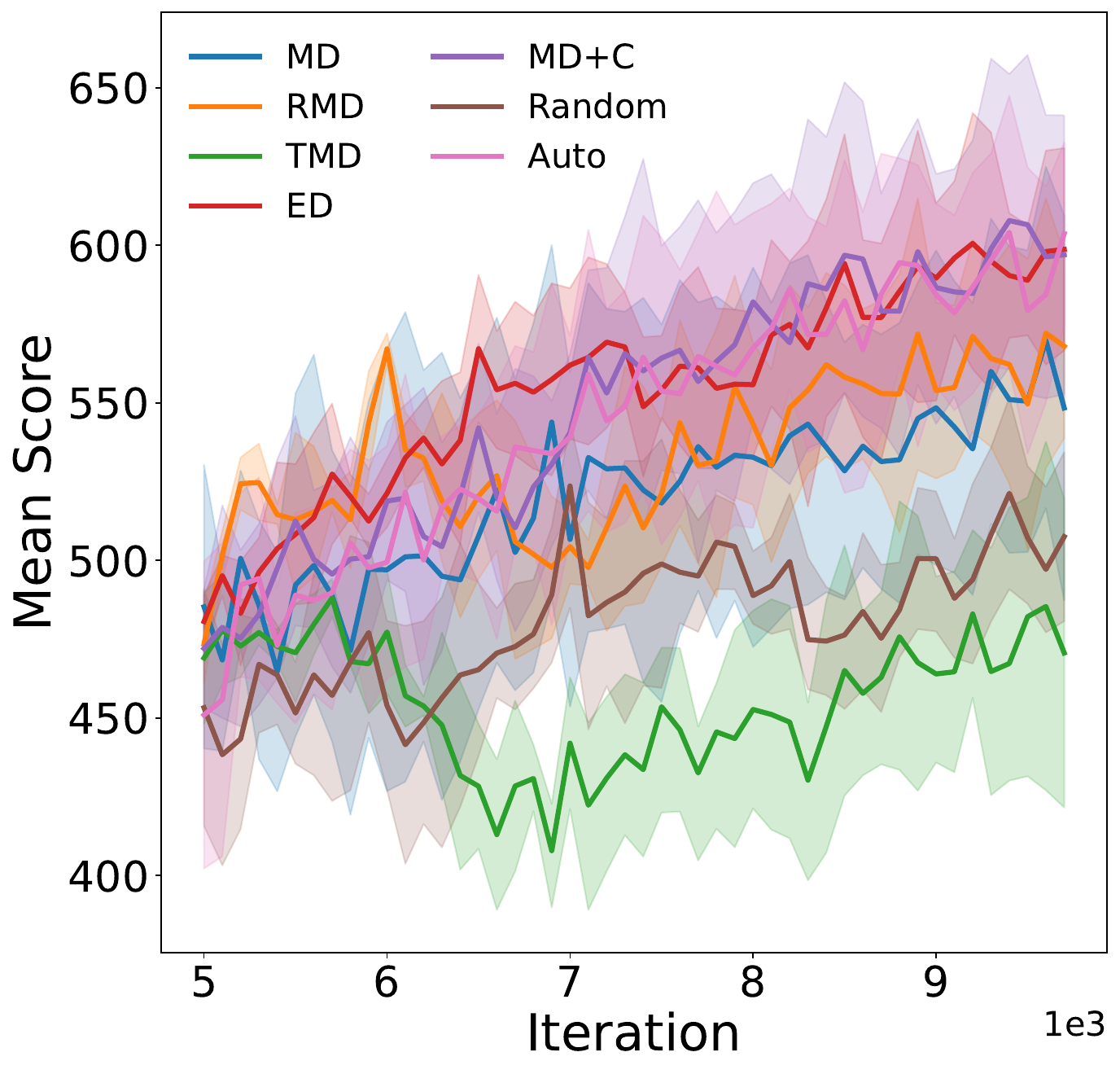}}
	\subfigure{\includegraphics[width=0.19\textwidth]{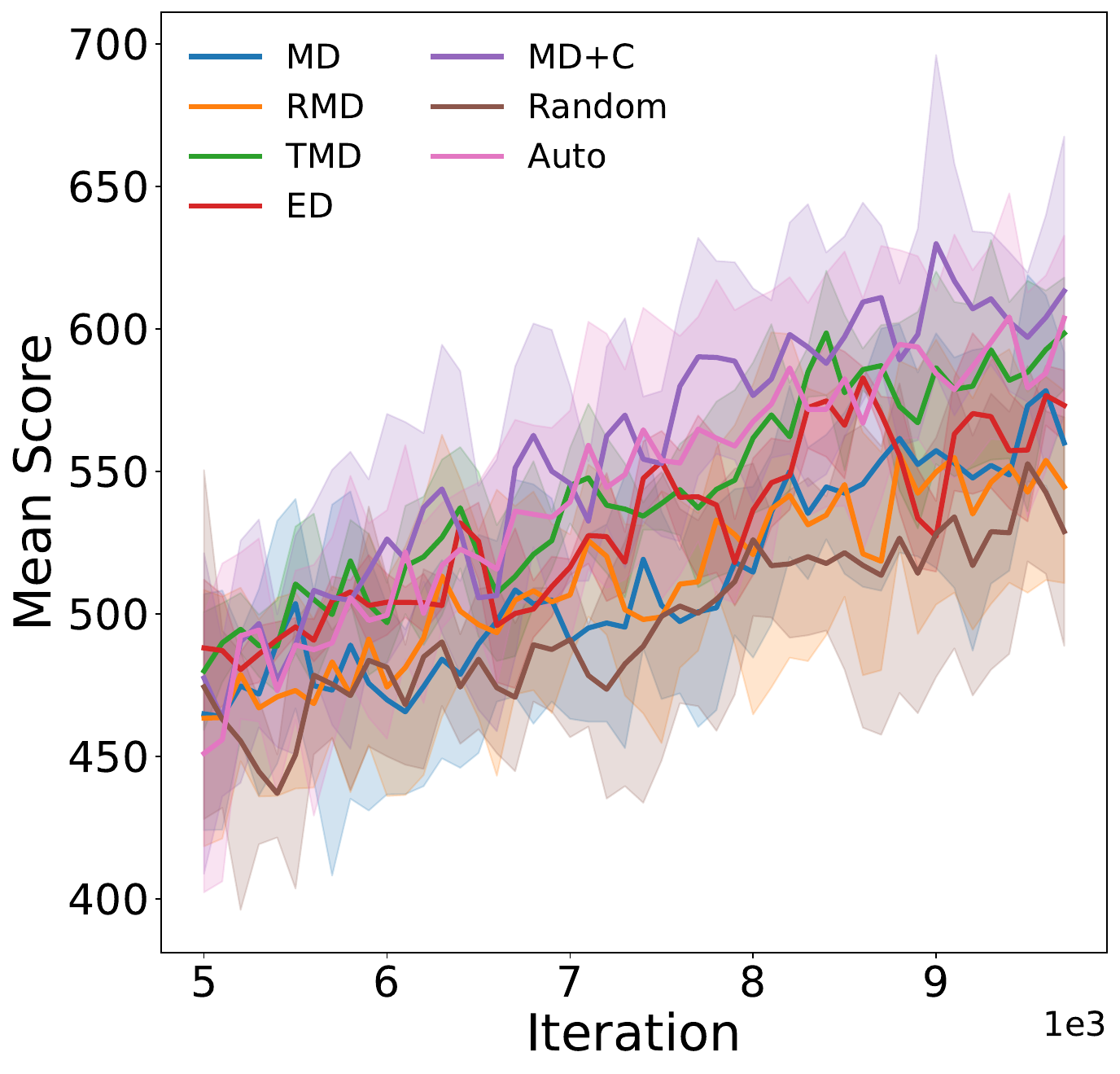}}
	\subfigure{\includegraphics[width=0.19\textwidth]{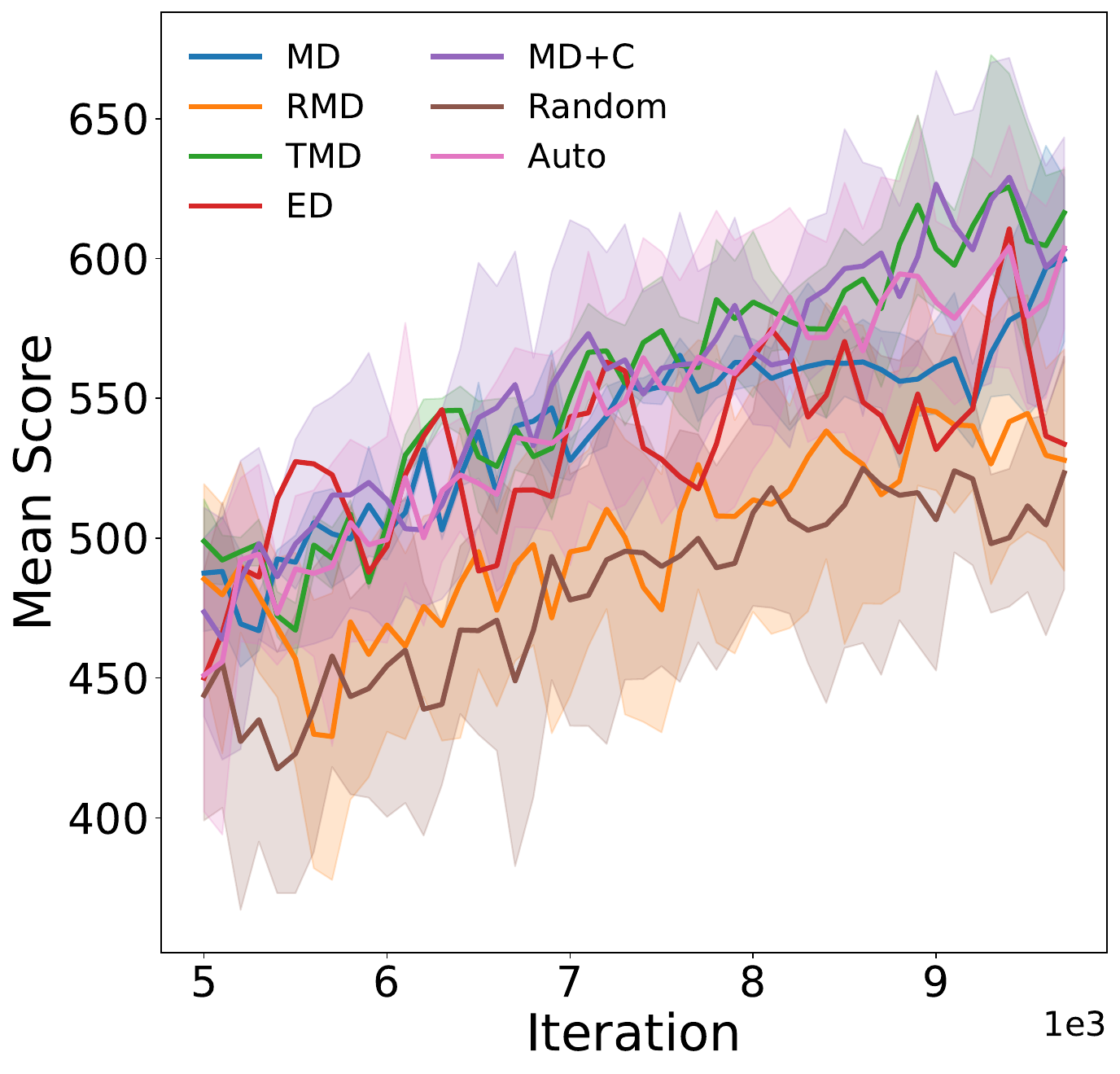}}
	\subfigure{\includegraphics[width=0.19\textwidth]{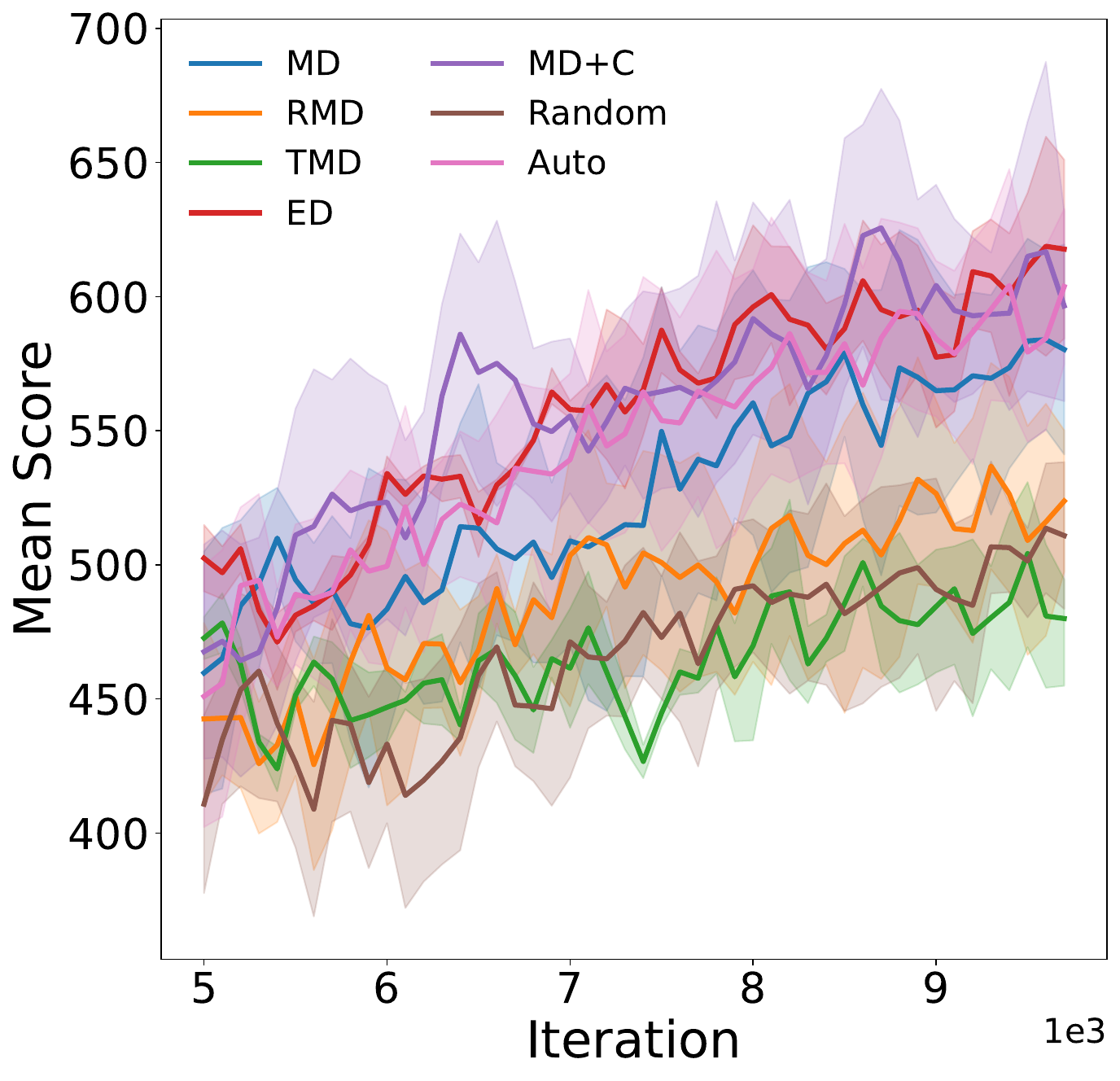}}
    \setcounter{subfigure}{0}
	\subfigure{\includegraphics[width=0.19\textwidth]{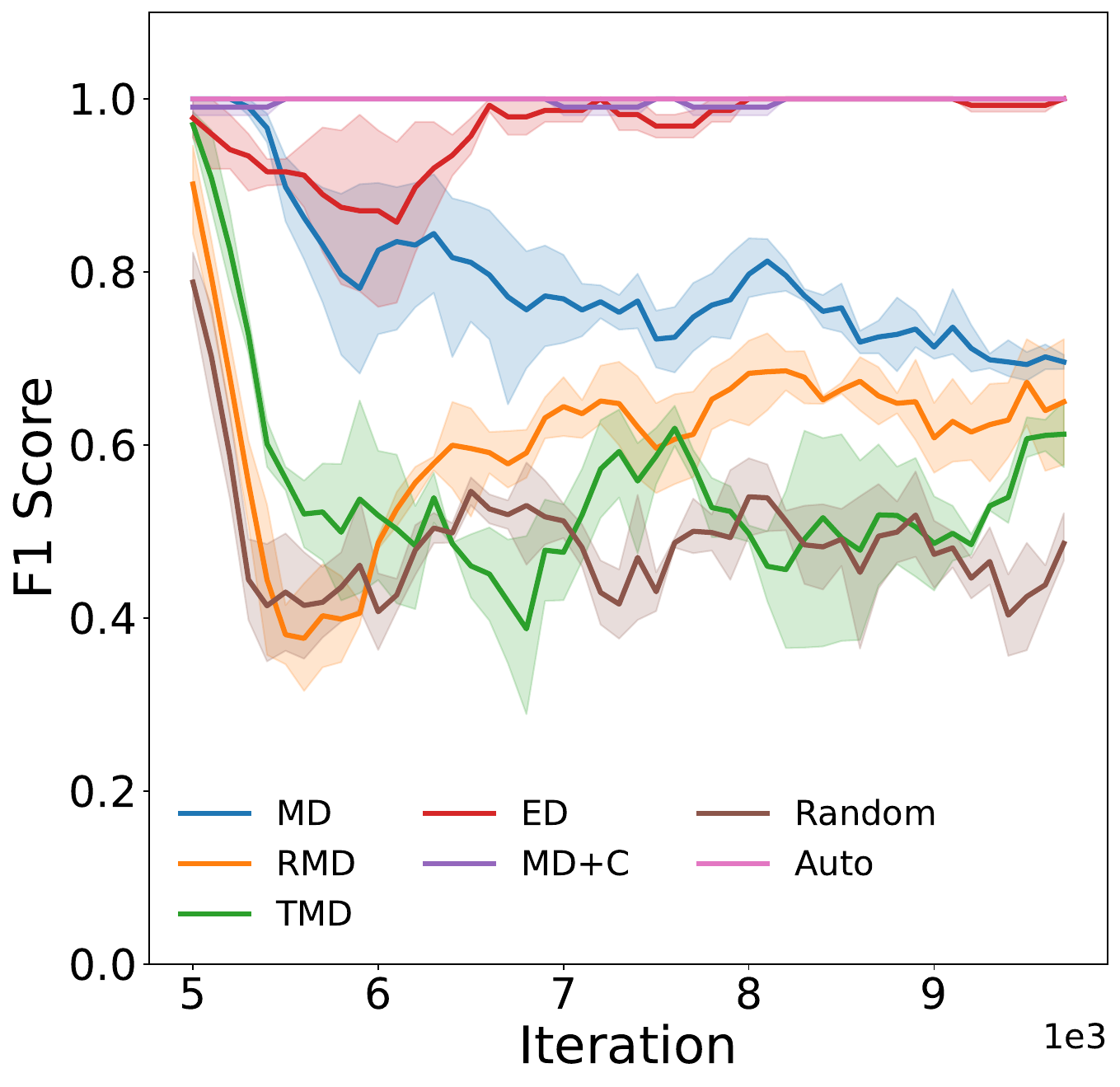}}
	\subfigure{\includegraphics[width=0.19\textwidth]{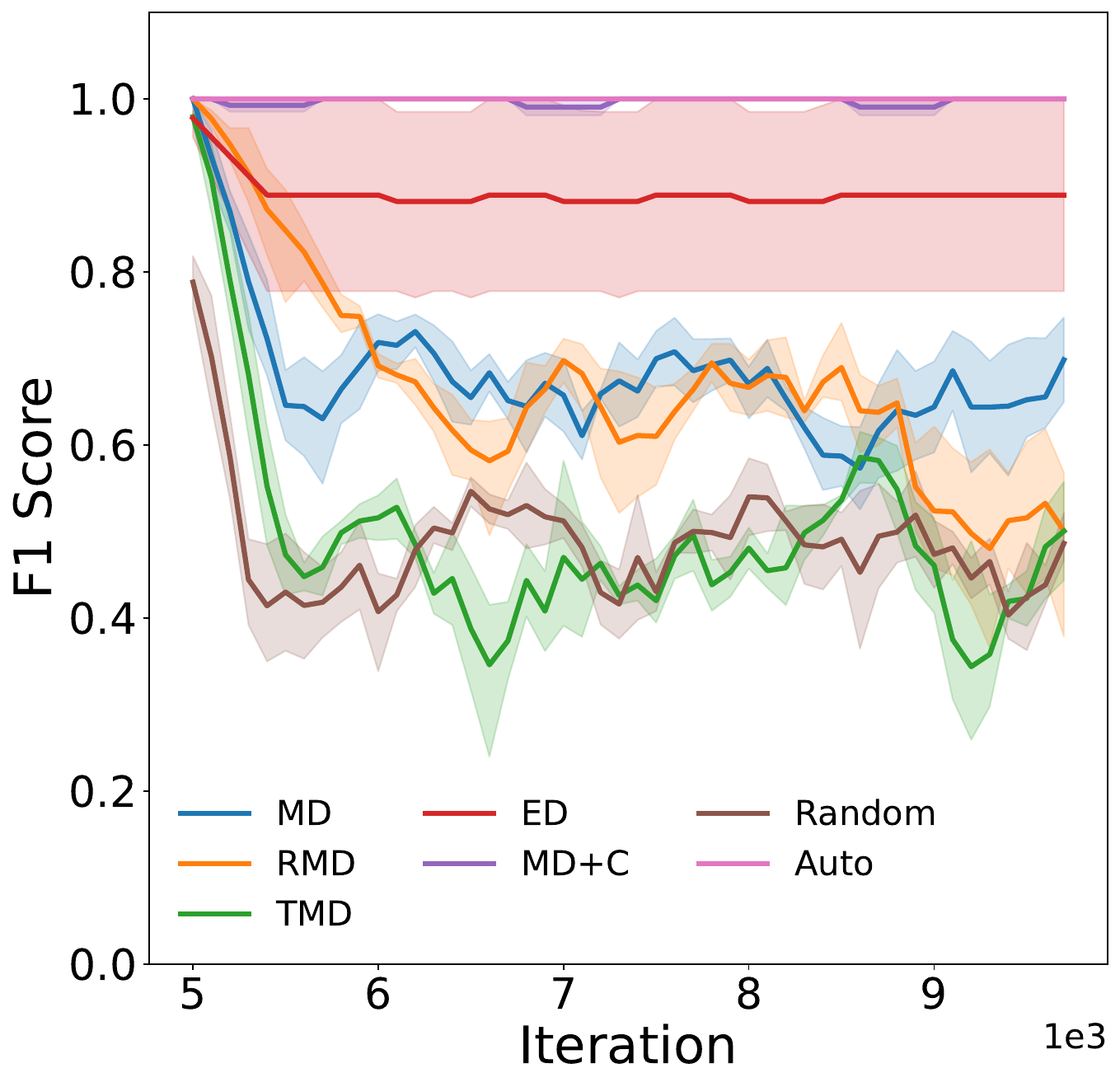}}
	\subfigure{\includegraphics[width=0.19\textwidth]{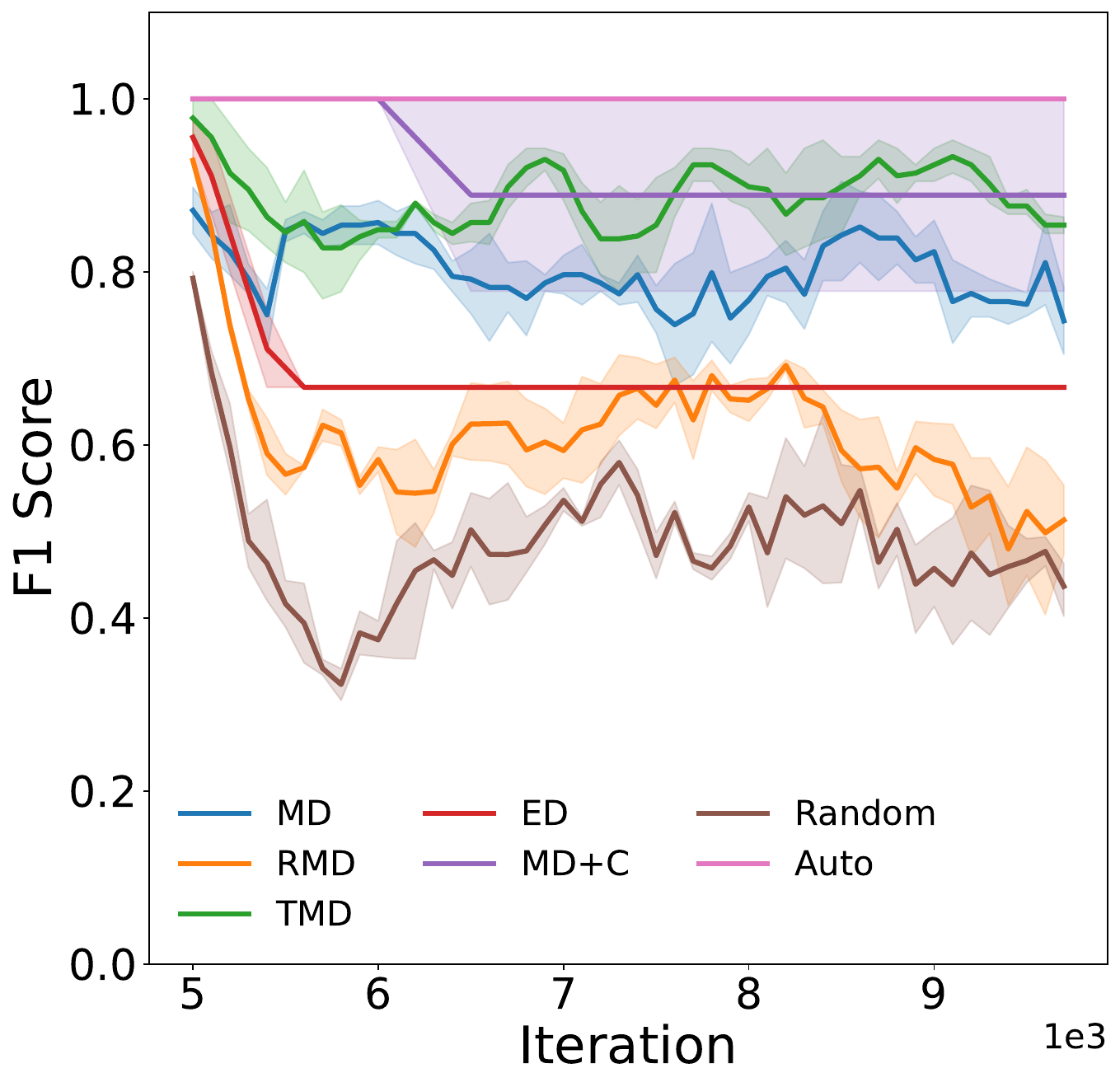}}
	\subfigure{\includegraphics[width=0.19\textwidth]{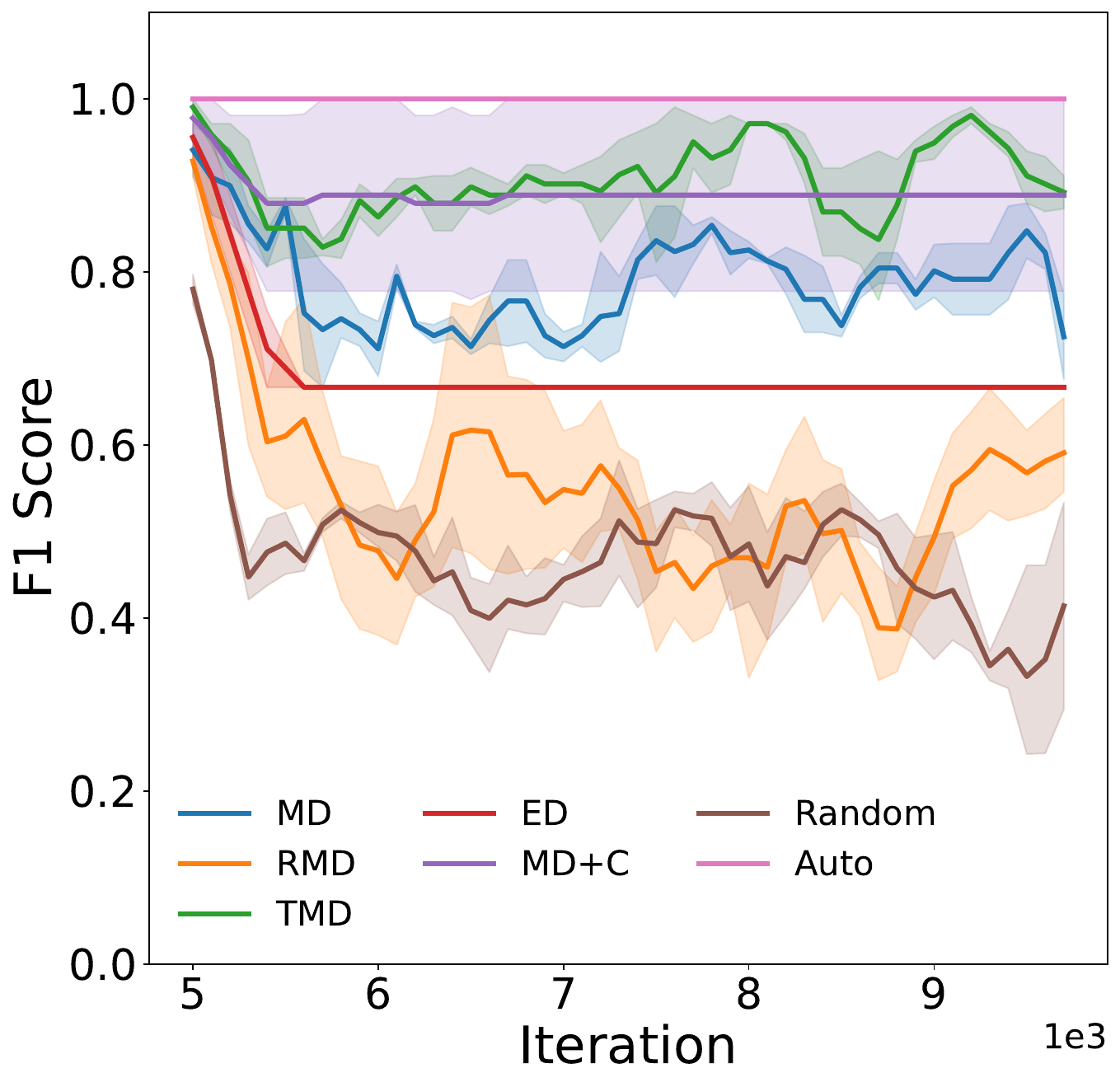}}
	\subfigure{\includegraphics[width=0.19\textwidth]{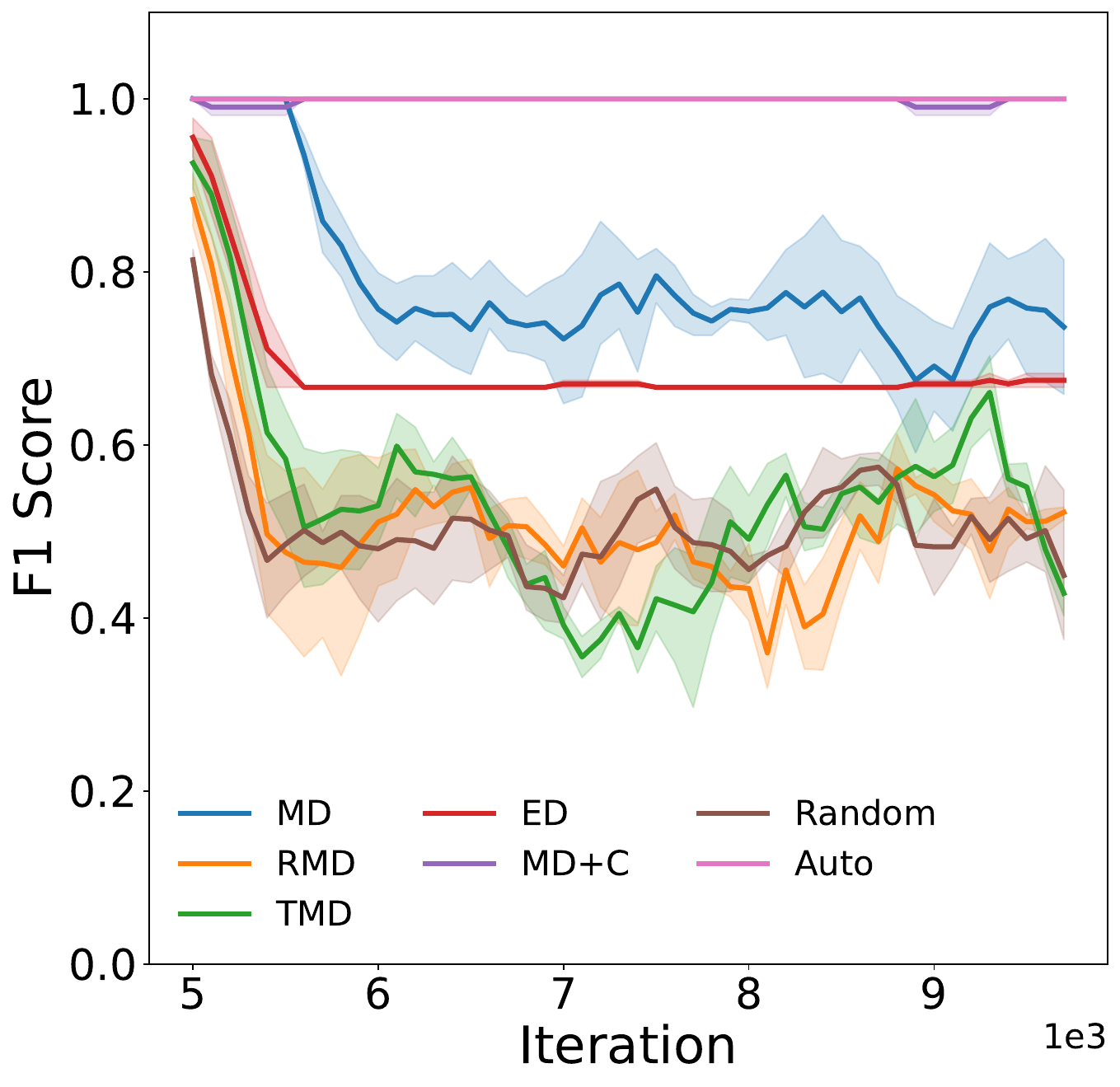}}
     \setcounter{subfigure}{0}
	\subfigure[Gaussian std=1]{\includegraphics[width=0.19\textwidth]{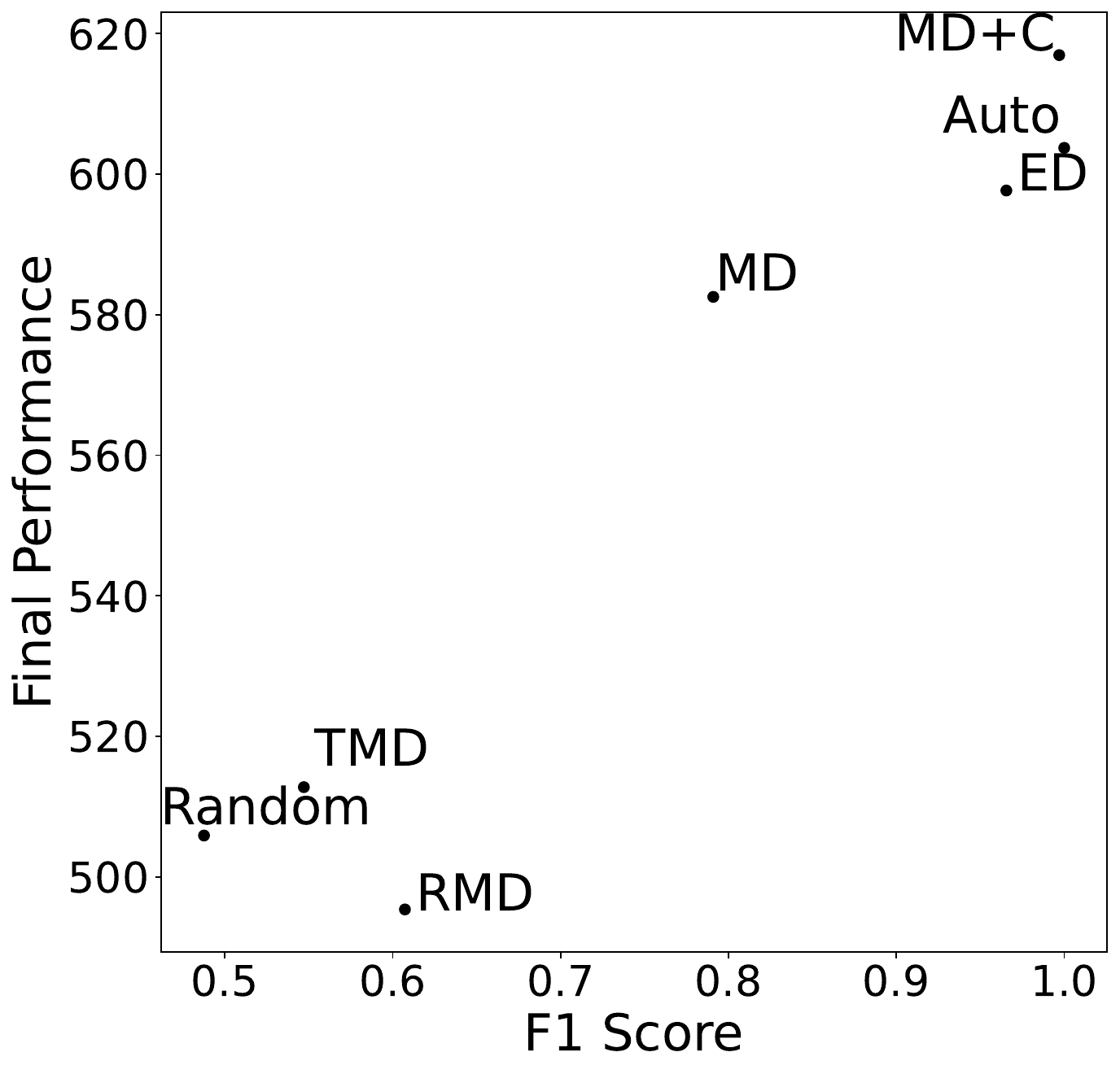}}
	\subfigure[Gaussian std=0.3]{\includegraphics[width=0.19\textwidth]{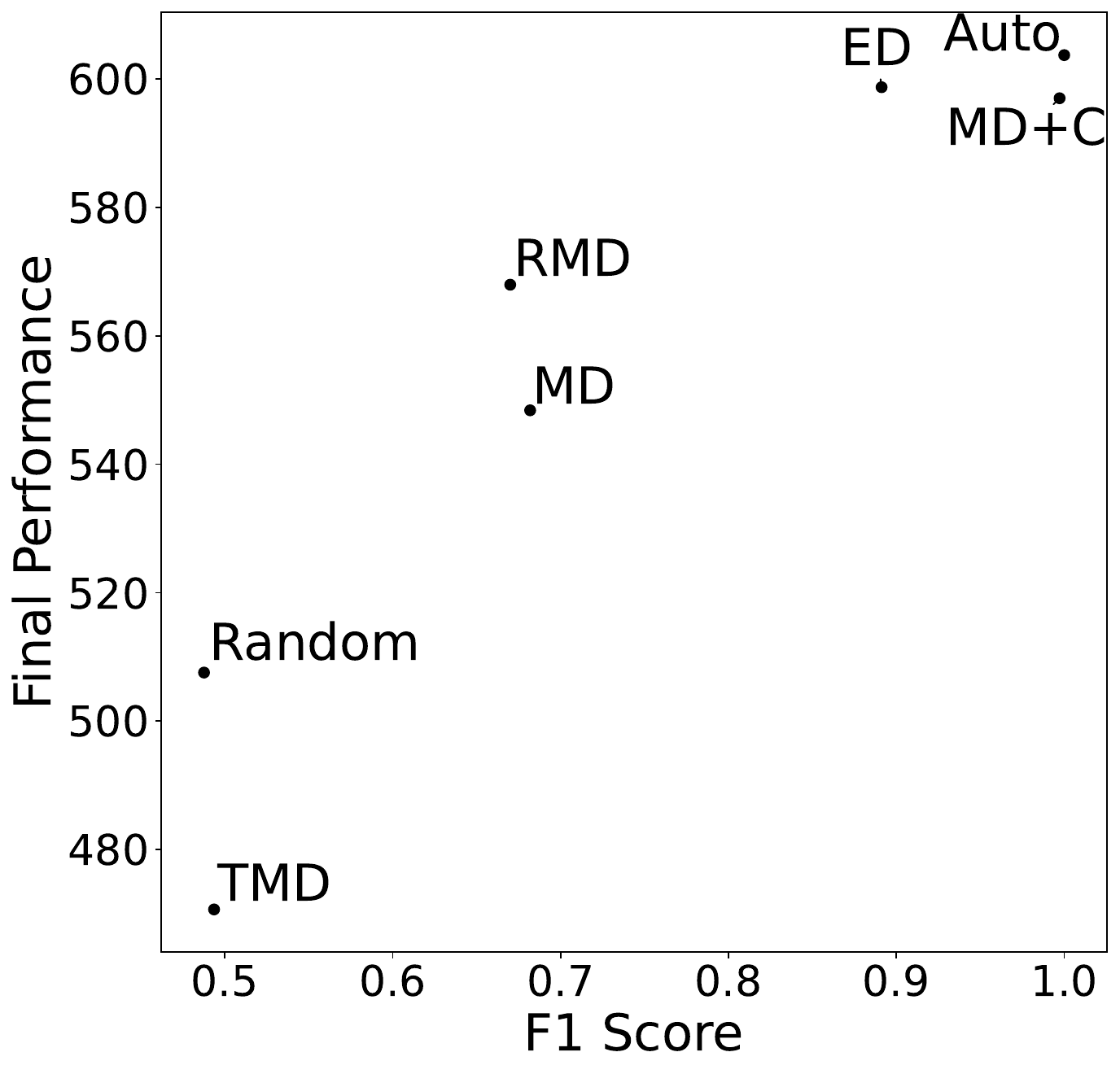}}
	\subfigure[OOD Breakout]{\includegraphics[width=0.19\textwidth]{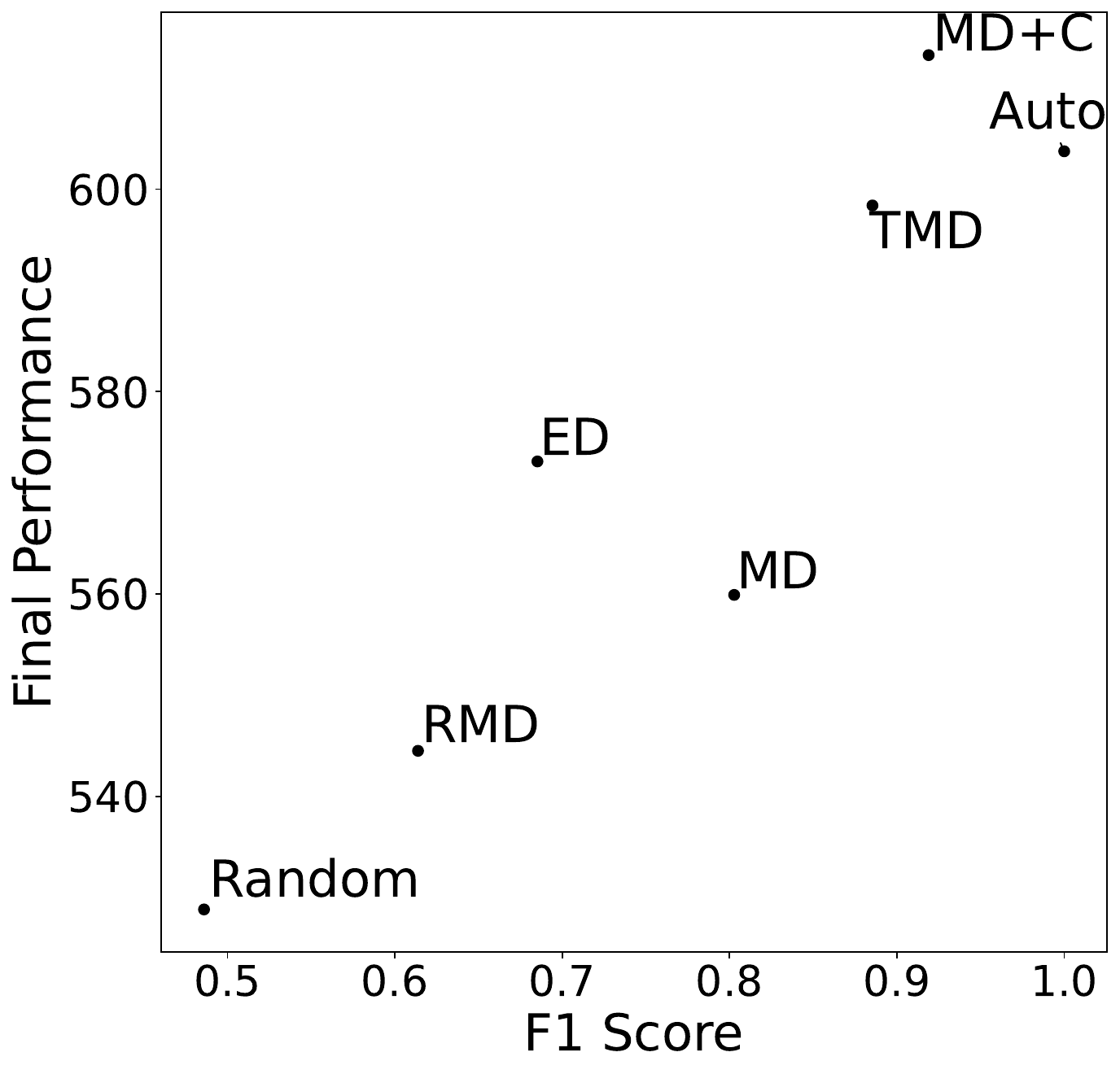}}
	\subfigure[OOD Asterix]{\includegraphics[width=0.19\textwidth]{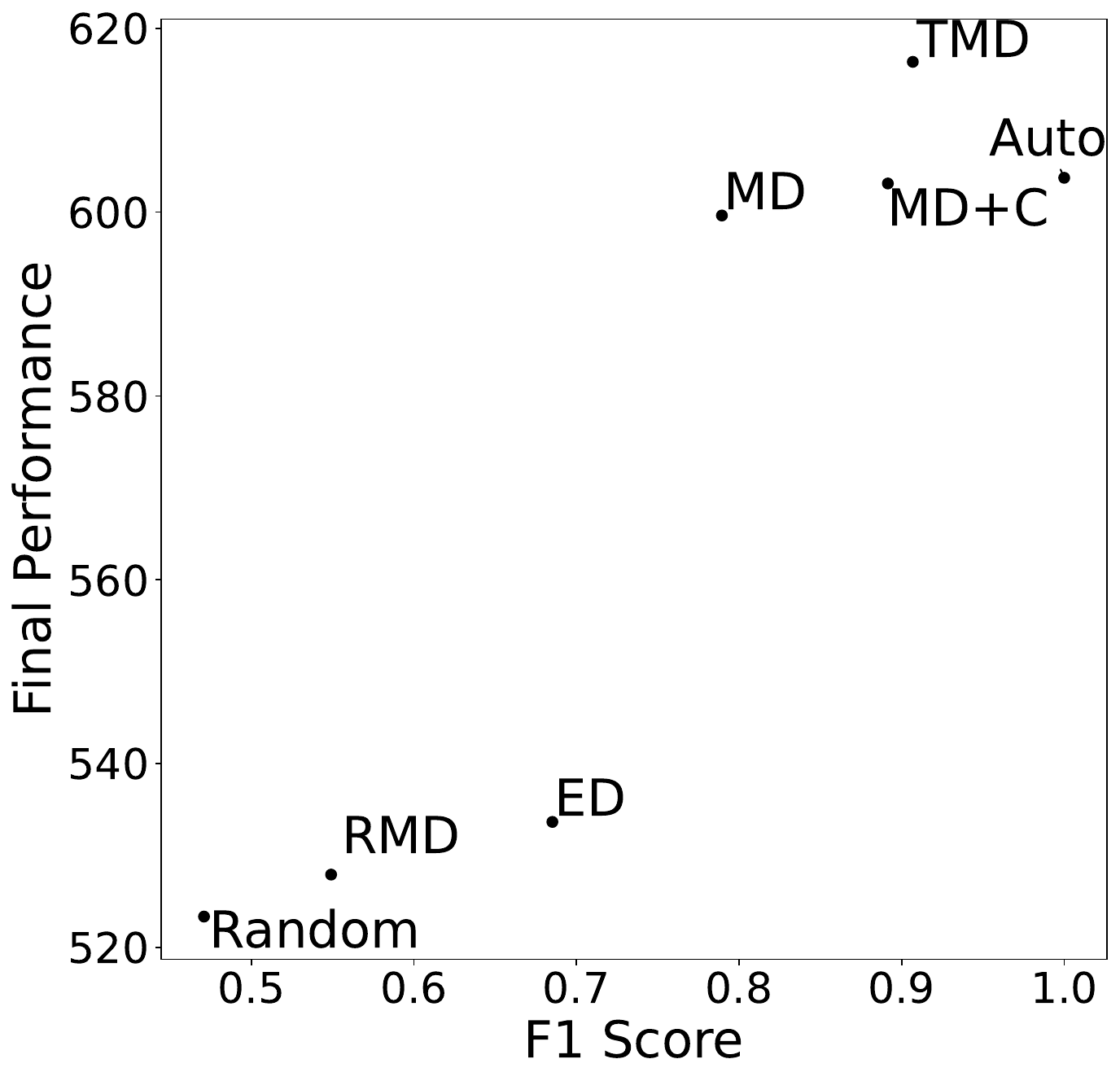}}
	\subfigure[Adversarial]{\includegraphics[width=0.19\textwidth]{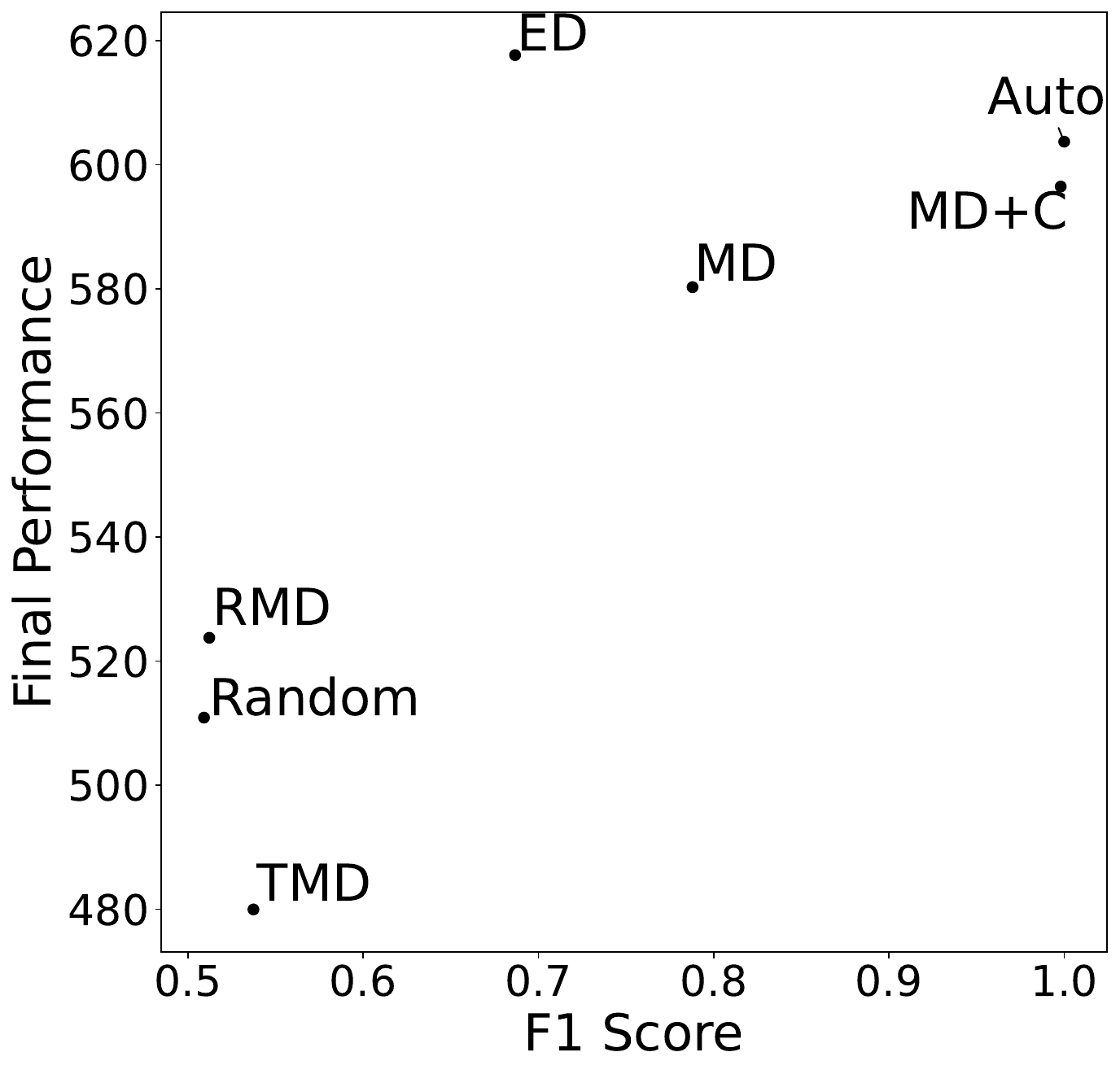}}
	\caption{Detection performance across various state outliers in the online training on SpaceInvaders.}
	\label{fig:SpaceInvaders_online_full}
\end{figure*}

\begin{figure*}[htbp]
	\centering
	\subfigure{\includegraphics[width=0.19\textwidth]{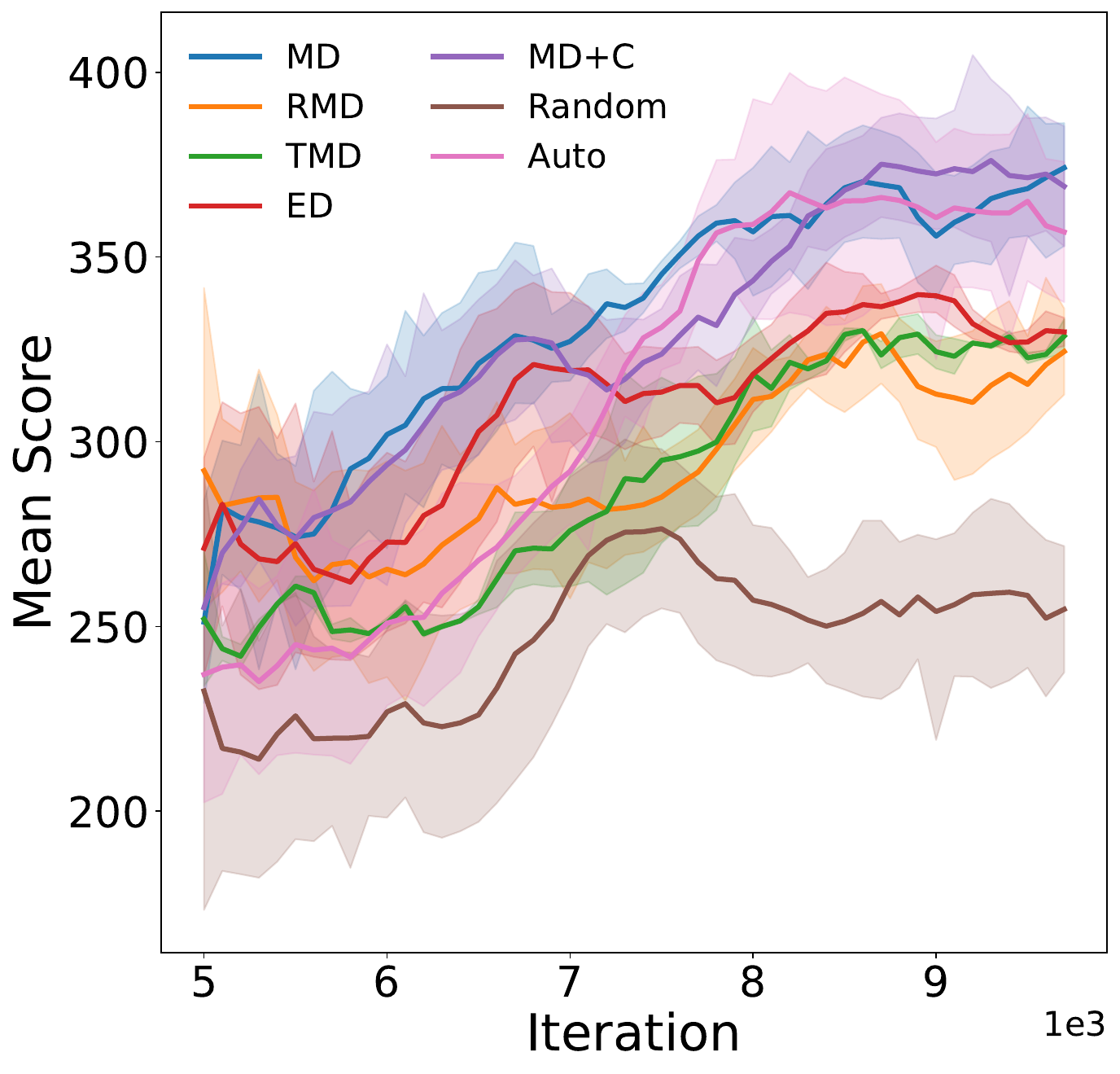}}
	\subfigure{\includegraphics[width=0.19\textwidth]{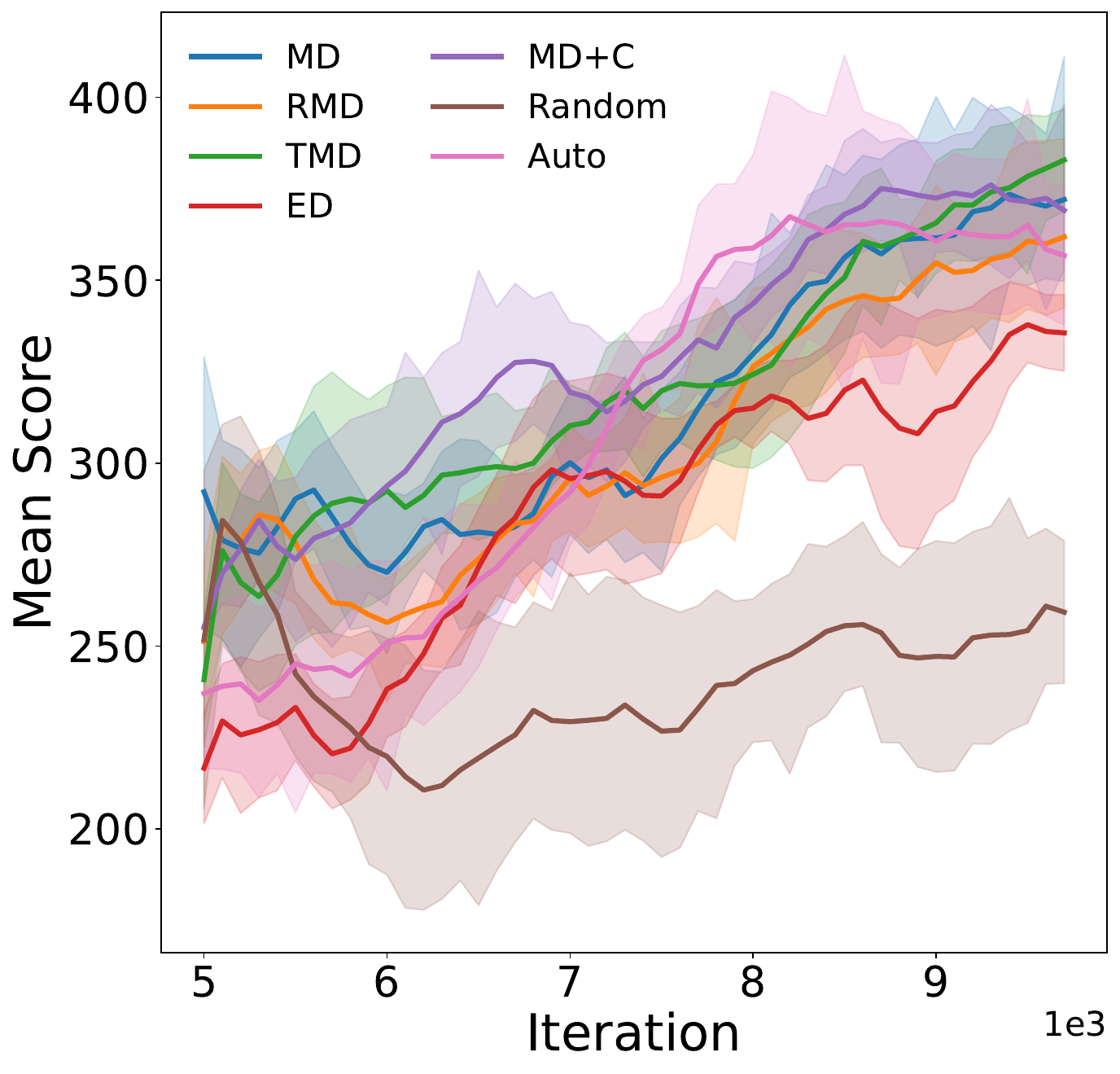}}
	\subfigure{\includegraphics[width=0.19\textwidth]{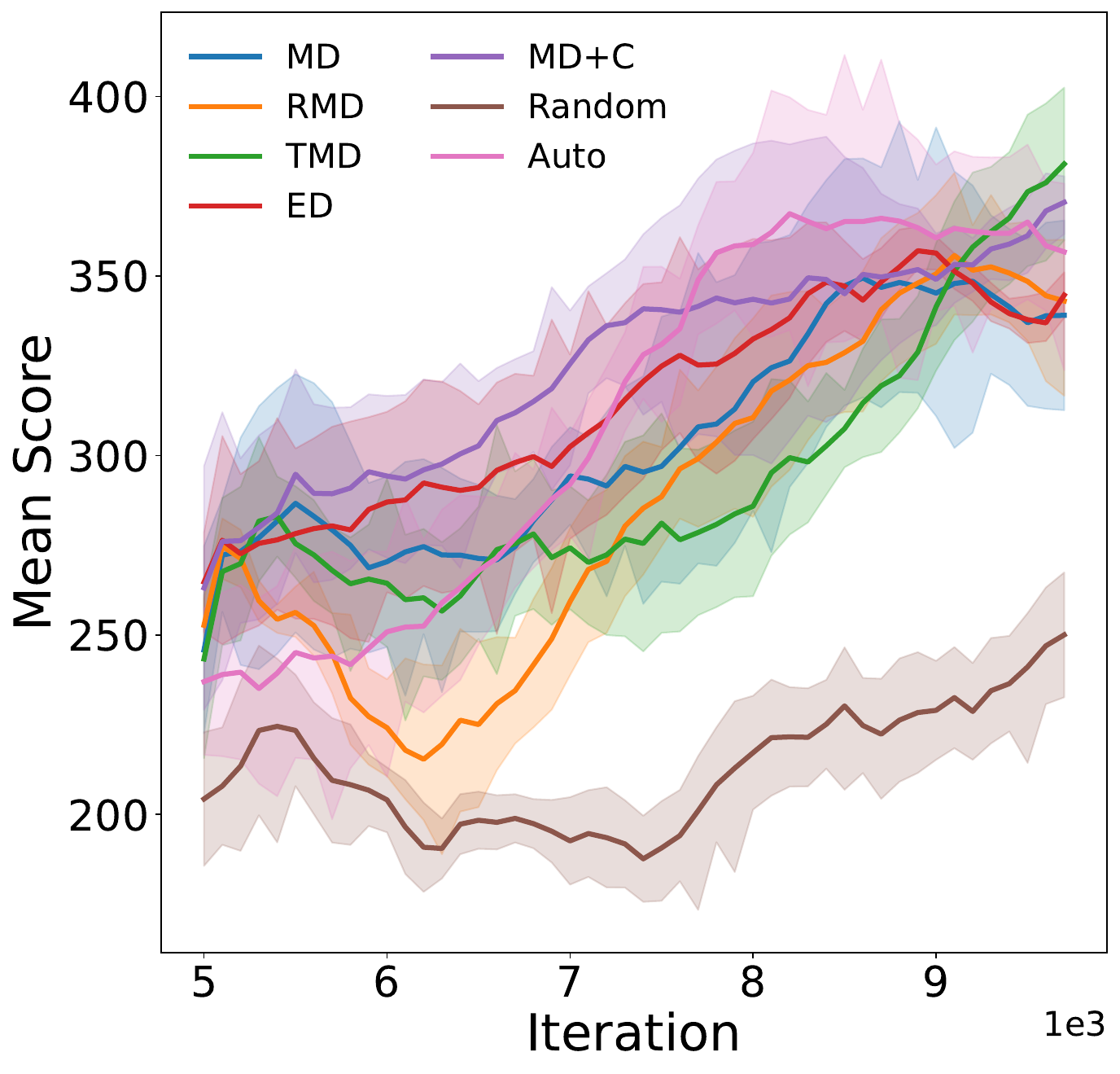}}
	\subfigure{\includegraphics[width=0.19\textwidth]{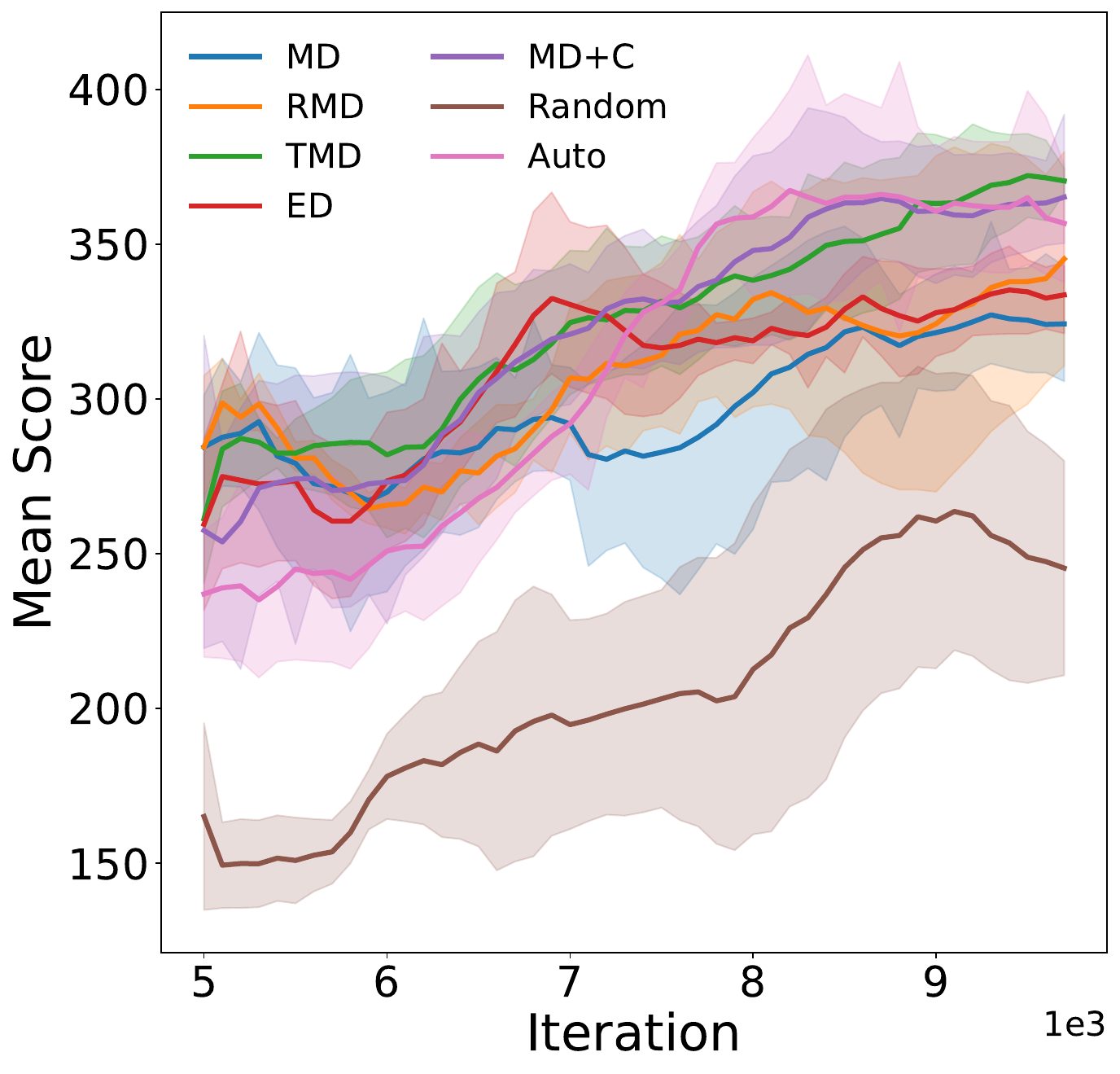}}
	\subfigure{\includegraphics[width=0.19\textwidth]{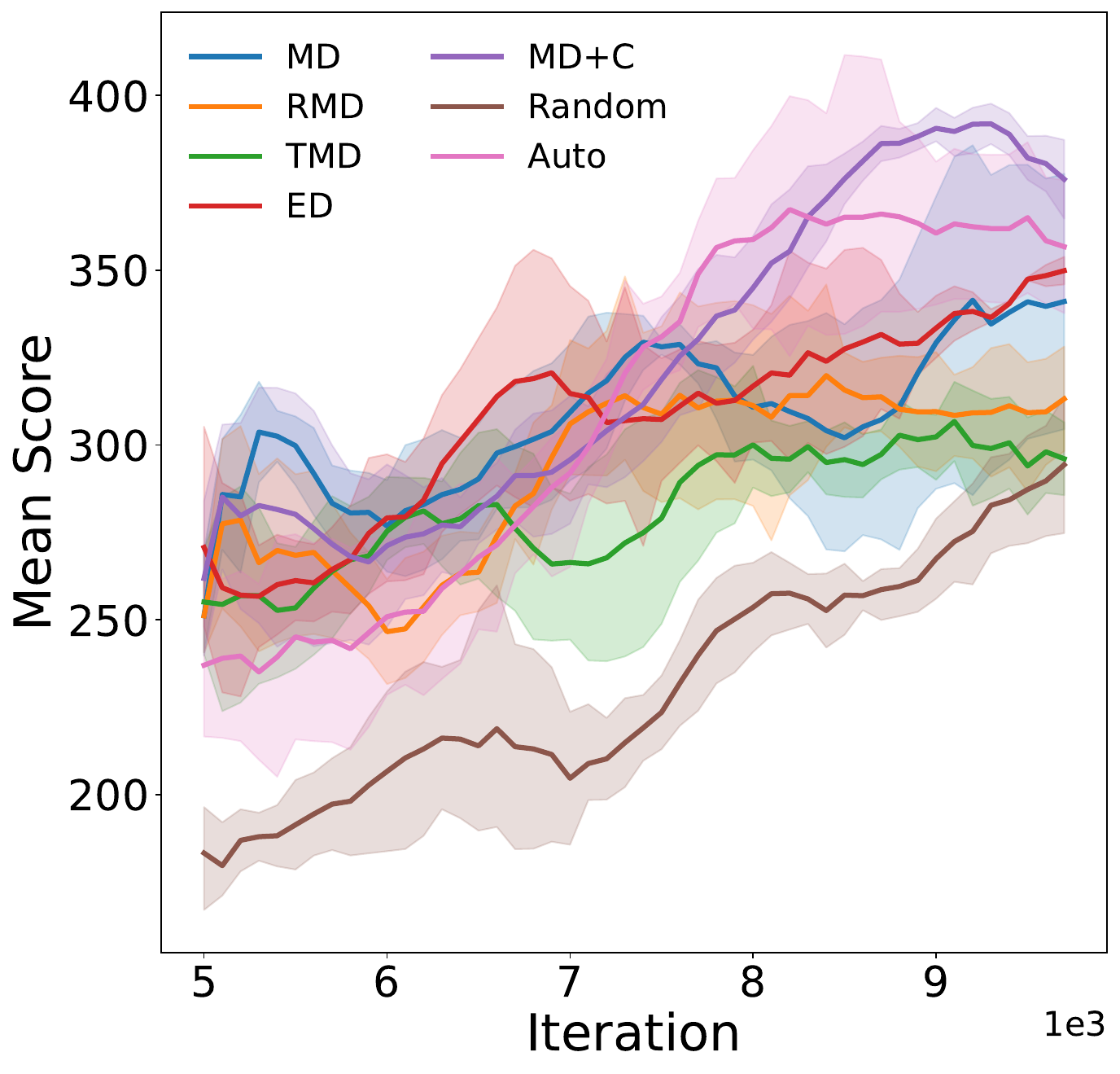}}
    \setcounter{subfigure}{0}
	\subfigure{\includegraphics[width=0.19\textwidth]{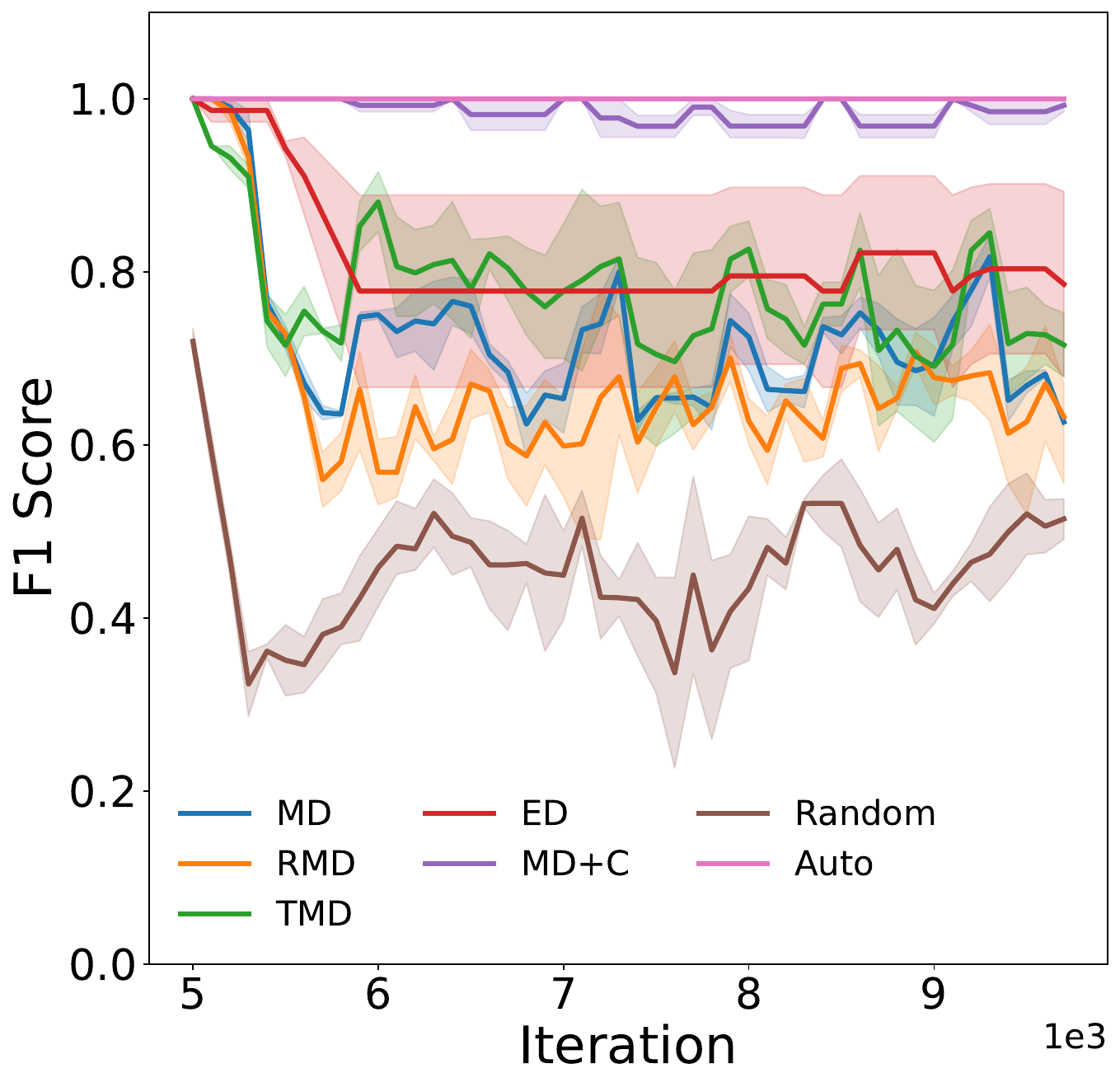}}
	\subfigure{\includegraphics[width=0.19\textwidth]{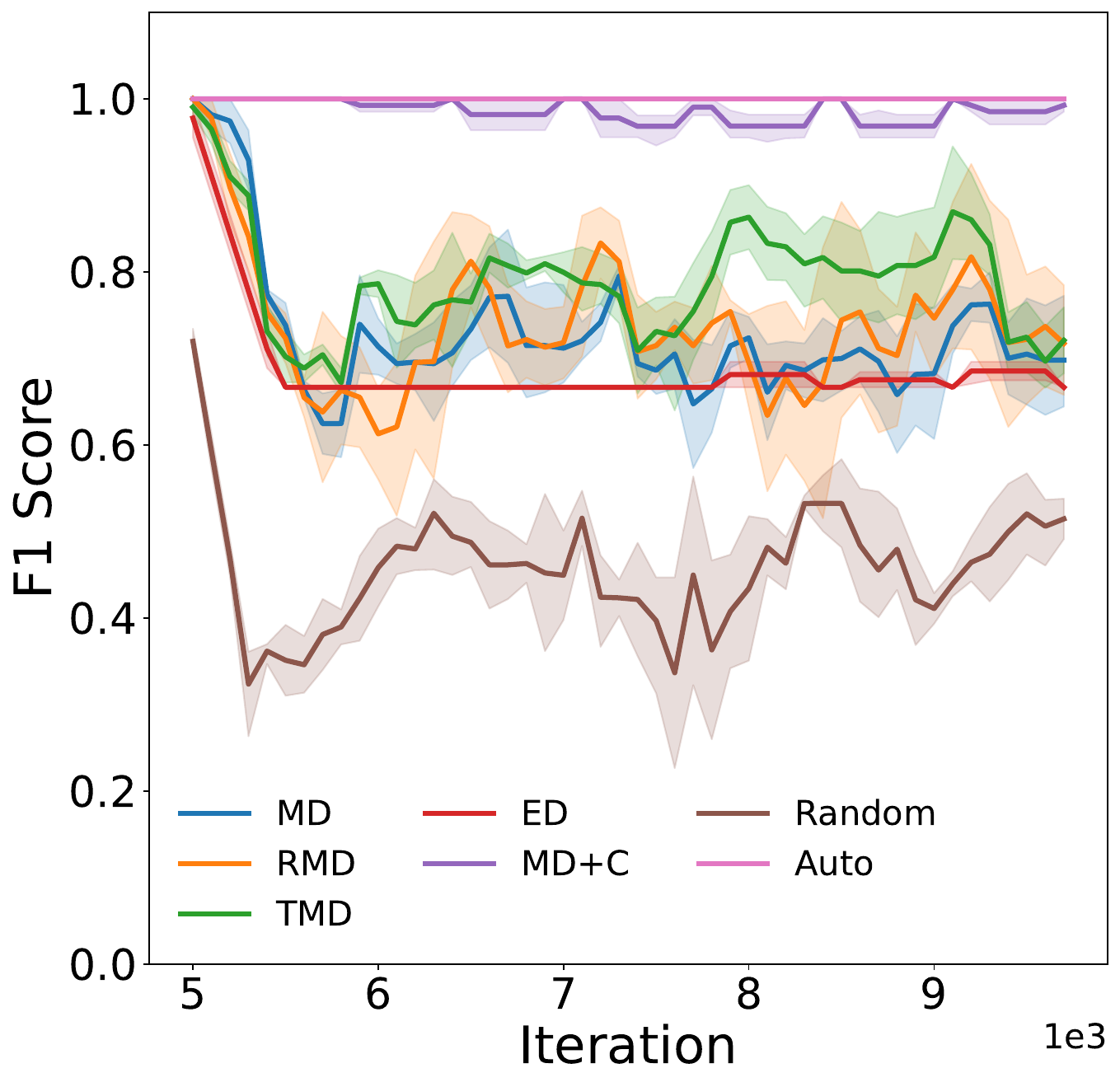}}
	\subfigure{\includegraphics[width=0.19\textwidth]{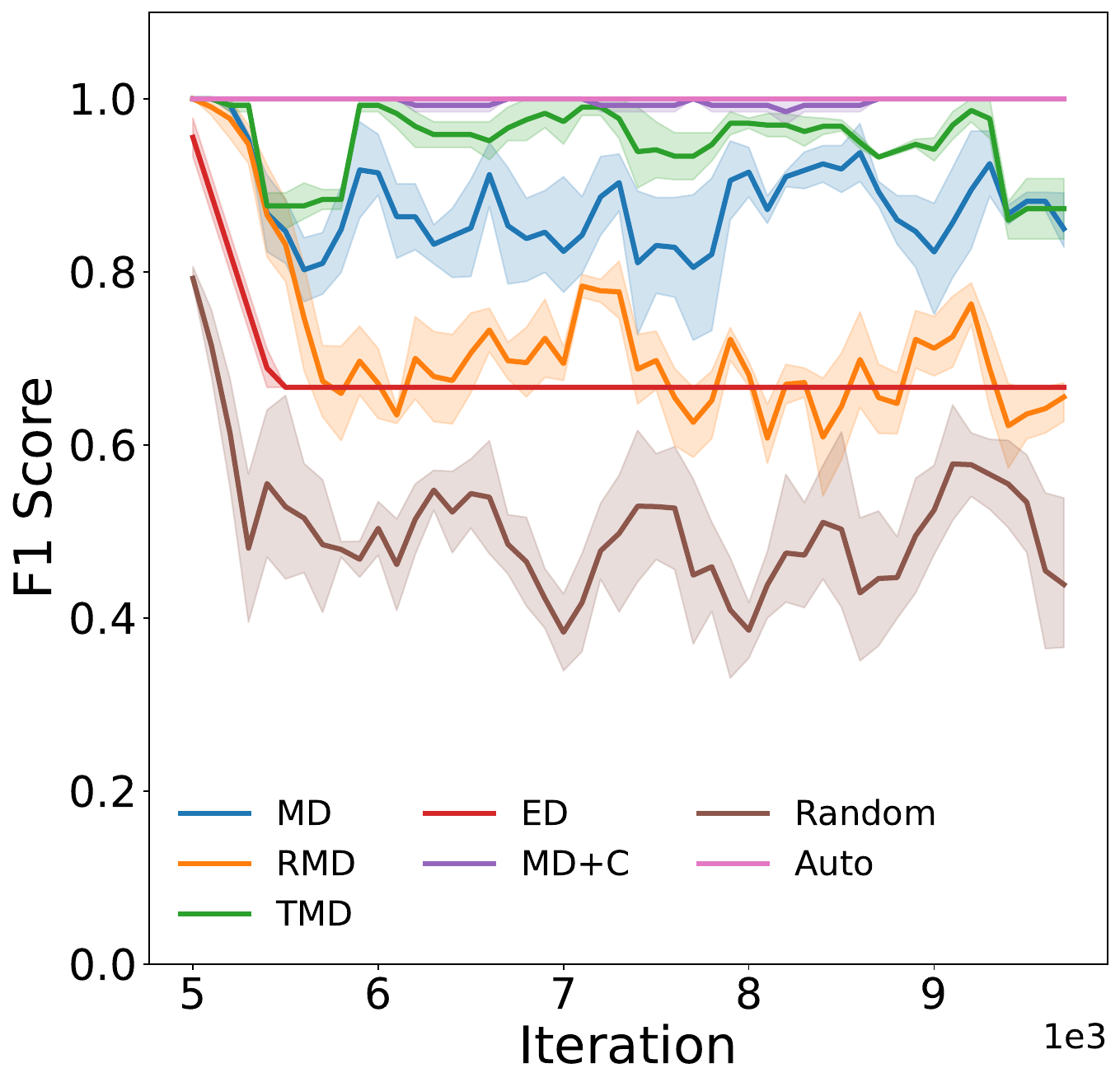}}
	\subfigure{\includegraphics[width=0.19\textwidth]{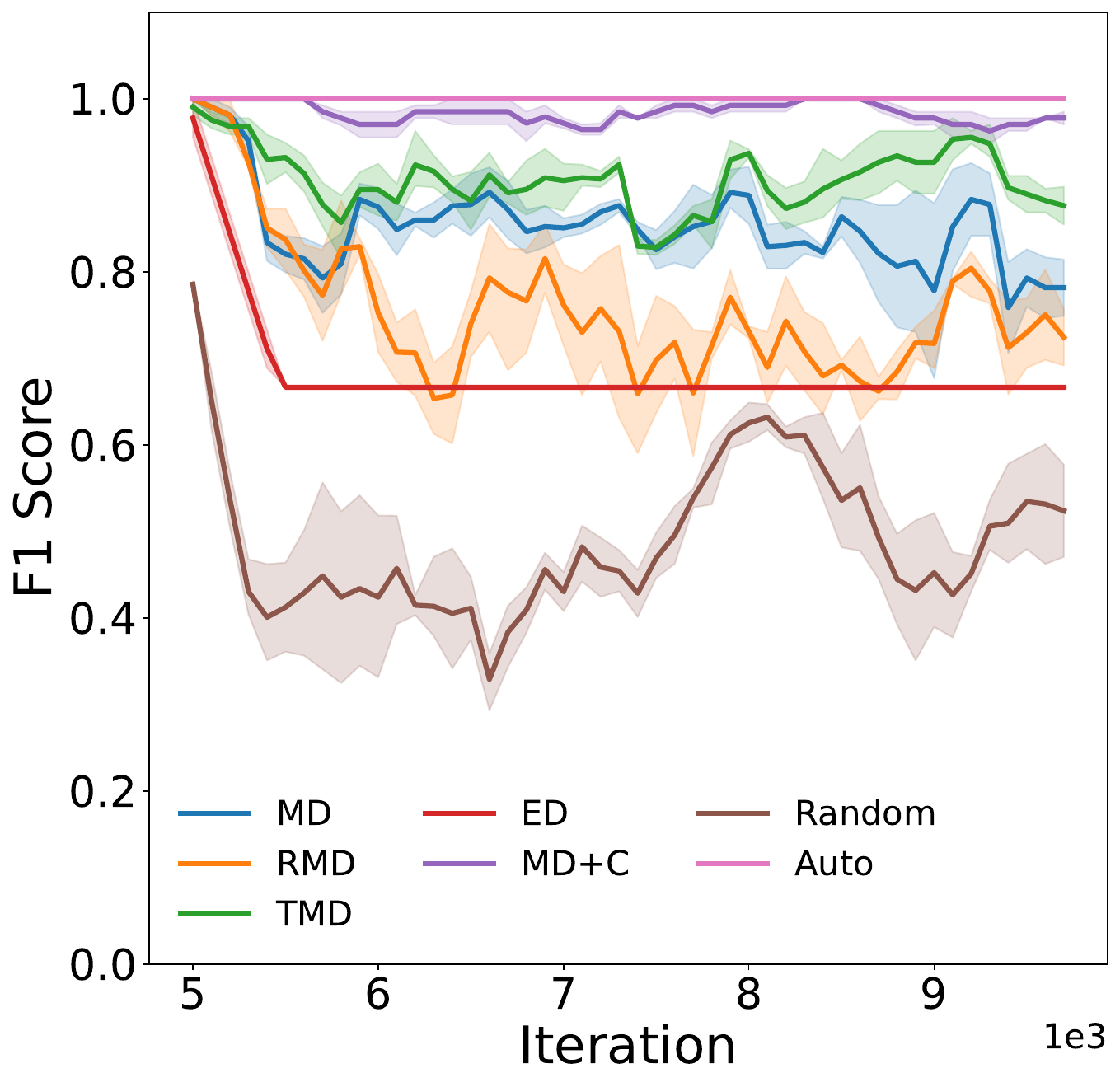}}
	\subfigure{\includegraphics[width=0.19\textwidth]{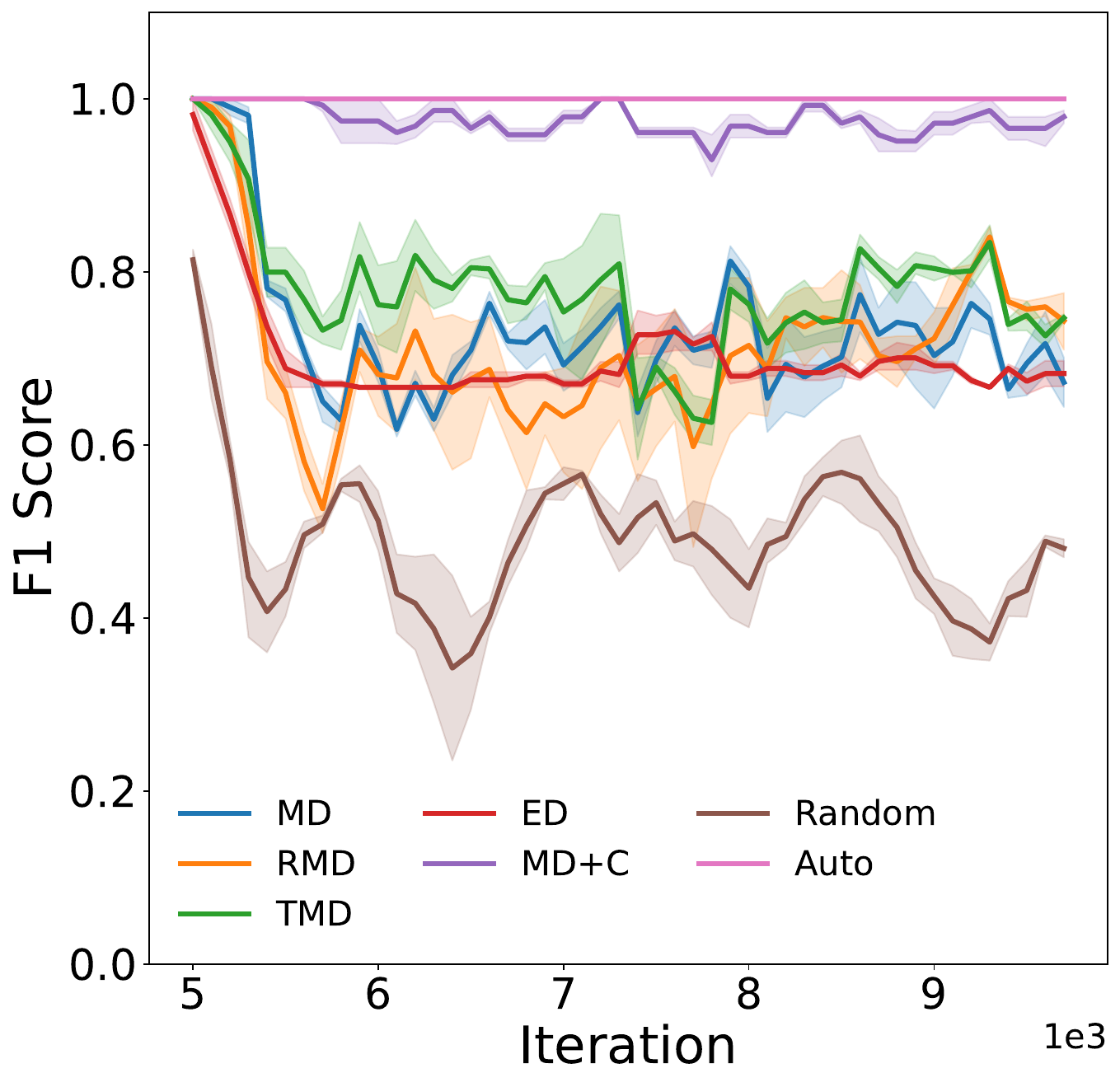}}
     \setcounter{subfigure}{0}
	\subfigure[Gaussian std=1]{\includegraphics[width=0.19\textwidth]{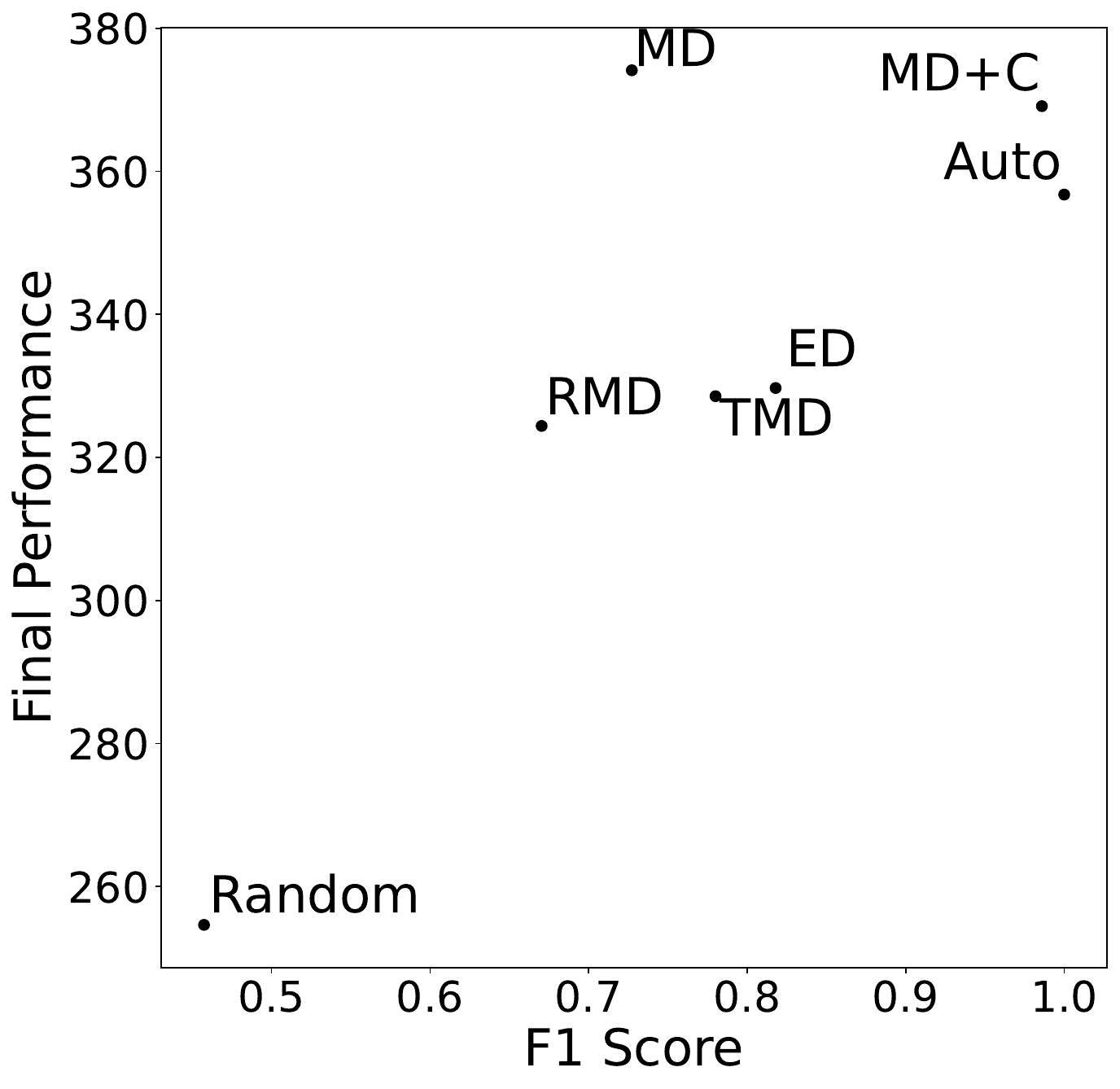}}
	\subfigure[Gaussian std=0.3]{\includegraphics[width=0.19\textwidth]{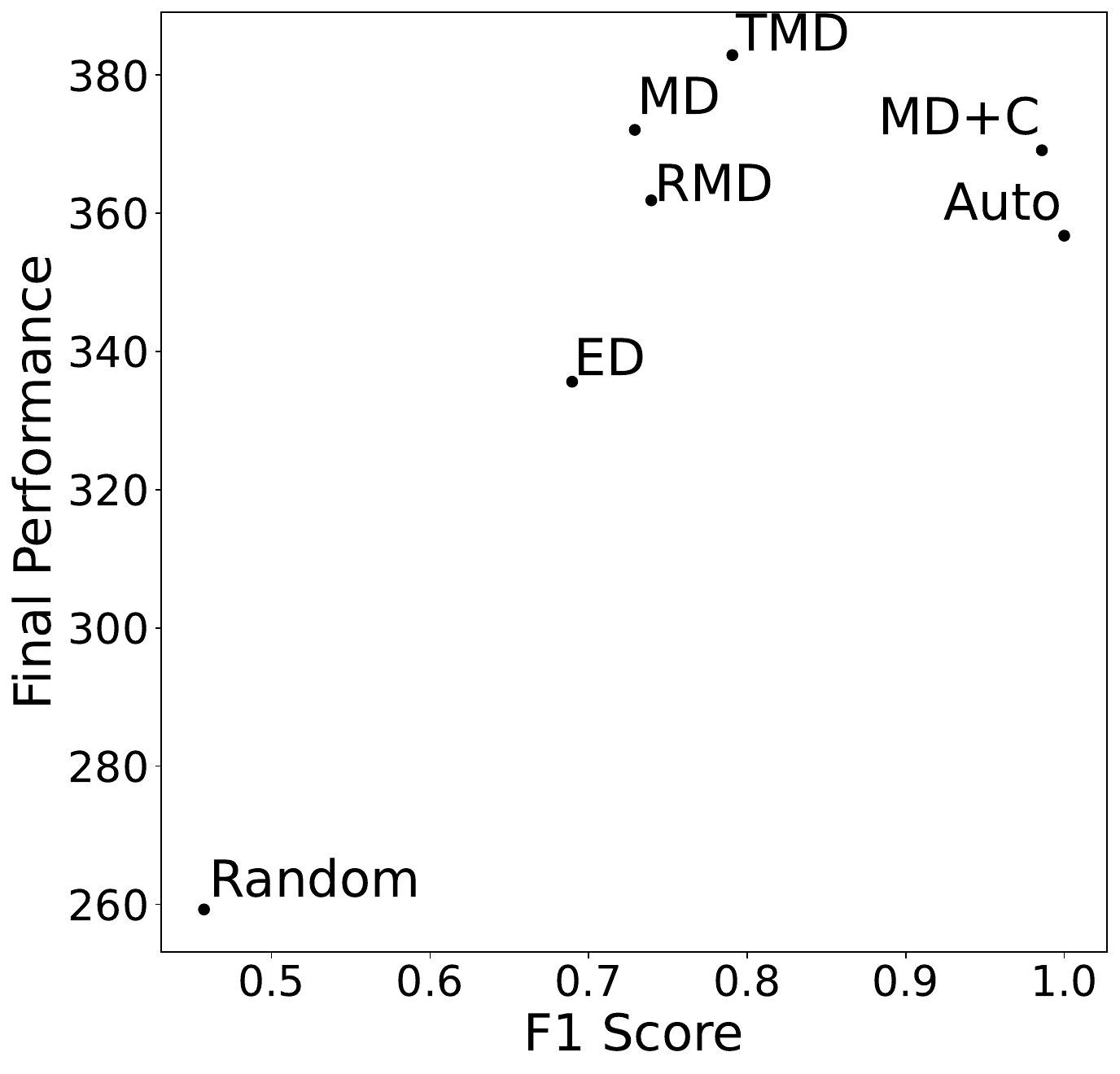}}
	\subfigure[OOD Tutankham]{\includegraphics[width=0.19\textwidth]{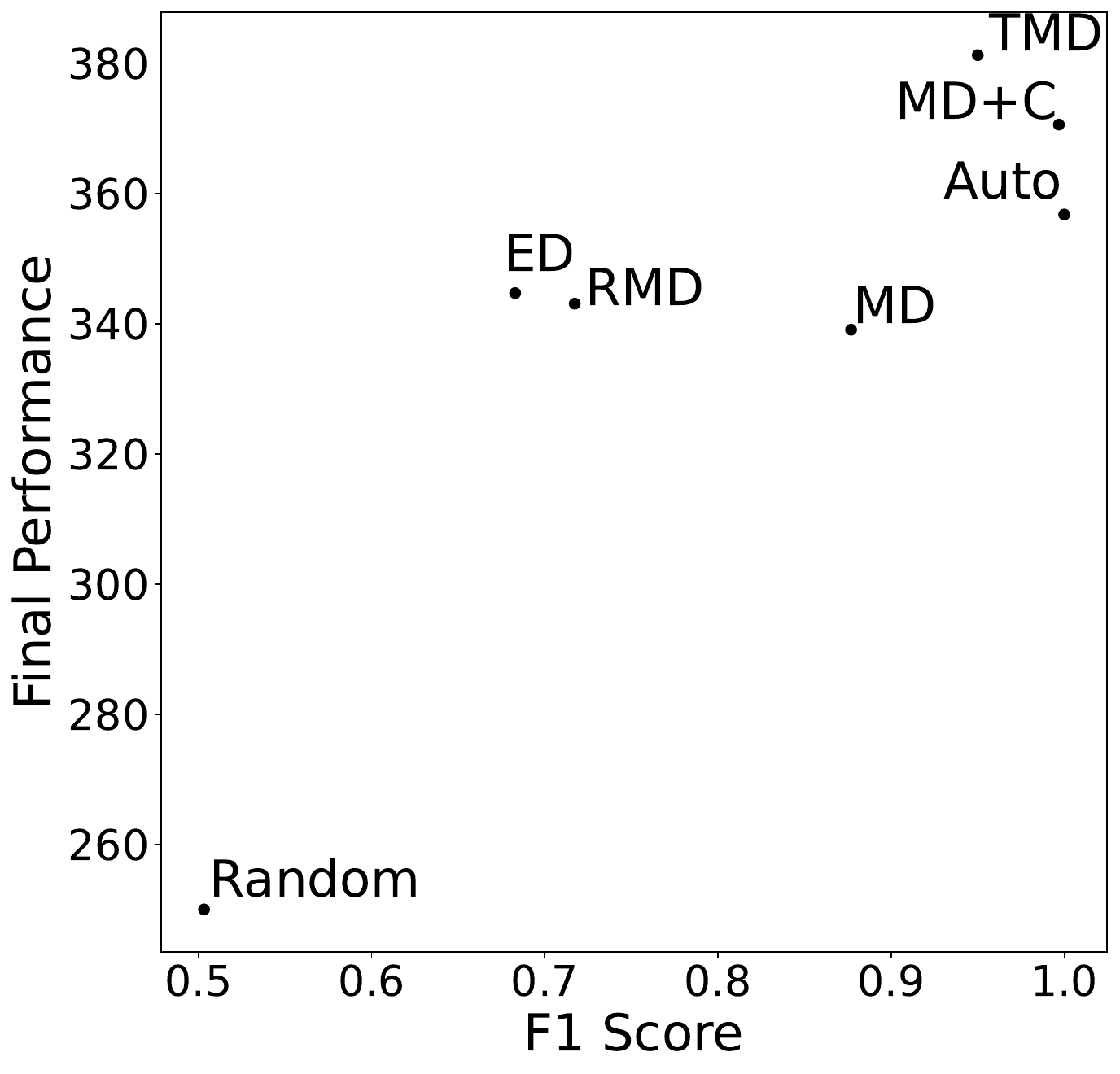}}
	\subfigure[OOD FishingDerby]{\includegraphics[width=0.19\textwidth]{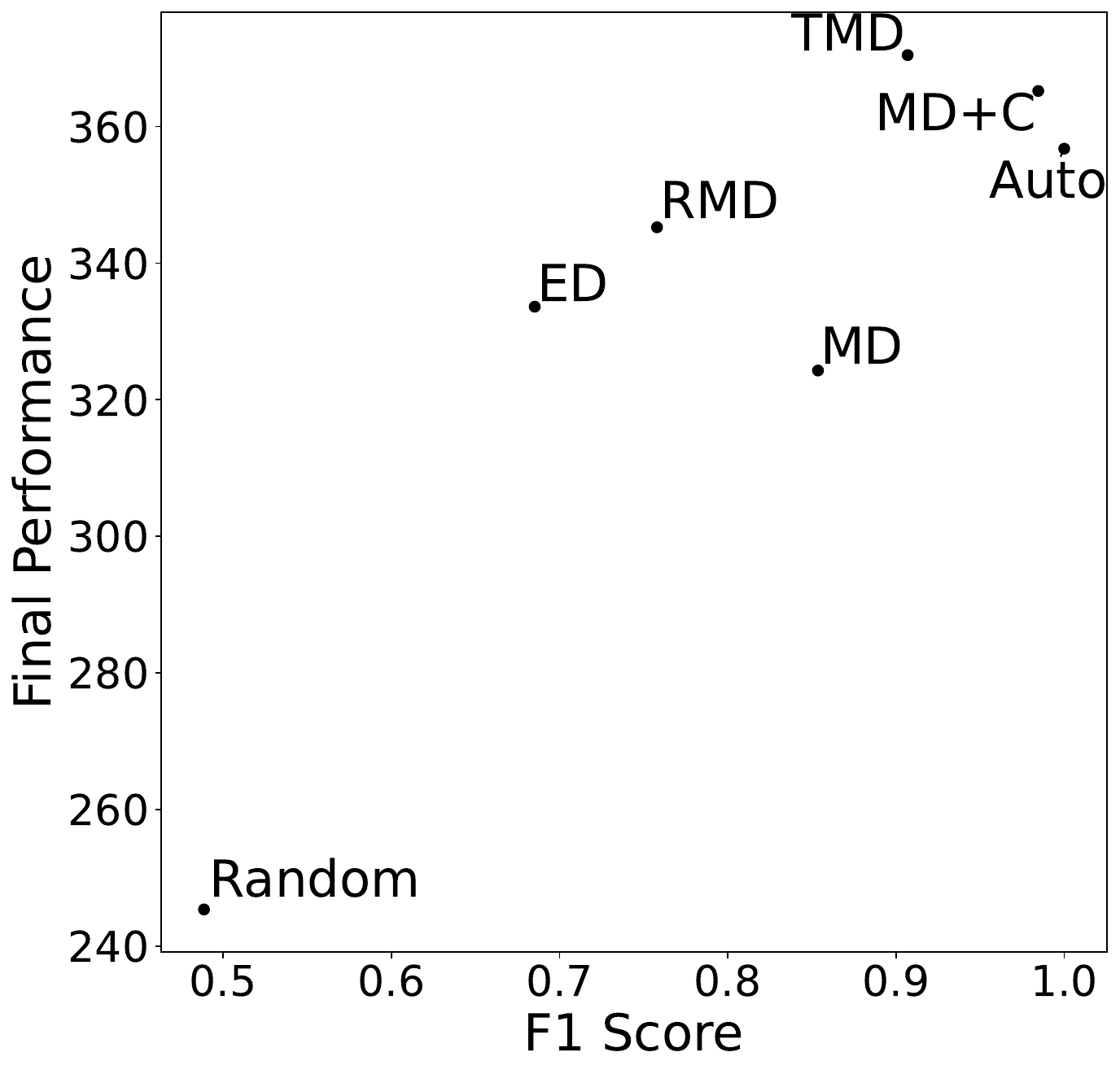}}
	\subfigure[Adversarial]{\includegraphics[width=0.19\textwidth]{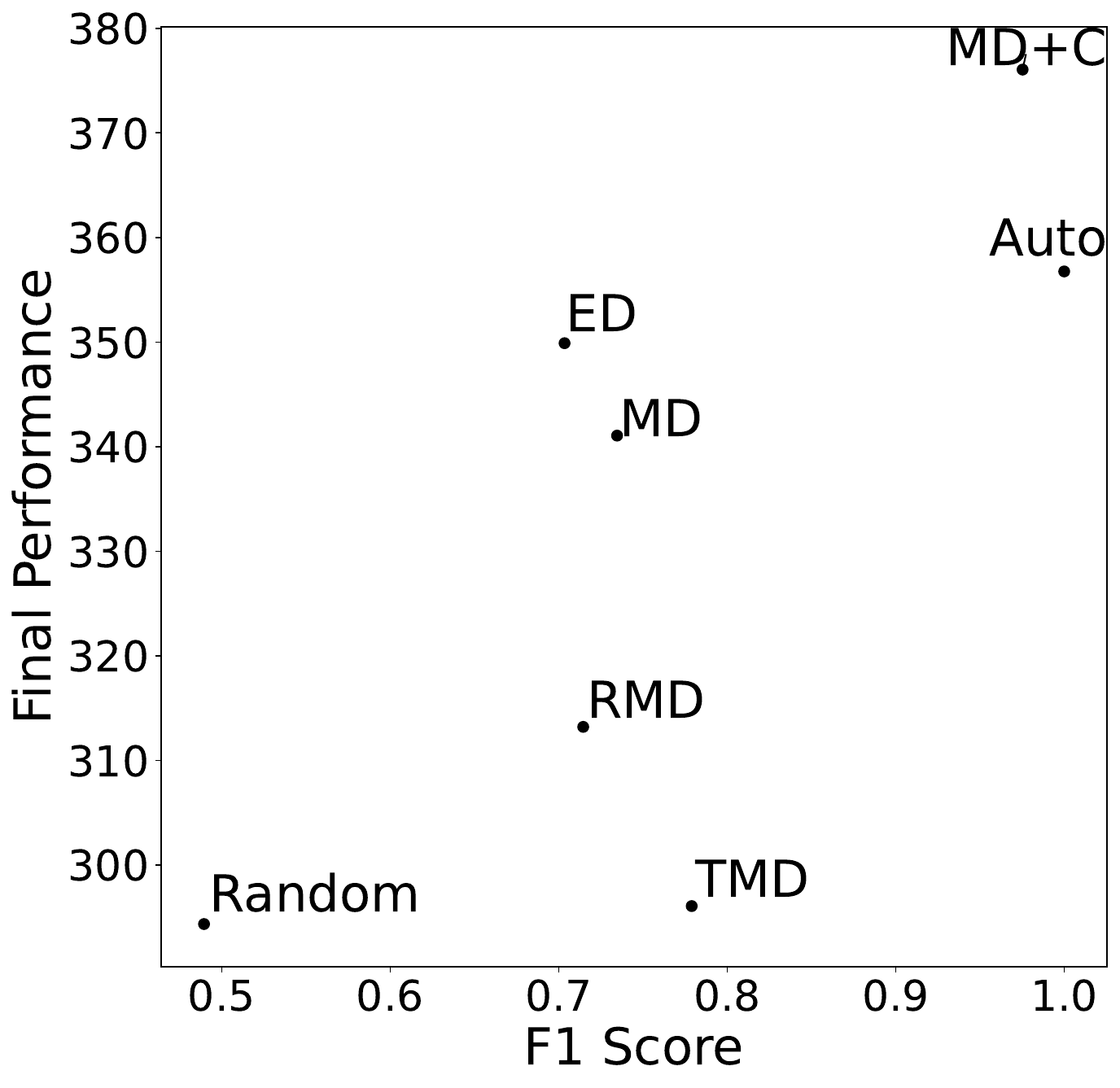}}
	\caption{Detection performance across various state outliers in the online training on Enduro.}
	\label{fig:Enduro_online_full}
\end{figure*}

\begin{figure*}[htbp]
	\centering
	\subfigure{\includegraphics[width=0.19\textwidth]{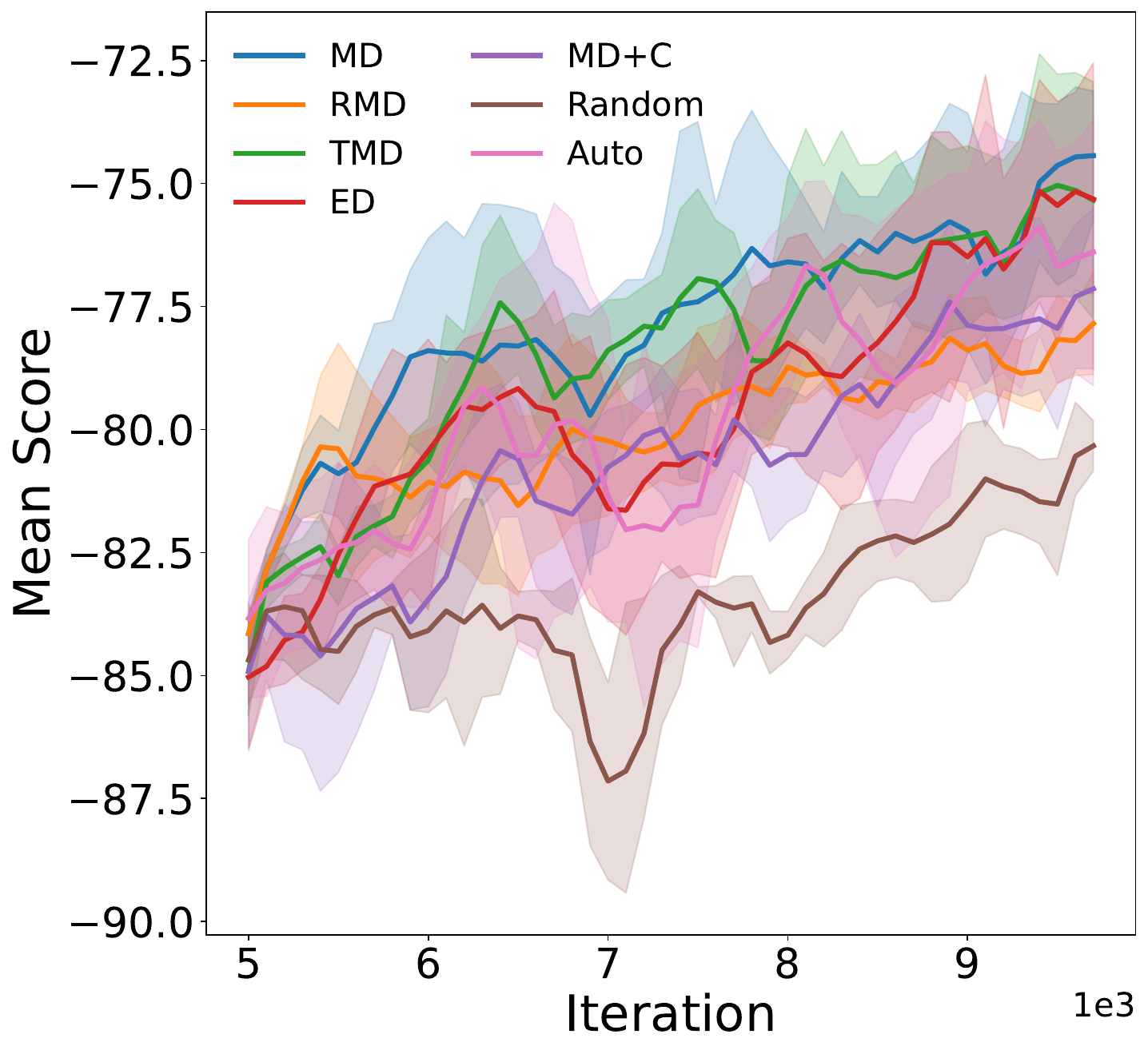}}
	\subfigure{\includegraphics[width=0.19\textwidth]{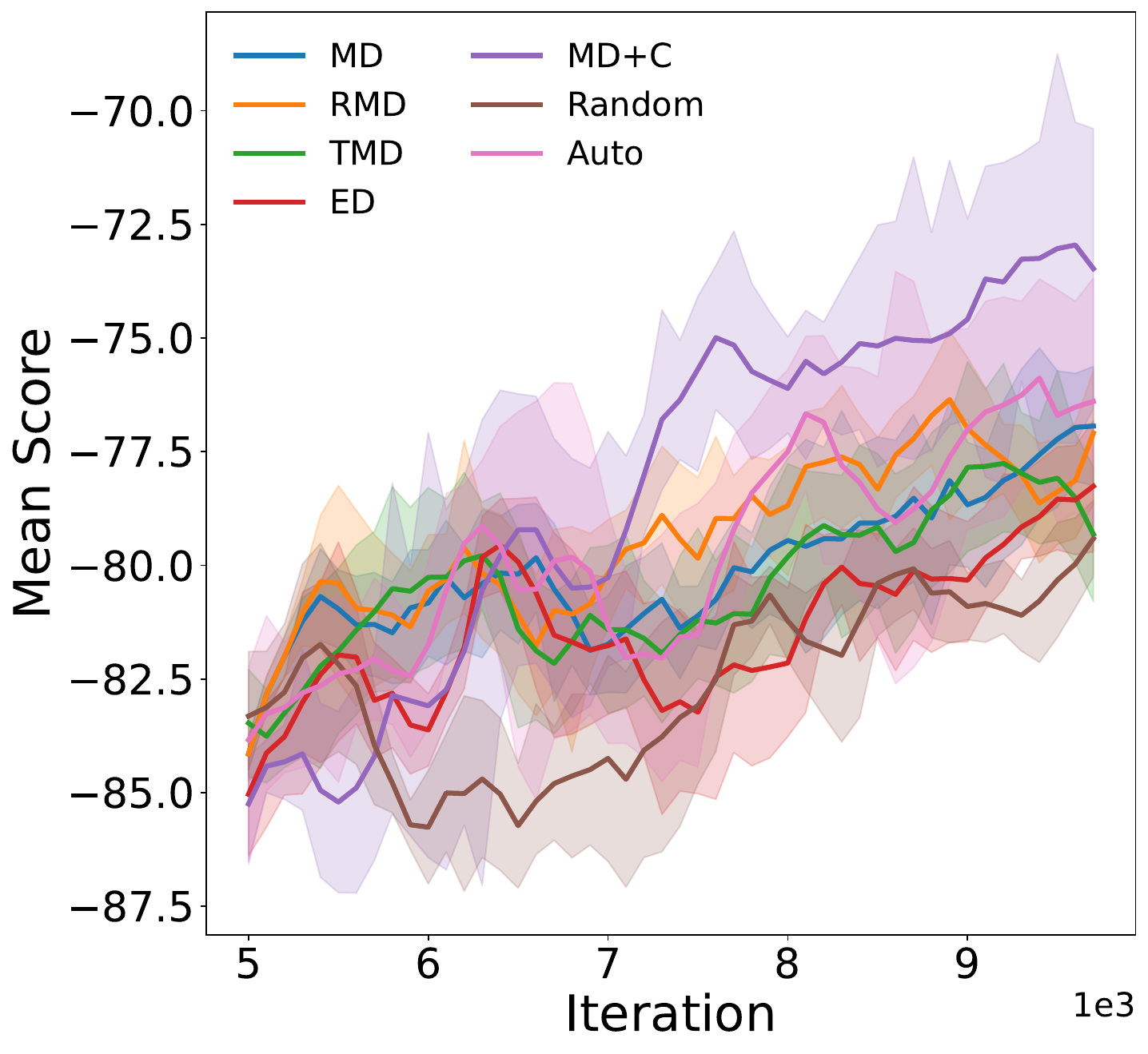}}
	\subfigure{\includegraphics[width=0.19\textwidth]{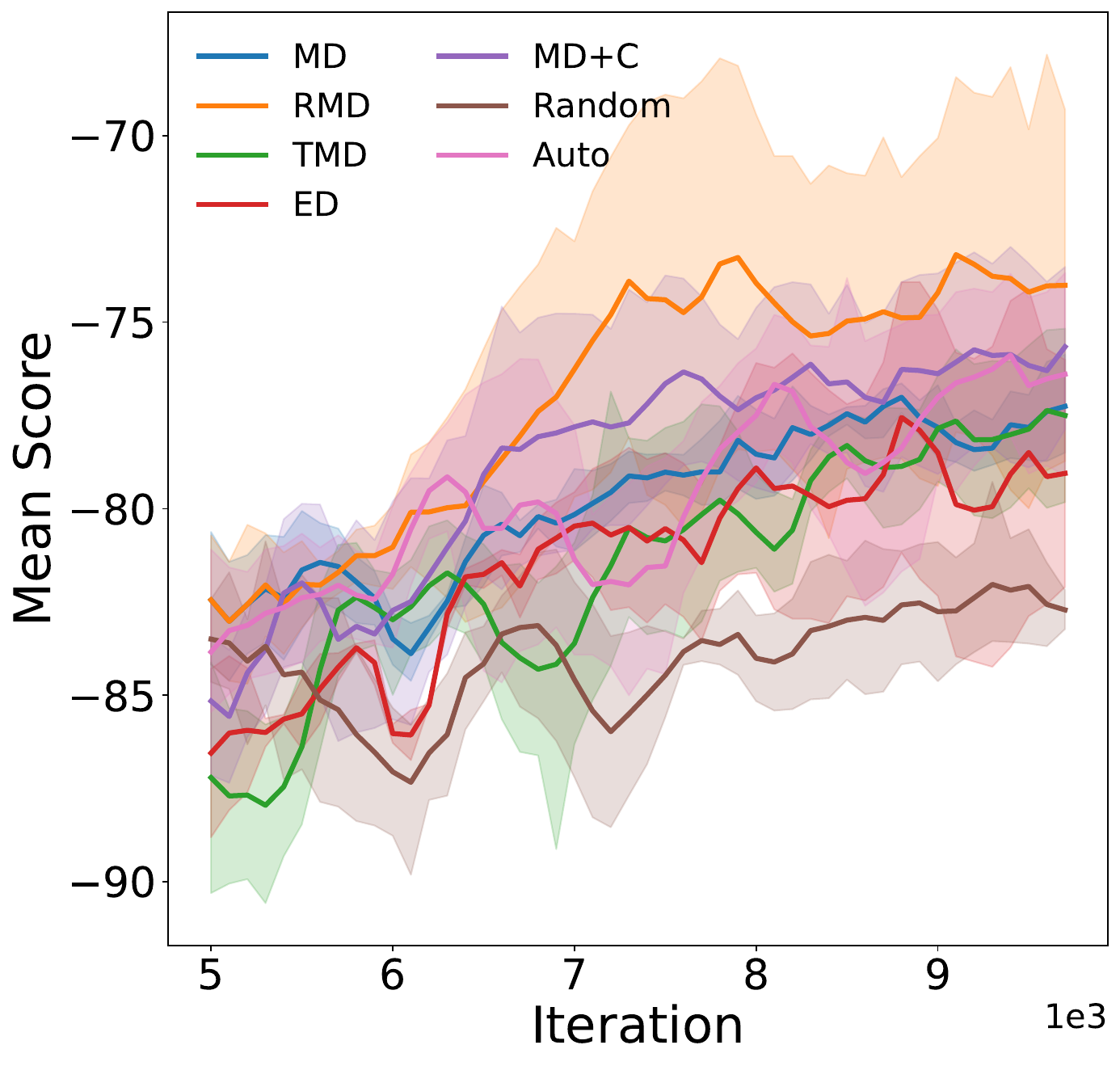}}
	\subfigure{\includegraphics[width=0.19\textwidth]{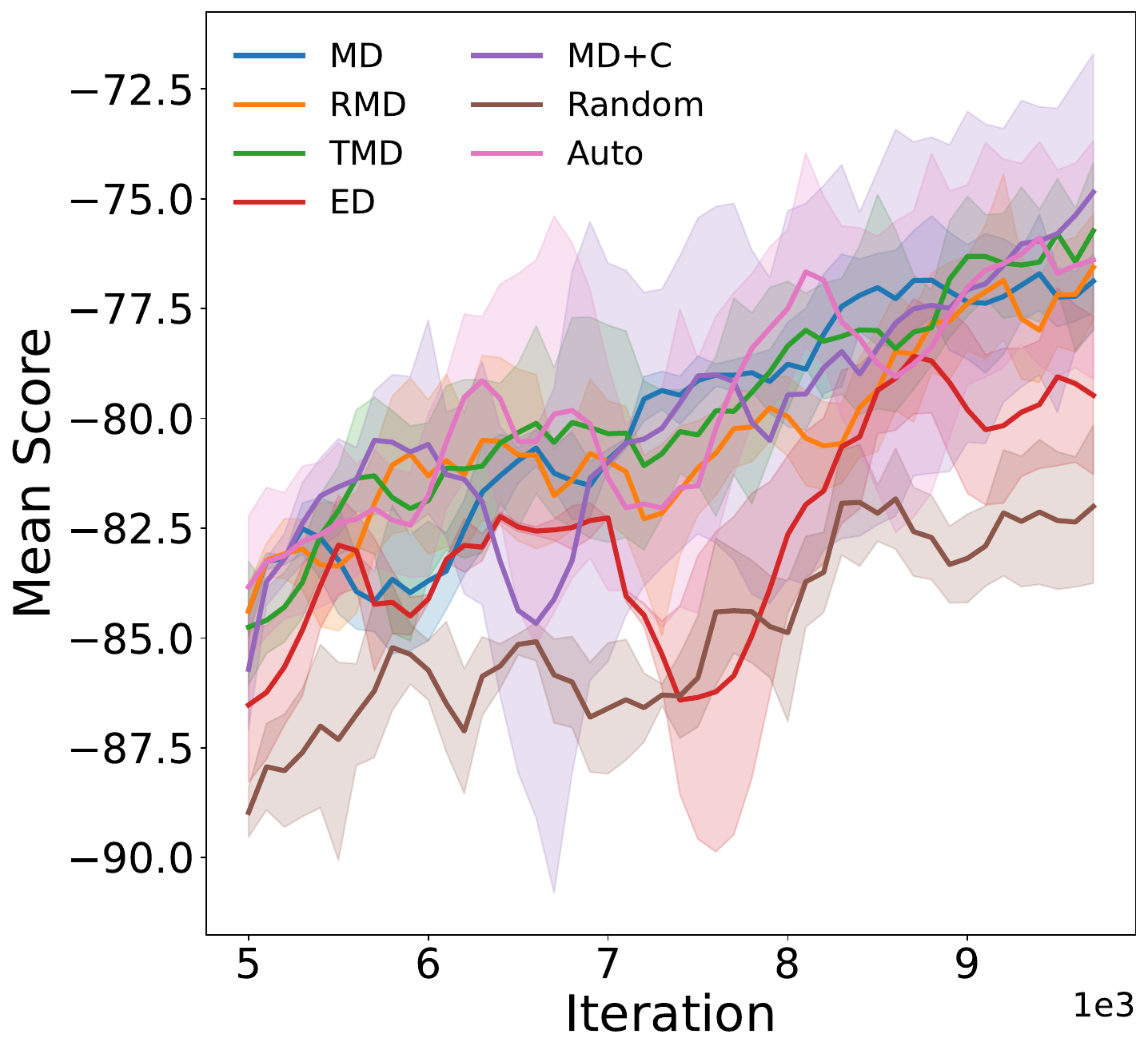}}
	\subfigure{\includegraphics[width=0.19\textwidth]{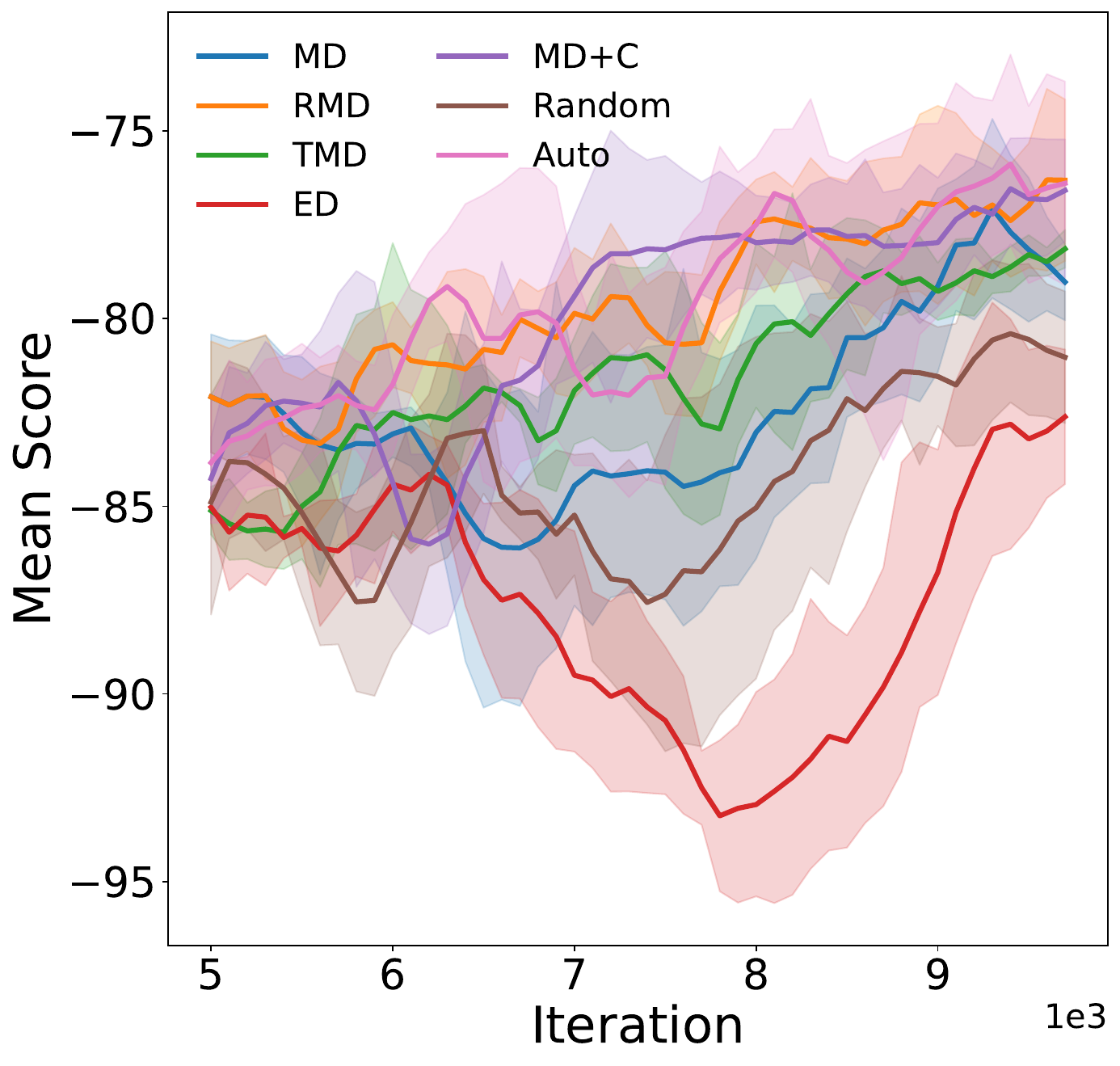}}
    \setcounter{subfigure}{0}
	\subfigure{\includegraphics[width=0.19\textwidth]{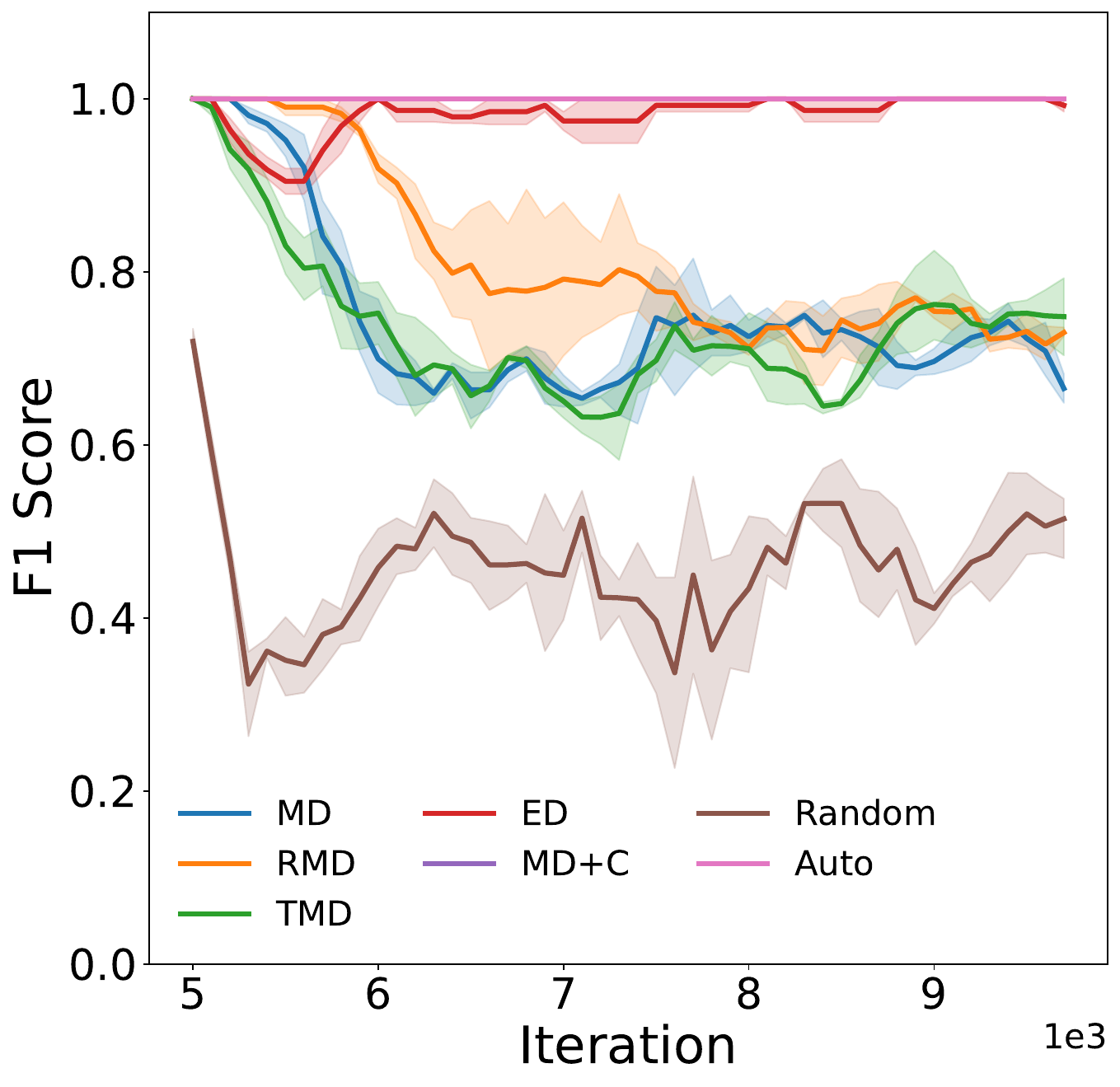}}
	\subfigure{\includegraphics[width=0.19\textwidth]{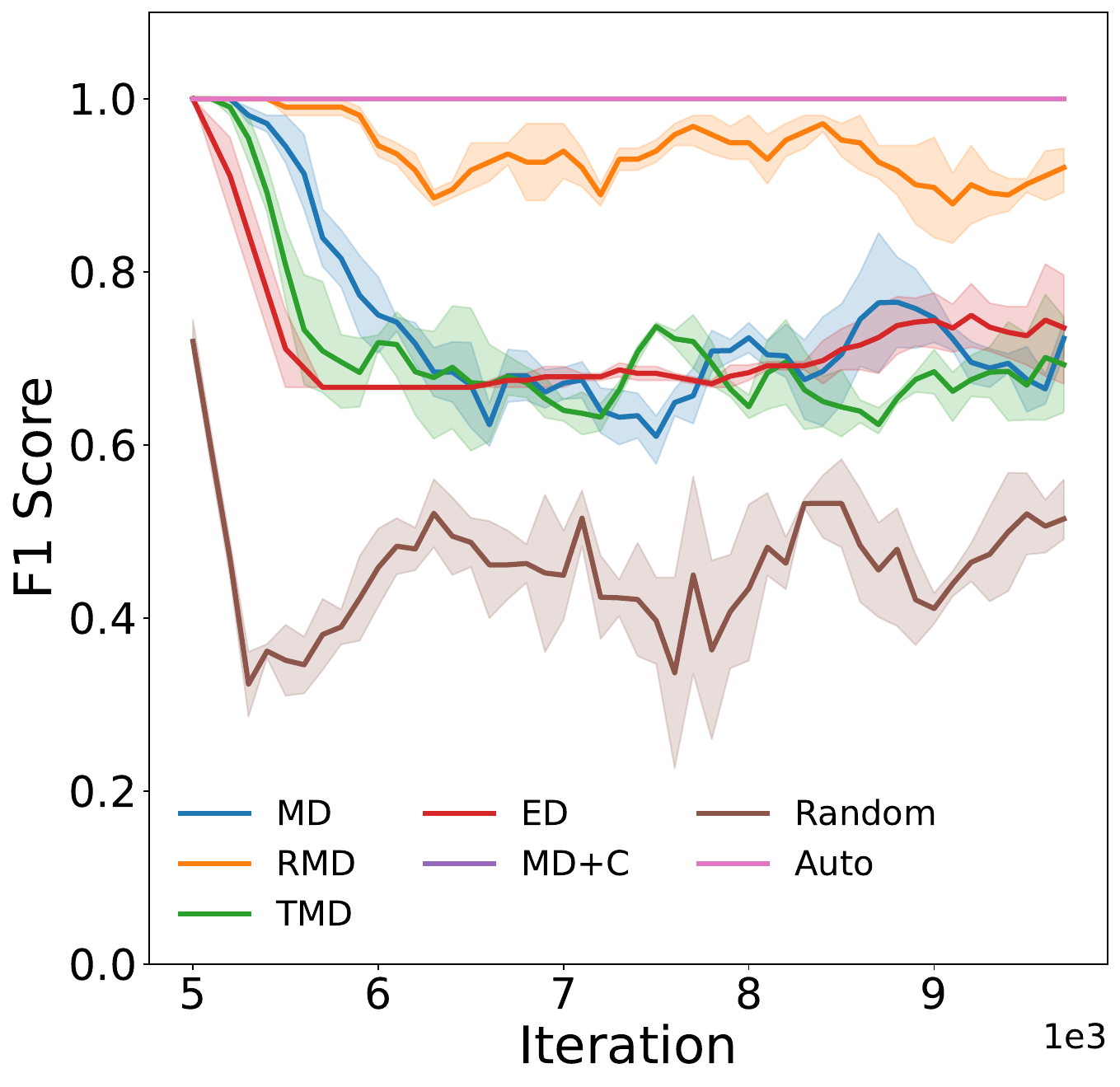}}
	\subfigure{\includegraphics[width=0.19\textwidth]{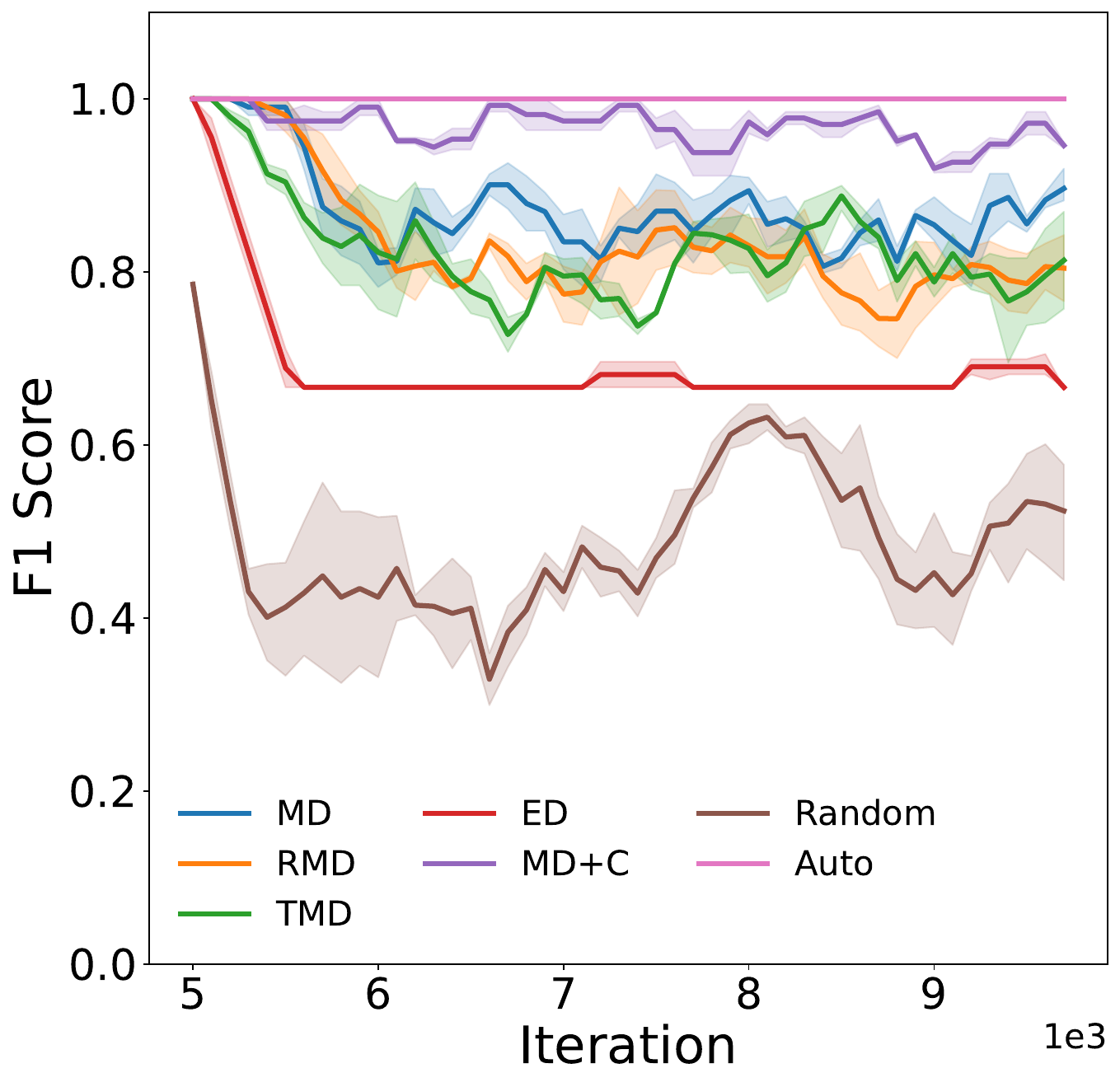}}
	\subfigure{\includegraphics[width=0.19\textwidth]{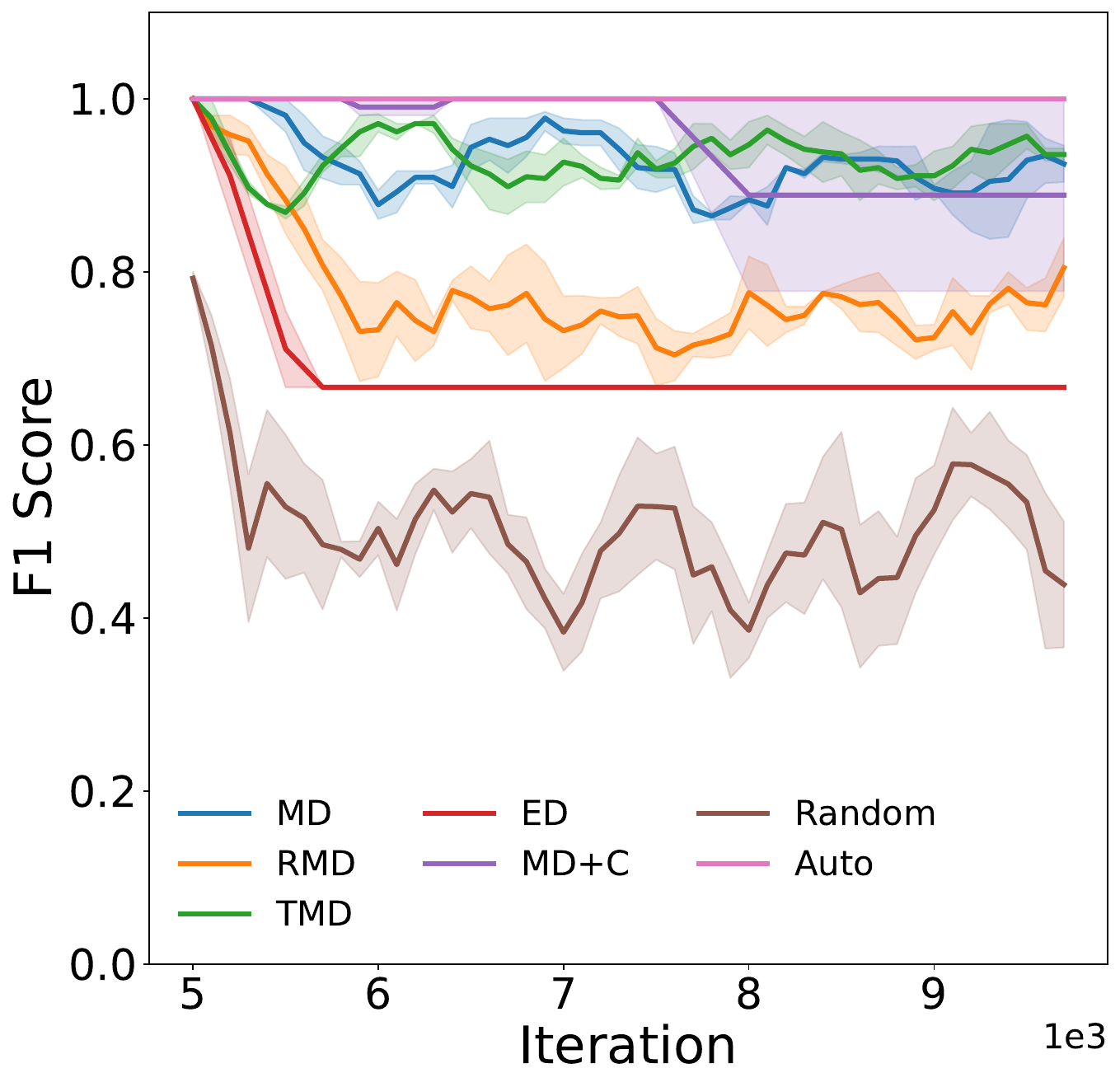}}
	\subfigure{\includegraphics[width=0.19\textwidth]{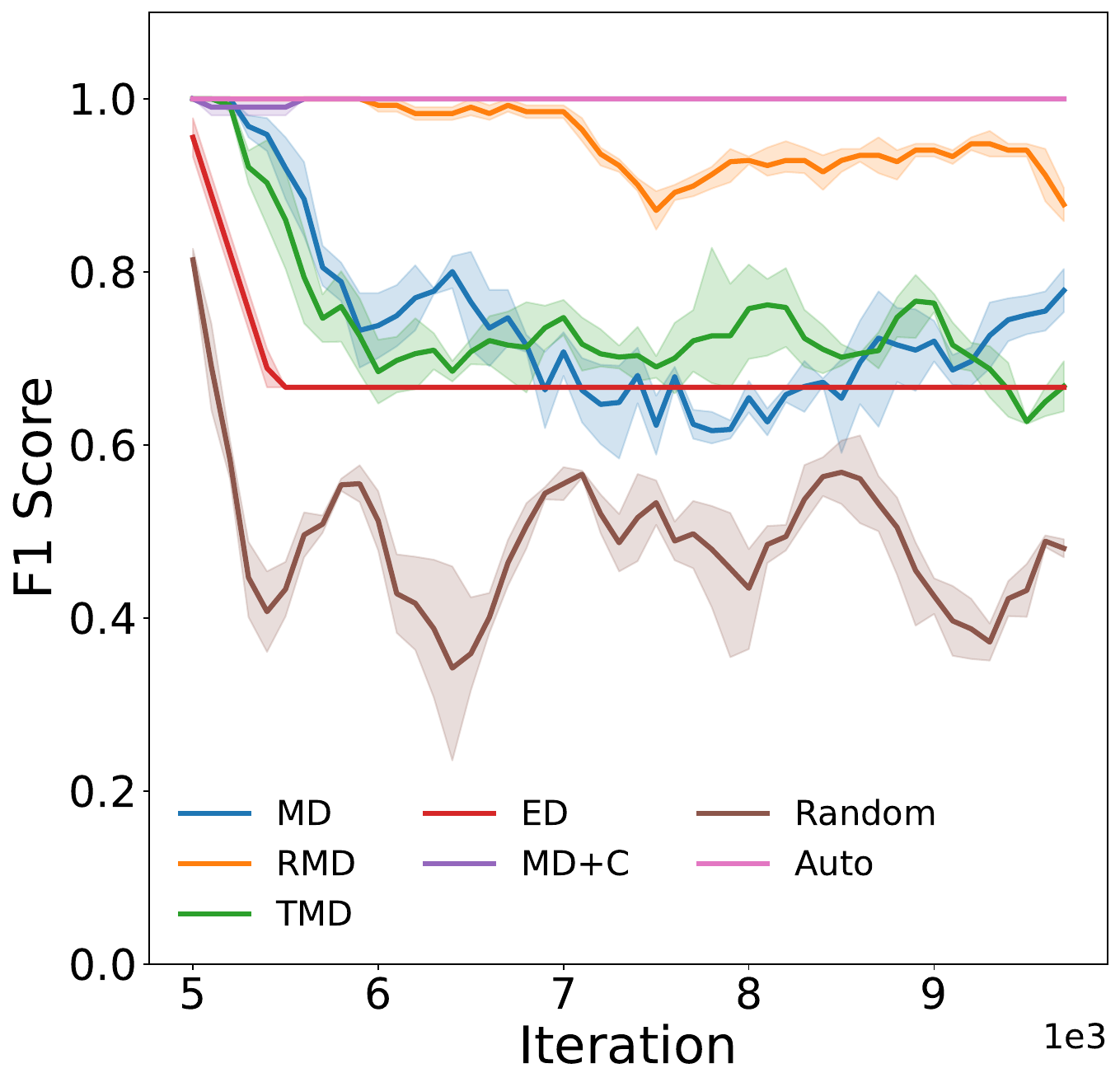}}
     \setcounter{subfigure}{0}
	\subfigure[Gaussian std=1]{\includegraphics[width=0.19\textwidth]{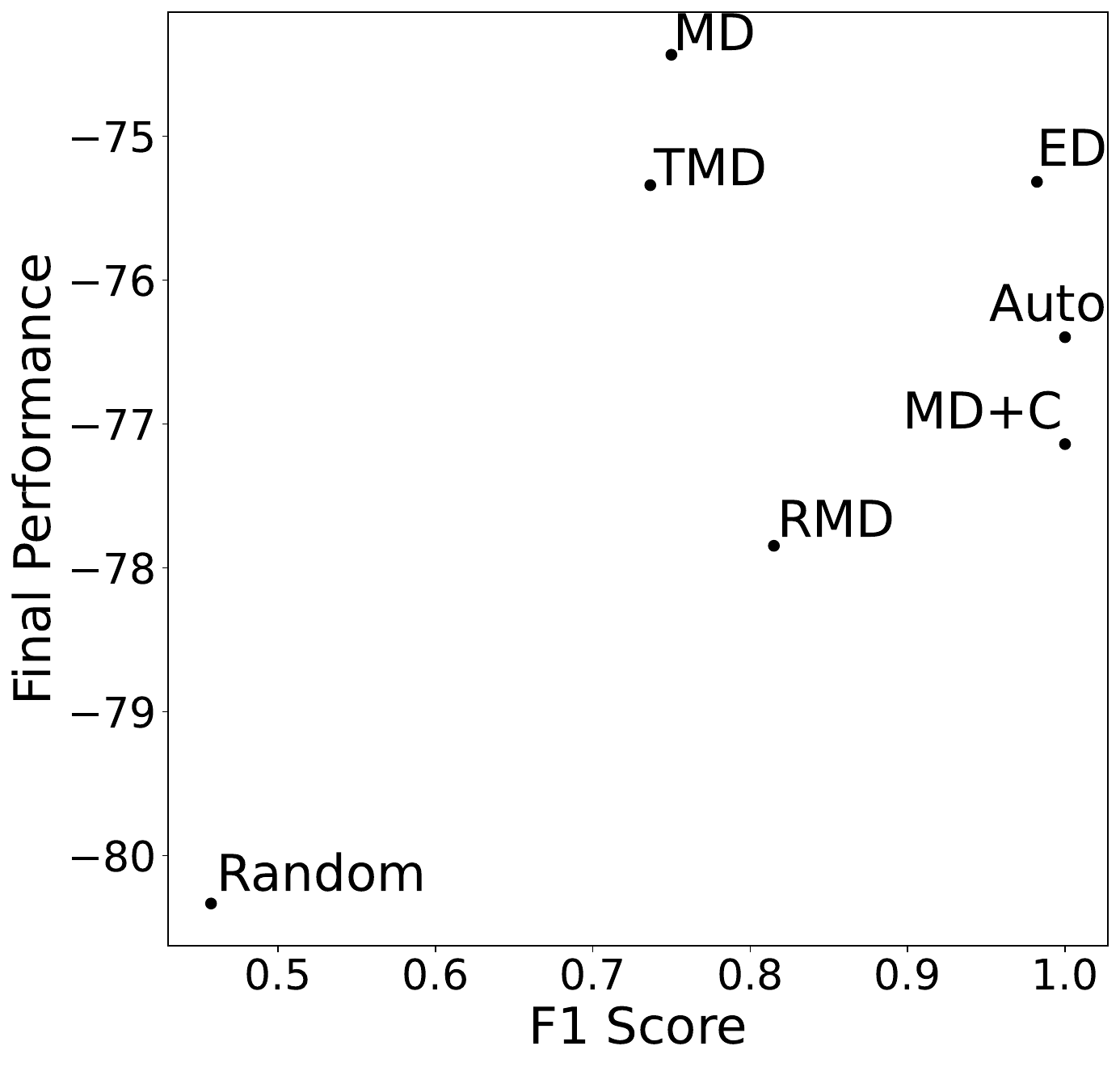}}
	\subfigure[Gaussian std=0.3]{\includegraphics[width=0.19\textwidth]{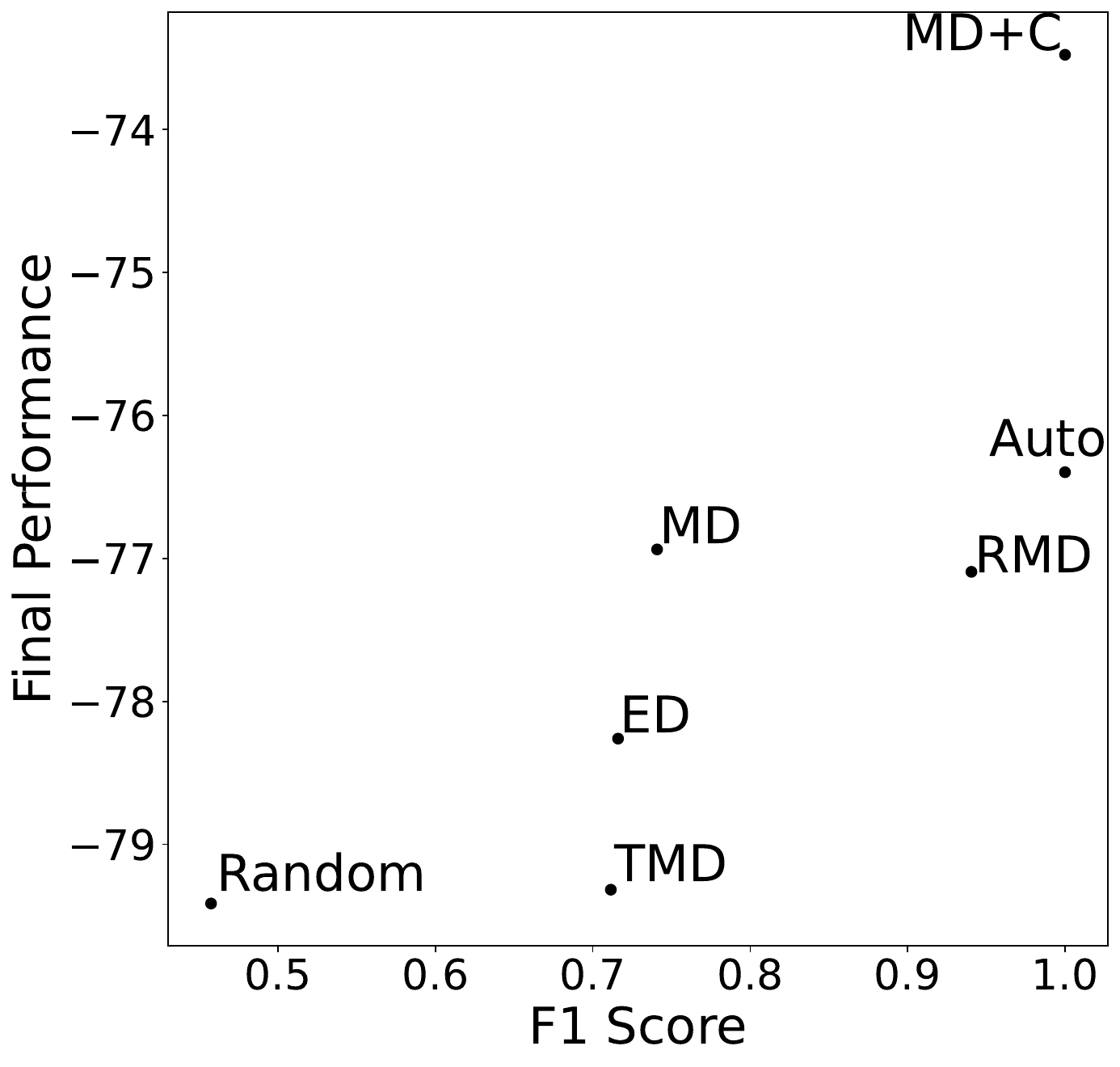}}
	\subfigure[OOD Enduro]{\includegraphics[width=0.19\textwidth]{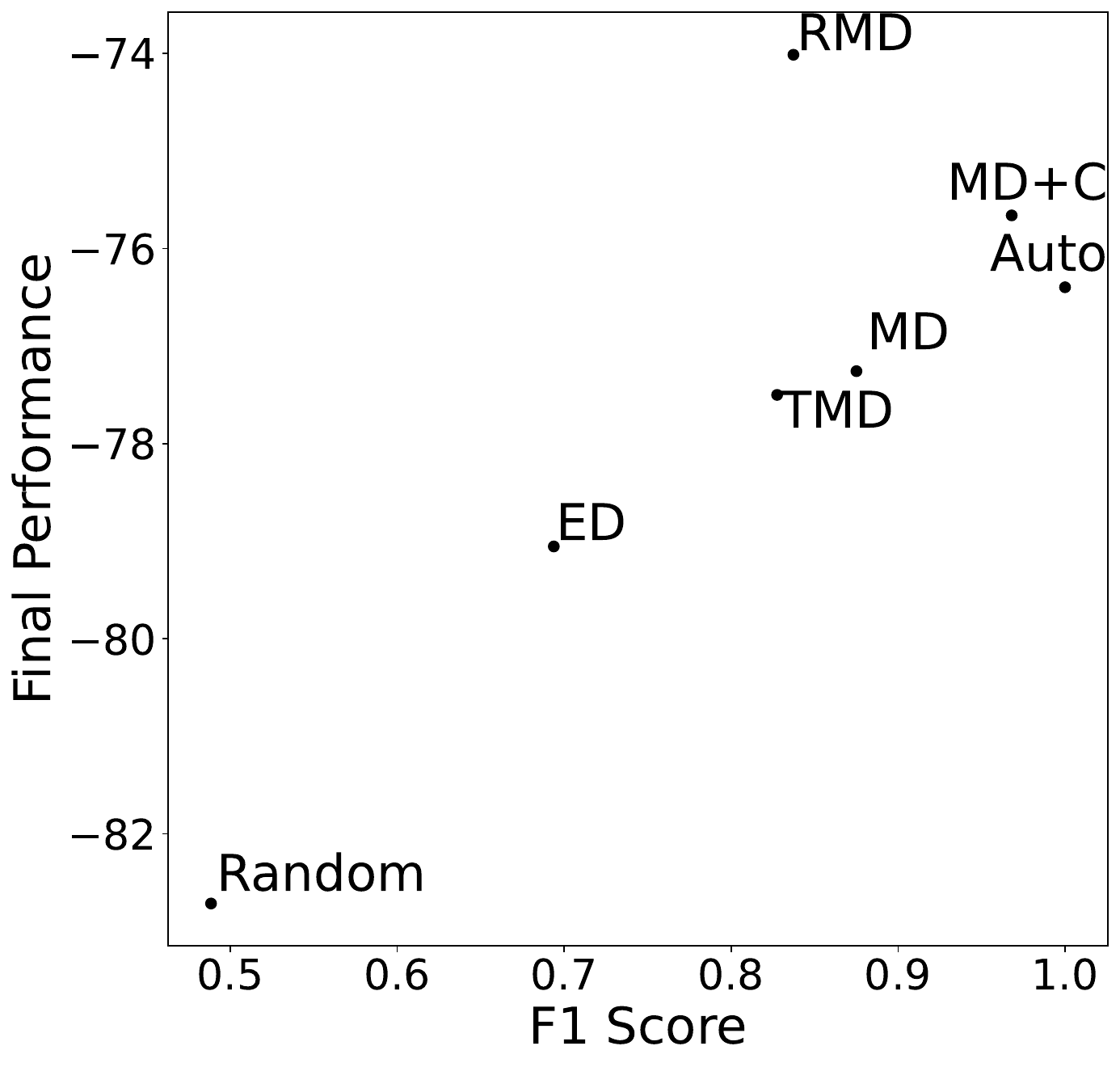}}
	\subfigure[OOD Tutankham]{\includegraphics[width=0.19\textwidth]{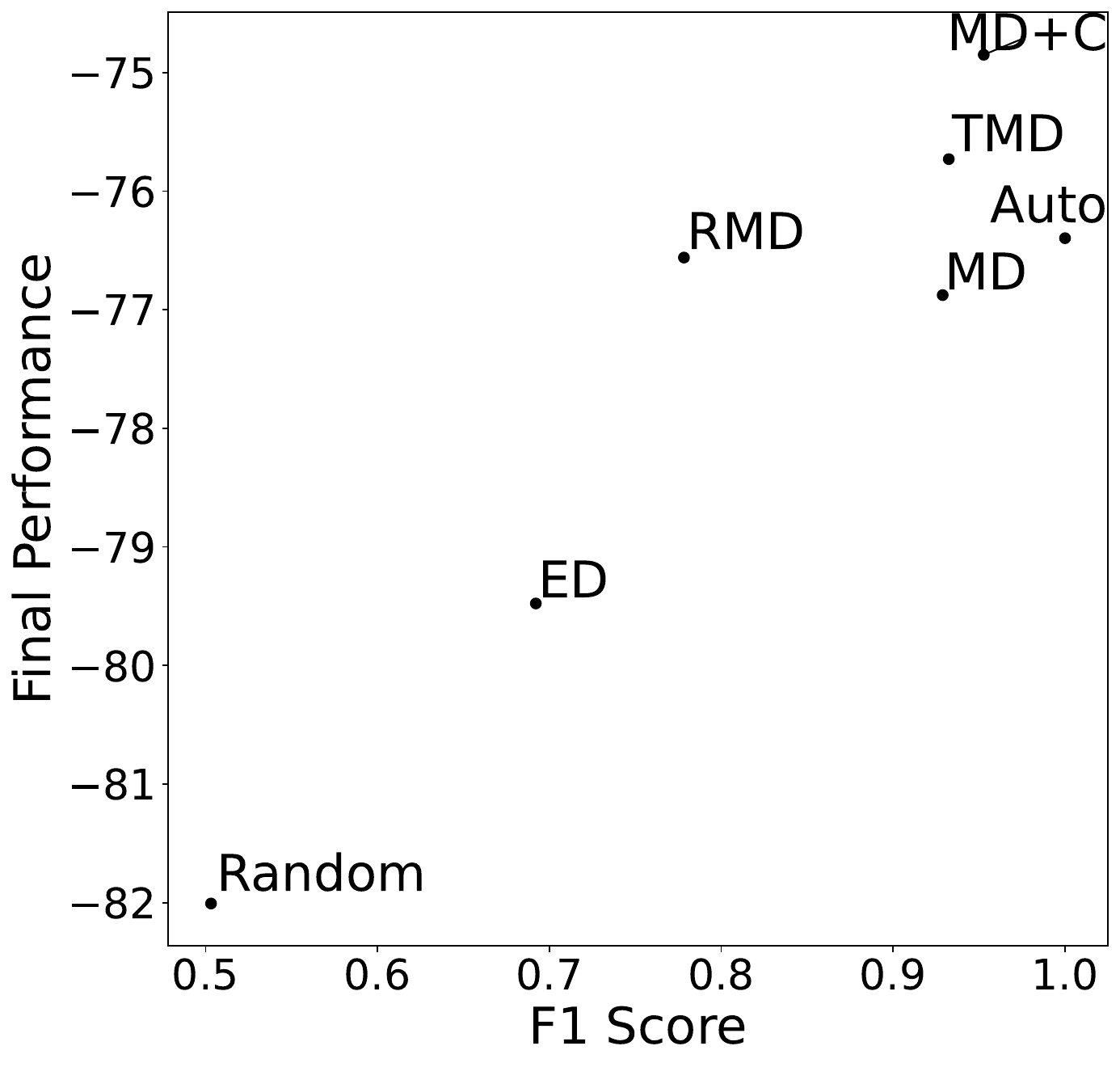}}
	\subfigure[Adversarial]{\includegraphics[width=0.19\textwidth]{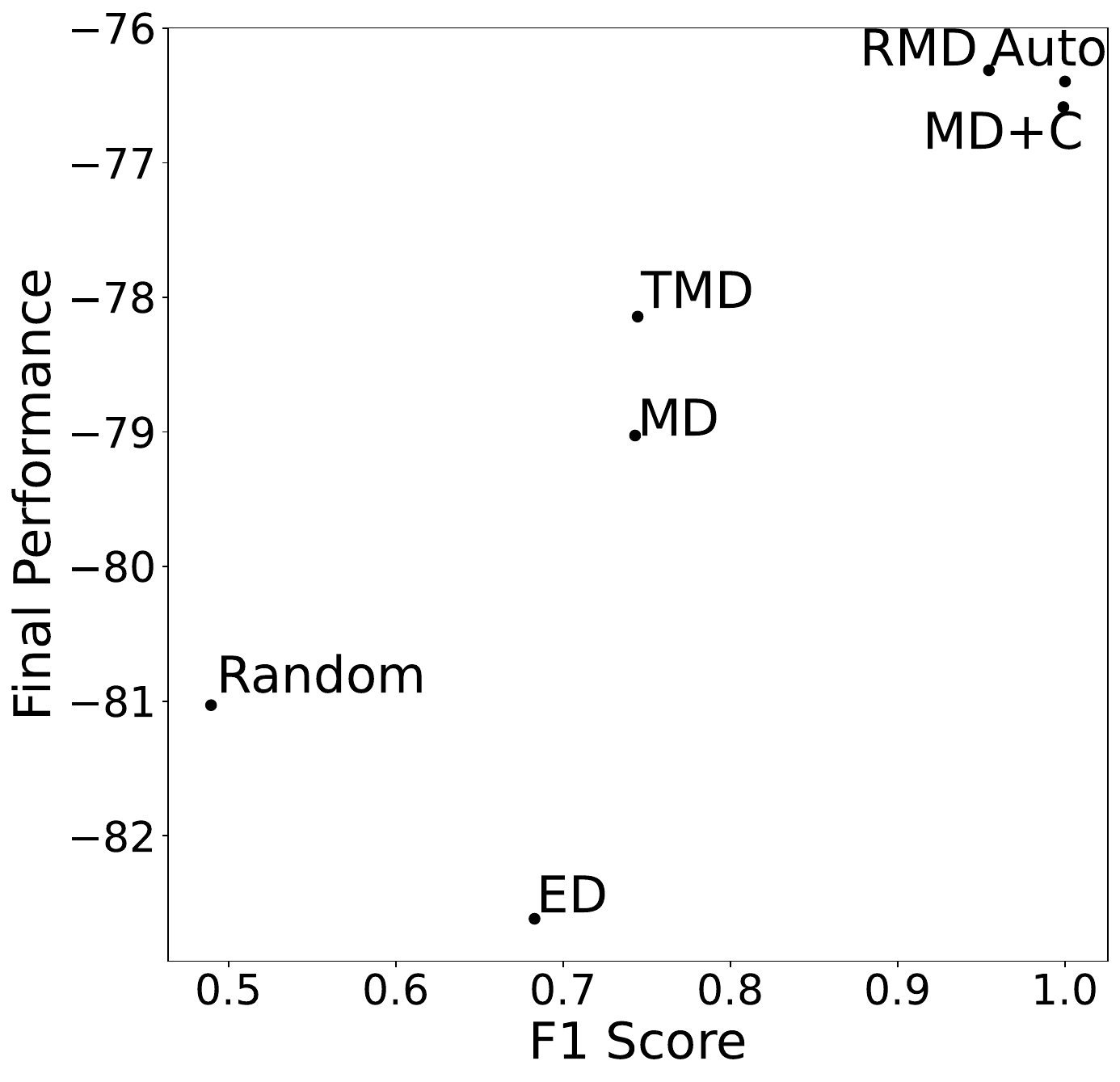}}
	\caption{Detection performance across various state outliers in the online training on FishingDerby.}
	\label{fig:FishingDerby_online_full}
\end{figure*}

\begin{figure*}[htbp]
	\centering
	\subfigure{\includegraphics[width=0.19\textwidth]{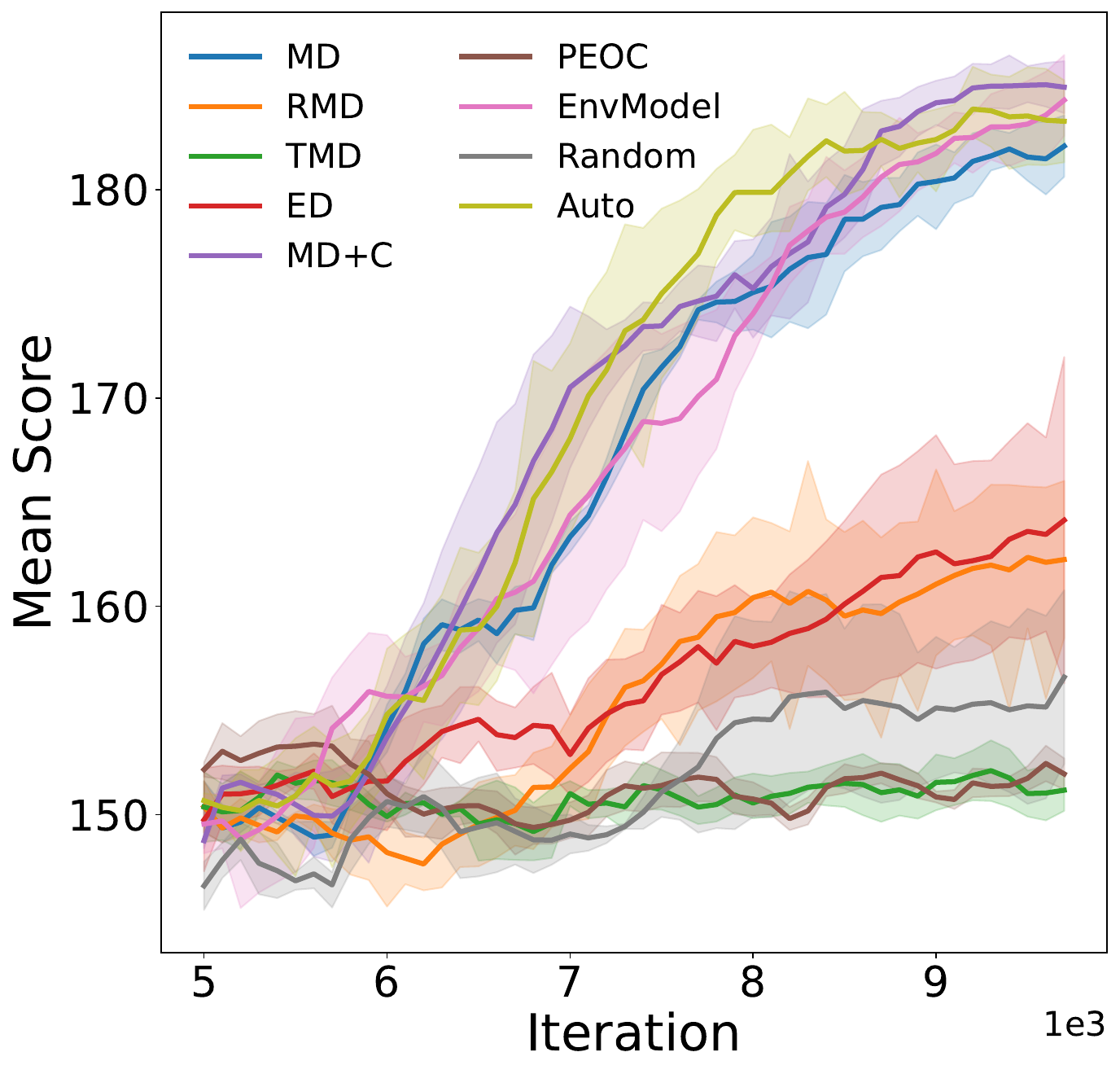}}
	\subfigure{\includegraphics[width=0.19\textwidth]{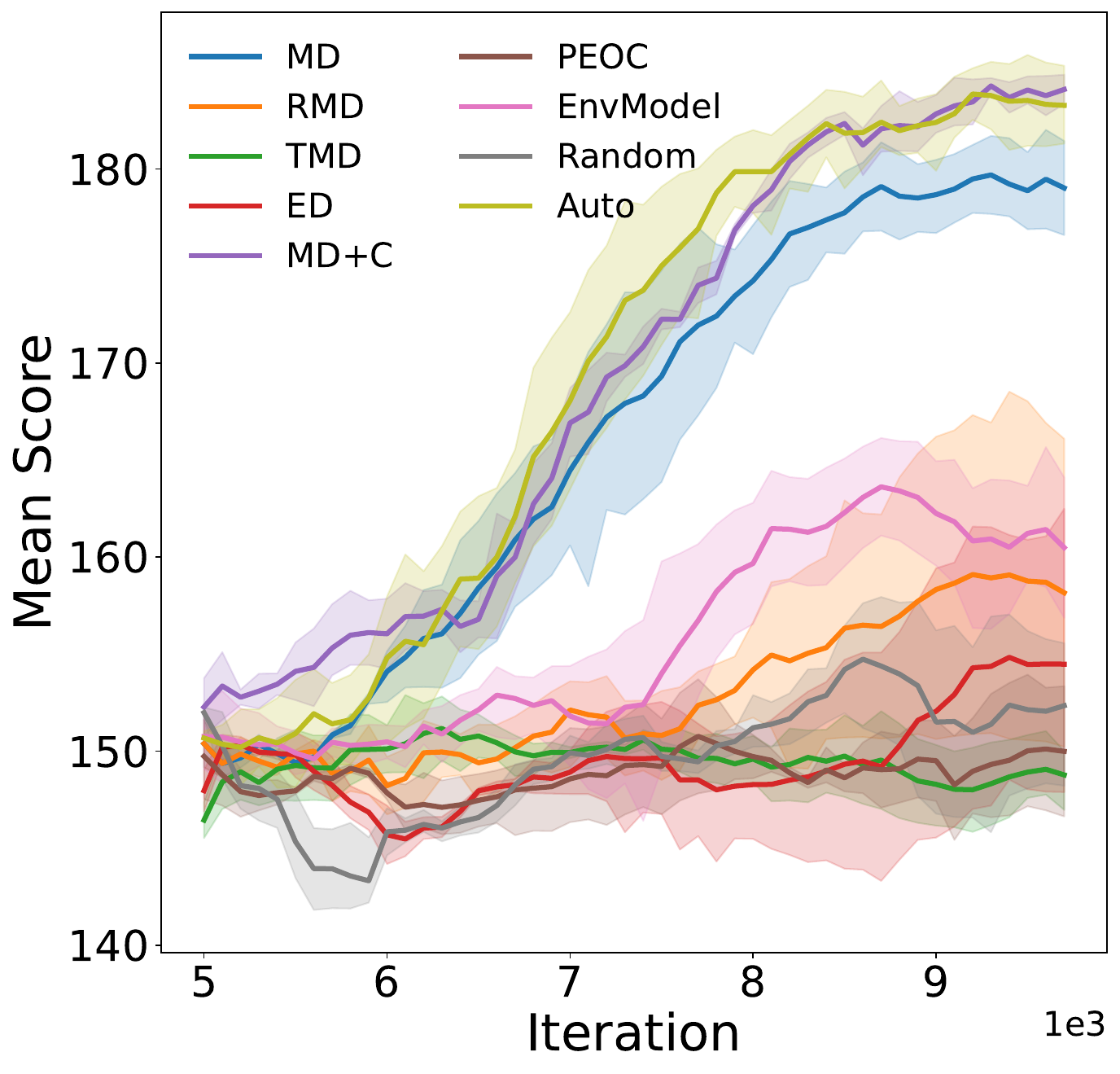}}
	\subfigure{\includegraphics[width=0.19\textwidth]{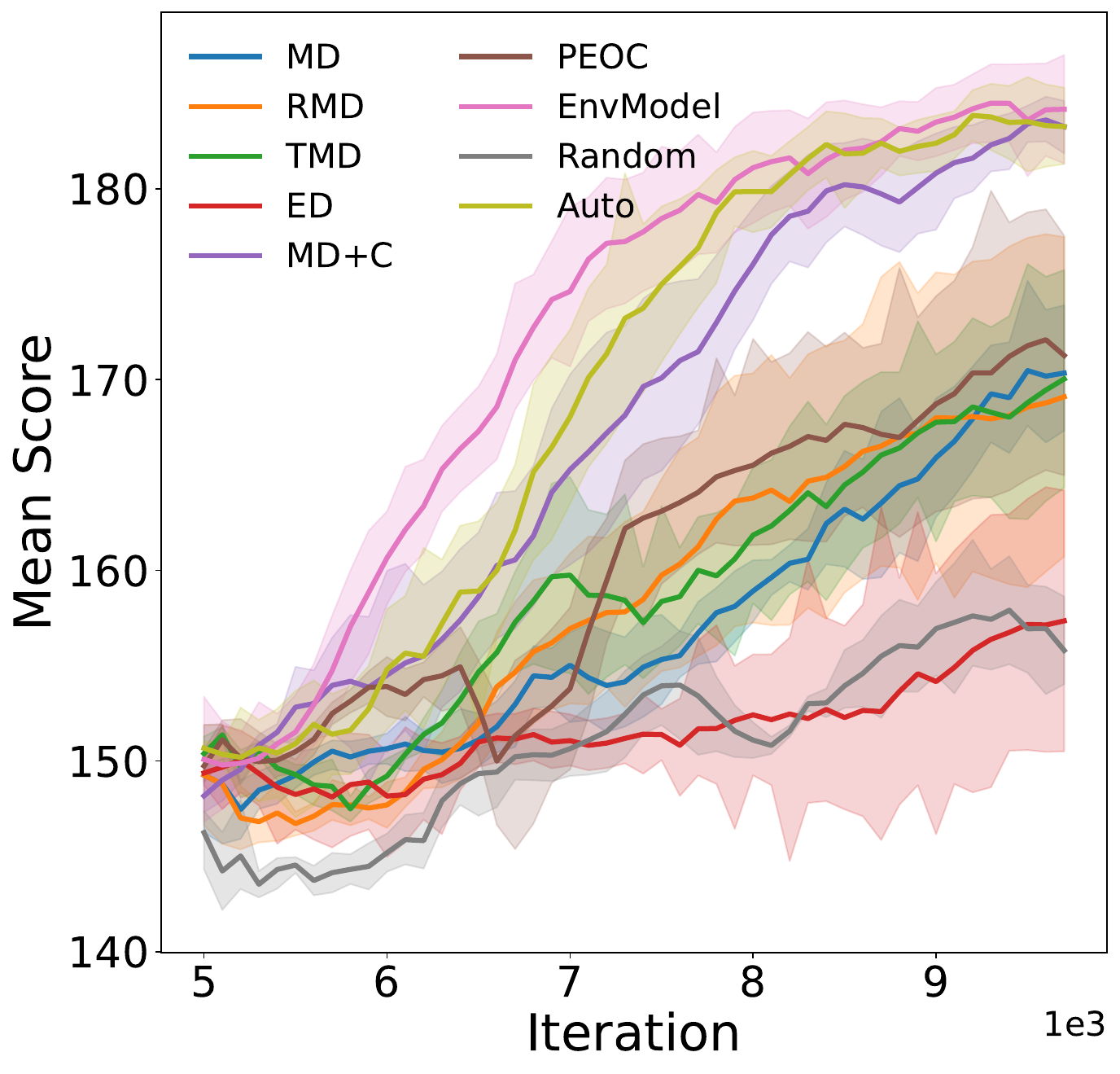}}
	\subfigure{\includegraphics[width=0.19\textwidth]{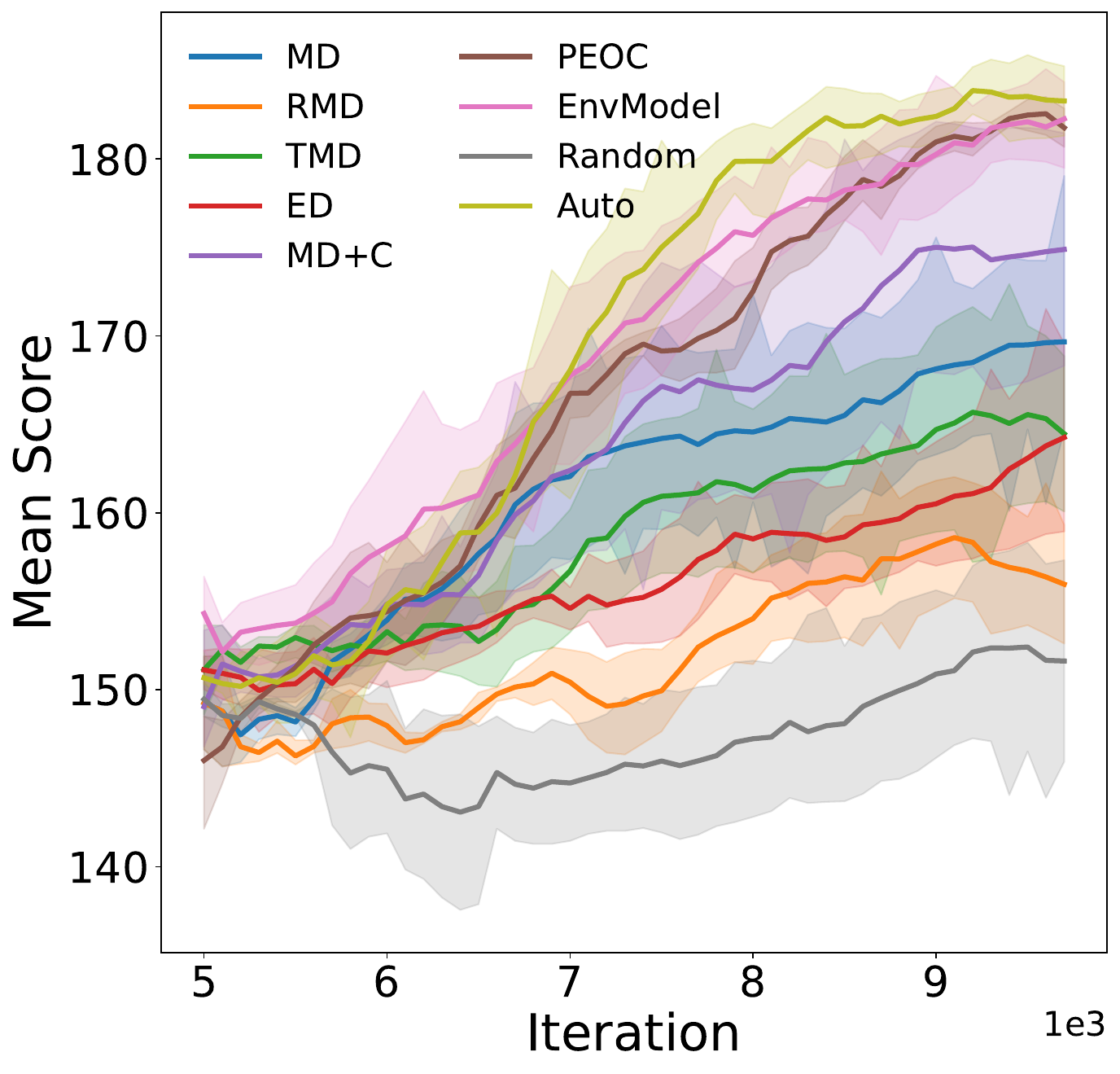}}
	\subfigure{\includegraphics[width=0.19\textwidth]{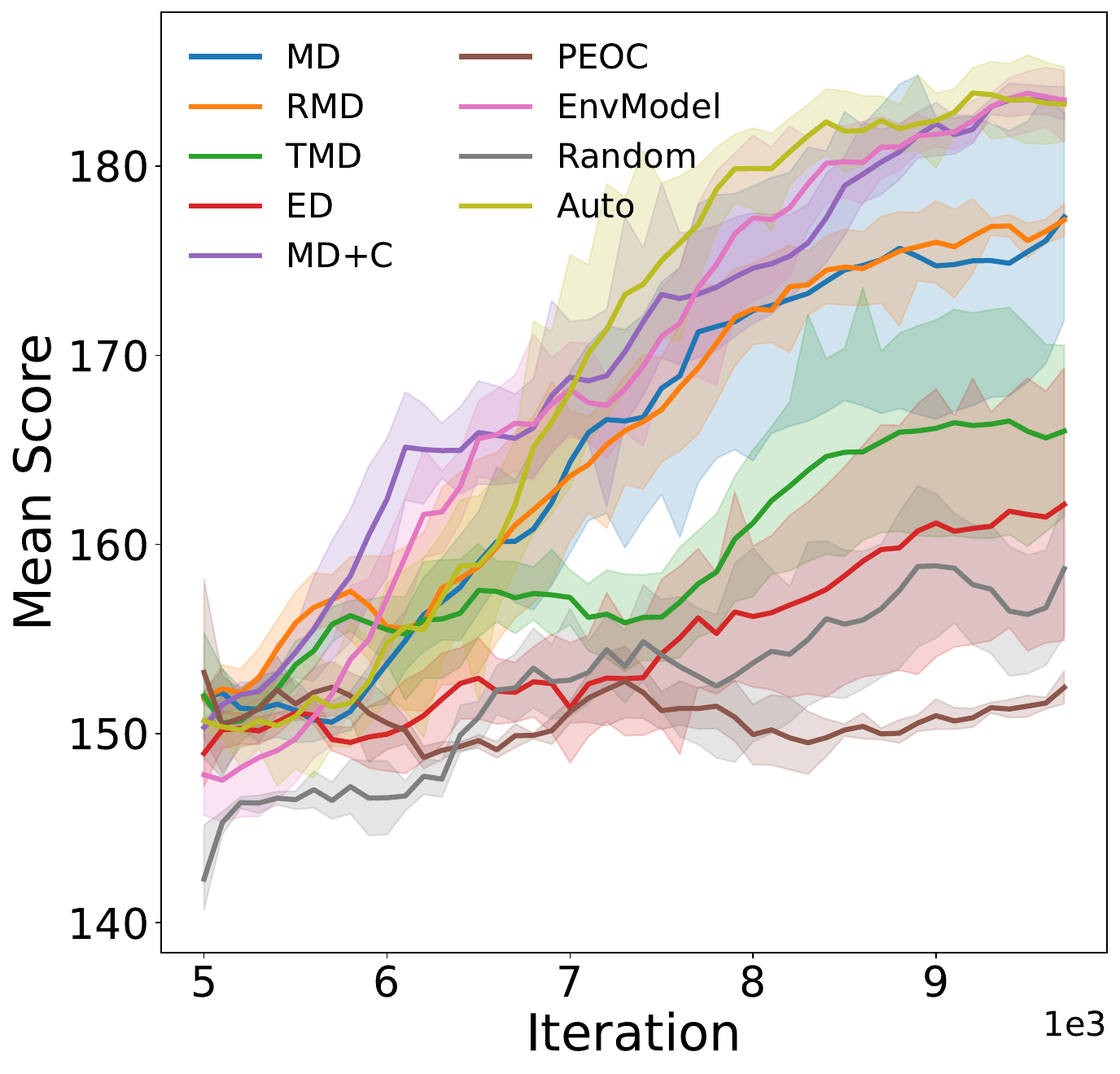}}
    \setcounter{subfigure}{0}
	\subfigure{\includegraphics[width=0.19\textwidth]{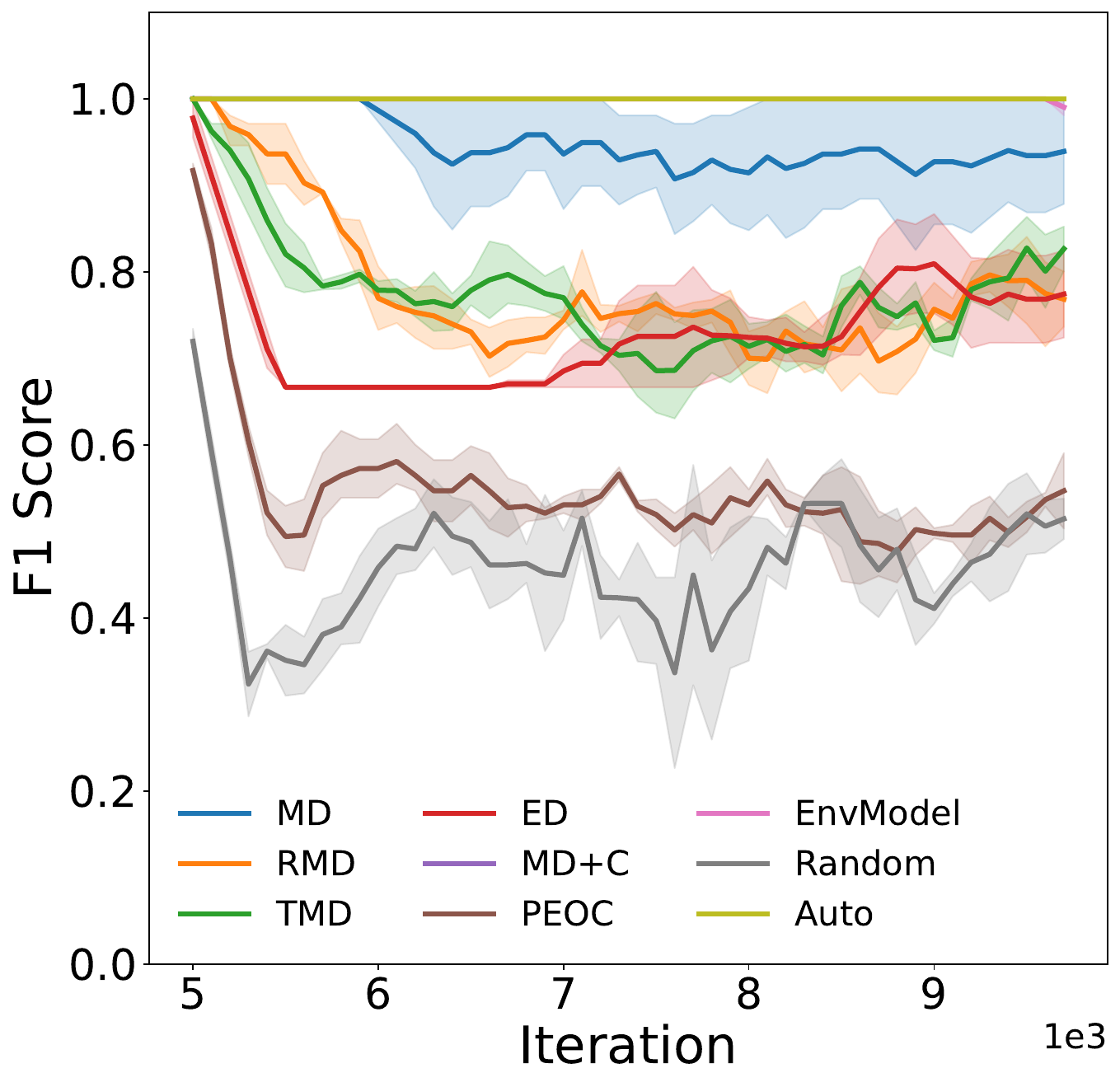}}
	\subfigure{\includegraphics[width=0.19\textwidth]{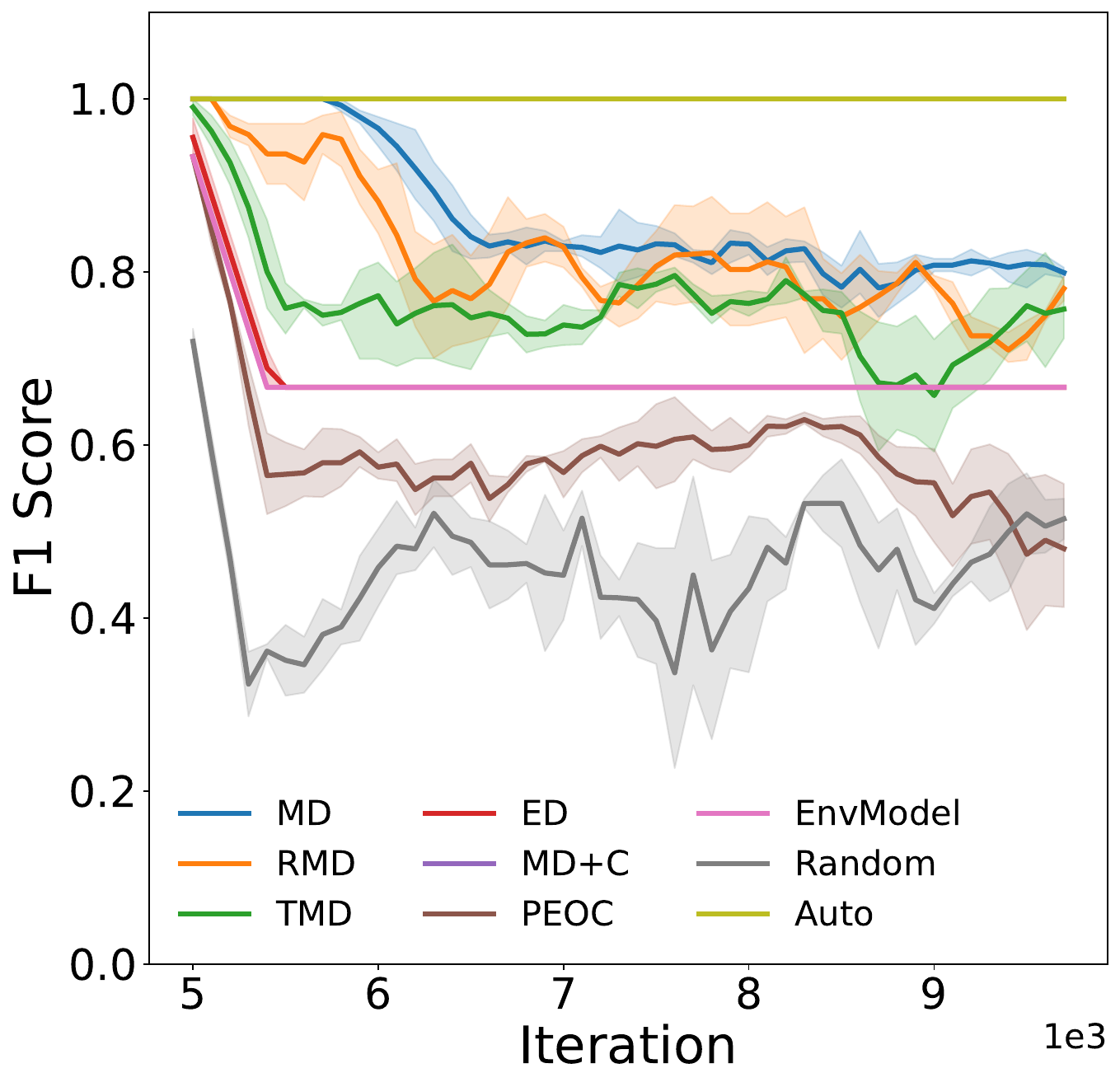}}
	\subfigure{\includegraphics[width=0.19\textwidth]{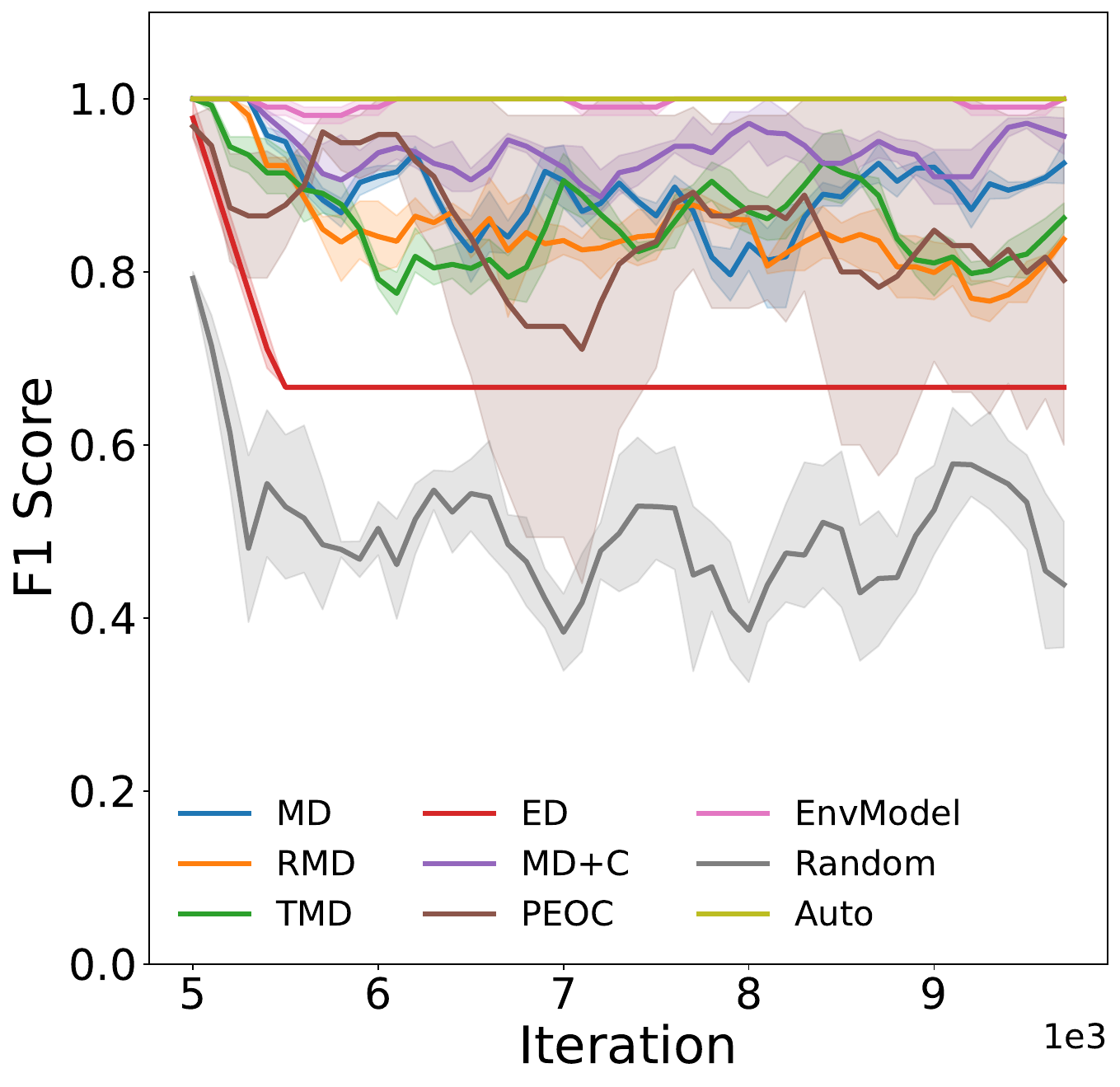}}
	\subfigure{\includegraphics[width=0.19\textwidth]{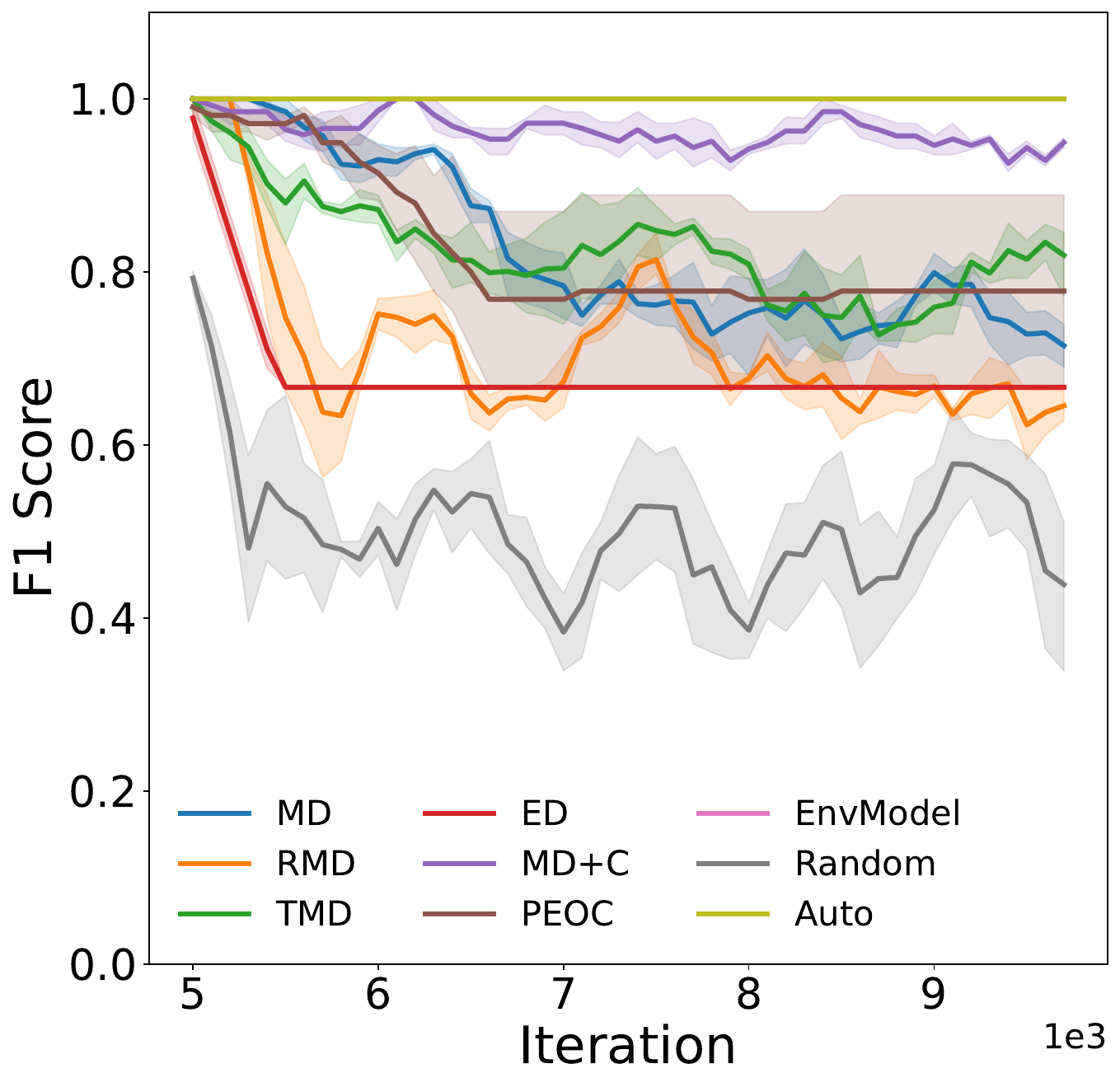}}
	\subfigure{\includegraphics[width=0.19\textwidth]{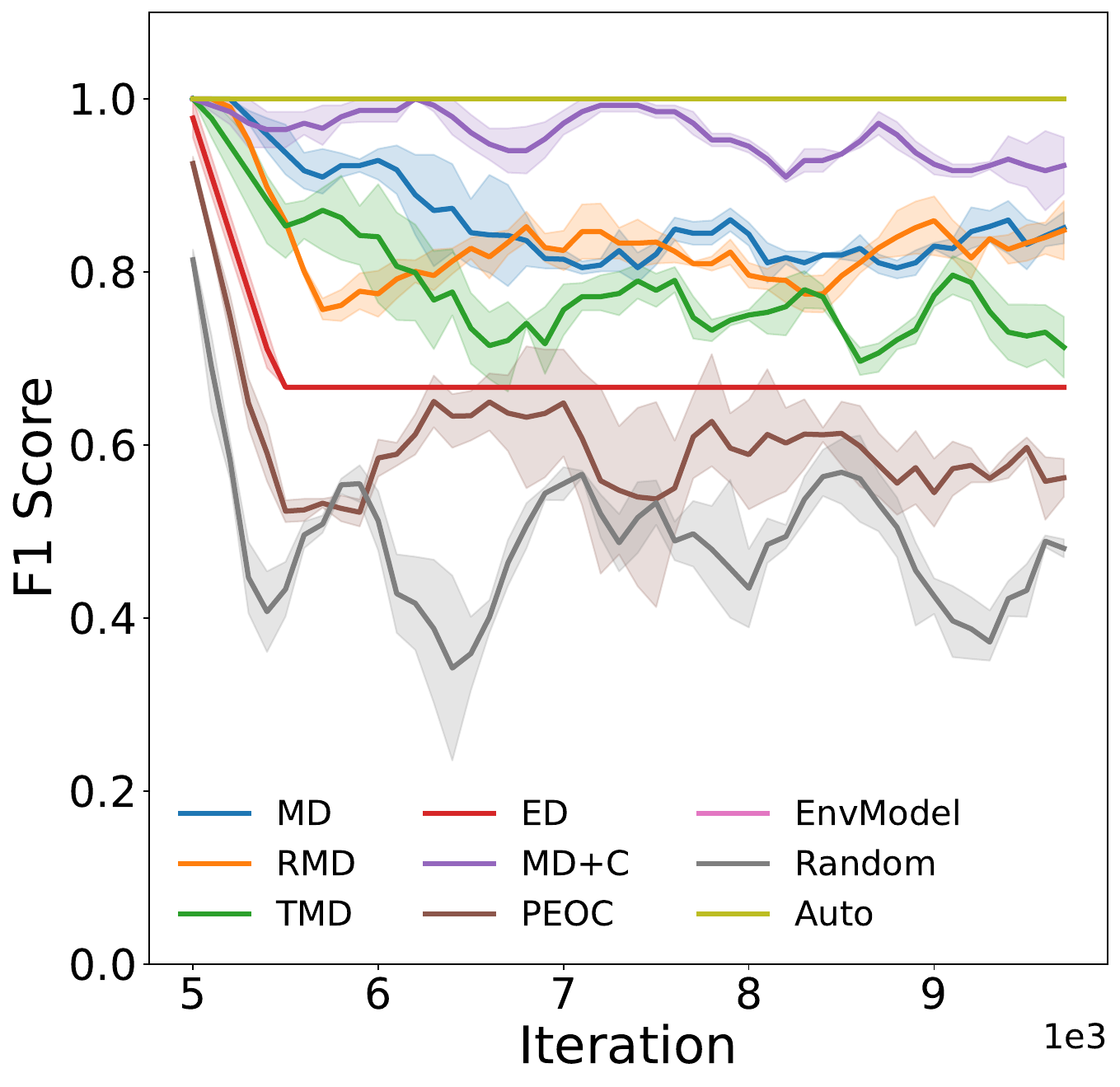}}
     \setcounter{subfigure}{0}
	\subfigure[Gaussian std=1]{\includegraphics[width=0.19\textwidth]{pic/relation_Tutankham_21.pdf}}
	\subfigure[Gaussian std=0.3]{\includegraphics[width=0.19\textwidth]{pic/relation_Tutankham_22.pdf}}
	\subfigure[OOD Enduro]{\includegraphics[width=0.19\textwidth]{pic/relation_Tutankham_23.pdf}}
	\subfigure[OOD FishingDerby]{\includegraphics[width=0.19\textwidth]{pic/relation_Tutankham_24.pdf}}
	\subfigure[Adversarial]{\includegraphics[width=0.19\textwidth]{pic/relation_Tutankham_25.pdf}}
	\caption{Detection performance across various state outliers in the online training on Tutankham.}
	\label{fig:Tutankham_online_full}
\end{figure*}

% \textbf{Main Results.}
As a supplement to the results on the main pages, we provide the whole results on two feature-input tasks and six Atari games from~\cref{fig:CartPole_online_full} to~\cref{fig:Tutankham_online_full}.
% In addition to Figures~\ref{fig:FishingDerby_online_full} and \ref{fig:Tutankham_online_full} on FishingDerby and Tutankham games, the main result on Enduro game in the training phase is also provided in  Figure~\ref{fig:Enduro_online_full}.
The "Mean Score" in the first row indicates the accumulated rewards of PPO, and the "F1 Score" in the second row shows the detection performance during RL training. The F1 score is computed based on precision and recall. 
The third row shows the relationship between the average F1 score and policy performance during training. 
We can find that higher detection accuracy is generally associated with better policy performance. We also find that the cumulative reward is not strongly correlated with detection ability in some games. A high detection accuracy may only improve the cumulative reward to a small degree. This suggests that we need more metrics to measure the effect of our detection performance more effectively.
Hyperparameters in our methods are shown in~\cref{appendix:hyperparameter}.

\begin{table}[htbp]
	\centering
	\scalebox{1}{
		\begin{tabular}{cc}\toprule[2pt]
			\specialrule{0pt}{1pt}{1pt}
			Hyperparameter & Value  \\ 
			\hline\specialrule{0pt}{1pt}{1pt}
			Confidence level (1-$\alpha$) & 1-0.05  \\\specialrule{0pt}{1pt}{1pt}
			Moving window size ($m$) & 5120  \\\specialrule{0pt}{1pt}{1pt}
			Sample size ($N_c$)  & 2560 \\\specialrule{0pt}{1pt}{1pt}
			Iteration ($K$) & $\approx$ 10000 (1e7 steps in total) \\\specialrule{0pt}{1pt}{1pt}
			Environment number ($N$) & 8  \\\specialrule{0pt}{1pt}{1pt}
			Horizon ($T$)     & 128   \\\specialrule{0pt}{1pt}{1pt}
			% 		Feature dimension ($k$) & 50 \\\specialrule{0pt}{1pt}{1pt}
			\specialrule{0pt}{1pt}{1pt}\bottomrule[2pt]
		\end{tabular}
	}
	\caption{Hyper-parameters in the training phase. RL-related parameters are the same as those of the PPO algorithm.}
	\label{appendix:hyperparameter}
\end{table}

\clearpage
\subsection{Ablation Study on Double Anomaly Detectors}\label{appendix:Training Phase_ablation_double}
\begin{figure*}[htbp]
	\centering
	\subfigure[Gaussian std=1.]{\includegraphics[width=0.35\textwidth]{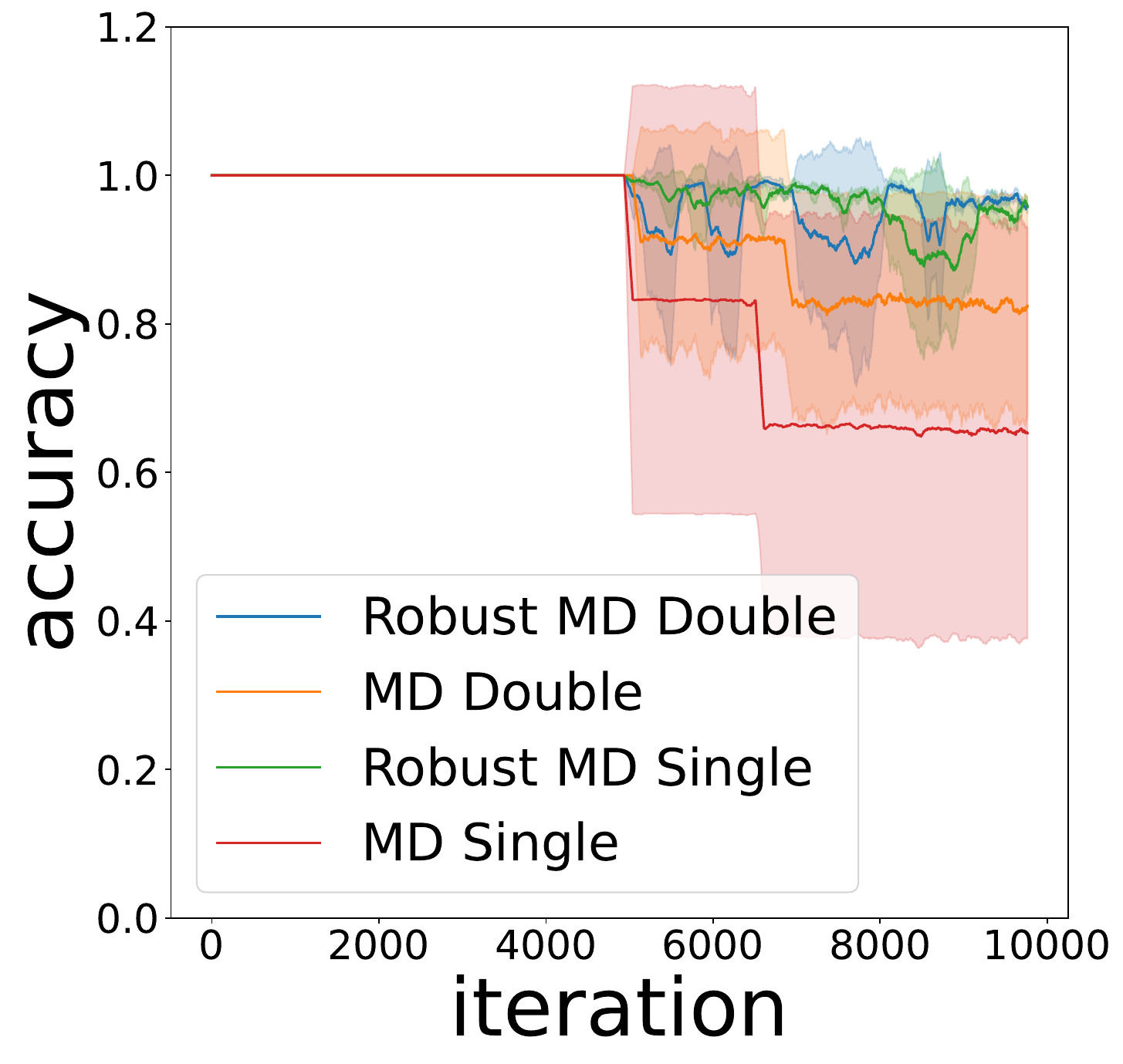}}
	\subfigure[OOD Asterix.]{\includegraphics[width=0.35\textwidth]{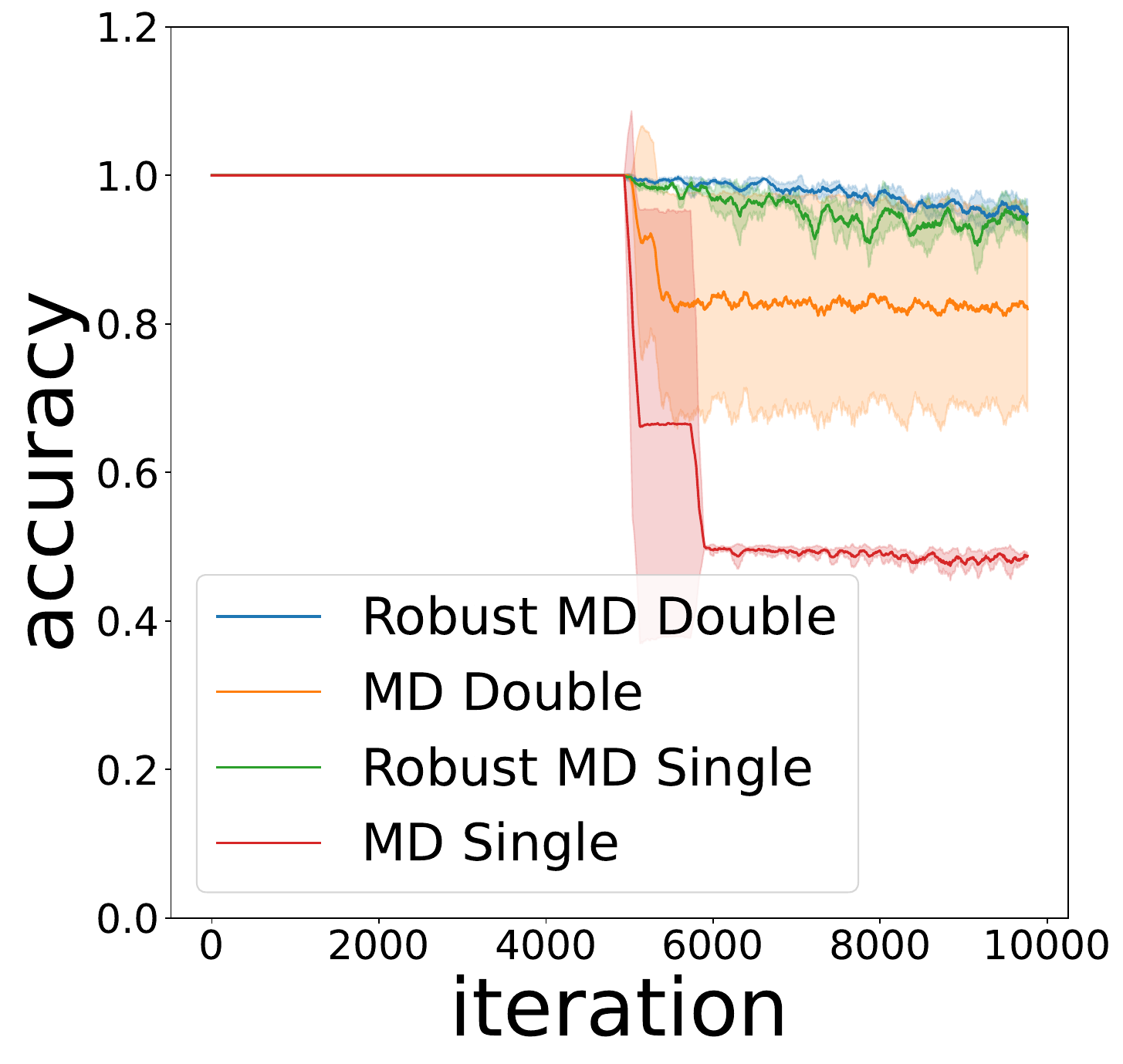}}
	\vskip -0.1in
	\caption{The detection accuracy with and without double self-supervised detectors on Breakout with random and OOD outliers on Breakout.}
	\label{fig:ablation_study_double}
\end{figure*}

% \textbf{Ablation Study on Double Anomaly Detectors.}
\cref{fig:ablation_study_double} reveals that double self-supervised detectors can help adjust the detection errors and improve the detection accuracy compared with the single detector. MD with double detectors outperforms MD with a single detector significantly, although RMD with double detectors is comparable to RMD with a single detector. 

\subsection{Ablation Study on Number of Noisy Environments}\label{appendix:Training Phase_ablation_noises}
\begin{figure*}[htbp]
	\centering
	\subfigure[Gaussian std=0.3]{\includegraphics[width=0.35\textwidth]{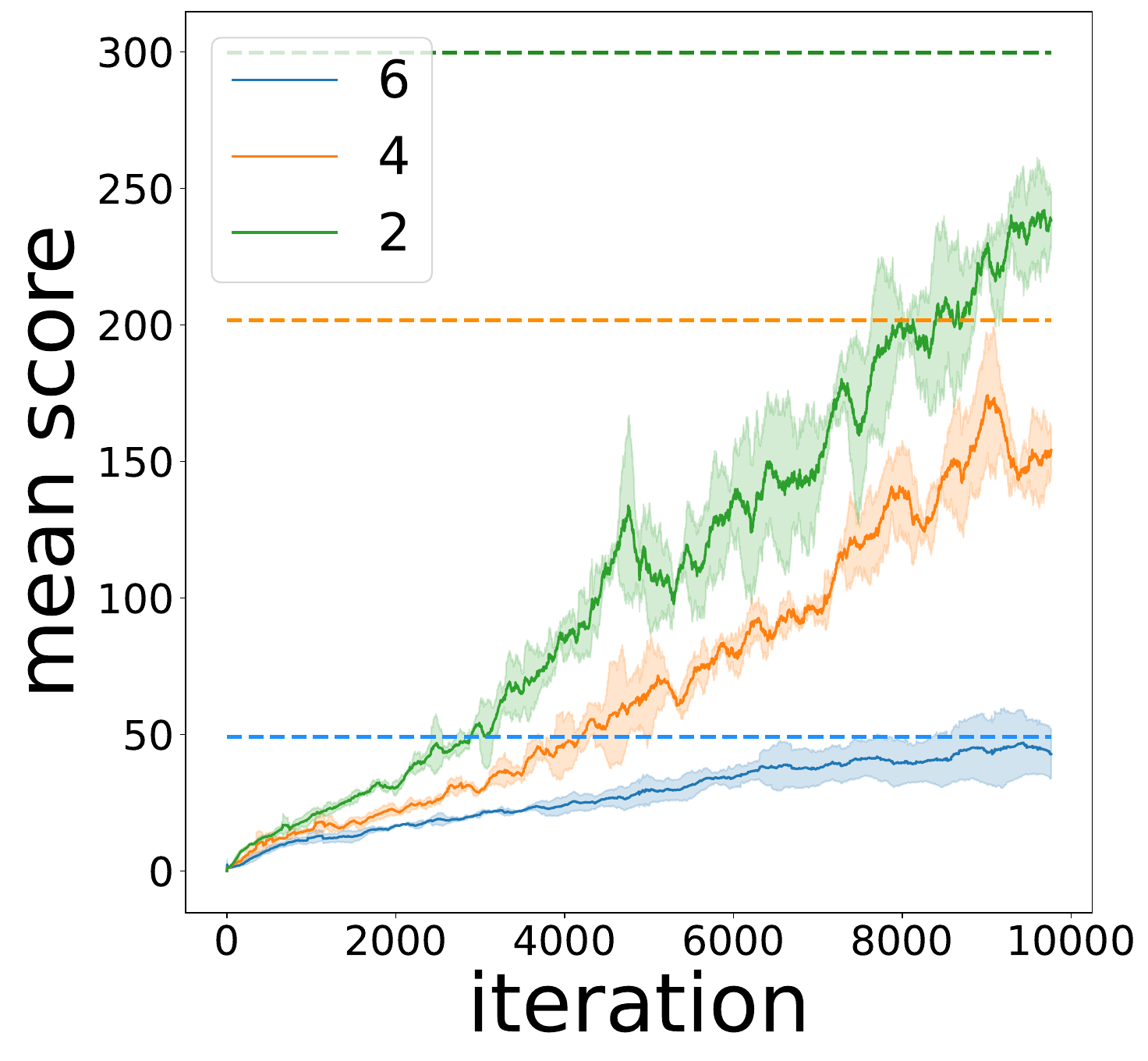}}
	\subfigure[OOD Asterix.]{\includegraphics[width=0.35\textwidth]{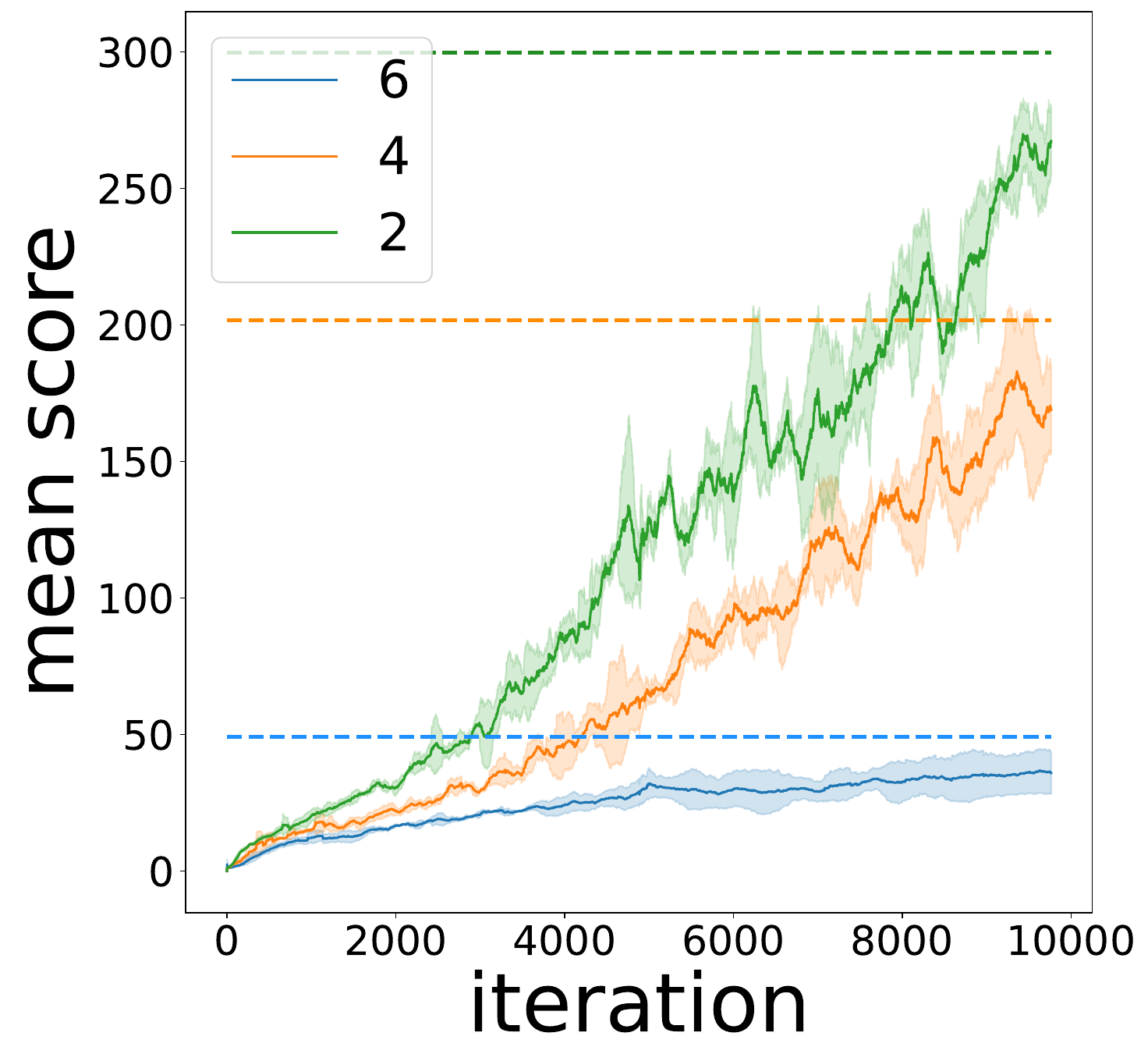}}
	\vskip -0.1in
	\caption{Training performance under Robust MD detection under different proportions of outlier exposure on Breakout (2, 4, 6 out of 8 environments).}
	\label{fig:Ablation_study}
\end{figure*}

% \textbf{Ablation Study on Number of Noisy Environments.}
We train PPO in two, four, or six noisy environments with random and OOD outliers among all eight parallel environments. We use PCA to reduce the feature vectors to 50 dimensions and estimate the detector using Robust MD.  \cref{fig:Ablation_study} illustrates that compared with the \textbf{Auto} baseline, our RMD method is robust when encountering different ratios of outliers, especially with a higher contamination ratio. 
The dashed lines in different colors represent \textbf{Auto} baselines that correspond to the different number of noisy environments. The training performance with our detection method gradually approaches the ideal baselines, i.e., \textbf{Auto}.

% \clearpage
% \section{Results in Classical Control Environments}

% We also deploy the PPO algorithm in two classical control environments, including Cartpole and MountainCar.

\end{document}